\documentclass[10pt,logo,copyright]{nvidiatechreport}
\usepackage[authoryear,round]{natbib}

\usepackage[utf8]{inputenc} 
\usepackage[T1]{fontenc}    

\usepackage{parskip}        
\usepackage{url}            
\usepackage{booktabs}       
\usepackage{amsfonts}       
\usepackage{nicefrac}       
\usepackage{microtype}      
\usepackage{xcolor}         
\usepackage[dvipsnames]{xcolor} 
\usepackage{graphicx}
\usepackage{animate}        
\usepackage{subcaption}
\usepackage{tabularx}
\usepackage{makecell}
\usepackage{adjustbox}
\usepackage{setspace}
\newcolumntype{M}[1]{>{\centering\arraybackslash}m{#1}}
\usepackage{float}
\usepackage{placeins}
\usepackage{tikz}
\usetikzlibrary{positioning,shapes,shapes.geometric,arrows, decorations.pathreplacing, backgrounds}
\usepackage{amsmath,amsfonts,bm, bbm,leftindex}
\usepackage{multirow}
\usepackage{comment}
\usepackage{gensymb}
\usepackage{lipsum}
\usetikzlibrary{arrows.meta, positioning, fit}
\usepackage[para]{threeparttable}
\usetikzlibrary{tikzmark}
\usepackage{tasks}

\usepackage[most]{tcolorbox}
\usepackage{fancyvrb}
\usepackage{fvextra}
\usepackage{dashrule}
\usepackage{nicematrix}

\usepackage{multicol}

\makeatletter
\renewcommand{\paragraph}{%
  \@startsection{paragraph}{4}{\z@}%
    {1\baselineskip \@plus .2\baselineskip}
    {-0.4em}
    {\normalfont\normalsize\bfseries}%
}
\makeatother

\def\eqref#1{equation~\ref{#1}}

\def\1{\bm{1}}

\newcommand{\txtvar}{l}
\newcommand{\vidvar}{v}
\newcommand{\actvar}{a}
\newcommand{\audiovar}{s}

\newcommand{\txttok}{\txtvar}

\newcommand{\vidtok}{\vidvar}

\newcommand{\acttok}{\actvar}

\newcommand{\audiotok}{\audiovar}

\newcommand{\noisyvidvar}{\tilde{\vidvar}}
\newcommand{\noisyactvar}{\tilde{\actvar}}
\newcommand{\noisyaudiovar}{\tilde{\audiovar}}

\newcommand{\noisyvidtok}{\noisyvidvar}

\newcommand{\noisyacttok}{\noisyactvar}

\newcommand{\noisyaudiotok}{\noisyaudiovar}

\newcommand{\txttokseq}{\txttok_{1},\ldots,\txttok_{n}}

\newcommand{\eostok}{\langle\text{EOS}\rangle}
\newcommand{\bovtok}{\langle\text{BOG}\rangle}

\DeclareMathAlphabet{\mathsfit}{\encodingdefault}{\sfdefault}{m}{sl}
\SetMathAlphabet{\mathsfit}{bold}{\encodingdefault}{\sfdefault}{bx}{n}

\newcommand{\preseq}{\mathbf{S}_{\mathrm{AR}}}

\newcommand{\E}{\mathbb{E}}

\makeatletter
\let\save@mathaccent\mathaccent
\newcommand*\if@single[3]{%
  \setbox0\hbox{${\mathaccent"0362{#1}}^H$}%
  \setbox2\hbox{${\mathaccent"0362{\kern0pt#1}}^H$}%
  \ifdim\ht0=\ht2 #3\else #2\fi
  }
\newcommand*\rel@kern[1]{\kern#1\dimexpr\macc@kerna}
\newcommand*\widebar[1]{\@ifnextchar^{{\wide@bar{#1}{0}}}{\wide@bar{#1}{1}}}
\newcommand*\wide@bar[2]{\if@single{#1}{\wide@bar@{#1}{#2}{1}}{\wide@bar@{#1}{#2}{2}}}
\newcommand*\wide@bar@[3]{%
  \begingroup
  \def\mathaccent##1##2{%
    \let\mathaccent\save@mathaccent
    \if#32 \let\macc@nucleus\first@char \fi
    \setbox\z@\hbox{$\macc@style{\macc@nucleus}_{}$}%
    \setbox\tw@\hbox{$\macc@style{\macc@nucleus}{}_{}$}%
    \dimen@\wd\tw@
    \advance\dimen@-\wd\z@
    \divide\dimen@ 3
    \@tempdima\wd\tw@
    \advance\@tempdima-\scriptspace
    \divide\@tempdima 10
    \advance\dimen@-\@tempdima
    \ifdim\dimen@>\z@ \dimen@0pt\fi
    \rel@kern{0.6}\kern-\dimen@
    \if#31
      \overline{\rel@kern{-0.6}\kern\dimen@\macc@nucleus\rel@kern{0.4}\kern\dimen@}%
      \advance\dimen@0.4\dimexpr\macc@kerna
      \let\final@kern#2%
      \ifdim\dimen@<\z@ \let\final@kern1\fi
      \if\final@kern1 \kern-\dimen@\fi
    \else
      \overline{\rel@kern{-0.6}\kern\dimen@#1}%
    \fi
  }%
  \macc@depth\@ne
  \let\math@bgroup\@empty \let\math@egroup\macc@set@skewchar
  \mathsurround\z@ \frozen@everymath{\mathgroup\macc@group\relax}%
  \macc@set@skewchar\relax
  \let\mathaccentV\macc@nested@a
  \if#31
    \macc@nested@a\relax111{#1}%
  \else
    \def\gobble@till@marker##1\endmarker{}%
    \futurelet\first@char\gobble@till@marker#1\endmarker
    \ifcat\noexpand\first@char A\else
      \def\first@char{}%
    \fi
    \macc@nested@a\relax111{\first@char}%
  \fi
  \endgroup
}
\makeatother

\usepackage{tikz}
\usetikzlibrary{arrows.meta,positioning}
\usepackage{pifont}
\usepackage{diagbox}
\usepackage{pgfplots}
\pgfplotsset{compat=1.18}
\usepackage{colortbl}
\usepackage{graphicx}
\usepackage{wrapfig}

\usepackage[nameinlink]{cleveref}
\crefname{equation}{Eq.}{Eqs.}
\crefname{figure}{Fig.}{Figs.}
\crefname{section}{Sec.}{Sec.}
\crefname{appendix}{App.}{App.}
\crefname{table}{Tab.}{Tabs.}
\crefname{algorithm}{Algo}{Algo}
\crefname{thm}{Thm}{Thm}
\Crefname{thm}{Thm}{Thm}
\crefname{prop}{Prop}{Prop}

\definecolor{darkred}{rgb}{0.7, 0.0, 0.0}
\definecolor{rowours}{HTML}{E8F4E0}
\definecolor{columnours}{HTML}{E8F4E0}

\newcommand{\crefnames}[3]{%
  \@for\next:=#1\do{%
    \expandafter\crefname\expandafter{\next}{#2}{#3}%
  }%
}

\title{Cosmos 3: Omnimodal World Models for Physical AI}

\author{NVIDIA\footnote{Contributors and acknowledgments are listed in Appendix~\ref{sec::contributors}.}}

\begin{abstract}
We introduce Cosmos 3, a family of omnimodal world models designed to jointly process and generate language, image, video, audio, and action sequences within a unified mixture-of-transformers architecture. By supporting highly flexible input-output configurations, Cosmos 3 seamlessly unifies critical modalities for Physical AI---effectively subsuming vision-language models, video generators, world simulators, and world-action models into a single framework. Our evaluation demonstrates that Cosmos 3 establishes a new state-of-the-art across a diverse suite of understanding and generation tasks, demonstrating omnimodal world models as scalable, general-purpose backbones for embodied agents. Our post-trained Cosmos 3 models were ranked as the best open-source Text-to-Image and Image-to-Video models by Artificial Analysis, and the best policy model by RoboArena at the time the technical report was written. To accelerate open research and deployment in Physical AI, we make our code, model checkpoints, curated synthetic datasets, and evaluation benchmark available under the Linux Foundation's \href{https://openmdw.ai/license/1-1/}{OpenMDW-1.1} License at \href{https://github.com/nvidia/cosmos}{github.com/nvidia/cosmos} and \href{https://huggingface.co/collections/nvidia/cosmos3}{huggingface.co/collections/nvidia/cosmos3} . The project website is available at \href{https://research.nvidia.com/labs/cosmos-lab/cosmos3}{research.nvidia.com/labs/cosmos-lab/cosmos3} .
\end{abstract}

\begin{document}
\maketitle
\abscontent

\vspace{8mm}

\begin{tcolorbox}[
    colback=black!2,
    colframe=nvidiagreen!75!white,
    boxrule=0.4pt,
    arc=1pt,
    left=4pt,
    right=4pt,
    top=4pt,
    bottom=4pt,
    title=Open-Source Code
]
\small
\begin{tabular}{p{4.5cm}l}
Cosmos & {\normalfont\scriptsize\href{https://github.com/nvidia/cosmos}{github.com/nvidia/cosmos}} \\
Cosmos-Framework & {\normalfont\scriptsize\href{https://github.com/nvidia/cosmos-framework}{github.com/nvidia/cosmos-framework}} \\
\end{tabular}
\end{tcolorbox}

\begin{tcolorbox}[
    colback=black!2,
    colframe=nvidiagreen!75!white,
    boxrule=0.4pt,
    arc=1pt,
    left=4pt,
    right=4pt,
    top=4pt,
    bottom=4pt,
    title=Open-Weight Model Checkpoint
]
\small
\begin{tabular}{p{4.5cm}l}
Cosmos3-Super & {\normalfont\scriptsize\href{https://huggingface.co/nvidia/Cosmos3-Super}{huggingface.co/nvidia/Cosmos3-Super}} \\
Cosmos3-Nano & {\normalfont\scriptsize\href{https://huggingface.co/nvidia/Cosmos3-Nano}{huggingface.co/nvidia/Cosmos3-Nano}} \\
Cosmos3-Super-Text2Image & {\normalfont\scriptsize\href{https://huggingface.co/nvidia/Cosmos3-Super-Text2Image}{huggingface.co/nvidia/Cosmos3-Super-Text2Image}} \\
Cosmos3-Super-Image2Video & {\normalfont\scriptsize\href{https://huggingface.co/nvidia/Cosmos3-Super-Image2Video}{huggingface.co/nvidia/Cosmos3-Super-Image2Video}} \\
Cosmos3-Nano-Policy-DROID & {\normalfont\scriptsize\href{https://huggingface.co/nvidia/Cosmos3-Nano-Policy-DROID}{huggingface.co/nvidia/Cosmos3-Nano-Policy-DROID}} \\
\end{tabular}
\end{tcolorbox}

\begin{tcolorbox}[
    colback=black!2,
    colframe=nvidiagreen!75!white,
    boxrule=0.4pt,
    arc=1pt,
    left=4pt,
    right=4pt,
    top=4pt,
    bottom=4pt,
    title=Open Synthetic Dataset
]
\small
\begin{tabular}{p{4.5cm}l}
SDG-PhyxSim & {\normalfont\scriptsize\href{https://huggingface.co/datasets/nvidia/PhysicalAI-WorldModel-Synthetic-Physical-Interaction-Scenes}{huggingface.co/datasets/nvidia/PhysicalAI-WorldModel-Synthetic-Physical-Interaction-Scenes}} \\
SDG-RobotSim & {\normalfont\scriptsize\href{https://huggingface.co/datasets/nvidia/PhysicalAI-WorldModel-Synthetic-Embodied-Robot-Scenes}{huggingface.co/datasets/nvidia/PhysicalAI-WorldModel-Synthetic-Embodied-Robot-Scenes}} \\
SDG-DriveSim & {\normalfont\scriptsize\href{https://huggingface.co/datasets/nvidia/PhysicalAI-WorldModel-Synthetic-Autonomous-Driving-Scenarios}{huggingface.co/datasets/nvidia/PhysicalAI-WorldModel-Synthetic-Autonomous-Driving-Scenarios}} \\
SDG-SynHuman & {\normalfont\scriptsize\href{https://huggingface.co/datasets/nvidia/PhysicalAI-WorldModel-Synthetic-Digital-Human-Scenes}{huggingface.co/datasets/nvidia/PhysicalAI-WorldModel-Synthetic-Digital-Human-Scenes}} \\
SDG-Warehouse & {\normalfont\scriptsize\href{https://huggingface.co/datasets/nvidia/PhysicalAI-WorldModel-Synthetic-Warehouse-Operations-Scenes}{huggingface.co/datasets/nvidia/PhysicalAI-WorldModel-Synthetic-Warehouse-Operations-Scenes}} \\
\end{tabular}
\end{tcolorbox}

\begin{tcolorbox}[
    colback=black!2,
    colframe=nvidiagreen!75!white,
    boxrule=0.4pt,
    arc=1pt,
    left=4pt,
    right=4pt,
    top=4pt,
    bottom=4pt,
    title=Open Evaluation Benchmark
]
\small
\begin{tabular}{p{4.5cm}l}
Cosmos-HUE & {\normalfont\scriptsize\href{https://huggingface.co/datasets/nvidia/Cosmos-HumanEval-v1}{huggingface.co/datasets/nvidia/Cosmos-HumanEval-v1}} \\
\end{tabular}
\end{tcolorbox}

\newpage
\tableofcontents
\newpage

\section{Introduction}
\label{sec::intro}

Physical AI agents perceive, reason, and take actions to interact with the real world. However, training such agents directly in the real world is slow, expensive, and could be dangerous. To overcome these bottlenecks, we must construct a training facility to enable safe and scalable learning in simulated worlds, where Physical AI agents acquire two fundamentally coupled capabilities: \textit{understanding} and \textit{generation}. Understanding allows an agent to infer latent representations, semantics, and dynamics from partial observations, and generation empowers the agent to predict and simulate plausible futures, anticipating how the world evolves and how the agent should take actions in response. Prior work has largely treated these two pillars in isolation, leading to separate discriminative models for perception and reasoning, such as Vision-Language Models (VLMs); generative models for world simulation, such as Video Generation Models and Forward Dynamics Models; and action-prediction models, such as Vision-Language-Action Models (VLAs) and World-Action Models (WAMs).

We argue that this paradigm separation is fundamentally limiting: understanding requires reasoning about the future evolution of the world and the consequences of actions, while generation relies on a compact, structured representation of the world and agent behaviors. Unifying them into a single scalable framework is therefore essential for Physical AI. Consider a general home robot instructed to clean a dining table after dinner. Under the current paradigm, the robot must stitch together a disjointed suite of models: a VLM to locate dishware and generate an executable plan, a VLA or WAM to generate action sequences, and a Forward Dynamics Model or ``World Model'' to simulate and evaluate future states. This fragmented architecture is suboptimal and computationally wasteful. Can we instead design a single, unified model that natively addresses all essential capabilities for Physical AI agents?

\begin{figure}[!tbh]
  \centering
  \resizebox{\linewidth}{!}{\includegraphics{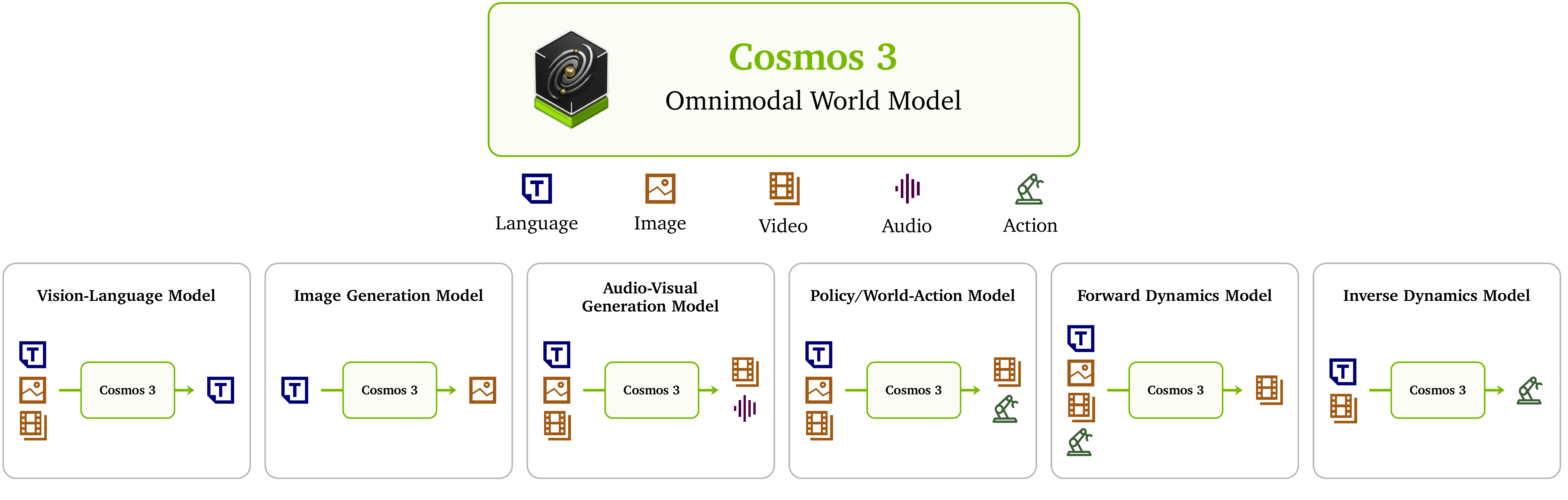}}
  \caption{\textbf{Cosmos 3 serves as a general-purpose backbone for Physical AI.} By jointly modeling language, image, video, audio, and action for both understanding and generation, Cosmos 3 unifies a wide range of model classes within a single network architecture, including vision-language models, image generation models, audio-visual generation models, policy or world-action models, forward dynamics models, and inverse dynamics models.}
  \label{fig:overview}
\end{figure}

We introduce Cosmos 3, a family of omnimodal world models that jointly model language, image, video, audio, and action for both understanding and generation.
Serving as a general-purpose backbone for Physical AI, Cosmos 3 unifies a wide array of distinct model classes into a single framework (\cref{fig:overview}).
Depending on the input-output configuration, Cosmos 3 seamlessly transitions between multiple operational modes: it can operate as a vision-language model for multimodal understanding and reasoning; a text-to-image generator, a video generator for text-to-video synthesis, image animation (image-to-video), future prediction (video-to-video), or synchronous audio-video generation; a world-action model for joint action prediction and environmental simulation.
By unifying perception, simulation, and execution without architectural modifications, Cosmos 3 eliminates the need for fragmented, task-specific pipelines, enabling scalable learning through shared representations and joint multi-task supervision.

\begin{figure}[t]
  \centering
  \resizebox{0.6\linewidth}{!}{\includegraphics{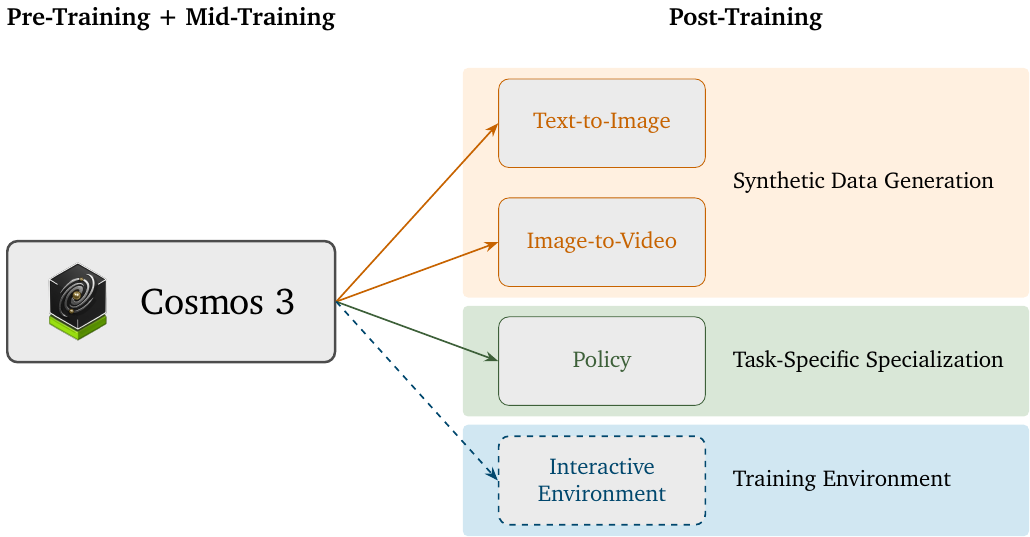}}
  \vspace{-0.5em}
  \caption{\textbf{Cosmos 3 offers a strong starting point for training Physical AI agents.} Cosmos 3 can be post-trained on target data for distinct applications without architectural modifications. In this paper, we demonstrate how we post-train Cosmos 3 for better synthetic data generation (\cref{sec::t2i_post_train} and~\cref{sec::i2v_post_train}) and better robot policy (\cref{sec::robot_policy_post_train}). In the future, we expect Cosmos 3 to play an essential role in generating high-quality, complex environments for training Physical AI agents.}
  \label{fig:cosmos_platform}
\end{figure}

Scaling training data and environments for Physical AI agents remains a persistent bottleneck. Cosmos 3 offers a strong starting point to address this challenge in three ways: (i) synthetic data generation, (ii) task-specific specialization, and (iii) training environment (\cref{fig:cosmos_platform}).
In the near term, Cosmos 3 synthesizes high-fidelity, diverse visual data to enhance training for Physical AI agents. We demonstrate how we can post-train Cosmos 3 into a better synthetic data generator in~\cref{sec::t2i_post_train} and~\cref{sec::i2v_post_train}.
Since agents perceive and interact with environments through diverse embodiments and tasks, Cosmos 3 supports task- and embodiment-specific specialization on top of a shared model. As a powerful mid-training model for Physical AI, Cosmos 3 establishes a better starting point by modeling general world dynamics and action priors while remaining highly amenable to downstream adaptation. In practice, the model can be post-trained on target data for distinct applications without architectural modifications, enabling data-driven specialization that retains a common world representation thanks to its omnimodal design.
\cref{sec::robot_policy_post_train} describes how we post-train Cosmos 3 into a highly capable world-action model on DROID.
In the long term, Cosmos 3 is positioned to generate high-quality, complex training environments for Physical AI agents. To accelerate open research and deployment in Physical AI, we release our code, model checkpoints, curated synthetic datasets, and an evaluation benchmark under the OpenMDW-1.1 License at \href{https://github.com/nvidia/cosmos}{github.com/nvidia/cosmos} and \href{https://huggingface.co/collections/nvidia/cosmos3}{huggingface.co/collections/nvidia/cosmos3} .

\newlength{\resultsOverviewModelColWidth}
\newlength{\resultsOverviewCornerHeight}
\newcommand{\resultsOverviewCornerCell}{%
    \begin{tikzpicture}[baseline=(current bounding box.center)]
        \path[use as bounding box] (0,0) rectangle (\resultsOverviewModelColWidth,\resultsOverviewCornerHeight);
        \draw[line width=0.4pt] (0,\resultsOverviewCornerHeight) -- (\resultsOverviewModelColWidth,0);
        \node[anchor=north east,inner sep=0pt,xshift=-0.2em,yshift=-0.1ex] at (\resultsOverviewModelColWidth,\resultsOverviewCornerHeight) {\textbf{Capability}};
        \node[anchor=south west,inner sep=0pt,xshift=0.2em,yshift=0.1ex] at (0,0) {\textbf{Model}};
    \end{tikzpicture}%
}

\begin{table*}[t]
    \centering
    \footnotesize
    \caption{\textbf{Cosmos 3 results overview.} Cosmos 3 consistently outperforms specialized open-source baselines across all capabilities. Detailed results can be found in~\cref{sec::results}. In the table, $^\ast$ denotes post-trained Cosmos 3 variants; $\dagger$ denotes closed models; gray-colored cells in each row indicate the model does not possess the corresponding capabilities.}
    \label{tab:results_overview}
    \settowidth{\resultsOverviewModelColWidth}{Lingbot-World-Base (C)}
    \setlength{\resultsOverviewCornerHeight}{2.25\baselineskip}
    \setlength{\tabcolsep}{4pt}
    \resizebox{\textwidth}{!}{%
    \begin{tabular}{l|cccc|cccccc}
        \toprule
        \multirow{2}{*}{\resultsOverviewCornerCell} & \multicolumn{4}{c|}{\textbf{Reasoning}} & \multicolumn{6}{c}{\textbf{Generation}} \\
        \cmidrule(lr){2-5} \cmidrule(lr){6-11}
        & \textbf{General} & \textbf{Robotics} & \textbf{Smart infra.} & \textbf{Driving} & \textbf{Text2Image} & \textbf{Text2Video} & \textbf{Image2Video} & \textbf{Audio} & \textbf{FD: Robot} & \textbf{Policy: Robot} \\
        \midrule
        \rowcolor{rowours}
        \textbf{Cosmos3-Super} & \underline{73.7} & \underline{57.8} & \textbf{62.6} & \textbf{79.3} & \textbf{91.36$^{\ast}$} & \textbf{80.0} & \textbf{82.8} & 7.31 & \textbf{26.0$^{\ast}$} & - \\
        \rowcolor{rowours}
        \textbf{Cosmos3-Nano} & 69.6 & 55.1 & \underline{61.0} & \underline{76.0} & 84.61 & \underline{79.4} & \underline{82.7} & \underline{7.34} & \underline{25.5$^{\ast}$} & \textbf{39.7$^{\ast}$} \\
        \midrule
        Gemini 3.1 Pro$^\dagger$ & \textbf{77.5} & \textbf{58.2} & 58.6 & 47.2 & \cellcolor{black!5} & \cellcolor{black!5} & \cellcolor{black!5} & \cellcolor{black!5} & \cellcolor{black!5} & \cellcolor{black!5} \\
        Qwen3-VL-32B & 72.8 & 52.6 & 56.1 & 40.7 & \cellcolor{black!5} & \cellcolor{black!5} & \cellcolor{black!5} & \cellcolor{black!5} & \cellcolor{black!5} & \cellcolor{black!5} \\
        Qwen3-VL-8B & 68.9 & 48.5 & 52.7 & 46.4 & \cellcolor{black!5} & \cellcolor{black!5} & \cellcolor{black!5} & \cellcolor{black!5} & \cellcolor{black!5} & \cellcolor{black!5} \\
        Gemma-4-31B & 69.8 & 51.0 & 51.3 & 36.6 & \cellcolor{black!5} & \cellcolor{black!5} & \cellcolor{black!5} & \cellcolor{black!5} & \cellcolor{black!5} & \cellcolor{black!5} \\
        Gemma-4-E4B & 53.1 & 39.3 & 29.4 & 26.0 & \cellcolor{black!5} & \cellcolor{black!5} & \cellcolor{black!5} & \cellcolor{black!5} & \cellcolor{black!5} & \cellcolor{black!5} \\
        Gemini 3 Pro Image$^\dagger$ & \cellcolor{black!5} & \cellcolor{black!5} & \cellcolor{black!5} & \cellcolor{black!5} & \underline{90.85} & \cellcolor{black!5} & \cellcolor{black!5} & \cellcolor{black!5} & \cellcolor{black!5} & \cellcolor{black!5} \\
        Qwen-Image-2512 & \cellcolor{black!5} & \cellcolor{black!5} & \cellcolor{black!5} & \cellcolor{black!5} & 84.25 & \cellcolor{black!5} & \cellcolor{black!5} & \cellcolor{black!5} & \cellcolor{black!5} & \cellcolor{black!5} \\
        Veo-3.1$^\dagger$ & \cellcolor{black!5} & \cellcolor{black!5} & \cellcolor{black!5} & \cellcolor{black!5} & \cellcolor{black!5} & 79.1 & 82.6 & \textbf{7.45} & \cellcolor{black!5} & \cellcolor{black!5} \\
        Wan2.2-A14B & \cellcolor{black!5} & \cellcolor{black!5} & \cellcolor{black!5} & \cellcolor{black!5} & \cellcolor{black!5} & 78.0 & 81.3 & \cellcolor{black!5} & \cellcolor{black!5} & \cellcolor{black!5} \\
        Ctrl-World & \cellcolor{black!5} & \cellcolor{black!5} & \cellcolor{black!5} & \cellcolor{black!5} & \cellcolor{black!5} & \cellcolor{black!5} & \cellcolor{black!5} & \cellcolor{black!5} & 23.0 & \cellcolor{black!5} \\
        $\pi_{0.5}$ & \cellcolor{black!5} & \cellcolor{black!5} & \cellcolor{black!5} & \cellcolor{black!5} & \cellcolor{black!5} & \cellcolor{black!5} & \cellcolor{black!5} & \cellcolor{black!5} & \cellcolor{black!5} & \underline{28.1} \\
        \bottomrule
    \end{tabular}%
    }
\end{table*}

We evaluate Cosmos 3 and its post-trained variants on a wide range of benchmarks, covering essential understanding and generation capabilities for Physical AI. \cref{tab:results_overview} provides a summary of our benchmark results, detailed in~\cref{sec::results}. As shown in the table, Cosmos 3 establishes a new state-of-the-art across most capabilities, being highly competitive or outperforming specialized models.

The technical details of Cosmos 3 are organized as follows.
\cref{sec::model} introduces the model architecture, including encoders for all modalities, the arrangement of multimodal tokens to enable different generation modes, the Mixture-of-Transformers (MoT) backbone, multimodal position embedding, and model variants.
\cref{sec::data} outlines the training data for the reasoner and generator training.
\cref{sec::training} details the training recipes for reasoner and generator.
\cref{sec::infrastructure} describes infrastructure, including data, training, serving, and evaluation.
\cref{sec::results} presents our experimental results.
\cref{sec::related_work} discusses related work and \cref{sec::conclusion} concludes the paper.

\section{Model Architecture}
\label{sec::model}

Cosmos 3 is capable of processing multimodal inputs and generating multimodal outputs. Beyond language, vision (image and video), and audio, Cosmos 3 treats action as a core modality, introducing a dedicated class of action tokens. These action tokens bridge the physical world with language-based reasoning and video-based world modeling, linking directly to physically grounded control signals for real-world interaction. Cosmos 3 integrates modality-specific encoders to project different modalities into a unified representation space, which is then processed by a Mixture-of-Transformers (MoT) backbone. During inference, language tokens are generated via next-token prediction, while other modalities are generated through iterative denoising.

\subsection{Encoders}
Given an input sequence of language, vision, audio, and action, the first step is to embed them into a unified representation space using modality-specific encoders. To enable the shared transformer parameters and positional embeddings to distinguish between different modalities, we add a learnable, modality-specific embedding vector to each non-language modality before feeding it into the MoT backbone.

\subsubsection{Image and Video}
We adopt two separate encoders for visual input. For visual understanding, we use a ViT encoder pre-trained with vision-language alignment. For visual generation, we use the video VAE encoder from Wan2.2-TI2V-5B~\citep{wan2025}. The ViT encoder has a $16\times16$ patch size, followed by a two-layer MLP that merges $2\times2$ tokens and projects them into the latent space of the transformer. Following Qwen3-VL~\citep{qwen3vl2025}, we also aggregate visual features from ViT via DeepStack~\citep{meng2024deepstack} and insert text–based video timestamps interleaved with video frames~\citep{chen2024timemarker}. The VAE compresses the input video temporally by $4\times$ and spatially by $32\times32$, implemented as $16\times16$ spatial compression followed by a $2\times2$ patch merge. We use a linear layer to project each VAE token into the transformer's hidden dimension before feeding the latents into the MoT backbone. The ViT encoder for understanding is jointly trained with the backbone, while the VAE encoder for generation is kept frozen during training.

\subsubsection{Audio}
For audio generation, we adopt the audio VAE architecture from~\citet{lee2024etta}. The raw stereo audio sampled at 48\,kHz is encoded with a hop size of 1920 samples, resulting in 25 tokens per second of audio. The audio VAE is frozen during training. As with the other non-text modalities, audio tokens are projected into the transformer's hidden dimension using a linear layer before entering the MoT backbone.

\subsubsection{Action}
\label{sec:action}
We support action modeling across diverse embodiments, including autonomous vehicles, camera motion, robots, and egocentric human motion (head and hands). Since each domain exposes its own native control space---such as joint trajectories, steering commands, body poses, or camera transformations---we map them into a unified action interface that enables consistent multimodal reasoning, generation, and policy learning across domains.

\begin{figure}[t]
    \centering
    \resizebox{\linewidth}{!}{\includegraphics{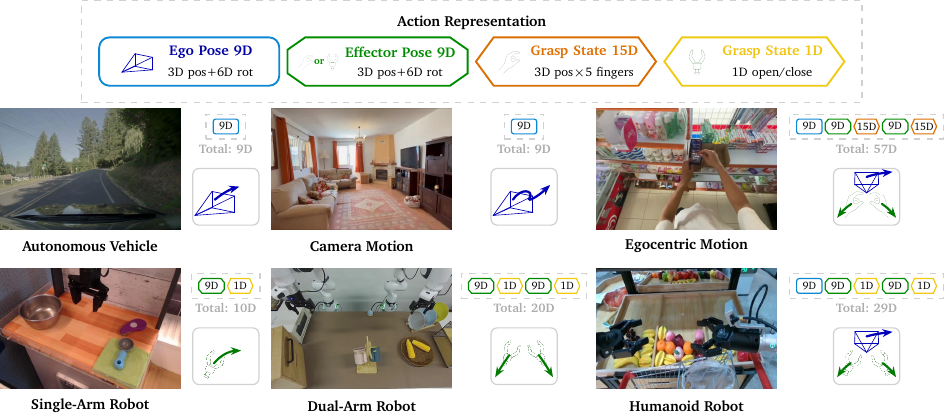}}
    \caption{\textbf{Unified action representation.} We map heterogeneous embodiment controls into compact action vectors built from shared geometric components. Ego and effector motions are encoded as relative-pose pseudo-actions using 3D translation and 6D rotation (an over-parameterized rotation representation by Zhou~\citet{zhou2019continuity}, as the degree of freedom of rotation is 3), while grasp states directly encode the current manipulation state, such as fingertip positions for hands or gripper open/close values for robots. Domain-aware input and output projections handle heterogeneous action-vector lengths while preserving the shared semantic space.}
    \label{fig:action_representation}
\end{figure}

\paragraph{Action representations.}
We use actions to denote causal variables that induce changes in the world state.
Given consecutive video tokens, an action token $\acttok_{t}$ represents the transition from the previous state $\vidtok_{t{-}1}$ to the current state $\vidtok_{t}$.
Each embodiment source is transformed into a compact representation that captures a shared underlying geometric structure across different action domains, as illustrated in~\cref{fig:action_representation}.
At a high level, actions can include up to three components: ego poses for the agent's main observation frame, effector poses for the agent's effectors, and grasp states for the manipulation state.
To avoid embodiment-specific controller details such as Proportional–Integral–Derivative (PID) parameters or low-level actuation interfaces, ego and effector poses are represented as pseudo-actions derived from state differences.
For consecutive $\mathrm{SE}(3)$ poses $\mathbf{T}_{t-1}$ and $\mathbf{T}_t$, we represent motion as the relative transform $\Delta \mathbf{T}_t = \mathbf{T}_{t-1}^{-1}\mathbf{T}_t$.
We use the 6D representation following~\citet{zhou2019continuity} and the OpenCV convention for rotations where the z-axis is along the fingers/grippers and x-axis is to the right.
Grasp states, however, are treated differently: rather than representing temporal differences, they directly encode the current manipulation state at time t.

For cameras and autonomous vehicles, actions are represented by ego poses only, without any effector poses or grasp states. For egocentric data, we use head-camera pose deltas as ego poses, wrist-pose deltas as effector poses, and fingertip positions in each wrist frame as grasp states~\citep{yang2025egovla}. For robotic data, we use head-camera pose deltas as ego poses, end-effector flange-pose deltas as effector poses~\citep{lyu2026lda}, and continuous gripper open/close values as grasp states.

\paragraph{Action tokenization.} Our action representation maps diverse embodiments into a shared latent action space while preserving embodiment-specific structure and semantics.
We therefore use domain-aware input and output projection layers with separate weight matrices for each embodiment domain~\citep{zheng2026xvla}, while sharing the MoT backbone.
For an input $\mathbf{x} \in \mathbb{R}^{d_{\text{in}}^{(k)}}$, such as an egocentric action vector concatenating the head-pose delta, left and right wrist-pose deltas, and fingertip coordinates, and domain identifier $k \in \{1, \ldots, K\}$, the input projection is:
\begin{equation}
    \mathbf{z} = \mathbf{W}_{\mathrm{in}}^{(k)} \mathbf{x} + \mathbf{b}_{\mathrm{in}}^{(k)}
\end{equation}

where $\mathbf{z} \in \mathbb{R}^{d_{\text{model}}}$ is the latent action token, $\mathbf{x}$ denotes the normalized action vector, and $\mathbf{W}_{\mathrm{in}}^{(k)} \in \mathbb{R}^{d_{\text{model}} \times d_{\text{in}}^{(k)}}$ and $\mathbf{b}_{\mathrm{in}}^{(k)} \in \mathbb{R}^{d_{\text{model}}}$ are the domain-specific input projection matrix and bias.

To decode the tokens back to the original action space, we use a domain-specific output projection:
\begin{equation}
    \mathbf{x} = \mathbf{W}_{\mathrm{out}}^{(k)} \mathbf{z} + \mathbf{b}_{\mathrm{out}}^{(k)}
\end{equation}
where $\mathbf{W}_{\mathrm{out}}^{(k)} \in \mathbb{R}^{d_{\text{in}}^{(k)} \times d_{\text{model}}}$ and $\mathbf{b}_{\mathrm{out}}^{(k)} \in \mathbb{R}^{d_{\text{in}}^{(k)}}$ are the domain-specific output projection matrix and bias.
All projection parameters are initialized from scratch and optimized jointly with the MoT backbone.
We convert the predicted 6D rotation back to a $3\times3$ $\mathrm{SO}(3)$ rotation matrix using singular value decomposition (SVD).

\subsection{Token Arrangement and Generation Mode}
Cosmos 3 is a unified model that supports various modalities and tasks.
Different tasks can be formulated as interleaved multimodal sequences, each consisting of a series of segments from different modalities. Given a task, all segments are first encoded into embeddings using the modality-specific encoders described above. Once embedded, tokens from different modalities are packed using a unified format that applies across all tasks, which we describe next.

\subsubsection{Token Arrangement}
The input token sequence consists of two subsequences: an autoregressive (AR) subsequence followed by a diffusion (DM) subsequence. 

The \textbf{AR subsequence} is responsible for reasoning and understanding. It contains language tokens as well as video and image tokens embedded by the ViT encoder.
All AR tokens are routed to a dedicated set of parameters in the transformer decoder layers.

The \textbf{diffusion subsequence} follows the AR subsequence and contains video and image tokens from the VAE encoder, as well as audio and action tokens. During generation, the model iteratively denoises the noisy diffusion tokens to produce the corresponding clean tokens. Diffusion tokens are routed to a separate parameter set from that used by AR tokens, while still interacting with AR tokens through joint attention in each of the transformer decoder layers.

For any given task, we apply the same format to arrange these tokens: (1) autoregressive tokens are placed before diffusion tokens; (2) within the diffusion subsequence, for each modality, clean conditioning tokens are placed before noisy diffusion tokens; and (3) within both the conditioning and diffusion subsequence, tokens are ordered by vision, audio, and action modality.
By using this unified format, Cosmos 3 can support various generation tasks, which we detail below.

\subsubsection{Generation Mode}
\label{sec::generation_mode_def}

\begin{figure}[htb]
    \centering
    \resizebox{0.95\linewidth}{!}{\includegraphics{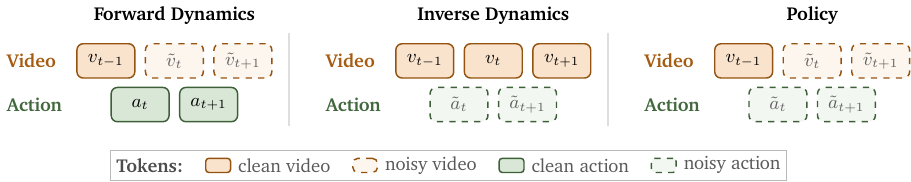}}
    \caption{\textbf{Action sequence configurations.} For a video-action data sample, Cosmos 3 constructs different training modes by varying which tokens are clean and which are noisy. The diagram shows a local temporal window in which action tokens lie between adjacent video tokens: $\acttok_{t}$ connects $\vidtok_{t{-}1}$ to $\vidtok_{t}$, and $\acttok_{t{+}1}$ connects $\vidtok_{t}$ to $\vidtok_{t{+}1}$. Forward dynamics mode denoises vision tokens conditioned on clean action tokens; inverse dynamics mode denoises action tokens conditioned on clean vision tokens; and video-action (policy) mode denoises both vision and action tokens. Language and special tokens are omitted for compactness.}
    \label{fig:action_modes}
\end{figure}

Cosmos 3 supports different modalities: language, vision, audio, and action.
We denote clean vision, audio, and action tokens as $\vidtok$, $\audiotok$, and $\acttok$, respectively, and their noisy counterparts with tildes: $\noisyvidtok$, $\noisyaudiotok$, and $\noisyacttok$.
Given these modalities, the supported generation modes are listed as follows:
\begin{itemize}
    \item \textbf{Language.} For language generation, the input contains only the autoregressive subsequence, and the generation-specific diffusion parameters are not activated. Image and video inputs, if present, are embedded by the ViT encoder and placed in the autoregressive subsequence. In this setting, Cosmos 3 operates like a standard VLM.
    \item \textbf{Text-to-Image.} In this mode, the autoregressive subsequence contains the language tokens, while the diffusion subsequence contains the noisy target image tokens embedded by the VAE encoder. The entire sequence of tokens becomes:
    \begin{equation}
    \mathbf{S}_{\mathrm{T2I}} = [\preseq,\; \noisyvidtok_{1}],
    \end{equation}
    where $\preseq \triangleq [\txttokseq, \eostok, \bovtok]$ is the AR prefix shared by all modes below ($\txttokseq$ are the language tokens; $\eostok$ and $\bovtok$ are the end-of-sentence and begin-of-generation special tokens), and $\noisyvidtok_{1}$ is the noisy image token.
    \item \textbf{Text-to-Video (+Audio).} This mode is similar to Text-to-Image, but the diffusion subsequence contains the noisy target video tokens instead. When audio is (optionally) generated jointly, noisy audio tokens are appended after the noisy vision tokens. In summary, the packed sequence becomes:
    \begin{equation}
    \mathbf{S}_{\mathrm{T2V+Audio}} = [\preseq,\; \noisyvidtok_{1:N},\; \noisyaudiotok],
    \end{equation}
    where $N$ is the number of latent video frames.

    \item \textbf{Image-to-Video/Video-to-Video (+Audio).} This mode introduces an initial conditioning image or a number of initial video frames, and the model generates the complete continuation conditioned on them and the text prompt. In the diffusion subsequence, the clean conditioning image or video tokens are followed by the noisy target video tokens:

    \begin{equation}
    \mathbf{S}_{\mathrm{V2V}} = [\preseq,\; \vidtok_{1:P},\; \noisyvidtok_{P+1:N}],
    \end{equation}
    where $P$ is the number of conditioning latent frames. When $P=1$, the task becomes Image-to-Video, while $P>1$ corresponds to Video-to-Video. When audio is also generated, the audio tokens are appended similarly to the Text-to-Video case.

     \item \textbf{Video transfer.} In this task, the input consists of a control video (\eg, edge, or depth) together with a text description, and the model generates the corresponding RGB video. The token layout is similar to that of Video-to-Video, with the control-video tokens used as conditioning tokens and the RGB video tokens used as noisy target tokens:

     \begin{equation}
    \mathbf{S}_{\mathrm{Transfer}} = [\preseq,\; \vidtok^{\mathrm{ctrl}}_{1:N},\; \noisyvidtok_{1:N}],
    \end{equation}
    where $\vidtok^{\mathrm{ctrl}}_{1:N}$ are the clean VAE-encoded tokens of the control video.

    \item \textbf{Action.}
    Cosmos 3 supports three generation modes for action---forward dynamics, inverse dynamics, and joint video-action prediction (policy). For a trajectory with consecutive video tokens, each action token $\acttok_{t}$ represents the transition from $\vidtok_{t{-}1}$ to $\vidtok_{t}$.
Forward dynamics predicts future visual states conditioned on observed context and clean action tokens, while inverse dynamics infers the action tokens that explain an observed visual transition.
In policy mode, the model jointly predicts action and video tokens, enabling it to generate both the intervention and its expected visual consequence under the same sequence model.
The conditional directions are summarized in~\cref{fig:action_modes}.

\end{itemize}

\subsection{Mixture-of-Transformers (MoT) Architecture}
\label{subsec::mot}

Cosmos 3 adopts a \textbf{Mixture-of-Transformers (MoT)} architecture that processes a unified sequence of tokens from different modalities. At the layer level, each transformer decoder layer contains two sets of parameters: one for reasoning tasks, which processes tokens from the AR subsequence (reasoner), and one for generation tasks, which processes tokens from the diffusion subsequence (generator). Although Cosmos 3 shares similarities with unified generation models such as \cite{deng2025bagel} in its decoder-layer structure, it differs in its training strategy, positional embeddings, and overall capabilities.

\begin{figure}[t]
    \centering
    \resizebox{0.95\linewidth}{!}{\includegraphics{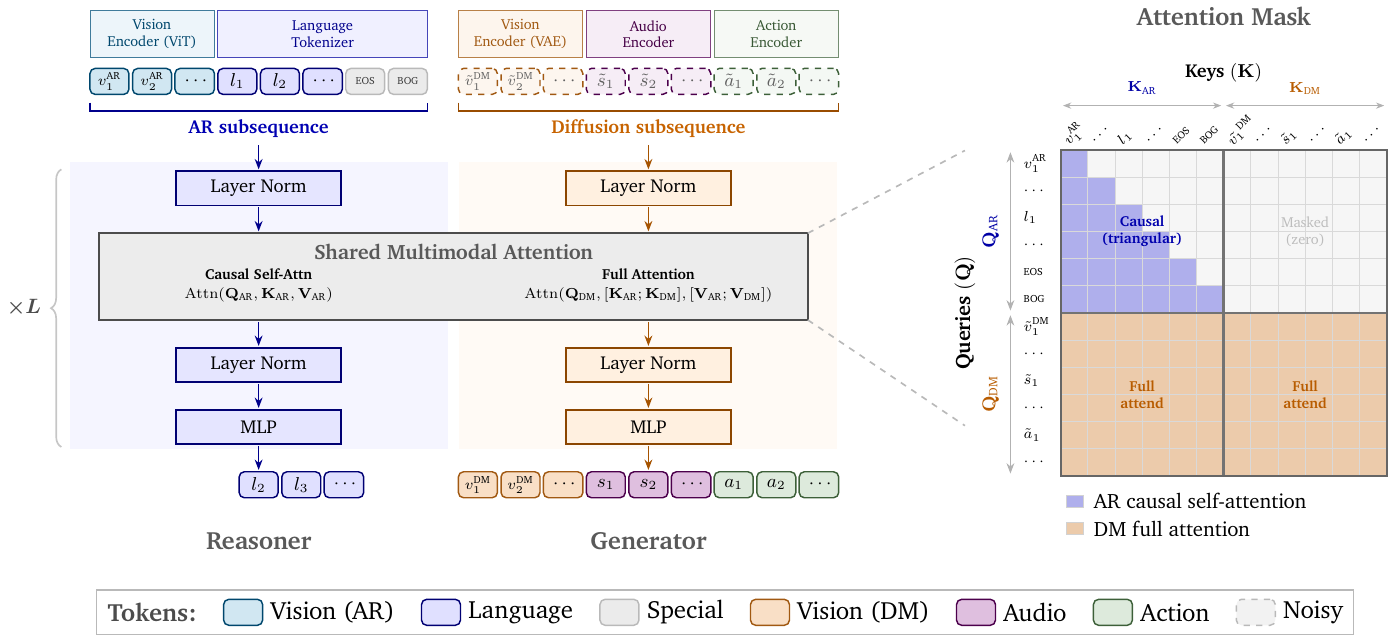}}
    \caption{
        \textbf{Mixture-of-Transformers (MoT) architecture of Cosmos 3.} \textbf{Left:} a single transformer operates on one token sequence comprising the autoregressive (\textcolor{blue!65!black}{\textbf{AR}}) and diffusion (\textcolor{orange!75!black}{\textbf{DM}}) subsequences: AR carries discrete text tokens and, optionally, ViT-encoded vision tokens, ending with \texttt{<EOS>} and a begin-of-generation token \texttt{<BOG>}, while DM carries continuous tokens from their respective encoders, noise-perturbed during training. Here we visualize all input tokens as noisy for simplicity; for generation modes such as image-to-video or video transfer, clean conditioning tokens precede the noisy targets within DM; see \cref{sec::generation_mode_def}. 
        Within each transformer block, AR tokens and DM tokens are processed by independent LayerNorms and MLPs (all co-initialized from a pre-trained VLM) and meet only at a shared self-attention operator. Let $\mathbf{Q}$, $\mathbf{K}$, and $\mathbf{V}$ be query, key, and value vectors in attention, where the subscript indicates which tower it is in. 
        $\mathbf{Q}_{\mathrm{AR}}$ attends causally over
        $\mathbf{K}_{\mathrm{AR}}, \mathbf{V}_{\mathrm{AR}}$ only, while
        $\mathbf{Q}_{\mathrm{DM}}$ attends bidirectionally over the concatenated
        $[\mathbf{K}_{\mathrm{AR}};\mathbf{K}_{\mathrm{DM}}]$ and
        $[\mathbf{V}_{\mathrm{AR}};\mathbf{V}_{\mathrm{DM}}]$. 
        In this way, diffusion is conditioned on the AR context, while AR remains autoregressively self-contained. Outputs are next-token predictions for
        Reasoner and denoised tokens for Generator (trained in practice with a flow-matching objective predicting velocity; we show the clean target here for clarity). \textbf{Right:} the attention mask, causal for AR and full for diffusion.
    }
    \label{fig:mot_architecture}
\end{figure}

\subsubsection{Dual-Tower Layer Structure}
\label{sec:cosmos3_mot}

A standard transformer decoder layer consists of a self-attention operation, a feed-forward network, and some normalization layers. Instead of processing all token types with the same parameters, the MoT design uses two pathways, as shown in~\cref{fig:mot_architecture}. Each pathway is a standard transformer layer with its own parameters, including layer normalization modules, attention projection matrices, and feed-forward networks. The two pathways are both initialized from the weights of a pre-trained Vision-Language Model (VLM), allowing Cosmos 3 to inherit strong language and visual reasoning capabilities while learning to generate high-fidelity videos. During both training and inference, the AR subsequence at the front is routed to the reasoner tower, while the diffusion subsequence at the back is routed to the generator tower.

\subsubsection{Dual-Stream Joint Attention}
Although the two towers use independent parameters, tokens from the diffusion subsequence interact with the AR subsequence through a dual-stream joint attention operation. Here we denote the query, key, and value vectors of the AR and diffusion subsequences as $\mathbf{Q}_\text{AR}$, $\mathbf{K}_\text{AR}$, $\mathbf{V}_\text{AR}$, $\mathbf{Q}_\text{DM}$, $\mathbf{K}_\text{DM}$, and $\mathbf{V}_\text{DM}$, respectively.

\paragraph{Autoregressive subsequence attention.}
Tokens in the AR subsequence attend only to tokens within the AR subsequence using \emph{causal self-attention}; that is, each token can attend only to preceding tokens in the same sequence. This is fully consistent with the autoregressive property inherited from the VLM backbone, allowing the model to preserve the text-generation capability of the pre-trained VLM:
\begin{equation}
    \mathbf{O}_\text{AR} = \operatorname{Attn}_\text{causal}\!\bigl(
        \mathbf{Q}_\text{AR},\; \mathbf{K}_\text{AR},\; \mathbf{V}_\text{AR}
    \bigr).
    \label{eq::ar_attn}
\end{equation}

\paragraph{Diffusion subsequence attention.}
Tokens in the DM subsequence use \emph{full bidirectional attention}, with the union of AR and DM tokens serving as the keys and values. This allows each diffusion token to freely attend to the text prompts from the autoregressive subsequence, as well as to all other conditional and diffusion tokens in the sequence, thereby maintaining temporal and spatial consistency:
\begin{equation}
    \mathbf{O}_\text{DM} = \operatorname{Attn}_\text{full}\!\bigl(
        \mathbf{Q}_\text{DM},\;
        [\mathbf{K}_\text{AR};\,\mathbf{K}_\text{DM}],\;
        [\mathbf{V}_\text{AR};\,\mathbf{V}_\text{DM}]
    \bigr),
    \label{eq::df_attn}
\end{equation}
where $[\cdot \,; \cdot]$ denotes concatenation along the sequence dimension. We note that AR tokens are never updated based on DM tokens, preserving the causal integrity of the conditioning pathway.

\subsection{Multimodal Position Embedding}
Position embeddings inject temporal and spatial structure into the attention mechanism, encouraging tokens to attend more strongly to semantically and geometrically relevant tokens, often nearby in space or time. Since Cosmos 3 jointly models language, vision, audio, and action tokens within a unified attention framework, designing a position-embedding scheme that generalizes consistently across modalities is inherently challenging. Inspired by 3D Multimodal RoPE (MRoPE)~\citep{bai2025qwen3}, we design a 3D MRoPE with absolute temporal indexing to align video, audio, and action tokens along the same physical temporal axis. The original 3D MRoPE divides the hidden dimension of each attention head into temporal, height, and width components, where the temporal component records only the discrete token index. This design is sufficient for image and video understanding tasks, but it is inadequate for our setting, where video, audio, and action tokens may be generated simultaneously at different frame or sampling rates. In this case, tokens from different modalities must be aligned to an absolute physical temporal axis. We first introduce the base formulation, which follows the original 3D MRoPE design, and then describe our extensions and modifications, especially our absolute temporal modulation, which aligns the absolute temporal axis.

\subsubsection{Position Index Allocation}
\paragraph{Autoregressive tokens.} For backward compatibility with language generation and image/video understanding models, position indices for all language tokens and ViT-encoded media tokens in the AR subsequence follow the original 3D MRoPE design. For language tokens, $t = h = w$ is set to the same monotonically increasing value, reducing 3D MRoPE to standard 1D RoPE behavior. For tokens from the ViT encoder, $t$ is shared by all tokens from the same frame, while the $h$ and $w$ indices vary independently according to the spatial location of each token. The allocation of the position index in the autoregressive subsequence is identical to the 3D MRoPE design in Qwen3-VL~\citep{bai2025qwen3}.

\begin{figure}[t]
    \centering
    \resizebox{\linewidth}{!}{\includegraphics{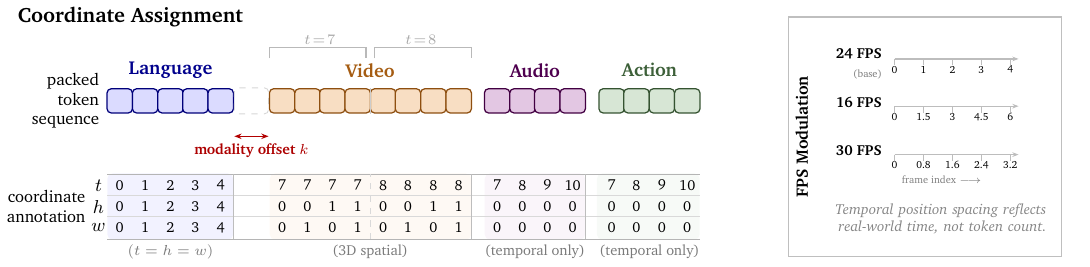}}
    \caption{\textbf{Illustrative coordinate assignment under 3D MRoPE.} \textbf{Left:} A packed token sequence containing language, video (two frames, $2\times2$ spatial grid each), audio, and action tokens. Each token receives a $(t, h, w)$ triplet. Language tokens use $t=h=w$; video tokens vary on all three axes; action and audio tokens use temporal coordinates only ($h=w=0$). A modality offset $k$ separates the text and vision temporal ranges. \textbf{Right:} FPS modulation maps frame indices to scaled temporal positions so that equal real-world durations occupy equal position ranges at 16, 24, and 30 FPS, where 24 FPS is our base frame-per-second.}
    \label{fig:mrope_coords}
\end{figure}

\paragraph{Diffusion tokens.} As illustrated in~\cref{fig:mrope_coords}, video tokens vary across all three axes: $t$ advances with the temporal latent frame index, while $h$ and $w$ tile over the spatial grid $(0 \ldots H{-}1,\; 0 \ldots W{-}1)$ independently per frame. Image tokens are treated as single-frame videos and vary only in $(h, w)$. Both spatial and temporal indices are reset to zero at the start of each vision segment, so the model treats $t$, $h$, and $w$ as absolute within-video coordinates rather than positions in the global sequence. For example, in the video transfer task where the user provides a text prompt together with controlled video frames such as depth maps, both the clean control-video tokens and the noisy generated-video tokens start from the temporal offset of the last token in the autoregressive subsequence. All \textbf{audio tokens} and \textbf{action tokens} only carry temporal coordinates. The spatial indices are set to zero ($h = w = 0$). For audio tokens, the temporal index advances with each audio hop; for action tokens, the temporal index advances with each sampling step.

\paragraph{Autoregressive and diffusion token margin.} In practice, we find that directly letting the diffusion tokens start from the temporal offset of the last autoregressive token leads to over-saturation and checkerboard artifacts in the initial video frames. This effect is especially pronounced in larger variants of Cosmos 3, such as the Super model. We hypothesize that this occurs because the last language token and the vision tokens from the first frame occupy adjacent temporal positions, resulting in nearly identical temporal embeddings. To address this issue, inspired by \citet{cao2025hunyuanimage}, we insert a fixed temporal gap between the autoregressive and diffusion subsequences, uniformly shifting the temporal indices of all the subsequent vision, audio, and action tokens. This creates a buffer in positional space that provides a clearer text-to-vision transition signal without requiring architectural changes or additional learnable embeddings. In all of our models, we set the gap to be $15000$.

\subsubsection{Absolute Temporal Modulation}
A single unit step along the temporal dimension may correspond to different physical time intervals across modalities or data sources. For example, when encoding videos at 60 FPS and 24 FPS, respectively, a temporal-index increment for 24-FPS video tokens corresponds to a physical time interval that is 2.5 times longer than that of 60-FPS video tokens. Similar discrepancies also arise for action and audio tokens, where different data sources may use different sampling rates. FPS modulation is designed to align tokens with different temporal resolutions onto a shared physical temporal axis by modulating the effective size of each temporal increment.

We first define the temporal steps per second (TPS) to characterize the physical temporal resolution. For video tokens, TPS is given by the video frame rate divided by the temporal compression factor, which is 4 in our case due to the video VAE encoder. For audio tokens, TPS is computed as $\mathrm{TPS}_{\mathrm{audio}} = \frac{48000}{1920} \approx 25$ (48\,kHz, 1920 hop size). For action tokens, TPS is exactly the sampling frequency of the action data.

We then associate a unit length along the temporal dimension with a base TPS, denoted as $\mathrm{TPS}_{\mathrm{base}}$. For tokens in a given diffusion subsequence, we compute their corresponding TPS. When the temporal index needs to be increased by one unit step, the temporal increment $\delta t$ with the modulation is computed as
\begin{equation}
    \delta t = \frac{\mathrm{TPS}_{\mathrm{base}}}{\mathrm{TPS}}.
    \label{eq:fps_modulation}
\end{equation}
Since video constitutes the majority of our training data, and 24 FPS is the most common frame rate in our setting, we set $\mathrm{TPS}_{\mathrm{base}} = \frac{24}{4} = 6$ where $4$ is our video tokenizer's temporal compression ratio.

\subsection{Model Variants}

Cosmos 3 is trained at three model scales: \textbf{Edge}, \textbf{Nano}, and \textbf{Super}, spanning a wide range of computational budgets from on-device deployment to large datacenter inference. \textbf{Edge} is a 4B-parameter model built upon a dense 2B-parameter transformer, \textbf{Nano} is a 16B-parameter model built upon a dense 8B-parameter transformer, and \textbf{Super} is a 64B-parameter model built upon a dense 32B-parameter transformer. All variants are initialized from pre-trained vision-language models (VLMs) and adopt the Mixture-of-Transformers (MoT) architecture described above.~\cref{tab::model_variants} summarizes the key architectural hyperparameters for each variant. Cosmos3-Nano and Cosmos3-Super models are released in this paper. Cosmos3-Edge model will be included in a later release.

\textbf{Cosmos3-Edge} uses the design of a 2B dense transformer of $28$ layers, $2048$ hidden size, $16$ attention heads, $8$ key-value heads, a head dimension of $128$, and $9216$ FFN dimension. We train the LLM from scratch using the Megatron codebase. The design of the LLM largely follows the Qwen3-1.7B architecture, with two notable differences: it removes QK normalization and uses ReLU-squared as the FFN activation, which is paired with the Edge FFN dimension reported in~\cref{tab::model_variants}.

\textbf{Cosmos3-Nano} adapts the Qwen3-VL 8B~\citep{qwen3vl2025} architecture, with $36$ layers in the LLM, a hidden size of $4096$, $32$ attention heads, 8 key-value heads, a head dimension of $128$, and a FFN dimension of $12{,}288$.

\textbf{Cosmos3-Super} adapts the Qwen3-VL 32B~\citep{qwen3vl2025} architecture, with $64$ layers in the LLM, a hidden size of $5120$, $64$ attention heads, $8$ key-value heads, a head dimension of $128$, and a FFN dimension of $25{,}600$.

\begin{table}[H]
\centering
\caption{
\textbf{Cosmos 3 MoT model variants.}
All models share the dual-tower MoT architecture. ``LLM Layers'' refers to the number of transformer decoder layers; each layer carries independent parameter sets for the reasoner and generator towers. \textbf{Edge} uses a dense 2B parameter transformer trained from scratch, while \textbf{Nano} and \textbf{Super} are initialized from pre-trained Qwen3-VL weights.}
\label{tab::model_variants}
\small
\begin{tabular}{lccccccc}
\toprule
\textbf{Variant} & \textbf{LLM Layers} & \textbf{Hidden Dim}
    & \textbf{Attn Heads} & \textbf{KV Heads} & \textbf{Head Dim}
    & \textbf{FFN Dim} \\
\midrule
Cosmos3-Edge &  28 & 2,048 & 16 & 8 & 128 & 9,216 \\
Cosmos3-Nano &  36 & 4,096 & 32 & 8 & 128 & 12,288 \\
Cosmos3-Super &  64 & 5,120 & 64 & 8 & 128 & 25,600 \\
\bottomrule
\end{tabular}
\end{table}

\section{Data}
\label{sec::data}

Training Cosmos 3 requires data for two complementary objectives: the Reasoner pathway learns to understand and reason about the world, while the Generator pathway learns to synthesize and simulate it, or act within it. Although both pathways share the same transformer and token representations, they rely on different types of training data. The Reasoner is trained on paired vision-language data, such as image-text and video-text pairs, to support tasks including question answering, spatial grounding, temporal reasoning, and action understanding. In contrast, the Generator is trained on large-scale multimodal corpora of images, videos, audio, and actions using reconstruction-based objectives rather than explicit annotations.

As a result, the two pathways follow different but complementary training curricula. Both adopt a multi-stage training strategy in which the data composition evolves over time. The Reasoner begins with broad vision-language pre-training and is later specialized through supervised fine-tuning on Physical AI tasks spanning robotics, autonomous driving, and spatial intelligence. This staged curriculum first establishes strong general capabilities before progressively introducing more specialized domain knowledge. The Generator begins with large-scale image, video, and audio pre-training, then progressively incorporates additional modalities such as actions, control-conditioned transfer, and targeted synthetic data to improve specific capabilities.

\subsection{Reasoner Data}
\label{subsec::data_reasoner}

\begin{figure}[H]
    \centering
    \begin{minipage}[t]{0.49\linewidth}
        \centering
        \resizebox{\linewidth}{!}{\includegraphics{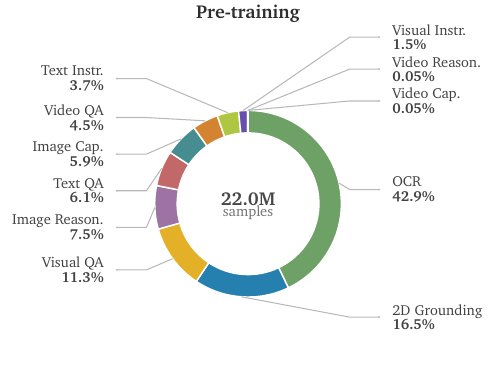}}
    \end{minipage}\hfill
    \begin{minipage}[t]{0.49\linewidth}
        \centering
        \resizebox{\linewidth}{!}{\includegraphics{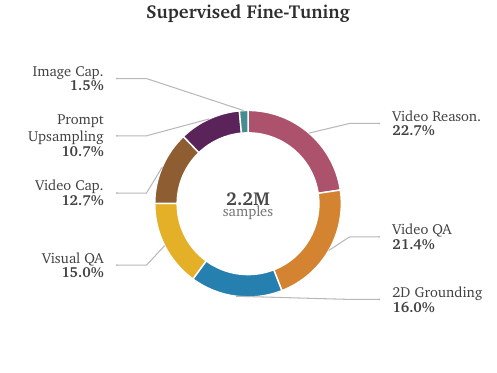}}
    \end{minipage}
    \vspace{-1.5\baselineskip}
    \caption{\textbf{Cosmos 3 Reasoner data composition by capability category.} We summarize the curated data mixture used to train Cosmos 3 Reasoner across the pre-training and supervised fine-tuning stages. The mixture contains 22.0M pre-training samples and 2.2M supervised fine-tuning samples spanning image--text, video--text, and text-only categories, with each ring showing the relative contribution of major capability streams such as OCR, visual question answering, reasoning, captioning, grounding, and instruction tuning.}
    \label{fig:reasoner_category_mix}
\end{figure}

Our reasoner data curriculum contains approximately $24.2$M samples: $22.0$M for pre-training and $2.2$M for supervised fine-tuning from domain-specific Physical AI datasets and synthetically generated data. \Cref{tab:data_reas_summary} summarizes the data modalities used in both stages. The pre-training stage is dominated by image--text and text-only data, providing broad general visual understanding. In contrast, the supervised fine-tuning stage shifts toward Physical AI specialization, with video--text samples comprising 50\% of the mixture to strengthen spatiotemporal understanding and capabilities in robotics, smart infrastructure, and autonomous vehicle domains. \Cref{fig:reasoner_category_mix} summarizes this mixture by capability category for the two stages.  

\begin{table}[h]
\centering
\renewcommand{\arraystretch}{1.5}
\caption{\textbf{Cosmos 3 Reasoner data curriculum by modality and training stage.} Table values represent the number of media samples (image or video) for Image-text and Video-text rows and the number of conversations for the text-only row.}
\label{tab:data_reas_summary}
\setlength{\tabcolsep}{7pt}
\small
\begin{tabular}{lrr}
\toprule
\textbf{Modality}
    & \textbf{Pre-training}
    & \textbf{Supervised Fine Tuning } \\
\midrule
Image-text      & 18,814,952 & 1,051,513 \\
Video-text      & 1,016,299 & 1,079,200 \\
Text only       & 2,170,762 & 40,960 \\
Total           & 22,002,013 & 2,171,673 \\
\bottomrule
\end{tabular}
\end{table}

\subsubsection{Pre-Training}
\label{subsubsec::data_reas_pretrain}

We build our pre-training data mixture with 19.7M samples sub-selected from the Nemotron Nano 2 data collection~\citep{nvidia2025nvidianemotronnanov2} and 2.3M additional samples curated to enhance math, video, spatial grounding, and instruction-following capabilities. See~\cref{tab:data_reas_summary} for details.
The datasets we source are fed through a two-stage data curation pipeline consisting of semantic deduplication followed by AI-judge quality filtering before inclusion in the final training mixture.

\paragraph{Semantic deduplication.}
The first stage removes multimodal near-duplicates at the conversation level, where a conversation denotes the complete training example: an image or video paired with its instruction-response text, or a text-only instruction-response sample when no media is present. For each conversation, we compute a joint embedding that combines the media representation, when available, with the associated instruction-response text representation. Image-text and text-only conversations are embedded using Qwen3-VL-Embedding-8B~\citep{qwen3vlembedding}, while video-text conversations are embedded using the Perception Encoder PE-Core-G14-448~\citep{bolya2025PerceptionEncoder}. The resulting concatenated embedding representation jointly captures visual and linguistic semantics, enabling the pipeline to distinguish visually similar samples with different task intents from truly redundant supervision. 

To scale duplicate detection to production-scale datasets, we use clustering. Image-text, video-text and text-only samples are first partitioned using K-means clustering. Near-duplicate groups are then identified within each cluster using cosine similarity in the conversation-embedding space. Samples with similarity above a high threshold of $0.95$ are removed. This hierarchical design makes large-scale duplicate detection tractable while preserving sensitivity to redundancy in both visual content and task semantics.

\paragraph{AI-judge quality filtering.}
The second stage applies an AI judge to assess annotation quality on the deduplicated corpus. We use Gemma-4 as the vision-language judge~\citep{gemma4modelcard}, specifically the Gemma-4-31B-it model~\citep{gemma4hf31bit}. The judge is prompted as a training-data auditor and assigns rubric-based integer scores from $1$ to $5$ across three primary quality dimensions:

\begin{itemize}
    \item \textit{Faithfulness:} whether all response claims are grounded in the provided image, video, or textual context.
    \item \textit{Completeness:} whether the response fully addresses the instruction without important omissions.
    \item \textit{Correctness:} whether the response is factually, logically, and task-level accurate.
\end{itemize}

Faithfulness is particularly important for Cosmos 3 because unsupported visual claims can teach the model to hallucinate physical states, object attributes, or temporal events. Meanwhile, Completeness filters under-specified or partial responses, while Correctness removes supervision whose final answer or reasoning is inconsistent with the input. In addition to scalar scores, the judge also produces short evidence-based rationales, enabling targeted spot checks and auditing of the filtering behavior.

\paragraph{Threshold-based dataset construction.}
We construct multiple judge-filtered dataset variants from the same deduplicated base corpus using a minimum-threshold rule over the three quality dimensions. A sample is retained only if its Completeness, Correctness, and Faithfulness scores all meet or exceed a specified threshold. In other words, every retained example must simultaneously satisfy the minimum quality requirement across all three dimensions.

This filtering strategy is intentionally stricter than averaging the scores. Samples with a severe failure mode in any single dimension are removed even if they score highly on the remaining criteria. For example, a response that is highly detailed and logically correct but contains unsupported visual claims will still be filtered out due to low Faithfulness. In practice, we use conservative thresholding to eliminate clearly low-quality supervision while minimizing excessive distribution shift across capability domains.

Multimodal deduplication removes $4.23\%$ of the data as near-duplicate supervision. The AI-judge score distribution reveals that quality failures are not uniform across dimensions. We also analyze retention by capability category to ensure that quality filtering improves the corpus without unintentionally collapsing the skill distribution. At stricter thresholds, pruning becomes strongly category-selective: for example, referring-expression grounding is removed most aggressively, while image captioning and visual question answering also decline substantially, as the threshold increases from $2$ to $5$. Here, the threshold denotes the minimum acceptable AI-judge score on each quality dimension: a sample is retained only if its Completeness, Correctness, and Faithfulness scores are all above the threshold. OCR data is comparatively robust with a threshold of $2$ but drops materially at $5$. These trends motivate using the lowest non-trivial judge threshold, \ie, $2$ for the primary mixture: it filters clear annotation failures while preserving the original coverage of reasoning, grounding, OCR, captioning, and VQA capabilities, thereby creating an optimal quality--quantity trade-off on the pre-training dataset. However, in the SFT stage, we use a threshold of $5$ to retain only the highest-confidence supervision examples, where annotation precision and response reliability are more critical than broad coverage. The AI-judge filter retains $78\%$ and $46\%$ of data at a threshold of $2$ and $5$, respectively.

The final pre-training mixture contains approximately 22M samples spanning OCR, grounding, question answering, reasoning, captioning, and instruction-following data, as shown in~\cref{tab:data_reas_summary}. Regarding the composition, OCR is the largest component, contributing $9.44$M samples ($42.9\%$), followed by 2D grounding with $3.62$M samples ($16.5\%$), visual QA with $2.48$M samples ($11.3\%$), and image reasoning with $1.66$M samples ($7.5\%$). The remaining mixture provides broad multimodal coverage through text QA ($1.35$M, $6.1\%$), image captioning ($1.30$M, $5.9\%$), video QA ($0.99$M, $4.5\%$), text instruction data ($0.82$M, $3.7\%$), visual instruction data ($0.34$M, $1.5\%$), and small amounts of video captioning and video reasoning data ($0.01$M each). This composition emphasizes strong image-text alignment, reading, and spatial grounding while retaining a lightweight video component that prepares the model for later supervised fine-tuning on temporal and video-reasoning tasks.

\subsubsection{Supervised Fine-Tuning}
\label{subsubsec::data_reas_posttrain}

In the supervised fine-tuning stage, we enhance the general spatial and temporal understanding capabilities and focus on curating data for the following three domains: autonomous vehicle, robotics, and smart infrastructure. In total, we train with 2.2M samples in this stage.

\paragraph{General spatial understanding.}
We enhance general spatial understanding through 2D and 3D grounding, augmented with both real and simulation data.

\textit{2D and 3D grounding.} 2D grounding supports Physical AI tasks that require localization (objects, regions, parts, and landmarks), pointing, counting, and multi-image correspondences. We curate samples spanning detection, referring expressions, OCR/layout, visual-prompted description and VQA, pointing, trajectories, counting, and dense grounding. Boxes and points are normalized and converted into a unified JSON format. We curate 2D grounding data from synthetic and existing data, with a majority coming from LocateAnything~\citep{wang2025locateanything} data, and we sample a high-quality subset from them. For 3D grounding data, we convert 3D scanned scenes into instruction-following training samples with camera-relative 3D boxes. Each object is annotated with label, center, dimensions, and orientation after canonicalization and intrinsic normalization~\citep{brazil2023omni3d}. Instead of chaining 2D grounding and 3D inference~\citep{gr3d,man2025locateanything3d}, we directly supervise the final structured boxes.

\textit{Real-world spatial understanding and grounding.} We combine several complementary QA types. Image-referring QA task requires pointing to a target object or to empty free space as normalized image coordinates given a natural-language expression~\citep{zhou2025roborefer}. Robotics spatial QA requires predicting placement points in free space and answers binary relative-position and reachability questions across ego-, world-, and object-centric reference frames~\citep{song2025robospatial}. We further include posed-scene multi-image QA, video-frame spatial QA, and multiple-choice questions that predict the relation between two objects from a fixed set (\eg, left of, behind, on/above), including perspective-substituted viewpoints~\citep{liu-etal-2025-multimodal-large}. Together, these cover object references, free space, cross-view correspondence, camera motion, size, distance, direction, routes, counting, and room-scale reasoning.

\textit{Simulator-grounded embodied spatial reasoning.} We bridge visual spatial QA and embodied action by deriving labels from executable simulator state. 
Answers are computed from cameras, object poses/boxes, depth, masks, visibility, and feasible regions, then checked by programmatic and VLM critics. The curriculum spans metric geometry, spatial frame, physical semantics, actionable grounding, viewpoint dynamics, and embodied composition, with multiple-choice question, numerical, point, box, binary, and text outputs.

\paragraph{General temporal understanding.}
We augment temporal capabilities along three axes: temporal event understanding, physical plausibility judgment, and structured spatiotemporal scene upsampling.

\textit{Temporal event understanding.}
We strengthen temporal and motion understanding with three complementary supervised data sources. First, human annotators create dense temporal captions: egocentric videos of everyday indoor tasks are labeled with atomic human-action descriptions and start/end timestamps, with actions averaging 1.8 seconds and 14.2 words. In addition, a broader video corpus of 55K videos (2.6K hours) provides 743K event triplets $(t_{\text{start}}, t_{\text{end}}, \text{caption})$ for event enumeration and query-conditioned localization. Second, to increase question diversity, we curate training data with the FoundationMotion pipeline~\citep{gan2025foundationmotion}, producing ten four-way multiple-choice questions per clip that probe action identity, temporal evolution, and fine-grained motion differences. Finally, we annotate camera motion patterns such as panning and zooming so the model can learn ego-camera movement.

\textit{Physical plausibility judgment.}
We strengthen Cosmos 3 Reasoner's ability to judge physical plausibility in generated videos with two complementary supervised sources. 
First, we incorporate annotations from the Cosmos human evaluation described in Appendix~\ref{sec::cosmos_hue_bench}, which contains 13.5K human-graded $(\text{video}, \text{question}, \text{answer})$ tuples from 1K generated videos, covering visual integrity, temporal stability, geometry, anatomy, motion plausibility, and physical commonsense with categorical Yes/No/Unclear answers. Second, we adopt VideoPhy-2~\citep{bansal2025videophy}, providing 3.4K action-centric videos spanning $200$ actions, each rated from $1$--$5$ for adherence to physical laws; its annotations cover conservation laws, gravity, collision dynamics, temporal causality, and spatial constraints, and are converted into supervised QA pairs where the model predicts a physics-adherence score. 

\textit{Structured spatiotemporal scene upsampling.} 
We curate paired data of under-specified user inputs and densely-structured captions. The input  may be a text prompt, an image, or both, and the target is the paired structured-caption annotation described in \Cref{subsubsection::imagevideo}. The Upsampler capability recovers spatial, temporal, and visual details that are implicit in the input, such as subject attributes, scene layout, camera framing, object interactions, and plausible future motion. To make the model robust to different request formats, we synthesize various instruction variants, with one variant being the canonical form, sampled more often, and by crossing different input-prompt lengths (\eg, how detailed a description should be; brief, detailed, \etc)  with several prompting styles (\eg, how the description should detail the scene; request, declarative, \etc). The resulting supervision encourages the model to follow compact, direct, and stylized user requests and generate detailed scene descriptions while preserving the same structured output contract. \Cref{sec::prompt_upsampling} describes using the capability for Cosmos 3 image/video generation.

\paragraph{Autonomous vehicle (AV).} We incorporate AV datasets spanning human-labeled and auto-labeled chain-of-thought reasoning, temporal event understanding, and 3D vehicle grounding.

\textit{Action CoT.} Human-labeled CoT data from internal driving logs contains more than 10K videos with explicit driving decisions, covering weather, lighting, road conditions, traffic rules, ego-vehicle behaviors, critical objects, and causal links between scene elements and ego behavior. To scale this signal, we auto-label about 1.1M additional decision-rich videos from internal logs. For each video, we identify the meta-action transition keyframe as the decision moment and use state-of-the-art VLMs, raw video, ego trajectory, dynamic states, and meta actions to produce structured decisions, critical components, and concise reasoning traces.

\textit{Temporal event localization.} We derive temporal event localization data from Nexar dashcam footage~\citep{dmoura2025nexar}, with more than 24K videos covering collisions, near-collisions, hard braking, harsh acceleration, and sharp cornering. Clips are sampled at 6 FPS and capped at 300 seconds. We augment human-/auto-labeled data with dense captions of the scene, agents, interactions, ego behavior, and spatiotemporal context.

\textit{3D vehicle grounding.} We train metric 3D vehicle grounding from the MADS dataset~\citep{ren2025cosmos}, which provides synchronized multi-camera sequences, world-scenario-map 3D annotations, camera intrinsics/extrinsics, and ego poses. We sample frames at ${\sim}1$\,FPS and create open-vocabulary detection and referring-grounding QA, asking the model to enumerate vehicles or localize instances by relative position, lane, motion state, or distance. Answers are camera-frame 3D boxes parameterized by position, size, roll, pitch, yaw, and category label, filtered to visible, lightly occluded objects within $100$\,m across diverse regions, lighting, and weather.

\paragraph{Robotics and embodied AI.} We curate data for action CoT in robot manipulation, embodied reasoning, and healthcare robotic surgery understanding.

\textit{Robot action Chain-of-Thought.} 
Action-CoT teaches Cosmos 3 Reasoner to turn a high-level embodied instruction and current frame into a 2D image-plane motion plan for robot end-effector control. Instead of free-form rationales, it structures reasoning through task-relevant locations, grounding points, move reasoning, and 2D waypoints, moving from perception to action in a compact, inspectable trace. The trace identifies manipulated objects, context objects, affordance points, and collision-free regions, localizes them as coordinates, and resolves the plan into pixel-space end-effector waypoints. Data is built with a modular pipeline because grounding, localization, and manipulation planning require different skills: Qwen3-VL-72B-Instruct~\citep{bai2025qwen3} generates grounding rationales and move reasoning, Molmo-7B~\citep{deitke2024molmo} localizes referring expressions, and motion-plan targets come from MolmoAct~\citep{lee2025molmoact} or tracked DROID~\citep{khazatsky2024droid} episodes. 

\textit{Embodied reasoning.} 
We strengthen Cosmos for embodied reasoning with temporal localization, task planning, and robotics embodied QA data. For robot manipulation, we target the MimicGen~\citep{mandlekar2023mimicgen} bottleneck of segmenting demonstrations into object-centric subtasks with precise timestamp boundaries, using 60 held-out videos for zero-shot evaluation and a supervised fine-tuning set of 3.6K Omniverse-rerendered videos across six tasks, with timestamps derived from object trajectories and joint kinematics and manually verified. For long-horizon planning, we curate 83K BEHAVIOR-1K~\citep{li2023behavior1k} samples that map a scene frame and candidate action list to the ground-truth action. We also add ERQA robotics QA from EO-Data-1.5M~\citep{eo1}, covering task planning, affordances, failure detection, physical commonsense, localization, referring, relations, and trajectory prediction.

\textit{Healthcare robotic surgery understanding.}
We curate a robotic-assisted surgery VQA dataset with 398K multi-turn conversations over 2.2M images. The data is collected from exocentric operating-room cameras, an egocentric robotic detail camera, and console displays inspired by ORQA~\citep{ozsoy2025specializedfoundationmodelsintelligent}. Most samples combine multiple viewpoints and include tracker metadata (tool states, 3D translations, Euler rotations) and robot metadata (surgical phase and step) as contextual hints. The tasks cover tool recognition, localization, yes/no classification, scene graphs, monitor-text transcription, personnel counting, distance/time estimation, action labeling, and surgical-step recognition.

\paragraph{Smart infrastructure.}
We curate three complementary smart-infrastructure data sources, covering warehouse spatial intelligence, dense pedestrian localization, and traffic and anomaly reasoning. 

\textit{Warehouse spatial intelligence.} We use PhysicalAI-Spatial-Intelligence-Warehouse~\citep{tang20259thaicitychallenge}, a synthetic Omniverse corpus spanning 44 warehouse scene collections and 40 camera views. From 93K RGB--D images and 873K QA pairs, we subsample 80K balanced examples covering object counting, metric distance, grounding, and binary spatial relations over pallets, boxes, forklifts, shelves, and operator zones.

\textit{Dense pedestrian localization.} We curate annotations with 208K images from 44 scenes and 5.6M manually labeled person boxes. All person bounding boxes are manually labeled by human annotators, and personally identifiable information is redacted via blurring prior to annotation and release to ensure subject anonymity.

\textit{Traffic and anomaly reasoning.} We combine synthetic ITS collision supervision, real traffic-event reasoning, and surveillance anomaly verification. CARLA~\citep{Dosovitskiy17} clips train binary collision prediction between marked vehicle pairs in unprotected-left-turn and T-bone scenarios, with Cosmos-Transfer2.5~\citep{cosmos_predict2p5} augmentation yielding 3.4K labeled pair queries. TAR~\citep{nvidia2026tar} contributes 3{,}6K traffic-camera videos (26 hours) and 44K annotations spanning QA, temporal reasoning, causal linkage, scene description, and summarization, with hierarchical CoT auto-labels cross-checked against human annotations. To broaden anomaly coverage beyond traffic, we also curate 1K internal surveillance clips for binary tailgating verification.
\subsection{Generator Data}
\label{subsec::data_generator}

\begin{figure}[!h]
  \centering
  \resizebox{0.95\linewidth}{!}{\includegraphics{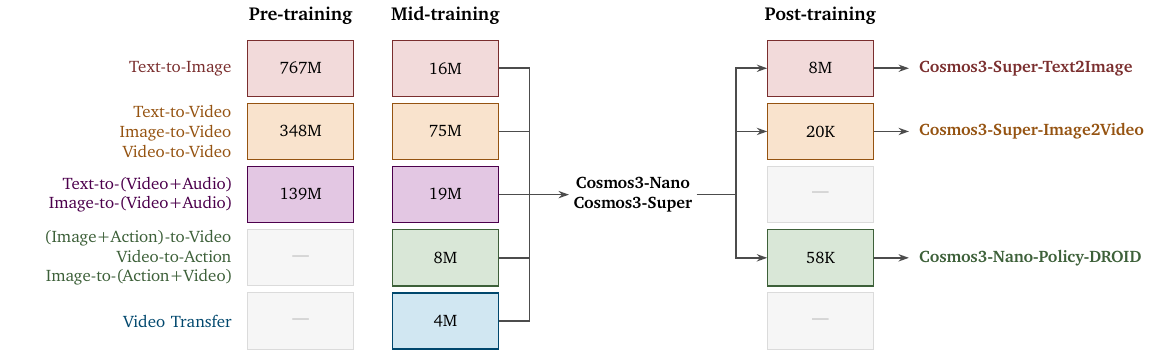}}
  \vspace{-0.5em}
  \caption{\textbf{Generator data curriculum.}
  Each row is a training mode; each column is a training stage.
  Colored cells show the number of training samples used at that stage; gray cells (---) indicate the mode is not active.
  The video row covers text-to-video, image-to-video, and video-to-video continuation; V2V uses clean conditioning-video prefixes and noisy future-video targets.
  Action and video transfer data are first introduced during mid-training.
  Mid-training yields the base \textbf{Cosmos3-Nano} and \textbf{Cosmos3-Super} models (shown between the Mid-training and Post-training columns), which then enter post-training. Post-training is conducted independently for each modality, yielding the specialized models listed on the right: \textbf{Cosmos3-Super-Text2Image}, \textbf{Cosmos3-Super-Image2Video}, and \textbf{Cosmos3-Nano-Policy-DROID}. We note that these specialized models share the exact same architecture with their corresponding mid-train models. }
  \label{fig:data_curriculum}
\end{figure}

Our Generator training follows a progressive multi-stage curriculum that introduces new modalities incrementally over the course of training, starting with images, videos, and audio during pre-training, and later incorporating actions and interleaved multimodal content during mid-training. Cosmos 3 is positioned as a good starting point for various Physical AI applications. To show its capabilities, we take the mid-trained checkpoints Cosmos3-Nano and Cosmos3-Super and post-train them to produce domain experts using specialized post-training datasets, including Cosmos3-Super-Text2Image, Cosmos3-Super-Image2Video, and Cosmos3-Nano-Policy-DROID. These models share the same architecture as their corresponding mid-trained models. \cref{fig:data_curriculum} summarizes the Generator training curriculum across modalities and stages.

\subsubsection{Image and Video}
\label{subsubsection::imagevideo}

Image and video data curation follows a set of carefully designed processing, annotation, and filtering pipelines: (1) collecting raw data and performing pre-processing; (2) computing embeddings and conducting deduplication; (3) categorizing samples and applying basic filtering; (4) annotating data; and (5) grouping samples into training-ready shards based on their resolution and duration. To improve generation quality for Physical AI scenarios and other challenging cases, we introduce synthetic data into the visual-generation data mixture. We organize the resulting data into pre-training data and higher-quality mid-training and post-training data.

\paragraph{Pre-training.}
In the pre-training stage, we use $767$M images and $347.7$M video clips processed from $7.8$B raw images and $3$B raw source videos. In the resulting corpus, $720$p and $480$p are the dominant resolutions for both images and videos. Specifically, $720$p accounts for $26.8\%$ of images and $36.4\%$ of videos, while $480$p accounts for $26.0\%$ of images and $30.8\%$ of videos. In addition, $25.2\%$ of images and $12.2\%$ of videos are at 1080p resolution or higher. The most common aspect ratio is 16:9, accounting for $52.0\%$ of images and $97.3\%$ of videos. For images, the second most common aspect ratio is 1:1, accounting for $25.2\%$ of the retained image corpus. The raw data is processed and filtered using the pipeline described below:

\begin{itemize}

\item \textit{Raw data collection and processing.} We collect billions of raw images and videos from diverse data sources, recording the raw media content together with associated metadata such as raw captions and descriptions. For videos, we additionally apply scene-change detection using TransNetV2~\citep{soucek2020transnetv2} to segment long videos into temporally consistent clips. We then use \texttt{ffmpeg cropdetect} to detect and remove black borders, and re-encode all video clips into a canonical format to standardize storage and ensure playback integrity.

\item \textit{Embedding and deduplication.} Raw data contains a large amount of repeated image and video content, and its concept distribution is often highly imbalanced. To remove duplicate content and establish a foundation for concept balancing and evaluation, we embed the media content into vectors. For images, we use Qwen3-VL-Embedding-8B~\citep{qwen3vlembedding}; for videos, we use nvidia/Cosmos-Embed1-448p~\citep{nvidia2025cosmosembed1}. We sample $147$M images and $400$M video clips from the full data corpus and independently run cuML KMeans with $20,000$ clusters for each data type \citep{cuml, kmeans}. We then assign each image or video clip to its nearest cluster and perform near-duplicate removal within each cluster based on cosine similarity scores.

\item \textit{Categorization and basic filtering.} We use a small suite of in-house VLM models for semantic tagging and quality filtering. Both image and video data are classified into $47$ hierarchical categories, including General and Physical AI domains. For image filtering, a dedicated model annotates attributes such as collage and produces aesthetic and photorealism scores. Images are retained only if their aesthetic score exceeds a predefined threshold. Images tagged as collage, watermark, white background, or NSFW are discarded. For synthetic images not intended for text rendering, we additionally filter them based on their photorealism score, retaining only those above a specified threshold. For video filtering, we use three continuous quality scores---DOVER aesthetic quality~\citep{dover}, DOVER technical quality, and VTSS training suitability~\citep{koala}, each on a 0--9 scale---together with approximately $100$ binary artifact tags. Major artifacts (split-screen layouts, rotated videos, static videos) lead to rejection, while minor artifacts (text overlays, motion blur, compression noise) are flagged but retained in the pre-training data.

\end{itemize}

\paragraph{Mid-training.} The mid-training stage aims to improve generation quality using carefully selected high-quality data and to equip the model with additional capabilities, including both domain-specific capabilities, such as Physical AI, and new tasks, such as video transfer. The data is drawn from three sources: (1) high-quality images and videos; (2) synthetic images and videos; and (3) video-transfer data.

\begin{itemize}
\item \textit{High-quality images and videos.} We select data using stricter filtering rules and sample data more heavily from hard-case concepts to mitigate the long-tailed distribution. For real images, samples must satisfy per-aspect-ratio resolution thresholds and a strict DOVER aesthetic-score cutoff. We also include synthetic and text-rendering subsets. Synthetic images, curated through careful rejection sampling, broaden coverage of uncommon visual concepts and object compositions, while text-rendering images address the underrepresentation of legible in-image text. The resulting mid-training image mixture has effective proportions of $60\%$ real images, $36\%$ synthetic images, and $4\%$ text-rendering images. For video, we similarly apply stricter resolution and aesthetic filtering to select clips from the pre-training pool, which constitute $46.0\%$ of the mid-training video mixture. We then incorporate additional high-quality, domain-specific clips from robotics, autonomous driving, human activity, and egocentric human-object interaction sequences, targeting embodied-AI and manipulation scenarios; these clips constitute another $43.9\%$ of the mixture. To further improve robustness on difficult and corner-case concepts, such as human motion, high-speed complex motion, and fine-grained manipulation, we collect additional capability-oriented data focused on these hard cases, which accounts for the remaining $10.1\%$ of mid-training videos.

\item \textit{Synthetic data.} Although our pre-training corpus is highly diverse, its concept distribution remains long-tailed. As a result, the model receives comparatively limited exposure to rare but important Physical AI domains and scenarios, such as robotics, autonomous driving, and warehouse environments. In these settings, the model often struggles to understand scene dynamics, physical interactions, and long-horizon behavior. To address these limitations, we construct a large-scale synthetic data corpus with the following subsets: 1) Physical-Interaction-Scenes \textit{(SDG-PhyxSim)} focusing on rigid-body collisions, articulated object dynamics, deformable materials, fluid dynamics, and optical effects; 2) Embodied-Robot-Scenes \textit{(SDG-RobotSim)} for manipulation and locomotion sequences across 6--8 robot embodiments and diverse task categories; 3) Autonomous-Driving-Scenarios \textit{(SDG-DriveSim)} covering both routine and corner-case traffic scenarios; 4) Digital-Human-Scenes \textit{(SDG-SynHuman)} designed to improve modeling of human dynamics, camera-motion priors, and multi-character interactions; and 5) Warehouse-Operation-Scenes \textit{(SDG-Warehouse)} for warehouse safety containing human-forklift interaction scenarios. Refer to Appendix~\ref{appendix:sdg_datasets} for more details on the construction of the synthetic data and analysis. We release all SDG datasets to support the community.

\item \textit{Video transfer data.} Transfer data equips the Generator with control-conditioned generation capabilities. Given a spatial control signal, such as an edge map, blurred frame, depth map, segmentation map, or world-scenario map, together with a text description, the model is trained to generate an RGB video. We select $3$M videos from the pre-training video pools, focusing on high-quality videos and physical-AI domains such as robotics and autonomous driving. For edge and blur control, we compute the control signals on the fly during training using Canny edge detection, Gaussian blur, and bilateral filtering with randomly sampled parameters. For depth and segmentation control, we pre-compute the control signals using Video Depth Anything~\citep{chen2025videodepth} and SAMv2~\citep{ravi2024sam}, respectively. For world-scenario-map control, we use the MADS dataset collected by Cosmos-Drive-Dreams~\citep{ren2025cosmos}. MADS contains $1.1$M samples, each with seven synchronized camera views: front-wide, front-tele, cross-left, cross-right, rear-left, rear-right, and rear-tele. The videos are recorded at $30$\,FPS and accompanied by per-camera world-scenario-map control inputs that encode lane lines, road boundaries, traffic lights, and dynamic 3D bounding boxes for vehicles and pedestrians. The dataset covers 14 geographic regions, including the United States, Germany, Japan, and the United Kingdom, across 25 country-duration partitions.
\end{itemize}

\paragraph{Post-training.} We construct post-training datasets for training domain-specialized Cosmos 3 models, such as Cosmos3-Super-Text2Image and Cosmos3-Super-Image2Video.

The post-training image corpus is a compact, carefully curated set drawn from three sources: synthetic images, text-rendering images, and high-quality real images. General web-scale pre-training data is excluded entirely in favor of high-fidelity content that directly targets generation quality and capability gaps.

The post-training video corpus is assembled from two components. The first is a subsampled subset of the pre-training video corpus.
This subset serve as a regularizer, preventing the model from overfitting to the narrower post-training distribution while preserving the broader visual knowledge acquired during pre-training. The second and primary component is a compact set of supervised fine-tuning (SFT) videos, curated specifically to close generation-quality gaps identified through systematic evaluation. The SFT video set consists of three sub-sources. The first is synthetic videos, which cover diverse visual concepts, motion types, and scene compositions that are difficult to source from real-world footage. The second is human-curated real SFT videos, selected and annotated by human curators to provide a direct quality signal for generation fidelity across key visual domains. The third is retrieved real videos from the pre-training corpus, selected via embedding similarity to ensure coverage of common failure cases.

\paragraph{Structured caption annotation.}
Caption quality is a critical factor in generation quality. A high-quality caption should faithfully describe the entities, attributes, relationships, and overall scene content in an image or video. For videos, captions should further capture temporal dynamics, including object motion, human actions, physical changes, interactions, and camera movement. To improve caption quality, we conducted multiple design iterations and adopted a structured JSON annotation format instead of dense free-form natural-language captions for all our data across all training stages. Our experiments show that free-form captions are often precise but incomplete: they tend to describe visible content accurately, yet omit important details in complex scenes. In contrast, a rich predefined structure encourages systematic coverage of objects, attributes, relationships, and scene-level information, improving recall while maintaining high precision.

Our structured format captures a broad set of visual attributes, including subjects, background, lighting, aesthetics, and cinematography. For videos, we additionally introduce fields for temporal dynamics, ranging from physical transformations and object interactions to complex human motion. We fine-tuned two \mbox{Qwen3-VL-8B} models on structured annotation data to serve as our in-house captioners for images and videos, respectively. Refer to Appendix~\ref{appendix:captioning_details} for more details on our captioning models and full structured caption schema.

To quantitatively and rigorously evaluate annotation quality, we designed a specialized caption-quality benchmark for both images and videos. This benchmark focuses on hard-to-caption examples, emphasizing domains such as Physical AI, where accurate descriptions of objects, spatial relationships, actions, and temporal dynamics are critical. For each model-generated caption, we compute precision and recall at the assertion level using two distinct approaches. Precision is evaluated directly against the source media to penalize hallucinations: a VLM decomposes the generated caption into atomic claims and verifies whether each claim is visually supported by the image or video itself. Recall, on the other hand, measures comprehensiveness and relies on human-curated ground truth. To enable a reliable and traceable recall evaluation, we decompose the visual content of videos or images into a list of atomic assertions covering entities, attributes, relationships, events, and other relevant details. An LLM then cross-references the generated caption against this ground-truth assertion list to determine which key details were successfully captured. This protocol allows us to evaluate not only the factual accuracy of the captions, but also whether they contain the critical visual information required to train high-quality generative models. On this benchmark, our structured annotation approach significantly improved recall while maintaining high precision.

\subsubsection{Audio}
\label{subsubsec::data_gen_audio}

Audio-video paired data teaches the Generator not only what sound should be present, but also when that sound should occur relative to visible events. Raw web video audio is challenging for this purpose: narration and voiceover often describe the video without being caused by it, while manually added background music (BGM) can mask the physical sounds produced by on-screen events. We therefore use audio differently across training stages. Pre-training preserves broad acoustic coverage, while mid-training constructs higher-precision audio-video pairs through an explicit selection policy for speech and non-speech audio.

\paragraph{Pre-training.}
The audio pre-training corpus is derived entirely from the pre-training video pool. In total, 138.9M pre-training clips contain usable audio tracks, covering a broad mixture of diegetic and non-diegetic speech, voiceover, BGM, ambient sound, music, and physical events. Of these clips, 62.5M are shorter than 30 seconds; for this subset, we use Qwen3-Omni-Captioner~\citep{xu2025qwen3omni} to generate synthetic audio descriptions. This stage favors scale and diversity, exposing the model to the long-tailed distribution of real video audio before applying stricter curation in mid-training.

\paragraph{Mid-training.}
The mid-training audio pool is filtered from the pre-training audio-video corpus to improve causal audio-visual alignment. The final pool contains 18.8M clips: 12.8M non-speech clips for environmental and physical sound generation, and 6M speech-synchronized clips for visually grounded speech generation. The curation pipeline is organized around a simple principle: keep speech only when it is synchronized with a visible face, remove off-screen speech from non-speech examples, and remove non-instrumental BGM when it would dominate the target audio.

\begin{itemize}
    \item \textit{Source separation.} For every candidate clip, SAM-Audio~\citep{shi2025samaudio} separates the original audio into a speech stem and a remaining stem. The speech stem is used to identify visually grounded speech, while the remaining stem provides a vocal-suppressed candidate for non-speech audio generation.
    \item \textit{Lip-sync scoring.} SyncNet~\citep{chung2016out} is run on the speech stem and original video to produce \texttt{has\_face} and \texttt{lip\_sync\_confidence}. We define \texttt{speech\_synced} as \texttt{has\_face} $=$ \texttt{True} and \texttt{lip\_sync\_confidence} $\geq 3.0$ (as shown in LatentSync~\citep{li2024latentsync}).
    \item \textit{Audio event detection.} FireRedASR2S~\citep{xu2026fireredasr2s} runs on the original audio to estimate \texttt{speech\_ratio}, the fraction of time labeled as speech or singing, and \texttt{music\_ratio}, the fraction labeled as music. We define \texttt{high\_music} as \texttt{music\_ratio} $\geq 0.1$.
    \item \textit{Instrument detection.} For all \texttt{high\_music} clips, including speech-synchronized ones, Qwen3-VL~\citep{qwen3vl2025} predicts \texttt{is\_music\_instrument}. This protects instrument-performance videos whose music is part of the visible event rather than removable BGM.
    \item \textit{Speech branch.} Clips satisfying \texttt{speech\_synced} form the speech mid-training pool. We keep the original audio unless the clip is \texttt{high\_music} and \texttt{is\_music\_instrument} is \texttt{False}; in that case, SAM-Audio removes music from the original waveform. This edited path is retained only if a second FireRedASR2S pass on the candidate audio reports \texttt{speech\_ratio} $\geq 0.05$ and \texttt{music\_ratio} $= 0$, and only if the candidate is not near-silent according to \texttt{max\_abs} $\geq 0.007$, \texttt{p50\_db} $\geq -80$, and \texttt{active\_ratio} $\geq 0.2$. This preserves lip-synchronized speech while suppressing non-instrumental BGM.
    \item \textit{Non-speech branch.} Clips not assigned to the speech-synchronized branch are curated for non-speech audio-visual generation. We first choose a base waveform: if \texttt{speech\_ratio} $\geq 0.05$, we use the SAM-Audio remaining stem to remove vocals; otherwise, we keep the original audio to preserve physical sounds. If the clip is \texttt{high\_music} and \texttt{is\_music\_instrument} is \texttt{False}, SAM-Audio music removal is applied to the chosen base waveform. Candidates derived from the remaining stem must pass a second FireRedASR2S check with \texttt{speech\_ratio} $< 0.05$; their \texttt{music\_ratio} must be $= 0$ when music removal was applied, and $< 0.1$ otherwise. Candidates derived from the original waveform with music removal must have \texttt{music\_ratio} $= 0$. Processed candidates must also pass the same non-silence thresholds used for the speech branch.
    \item \textit{Caption annotation.} Captions are tied to the final waveform used for training. If the selected waveform is the original audio, we retain the original audio description. If source separation or music removal changes the waveform, we re-caption the final candidate audio so that the text does not describe removed speech or accompaniment. For the speech-synchronized pool, we transcribe each selected speech track with Qwen3-ASR~\citep{shi2026qwen3asr}, then use GPT-OSS-120B~\citep{agarwal2025gptoss} to merge the transcript with the audio description. The resulting caption specifies both spoken content and non-linguistic acoustic context while remaining faithful to the audio paired with the video.
\end{itemize}
\subsubsection{Action}
\label{subsubsec::data_gen_action}

Actions provide the causal variables that connect observed world states across time. While video-only training teaches the generator to extrapolate likely motion, it does not expose the model to controllable interventions: the same initial observation may evolve differently under different robot commands, camera trajectories, vehicle routes, or human hand motions. We therefore introduce paired text-video-action data during mid-training so that Cosmos 3 can learn both directions of the world-action relationship: predicting future observations conditioned on actions, inferring the actions that explain an observed trajectory, and jointly generating actions and future video.

\begin{figure}[h]
    \centering
    \begin{minipage}[t]{0.48\linewidth}
        \vspace{0pt}
        \centering
        \resizebox{\linewidth}{!}{\includegraphics{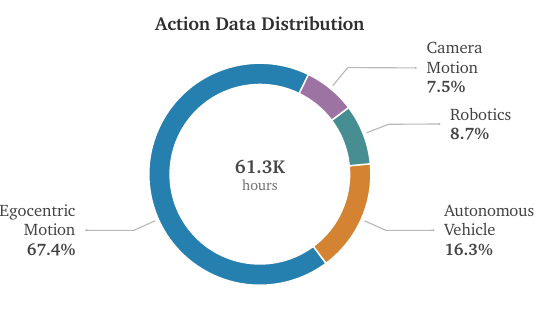}}
        \captionof{figure}{\textbf{Action data distribution.} Hours are aggregated over the four main action-data pillars in the final curated action mid-training set, which contains $8.4$M episodes and $61.3$K hours.}
        \label{fig:action_data_distribution}
    \end{minipage}
    \hfill
    \begin{minipage}[t]{0.48\linewidth}
        \vspace{0pt}
        \centering
        \captionof{table}{\textbf{Robotics data breakdown.} Grouped by robot embodiment.}
        \label{tab:robotics_breakdown}
        \scriptsize
        \setlength{\tabcolsep}{1pt}
        \renewcommand{\arraystretch}{1.3}
        \begin{tabular}{@{}llrrr@{}}
            \toprule
            \textbf{Embodiment} & \textbf{Data source} & \textbf{Tasks} & \textbf{Episodes} & \textbf{Hours} \\
            \midrule
            AgiBot & \cite{bu2025agibot} & 338 & 239.4K & 4.37K \\
            \cmidrule{1-5}
            Franka Panda & \begin{tabular}[t]{@{}l@{}}\cite{wu2024robomind};\\ \cite{khazatsky2024droid}\end{tabular} & 67.5K & 76.3K & 442 \\
            \cmidrule{1-5}
            Google Robot & \cite{brohan2022rt} & 599 & 87.2K & 351 \\
            \cmidrule{1-5}
            WidowX-250 & \cite{walke2023bridgedata} & 21.8K & 50.4K & 100.1 \\
            \cmidrule{1-5}
            UMI & \begin{tabular}[t]{@{}l@{}}\cite{lin2024data}; \cite{ha2024umilegs};\\ \cite{liu2024maniwav}; \cite{chi2024umi};\\ \cite{liu2025vitamin}; \cite{wu2024fastumi}\end{tabular} & 43 & 38.3K & 67 \\
            \cmidrule{1-5}
            UR & \cite{wu2024robomind} & 114 & 25.0K & 35 \\
            \midrule
            Total & -- & 90.4K & 516.7K & 5.36K \\
            \bottomrule
        \end{tabular}
    \end{minipage}
\end{figure}

\paragraph{Data statistics.} We focus action mid-training on four physical-AI pillars: egocentric motion, robotics, autonomous vehicles, and camera motion. The final curated data contains $8.4$M episodes and $61.3$K hours across these pillars, as summarized in \cref{fig:action_data_distribution}.

\begin{itemize}
  \item \textit{Egocentric motion.} Egocentric motion data contributes $41.3$K hours ($67.4\%$), making it the largest component. It comprises $1.7$M episodes from a proprietary dataset of bimanual hand manipulation captured with a head-mounted RGB camera.   Each frame is annotated with the synchronized head-camera pose and, for each hand, a 21-keypoint 3D pose~\citep{zimmermann2017learning,simon2017hand} that provides per-joint position and orientation in the camera coordinate frame, enabling the model to jointly learn egocentric ego-motion and fine-grained dexterous hand motion.
  \item \textit{Autonomous vehicle.} Autonomous vehicle data contributes $10.0$K hours ($16.3\%$), derived from high-quality, in-house driving logs collected using the NVIDIA Hyperion platform. The dataset is constructed by mining a large-scale corpus to match a target distribution spanning diverse driving scenarios. The selected scenarios cover a broad range of conditions, including diverse weather, lighting, and road conditions, as well as varied longitudinal and lateral maneuvers, rather than being limited to predominantly near-straight cruising. To align with other domains, we transform driving trajectories from the vehicle coordinate frame to the front-wide camera coordinate frame.
  \item \textit{Robotics.} Robotics data contributes $5.4$K hours ($8.7\%$), aggregated from open-source datasets. The subset contains $90.4$K tasks and $516.7$K episodes, as broken down by embodiment and source in \cref{tab:robotics_breakdown}. To avoid embodiment-specific controller details such as PID parameters or low-level actuation interfaces, we use pseudo-actions derived from state differences. We curate data from both successful and failed episodes so the model observes not only intended completions but also off-nominal action effects.
  \item \textit{Camera motion.} Camera motion data contributes $4.6$K hours ($7.5\%$), mined from our pre-training video dataset. We convert these videos into action trajectories by estimating camera poses with ViPE~\citep{huang2025vipe} and DepthAnything3~\citep{lin2025depthanything3}. To ensure data quality, we rigorously filter the dataset to remove clips with unreliable pose estimation, such as those exhibiting excessive jitter or abnormal camera intrinsics. All camera poses are kept in metric scale and converted to the unified action coordinate convention.
  This curation process yields a dataset of $1.9$M clips.
\end{itemize}

\paragraph{Data processing pipeline.}
We convert each source using the unified action tokenization described in \cref{sec:action}. To balance action magnitudes across embodiments after this conversion, we compute per-dimension normalizers from the training data and scale action channels to a comparable range of roughly $[-1,1]$. For data with multiple synchronized viewpoints, we concatenate the views into a canvas and store the camera layout in metadata, as shown in \cref{fig:action_multiview_packaging}. Rather than filtering out idle operations, we retain them and record the idle-step count in metadata, allowing downstream sampling to explicitly balance active and inactive segments.

\section{Training}
\label{sec::training}

We train Cosmos 3 in two main phases. First, the Reasoner is pre-trained on large-scale image--text and video--text corpora, and subsequently fine-tuned on a curated Physical AI mixture, producing a strong multimodal backbone for visual understanding and reasoning. Because the Reasoner and Generator share the same transformer block architecture, the trained Reasoner weights are then used to initialize the Generator, transferring semantic and world knowledge into a model capable of synthesizing pixels, audio, and actions. The Generator is trained using a progressive multi-stage curriculum. It begins with large-scale image, video, and audio pre-training, followed by mid-training that gradually introduces action and transfer data. Finally, the model is post-trained on smaller, carefully curated Physical AI datasets to improve downstream behavior, physical consistency, and action fidelity.

\subsection{Reasoner Training}

The Cosmos 3 Reasoner is trained in two stages: large-scale multimodal pre-training followed by supervised fine-tuning on curated Physical AI tasks. During pre-training, the model learns general multimodal representations from large-scale image--text and video--text corpora. Supervised fine-tuning then specializes the model for Physical AI domains, including robotics, autonomous driving, and smart infrastructure applications, while preserving the broad capabilities acquired during pre-training.

\subsubsection{Pre-Training}

Reasoner pre-training starts from a language model and a ViT encoder connected through a multimodal projector. For initialization, we experiment with both our internally pre-trained models described in Appendix~\ref{app:training_edge} and open-source Qwen3-VL models, using ours for the Edge model, Qwen3-VL-8B for the Nano model, and Qwen3-VL-32B for the Super model. Unlike previous approaches that perform a separate alignment stage by training only the projector while freezing the remaining VLM parameters~\citep{bai2025qwen3}, we found such staged alignment to be unnecessary and instead train all components jointly from the start of pre-training.

The model is trained using a next-token prediction objective over the large-scale multimodal corpus described in~\cref{subsubsec::data_reas_pretrain}. We train for two epochs over the full pre-training mixture using a no-replacement sampler that uniformly concatenates all datasets. Because many Physical AI applications require efficient reasoning and low-latency inference, we restrict training to sequences of at most $16$k tokens, with per-sample limits of $2048$ image tokens and $8192$ video tokens.

Following prior work~\citep{bai2025qwen3, wang2025internvl35}, we apply square-root normalized per-token loss weighting to balance the contributions of short and long sequences. We found that this normalization strategy significantly improved downstream benchmark scores and overall training stability.

Optimization uses AdamW with a peak learning rate of $5{\times}10^{-5}$ for the language model and projector, and $5{\times}10^{-6}$ for the ViT. All learning rates follow a cosine decay schedule to $0.1{\times}$ of the peak value after a $10\%$ linear warm-up phase. We use Adam coefficients $(\beta_1, \beta_2) = (0.9, 0.999)$. Training additionally uses weight decay of $0.05$ and gradient clipping with a global norm threshold of $1.0$.

\subsubsection{Supervised Fine-Tuning}

To adapt the model to downstream Physical AI tasks, we perform supervised fine-tuning on a curated high-quality multimodal mixture. Unlike pre-training, where datasets are sampled uniformly across epochs, supervised fine-tuning uses an importance-aware sampling strategy in which each dataset is assigned a fixed sampling budget based on its importance, quality, and scale. This allows optimization to focus on high-value downstream tasks while still maintaining diversity across domains and capabilities.

To prevent downstream specialization from degrading the model's general reasoning and visual understanding capabilities, we additionally mix in a filtered high-quality subset of pre-training data using a fixed 1:4 pre-training-to-SFT sampling-budget ratio. Retaining a small pre-training stream improves robustness, preserves instruction-following behavior, and maintains strong general-domain capabilities on several benchmarks. We also include a lightweight instruction-following dataset (800K samples) within the supervised fine-tuning mixture to further stabilize conversational and instruction-following capabilities during task adaptation.

Training is performed for 8200 iterations with a global batch size of 512. We use the AdamW optimizer with a peak learning rate of $1{\times}10^{-5}$ for the language model and projector, and $1{\times}10^{-6}$ for the ViT. All learning rates follow a cosine decay schedule to $0.1{\times}$ of the peak value after $1000$ steps of linear warm-up. We use Adam coefficients $(\beta_1, \beta_2) = (0.9, 0.95)$. We use weight decay of $0.1$ and gradient clipping with a global norm threshold of $1.0$.

\subsection{Generator Training}
\label{sec:generator-training-recipe}

The Cosmos 3 Generator is trained using a progressive multimodal curriculum designed to jointly model visual, auditory, and action-conditioned world dynamics across diverse resolutions, durations, and conditioning modalities. The training recipe emphasizes scalability, high-fidelity generation, and efficient long-context learning. During pre-training, the model learns general generative priors from large-scale data spanning images, videos, and audio. Subsequent training stages progressively introduce richer multimodal supervision, including actions and transfer sequences, enabling the model to learn temporally coherent world evolution and physically grounded interactions.

\paragraph{Training objective.} The Cosmos 3 generator is optimized under a rectified flow matching objective across all modalities. For a target latent from any modality, we construct a noisy latent via the straight-line interpolation $x_\sigma = \sigma \cdot \epsilon + (1 - \sigma) \cdot x_0$, where $x_0$ is the clean target, $\epsilon \sim \mathcal{N}(0, I)$, and $\sigma \in [0, 1]$ is the noise level. A single denoiser $v_\theta(x_\sigma, \sigma, c)$ is trained to predict the constant velocity $v^* = \epsilon - x_0$ via masked mean-squared error, where conditioning tokens (\eg, clean conditional frames in image-to-video tasks) are gated out of the loss.
We apply per-modality time sampling, drawing noise level $\sigma$ independently for each modality (images, videos, audio, and action).
Following Waver~\citep{waver}, we use logit-normal noise distribution for image, audio, and action batches and mode sampling for video batches. We found that using mode sampling yields better generation quality. We further map $t$ through a rectified-flow shift reparameterization $\sigma = s \cdot \bar{t} / (1 + (s-1) \cdot \bar{t})$ with $\bar{t} = 1 - t$, where $s \geq 1$ biases the marginal toward higher noise. 

\subsubsection{Pre-Training}

During the pre-training stage, we jointly train the model to generate images, videos, and audio across diverse resolutions and generation tasks. To support this, we employ a multi-resolution training strategy and optimize the model jointly over multiple generation tasks, including Text-to-Image, Text-to-(Video+Audio), Image-to-(Video+Audio), and Video-to-(Video+Audio).

\paragraph{Multi-resolution training.}
Rather than committing to a single output resolution, we train simultaneously across three resolution tiers (256p, 480p, 720p), five aspect ratios and variable number of frames, as shown in~\cref{tab:model-specs}. This exposes the model to high-fidelity content while encouraging resolution-agnostic representations. The training data is partitioned accordingly: the 256p stream draws from the full dataset (all native resolutions are eligible), the 480p stream is restricted to source material with native resolution at or above 480p, and the 720p stream uses only content at or above 720p, preserving sharpness and fine detail at the highest tier. Each resolution tier imposes a different maximum frame budget: up to 400 frames at 256p and 480p, and 300 frames at 720p. We restrict 720p to 300 frames due to the sequence length constraints. Training batches are composed across the four tiers using a 1:1:2:1 ratio for image-only, video-256p, video-480p, and video-720p samples, respectively. We find that this distribution provides a strong balance between high-fidelity learning and sample diversity, enabling the model to observe more training examples while still emphasizing higher-resolution content.
We use resolution-adaptive shift values: $s=1$ at 256p, $s=3$ at 480p, and $s=5$ at 720p.
\begin{table}[t]
\centering
\caption{\textbf{Image/Video Model Specifications.} Supported configurations for image and video modalities. Each row shows the FPS range, frame counts (video only), and image/video dimensions (w, h) for the five supported aspect ratios at each resolution.}
\small
\setlength{\tabcolsep}{6pt}
\resizebox{0.9\linewidth}{!}{%
\begin{tabular}{l cc ccccc}
\toprule
 & \multicolumn{2}{c}{\textbf{Video}} & \multicolumn{5}{c}{\textbf{Dimensions (w, h) by aspect ratio for images/videos}} \\
\cmidrule(lr){2-3} \cmidrule(lr){4-8}
\textbf{Resolution} & \textbf{FPS} & \textbf{\# frames} & \textbf{16:9} & \textbf{4:3} & \textbf{1:1} & \textbf{3:4} & \textbf{9:16} \\
\midrule
256p & 10--30 & 5--400 & (320, 192)  & (320, 256)  & (256, 256) & (256, 320)  & (192, 320) \\
480p & 10--30 & 5--400 & (832, 480)  & (736, 544)  & (640, 640) & (544, 736)  & (480, 832) \\
720p & 10--30 & 5--300 & (1280, 720) & (1104, 832) & (960, 960) & (832, 1104) & (720, 1280) \\
\bottomrule
\end{tabular}
}
\label{tab:model-specs}
\end{table}

\begin{figure}[t]
    \centering
    \begin{minipage}[c]{0.65\linewidth}
        \centering
        \resizebox{\linewidth}{!}{\includegraphics{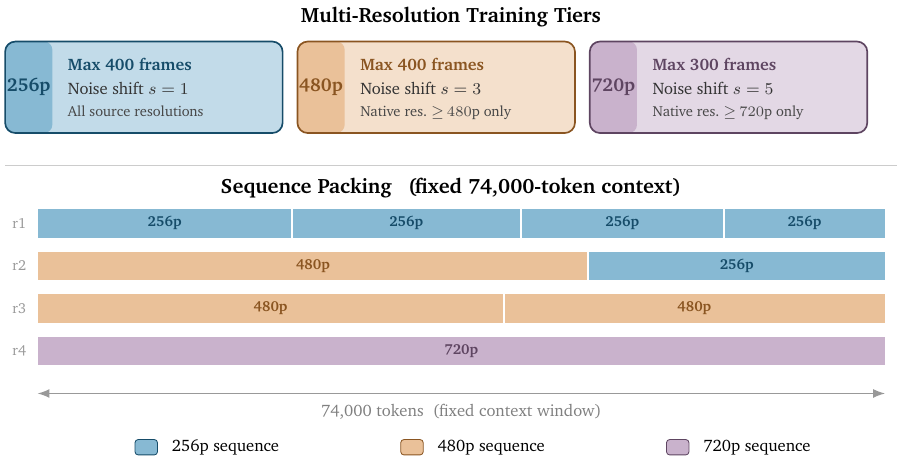}}
    \end{minipage}%
    \hfill
    \begin{minipage}[c]{0.33\linewidth}
        \centering
        {\scriptsize\bfseries Image--video pre-training data mixture\\ by resolution\par}
        \vspace{4pt}
        \resizebox{\linewidth}{!}{\includegraphics{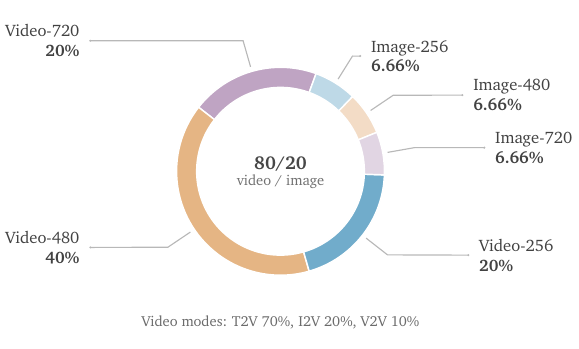}}
    \end{minipage}
    \captionsetup{justification=raggedright, singlelinecheck=false}
    \caption{%
        \textbf{Left:} \textbf{Multi-resolution training and sequence packing.}
        The three resolution tiers (256p, 480p, 720p) differ in their maximum frame budget,
        eligible source material, and rectified-flow noise-shift value; variable-length sequences
        from different tiers are packed together to fill a fixed 74,000-token context window,
        maximizing GPU utilization without padding.
        \textbf{Right:} \textbf{Data mixture used in generator pre-training.} We use joint image-video training, with videos sampled $80\%$ of the time and images the remaining $20\%$. Within each split, we train at multiple resolutions: 256p, 480p, and 720p. For video batches, we additionally sample uniformly among three conditioning modes---text-to-video, image-to-video, and video-to-video. The exact data mixture is shown in the right panel.
    }
    \label{fig:multiresolution_seqpacking}
\end{figure}

To prevent gratuitous recompilation overhead while supporting variable sequence lengths, we use token packing with a fixed budget of 74,000 tokens per sequence. Sequences at various resolutions are packed together to fill each batch, maximizing GPU utilization without padding (depicted in~\cref{fig:multiresolution_seqpacking}). 

\paragraph{Training modes.}
For a latent video tensor of shape $C\times T\times H \times W$, let $T_{\textrm{cond}}$ denote the number of conditional latent frames and $T_{\textrm{noised}}$ the number of noisy latent frames ($T = T_{\textrm{cond}} + T_{\textrm{noised}}$). During training, no noise is applied to the first $T_{\textrm{cond}}$ frames, which serve as conditional inputs; only the remaining $T_{\textrm{noised}}$ frames are noised, and the model learns to denoise them. Different choices of $T_{\textrm{cond}}$ and $T_{\textrm{noised}}$ yield different training modes. We use four generation modes---Text-to-Image, Text-to-Video, Image-to-Video, and Video-to-Video---distinguished solely by the number of conditioning visual frames prepended to each sample, with sampling ratios of $20\%$, $56\%$, $16\%$, and $8\%$, respectively. All modes use the structured JSON caption format described in~\cref{subsec::data_generator}.

\begin{itemize}
\item \textit{Text-to-Image (T2I).} Images are treated as a special case of videos with the temporal dimension restricted to $T=1$. In this mode, images are drawn randomly from all three resolution tiers and aspect ratios, then sequence-packed before being sent to the model. Since an image sample yields far fewer tokens than a video, a typical sequence contains many more samples than its video counterpart.
\item \textit{Text-to-Video (T2V).} In text-to-video training, $T_{\textrm{cond}}=0$. The model learns to denoise the entire video conditioned solely on text. Alongside the caption, the model receives duration, FPS, and timestamp metadata as additional fields in the JSON caption, enabling it to generate videos of specified length and temporal extent.
\item \textit{Image-to-Video (I2V).} For single-frame conditioning ($T_{\textrm{cond}}=1$), the first latent frame is held clean while subsequent frames are noised. The model learns to generate future frames consistent with both the initial frame and the caption.
\item \textit{Video-to-Video (V2V).} For multi-frame conditioning ($T_{\textrm{cond}}=2$), the model is conditioned on the first five frames of a video (equivalently, the first two latent frames) and learns to predict future frames consistent with both the conditioning frames and the input prompt.
\end{itemize}

\paragraph{FPS modulation.}
We train the model with varying FPS values, so the physical temporal spacing between tokens differs across samples: a clip sampled at 30~FPS packs frames more densely in real time than the same number of tokens sampled at 16~FPS. To reflect this, we modulate the temporal axis of 3D MRoPE position encodings by assigning temporal coordinates in proportion to real-world time rather than token index (see \cref{sec::model}), with a base rate of 24~FPS. Duration and FPS are also appended to the text prompt, allowing the model to be conditioned on specific temporal characteristics at inference time.

\paragraph{Optimization.}
Only the generation-specific parameters are updated during the generator pre-training. The reasoner tower remains frozen, preserving the language and visual understanding capabilities. We use FusedAdamW with learning rate $10^{-4}$, $(\beta_1, \beta_2) = (0.9, 0.99)$, weight decay $0.05$, and gradient clipping at norm~1.0. The learning rate schedule follows a linear decay with warmup, from the peak lr to a floor of $0.30\times$ over $n$ iterations. To enable classifier-free guidance, we use a text-dropout rate of $10\%$ across all modalities.

\paragraph{Tokens trained.} In the pre-training stage, Cosmos3-Nano was trained on $31.05$T tokens using $1024$ NVIDIA GB200 GPUs, while Cosmos3-Super was trained on $17.86$T tokens using $2048$ NVIDIA GB200 GPUs.

\subsubsection{Mid-Training}

Mid-training bridges the gap between broad pre-training and downstream deployment. At this point, the Generator has already learned general image, video, and audio generation from large-scale data, but the target Physical AI applications require stronger coverage of rare dynamics, embodied scenes, control interfaces, and high-quality visual domains. We therefore continue training from the pre-trained checkpoint with a curated mixture that both preserves the original visual generation modes and introduces new sources of supervision. The stage has two complementary objectives: \textit{domain specialization}, which increases exposure to high-value Physical AI domains, and \textit{multimodal integration}, which extends the model from visual and audio generation to action- and control-conditioned world modeling.

\paragraph{Domain specialization.}
While retaining its general knowledge, the model is exposed to highly curated specialized datasets to improve quality and reliability in application-critical Physical AI scenarios. For images, we use a 15.6M-sample mid-training pool that emphasizes high-quality real imagery while adding synthetic and text-rendering data to broaden concept coverage and preserve legible text generation. For videos, we incorporate 74.7M curated clips spanning robotics, autonomous driving, human activity, physics, and synthetic simulation data. These sources target failure modes that are underrepresented in generic web-scale pre-training, such as long-horizon interactions, fine-grained human and robot motion, physical object dynamics, and safety-critical driving or warehouse scenarios. By mixing these domain-focused datasets with the existing image and video training modes, mid-training improves Physical AI relevance without discarding the broad visual priors learned during pre-training, as described in~\cref{subsubsection::imagevideo}.

\paragraph{Multimodal integration.}
Mid-training expands the Generator from image, video, and audio generation into a unified Physical AI model that can also consume and synthesize action and control signals. We keep the same clean-prefix/noisy-target formulation used in pre-training for T2I, T2V, I2V, and V2V, so existing visual capabilities remain active while new modality-specific tokens are introduced in the diffusion subsequence. This lets action, audio, control, and video tokens share the same temporal coordinate system and two-way attention pattern described in~\cref{sec::model}. In addition to the pre-training modes, we add two additional families of multimodal supervision: action and video transfer.
\begin{itemize}
\item \textit{Action.}
We introduce paired text-video-action training data using the unified action representation in~\cref{sec:action}. The model is trained not only to predict future video conditioned on actions, but also to infer actions from observed trajectories and to jointly generate actions and visual futures. This teaches the Generator a causal interface between controllable interventions and world evolution.
\item \textit{Video transfer.}
We add control-conditioned transfer data in which clean control signals are provided as inputs and the model denoises the corresponding target image or video. The control signals include edge, blur, depth, and segmentation maps from high-quality video corpora, as well as world-scenario maps for driving scenes. This exposes the model to spatially grounded constraints while retaining text conditioning and visual generation quality.
\end{itemize}
The mixing ratios of different modalities are shown in \cref{tab:midtraining_modality_mix}.
\begin{table}[t]
    \centering
    \caption{\textbf{Generator mid-training data mixture.} After pre-training is done, we introduce new modalities (action and transfer) in the mid-training stage with the data ratios listed below.}
    \label{tab:midtraining_modality_mix}
    \resizebox{0.8\linewidth}{!}{%
    \begin{tabular}{@{}p{0.22\linewidth}p{0.58\linewidth}c@{}}
        \toprule
        \textbf{Training stream} & \textbf{Modes / Conditioning} & \textbf{Share} \\
        \midrule
        Image & T2I & 10\% \\
        Video & T2V, I2V, V2V & 32\% \\
        Video + Audio & T2(V+Audio), I2(V+Audio), V2(V+Audio) & 8\% \\
        Action & Forward dynamics, inverse dynamics, policy & 25\% \\
        General Transfer & Edge, blur, depth, and segmentation controls & 20\% \\
        Driving Transfer & World-scenario-map controls & 5\% \\
        \bottomrule
    \end{tabular}
    }
\end{table}

\paragraph{Multi-resolution training.}
Similar to pre-training, mid-training uses multi-resolution across 256p, 480p, and 720p within a fixed 74K context window. To better handle dynamics and reduce temporal and high-resolution artifacts, we increase rectified-flow shift values to $3$, $5$, and $10$ for 256p, 480p, and 720p, respectively.

\paragraph{Training objective.} Similar to pre-training, we use the rectified flow objective for all modalities. For action, we inherit the vision noise schedule. The total loss in mid-training is the sum of per-modality velocity MSEs weighted by modality-specific loss scales, with action losses scaled by $10\times$ to compensate for the smaller per-element MSE of normalized action vectors.

\paragraph{Optimization.}
Similar to pre-training, we use FusedAdamW with learning rate $10^{-4}$, weight decay $0.05$, gradient clipping at norm~1.0, and loss scale~10. The learning rate follows a LambdaLinear schedule with start factor $0.4$ and cycle length $100{,}000$.

\paragraph{Tokens trained.} In the mid-training stage, Cosmos3-Nano model was trained on $2.4$T tokens using $1024$ NVIDIA GB200 GPUs, while Cosmos3-Super model was trained on $1.9$T tokens using $2048$ NVIDIA GB200 GPUs.

\subsubsection{Text-to-Image Post-Training}
\label{sec::t2i_post_train}

To demonstrate the omnimodal capability of Cosmos3-Super, we further specialize the model into a text-to-image checkpoint, Cosmos3-Super-Text2Image. Our goal is to transfer the model's physically grounded world understanding to high-quality image generation, aiming for strong open-source T2I results while improving physical plausibility and scene-level alignment.

We perform text-to-image specialization using a two-stage SFT, following the common text-to-image foundation-model training paradigm that emphasizes semantic enhancement before preference-oriented refinement.

\begin{itemize}
\item \textit{Stage 1: broad T2I specialization.} We fine-tune the model for 20k training steps on the curated high-quality SFT dataset. The training mixture is sampled with a controlled ratio of $45\%$ general real image data, $40\%$ synthetic image data, and $15\%$ text-rendering-only data, balancing visual fidelity, caption alignment, and language retention. We use a base learning rate of $1\times10^{-4}$, 2k warmup iterations, and a linear learning-rate decay schedule, while keeping all other hyperparameters consistent with the Cosmos 3 mid-training stage.

\item \textit{Stage 2: high-quality refinement.} We perform a final 2k-step SFT pass using $470$k carefully curated ultra-high-quality image--caption pairs. This stage further improves visual aesthetics, prompt-following, text-rendering quality, and alignment with human preferences.

\item \textit{Resolution and context length.} For both stages, we use a fixed context window of 70k tokens and train only on images with a resolution higher than 720p.
\end{itemize}

Overall, Cosmos3-Super-Text2Image delivers strong text-to-image results across both semantic alignment and English text-rendering benchmarks. On UniGenBench, it achieves the best overall score among the evaluated models, reaching $91.36$ on the full benchmark (see \cref{tab:t2i_results}). With an agentic workflow, the model ranked top-1 among open-weight models on the Artificial Analysis Text-to-Image leaderboard (\cref{subsec::image_eval}). These results suggest that downstream T2I modality adaptation from Cosmos 3 is highly effective: it improves scene-level prompt alignment while preserving the model's physically grounded generation capability.

\subsubsection{Image-to-Video Post-Training}
\label{sec::i2v_post_train}

Image-to-Video capability is fundamentally important for comprehensive visual understanding. It probes the model's understanding of physical laws, object permanence, and intricate scene geometry, while also serving as a critical predictive mechanism for embodied AI and robot planning, where simulating plausible future frames yields an effective world model \citep{wiedemer2025video, chen2025largevideoplanner}. While Cosmos 3 is inherently designed to handle a diverse array of tasks natively, we utilize SFT to explicitly showcase and specialize its potential in the I2V domain. To demonstrate these capabilities, we employ the following procedure:

\begin{itemize}
    \item \textit{Data and training mixture.} We fine-tune the model using filtered pre-training data that have been refined for a more balanced topic diversity, augmented via an agentic workflow that identifies model weak spots to retrieve targeted examples from the pre-training set. This is combined with 1,000 high-quality manually curated videos and a dataset of approximately 20k synthetic video clips spanning diverse topics (accounting for roughly 6\% of the total tokens). While all video sequences are trained exclusively using the I2V formulation, our training mixture also incorporates 20\% T2I image tokens to preserve the model's semantic alignment.

    \item \textit{Resolution and duration.} We specialize the model for temporal generation at a targeted resolution of 480p and targeted duration of 189 frames, corresponding to roughly 8 seconds at 24fps. This configuration balances inference speed with temporal context, enabling fast, physically plausible video generation over a meaningful time horizon.

    \item \textit{Training schedule.} The I2V post-training stage runs for a duration of 10k iterations at a learning rate of $1\times10^{-5}$. The model processes roughly 50B tokens over the course of SFT.
\end{itemize}

Through post-training, Cosmos3-Super-Image2Video achieves leading quality in image-to-video generation. In particular, the model ranked top-1 among open-weight models on the Artificial Analysis Image-to-Video leaderboard (\cref{subsec::video_eval}). For details on the usage of this model, please refer to \cref{sec::gen_guide}.

\subsubsection{Robot Policy Post-Training}
\label{sec::robot_policy_post_train}

We conduct robot policy post-training to investigate whether our Cosmos 3 omnimodal world models can be extended into powerful robot policy models. Mid-training enables Cosmos 3 to model multimodal sequences, including language, visual observations, and actions, and to generate actions jointly with videos. We further customize it for robot policy learning by incorporating proprioceptive signals, reducing inference latency, and adapting the model to produce executable actions for closed-loop control.

As a pilot study, we use the DROID robot platform and dataset~\citep{khazatsky2024droid} due to its popularity and broad community adoption. The DROID platform uses a Franka Panda 7-DoF manipulator with a Robotiq 2F-85 parallel-jaw gripper to perform tabletop manipulation tasks in diverse real-world environments. The DROID dataset comprises 76k trajectories, 350 hours of interaction data, 86 tasks, and 564 scenes, providing substantial scale and broad task diversity for real-world robot policy learning. We ingest DROID at a high resolution of 360$\times$640, apply community-provided idle-frame filtering and failure-demonstration removal, and use random image augmentation during training.

We post-train Cosmos3-Nano-Policy-DROID by resuming from our mid-trained Cosmos3-Nano model, with a freshly initialized action encoder, action-decoding MLP, and action embedding tokens. We apply a 5$\times$ learning-rate multiplier to the action-related parameters to facilitate faster adaptation. The policy input consists of the current proprioceptive robot state and a three-view visual observation. Specifically, the wrist-view image, with a raw resolution of 360$\times$640, is placed above two external-view images, each with a raw resolution of 180$\times$320, which are concatenated side by side on the bottom left and bottom right. The resulting canvas is 540$\times$640. The policy is trained to predict 32 future absolute joint-position actions, along with auxiliary RGB video frames as additional outputs, operating at 15Hz. We use the official DROID short task instructions as the prompts during this post-training study. We use a learning rate of $2 \times 10^{-4}$ with other hyperparameters following the mid-training setup.

At inference time, we sample the model using 4 diffusion steps with a shifted noise schedule of 5. We also apply classifier-free guidance with CFG parallelism at a guidance scale of 3, and skip video-latent decoding to further reduce inference overhead. Together, these optimizations provide a significant inference speedup, enabling policy server deployment on 2 NVIDIA RTX Pro 6000 GPUs. The downstream joint-position controller is implemented using Franky~\citep{Schneider_franky_High-Level_Control} and executes the predicted 32 actions at 15Hz.

Overall, Cosmos3-Nano-Policy-DROID achieves strong results in robotic policy tasks. As detailed in \cref{subsec::action_benchmarks}, it ranked first on RoboLab~\citep{yang2026robolab}, RoboArena~\citep{atreya2025roboarena}, and MolmoSpaces~\citep{kim:arxiv2026} at the time of our leaderboard submissions, demonstrating the effectiveness of Cosmos3 as a foundation model backbone for robot policy learning.
\section{Infrastructure}
\label{sec::infrastructure}

\begin{figure}[t]
  \resizebox{\linewidth}{!}{\includegraphics{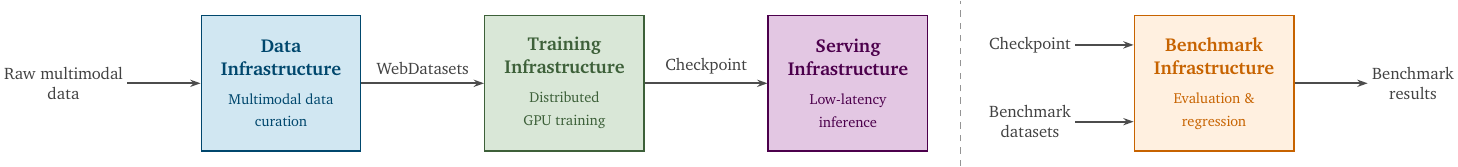}}
  \captionsetup{justification=raggedright, singlelinecheck=false}
  \caption{\textbf{Overview of the Cosmos 3 infrastructure stack.}
  The platform spans four pillars. \emph{Data Infrastructure} ingests raw
  multimodal streams and curates them into WebDataset-format training
  shards. \emph{Training Infrastructure} consumes those shards on NVIDIA
  GPU clusters with efficient parallelization, data loading, and
  checkpointing. The resulting checkpoints feed two parallel paths
  (separated by the dashed divider): \emph{Serving Infrastructure}
  deploys them for low-latency generation and reasoning inference, while
  \emph{Benchmark Infrastructure} evaluates the same checkpoints against
  standardized benchmark datasets to track regressions and enable
  systematic validation.}
  \label{fig:cosmos_infra_pipeline}
\end{figure}

In this section, we describe the integrated infrastructure stack designed to support the end-to-end lifecycle of Cosmos 3. As illustrated in~\cref{fig:cosmos_infra_pipeline}, the platform unified 4 core pillars:
\begin{itemize}
    \item \textit{Data engineering.} Ingests raw multimodal data and transforms it into curated datasets in the WebDataset format, optimized for scalable, distributed training.
    \item \textit{Large-scale training.} Maximizes NVIDIA GPU cluster utilization through highly efficient parallelization strategies, optimized data loading, rapid checkpointing, and collective communication primitives.
    \item \textit{Model serving.} Enables efficient, low-latency deployment and inference execution across both generative and reasoning workloads.
    \item \textit{Benchmarking \& validation.} Provides a unified evaluation framework to assess model capabilities across diverse tasks, enabling automated regression tracking and systematic model comparison.
\end{itemize}

\subsection{Data Infrastructure}
\label{subsec::data_infrastructure}
The Cosmos 3 training corpus is drawn from tens of billions of image and video candidates spanning diverse modalities, domains, and tasks. Operating at this scale demands a data infrastructure that can simultaneously (1) transform raw multimodal data into training-ready samples through large-scale distributed processing, (2) support embedding-based retrieval, clustering, and deduplication, and (3) enable interactive dataset visualization, inspection, and debugging. To meet these requirements, we developed \textbf{SILA} (Scalable Infrastructure for Large-scale data processing and Annotation), a scalable multimodal data infrastructure platform that consolidates storage, metadata management, distributed processing, semantic retrieval, and dataset visualization into a single extensible framework for large-scale data curation and management.

SILA is built around a clean separation between pipeline logic and infrastructure mechanics. Researchers declare typed processing stages---specifying the columns each stage consumes and the outputs it produces---while the platform transparently handles dataset sharding, distributed execution, fault tolerance, checkpointing, metadata updates, and asset registration. This abstraction makes it straightforward to incorporate new data sources, foundation models, and processing stages, including filtering, captioning, embedding generation, scoring, and tagging, without requiring researchers to develop distributed-systems expertise. The result is a platform that lets curation evolve at the pace of research: new signals can be added, recomputed, or replaced incrementally as models, quality criteria, and training recipes change.

\subsubsection{Large-Scale Data Processing}
Multimodal data curation is an iterative enrichment process rather than a single offline preprocessing pass. Raw text, image, and video samples are repeatedly transformed, filtered, annotated, and reprocessed as models, quality criteria, labels, and training recipes evolve. Scaling this workflow is challenging because the pipeline is both low-yield and highly iterative: only a small fraction of raw candidates ultimately survive into training, meaning that inefficient scans, copies, or model inference are disproportionately spent on samples that are later discarded. At the same time, stages such as ingestion, splitting and transcoding, embedding generation, deduplication, filtering, taxonomy tagging, captioning, and sharding repeatedly operate over the same samples. Supporting this workflow therefore requires efficient mechanisms for repeated transformations and incremental recomputation across billions of multimodal samples.

These challenges become even more pronounced under distributed execution. Curation workloads must run continuously on shared clusters with fragmented and dynamically changing GPU availability rather than assuming a single monolithic allocation. The infrastructure must coordinate many distributed workers, avoid duplicate computation, recover from failures, manage heterogeneous CPU- and GPU-bound stages, and continue processing unfinished work as resources become available. To support this execution model, SILA combines a unified data layer with fragment-level coordination and fault recovery, staged distributed execution, node-local model serving, opportunistic cluster utilization, and agent-friendly operational interfaces.

\paragraph{Unified data layer.}
SILA organizes data curation as a unified columnar Lance dataset~\citep{pace2025lance}, where each row represents a data sample and each typed column represents a curation signal such as a caption, tag, quality score, or annotation. This replaces the legacy table-per-pipeline architecture used in earlier infrastructure, Cosmos-Predict 1.0~\citep{cosmos_v1} and 2.5~\citep{cosmos_predict2p5}, where each pipeline wrote to its own Postgres table and outputs were later synchronized into Databricks through Change Data Capture (CDC). As the number of pipelines and metadata fields grew, the table-per-pipeline design required increasingly complex joins across large tables to reconstruct the state of a single sample. These joins became expensive at scale and made even simple operational queries difficult to express without detailed knowledge of join keys, table relationships, and pipeline-specific schemas. In contrast, SILA incrementally enriches the same logical sample by appending new typed columns to a shared Lance table. This unified representation naturally matches multimodal curation workloads, where most stages augment existing samples with additional metadata rather than creating new entities. By co-locating dataset contents, metadata, and processing state, the system can efficiently support large-scale scans, point lookups, incremental recomputation, and discovery of unfinished work directly from Lance fragment metadata without expensive startup joins.

\paragraph{Fragment-level coordination and fault recovery.}
Curation at this scale runs at high concurrency: many distributed workers within a single job, and many independent jobs in parallel, read from and write to the same continuously evolving Lance dataset. Without explicit coordination, workers must rely on expensive startup queries, randomized sampling, or post-hoc filtering of already processed samples to avoid overlap. These approaches delay job startup, permit duplicate work, and complicate recovery when long-running jobs are preempted or interrupted. SILA instead coordinates distributed curation directly at the Lance-fragment level. Workers discover unfinished fragments from Lance metadata and acquire time-limited leases before processing them, while the Lance dataset itself remains the source of truth for completion state. Lease ownership is maintained through periodic heartbeats; when heartbeats stop, the lease expires and another worker can reclaim the fragment, enabling automatic recovery from failures, preemption, or endpoint crashes without manual cleanup. Because a single fragment may contain many samples and require hours of model inference, SILA further partitions claimed fragments into smaller processing segments, writes completed segments as durable checkpoints, and atomically commits the full fragment back into the Lance through a single metadata update once all segments finish. This decouples the recovery unit from the visibility unit: interrupted jobs resume from completed segments while downstream readers observe only fully committed fragment outputs.

\paragraph{Staged Ray execution.}
Curation pipelines combine heterogeneous operations with very different resource profiles, including data loading, decoding, model inference, postprocessing, writing, and committing. If these operations are executed through a single undifferentiated control loop, fast upstream stages can accumulate intermediate outputs while slower inference or commit stages become bottlenecks. SILA instead executes each claimed fragment through a staged Ray pipeline engine~\citep{moritz2018raydistributedframeworkemerging}. Framework-managed stages handle loading, writing, and committing, while user-defined preprocess, compute, and postprocess stages execute in separate Ray actor pools with independently configured worker counts and resource requirements. Backpressure limits in-flight work across stage boundaries, preventing fast I/O-heavy stages from overwhelming slower downstream stages and forcing Ray object-store spill to disk.

\paragraph{Node-local model endpoints.}
Foundation-model curation workloads often require serving large captioning, embedding, tagging, or scoring models while many pipeline workers concurrently process data. Centralized inference services can become bottlenecks at scale, while requiring one large contiguous GPU allocation reduces the ability to exploit fragmented cluster availability. SILA instead launches node-local model-serving endpoints using systems such as vLLM~\citep{vllm} and passes local endpoint information directly to stage workers. Workers then invoke node-local services for inference, allowing model-heavy curation stages to scale across available nodes while decoupling data-parallel pipeline execution from model-serving placement.

\paragraph{Opportunistic cluster utilization.}
Large-scale curation must run continuously alongside training workloads on shared clusters, where GPU availability is often fragmented and dynamically changing. Instead of requiring one large monolithic allocation, SILA decomposes curation into fine-grained distributed jobs that can execute incrementally as resources become available. The system supports execution backends such as DGX Cloud Lepton~\citep{nvidia2026dgxcloudlepton} and Slurm~\citep{jette2023slurmarchitecture}, allowing pipelines to opportunistically utilize idle or partially available GPU capacity. This improves cluster utilization and enables continuous data processing without requiring large contiguous GPU reservations.

\paragraph{Agentic job orchestration.}
As AI agents become increasingly capable at tool use and long-horizon execution, SILA exposes large-scale curation workflows through agent-friendly operational interfaces, including reusable skills, command-line interfaces (CLIs), and structured job metadata. Long-running orchestration agents periodically monitor curation jobs, inspect logs and execution metadata, relaunch failed stages, and coordinate operational recovery automatically. Through this continuous monitoring loop, the agents can track pipeline progress, identify stalled or unhealthy workers, verify dataset coverage, trigger incremental recomputation when new models or filtering criteria are introduced, and notify engineers about failures, recoveries, and execution status as distributed jobs evolve over time.

Together, these design choices substantially improved the efficiency of large-scale curation. By eliminating expensive startup table joins and replacing randomized work selection with fragment-level discovery and coordination, SILA reduced job startup latency from 30--60 minutes to roughly 5 minutes, depending on the stage and model configuration. Combined with staged execution, checkpointing, and improved cluster utilization, the new infrastructure achieved a $10\times$ throughput increase over the previous architecture. In peak production windows, individual SILA stages processed billions of row-level annotations per day, reducing large captioning and curation campaigns from month-scale operations to week-scale iteration cycles.

Beyond improving scalability and throughput, SILA also simplifies pipeline development by hiding much of the operational complexity behind a dataset-centric interface. Pipeline authors specify only the input columns they consume and the output fields they produce, while the framework handles schema registration, column creation, fragment discovery, work coordination, checkpointing, and committing results back to the shared dataset. As a result, adding new captioning, scoring, tagging, or filtering stages no longer requires creating new storage tables, writing synchronization logic, or manually coordinating distributed workers. Researchers can instead iterate by incrementally adding new typed columns, reusing previously computed outputs, and recomputing only samples whose required fields are missing.

\subsubsection{Embedding Storage and Semantic Retrieval}

Semantic retrieval workloads require both vector similarity search and metadata-aware filtering over billions of multimodal data. In the previous architecture, storing high-dimensional embeddings directly in SQL tables significantly inflated table size, increasing I/O overhead for joins, scans, and operational queries. As a result, embeddings were exported into a separate vector database, while metadata used for pre-filtering, post-filtering, and result interpretation remained in relational storage. However, embeddings, metadata, and filtering criteria evolve continuously during curation, requiring frequent synchronization and migration between two systems whenever new embedding models, metadata fields, or search filters are introduced.

SILA instead stores embeddings directly alongside sample metadata in Lance, allowing LanceDB to build vector indexes over the primary dataset rather than requiring a separate vector database~\citep{lancedb2026vectorindexes}. Because embeddings are stored in Lance data files rather than inline relational rows, large embedding payloads do not inflate metadata tables or slow operational queries. By co-locating embeddings, metadata, and vector indexes within the same storage layer, SILA supports semantic retrieval, clustering, and deduplication directly over the curated dataset while keeping search results consistent with the latest curation state.

In production, SILA performs semantic retrieval, clustering, and deduplication over a 4096-dimensional embedding column covering tens of billions of rows using LanceDB IVF\_PQ indexes with cosine similarity. The deployed Approximate Nearest Neighbor (ANN) configuration uses 64K IVF partitions together with PQ-compressed embeddings to support billion-scale retrieval workloads efficiently. Because the vector indexes are built directly over the primary Lance datasets, semantic retrieval operates over the same storage layer that maintains the latest curation metadata and filtering state. This allows metadata-aware filtering, nearest-neighbor retrieval, and downstream dataset analysis to remain synchronized with continuously evolving embeddings, annotations, and curation outputs without requiring synchronization between separate vector and metadata systems.

\subsubsection{Dataset Visualization, Inspection, and Debugging}

At production scale, data curation must be observable as well as scalable: researchers need to understand pipeline coverage, track how curation outputs evolve over time, and diagnose why individual samples pass or fail quality criteria. SILA therefore treats visualization, inspection, and debugging as first-class components of the data infrastructure. Its tools operate directly over Lance tables, connecting aggregate pipeline progress, representative development subsets, sample-level inspection, and downstream analytical views within the same shared curation substrate.

\begin{itemize}

\item \textit{Development and pipeline validation.}
SILA provides utilities for constructing small development Lance tables from production datasets while preserving the schema and representative operational characteristics of the full corpus. Because these development tables closely mirror production data, the same pipelines can run unchanged in both environments, allowing researchers to validate correctness and execution behavior before launching large-scale production jobs.

\item \textit{Dataset inspection and progress analysis.}
To support large-scale curation monitoring, SILA exposes both aggregate progress analyzers and interactive inspection tools directly over Lance tables. Fragment-level metadata is used to estimate per-column coverage and pipeline completion without requiring full table scans, while interactive viewers allow researchers to sample rows, render media, and inspect the associated captions, scores, annotations, and schema metadata. Because Lance supports efficient random row-level access, these inspection workflows can operate directly over the primary curation tables, avoiding the expensive scans and limited row-level retrieval patterns common in traditional Parquet-based data lakes. This allows researchers to move quickly from pipeline-level progress monitoring to sample-level debugging within the same dataset.

\item \textit{Analytical querying integration.}
For large analytical workloads, SILA supports Online Analytical Processing (OLAP) queries used for large-scale aggregation, reporting, and dashboarding over the curated corpus. These queries are simpler to express because SILA stores each sample and its curation signals in a wide Lance table: captions, tags, scores, embeddings, filtering decisions, and processing state can be selected and filtered from one logical dataset rather than reconstructed through joins across pipeline-specific tables. After selecting the relevant columns and cohorts, analytical backends can compute the required aggregations for coverage reports, quality dashboards, dataset audits, and training-set analysis. SILA keeps the Lance table as the source of truth for curation state, media, metadata, embeddings, vector indexes, and row-level inspection, while treating downstream analytical execution as a deployment choice. In practice, SILA can scan the authoritative Lance tables to materialize query-ready snapshots or projections, including Parquet files~\citep{parquet2013}, and execute OLAP workloads using the available compute backend, such as Databricks, Spark clusters~\citep{zaharia2016apache}, or Slurm-backed batch jobs.

\end{itemize}

Across processing, retrieval, and inspection, SILA closes the loop on the three objectives that define the Cosmos 3 data infrastructure: transforming raw multimodal corpora into training-ready samples, organizing them for semantic retrieval and deduplication, and keeping them inspectable throughout the curation lifecycle. By unifying assets, curation signals, embeddings, vector indexes, and execution state within a single Lance-backed substrate, SILA turns data curation from a sequence of one-off preprocessing jobs into a continuously evolving production workflow. New models and quality criteria can be applied incrementally, distributed enrichment stages can recover and scale across shared heterogeneous clusters, and researchers can move seamlessly from corpus-level progress monitoring to sample-level debugging. This integrated workflow enables the Cosmos 3 training corpus to scale to tens of billions of multimodal candidates while remaining searchable, auditable, and continuously improvable.

\subsection{Training Infrastructure}
\label{subsec::training_infrastructure}

Cosmos 3 leverages a custom infrastructure platform engineered for scaling multimodal foundation-model training. This unified stack coordinates the end-to-end lifecycle for the Reasoner and Generator training. This lifecycle spans raw multimodal sample ingestion, training computation, and persistent checkpointing, and is structured around the stages described below.

\begin{itemize}
    \item \textit{Data loader.} The data loader ingests multimodal samples---images, videos, action, audio, text---at arbitrary native resolutions and aspect ratios.  It applies on-the-fly augmentation (\eg, resizing, spatial cropping, color jitter, and temporal video sub-sampling), tokenizes text conditions, and packs variable-length samples into batches. To hide I/O and pre-processing latency, the loader runs asynchronously in parallel worker processes and prefetches batches onto the device via a pinned-memory staging buffer.

    \item \textit{Distributed training.} Training is parallelized using a combination of Hybrid Sharded Data Parallelism (HSDP) and Context Parallelism (CP). This approach enables scaling to large model sizes and extended input sequence lengths. HSDP shards optimizer states, gradients, and model parameters within each replica group while replicating across groups. CP shards the sequence dimension across devices to handle massive context windows that would otherwise overflow a single GPU's memory capacity. These two strategies compose orthogonally and are dynamically configured per experiment to optimize for the target model size, sequence length, and cluster topology.

    \item \textit{Training loop.} Orchestrated in the style of TorchTitan~\citep{torchtitan}, the training loop executes standard forward, backward, optimization, and learning-rate-scheduling cycles. It natively supports various optimizers (\eg, AdamW and fused variants), schedulers (\eg, cosine with warmup, constant with warmup), and loss functions (cross-entropy loss for Reasoner text and EDM loss for Generator). Also incorporating on-the-fly variational encoders (\eg, the Wan2.2 VAE), the pipeline operates end-to-end on raw multimodal inputs. This design eliminates offline latent-extraction phases and ensures that augmentation, encoding, and training remain in lockstep across runs.

    \item \textit{Checkpoint saving.} Checkpoints are recorded at a configurable cadence and use an asynchronous, off-critical-path persistence mechanism to prevent disk and network I/O from stalling the training loop. Model parameters, optimizer states, and RNG/data-loader state are snapshotted on-device, handed off to a background writer, and serialized to remote storage while training continues uninterrupted.
\end{itemize}

Both Reasoner and Generator are trained with this unified framework, sharing a common trainer, parallelization architecture, optimizers, learning rate schedulers, tokenizers, data loaders, and monitoring utilities.

\subsubsection{Data Loader}
\label{sec::data-loader}

The data loader bridges persistent storage and the training loop: it ingests raw multimodal training data, applies on-the-fly augmentation, and forms the batches consumed by each training step. In Cosmos 3, the data loader is required to satisfy three concurrent requirements:
\begin{itemize}
    \item \textit{Pipeline saturation.} It must stream batches asynchronously to prevent the training loop from stalling or blocking on data I/O.
    \item \textit{Distributed load balancing.} It must emit balanced batches across distributed ranks to minimize cross-rank synchronization stalls and maximize aggregate GPU utilization.
    \item \textit{Distribution fidelity.} It must guarantee that the long-run modality and resolution mixtures strictly adhere to the configured target distribution.
\end{itemize}

In conventional LLM training, all three requirements have well-established remedies.
\begin{itemize}
  \item \textit{Pipeline saturation.} This is achieved by tuning the worker count, prefetch depth, and using pinned-memory staging buffers to overlap host-to-device transfers with compute.
  \item \textit{Load balancing.} This becomes trivial because every sample contributes a fixed number of tokens. Maintaining identical per-rank sample counts guarantees uniform per-rank compute and activation-memory profiles across ranks.
\end{itemize}

However, Cosmos 3 invalidates this recipe along all three axes. Since its joint training corpus spans highly heterogeneous modalities, the per-sample token counts vary by over two orders of magnitude, making token volume the primary driver of compute and memory costs. For example, a single 720p two-second video clip produces more tokens than dozens of short text captions combined. Under this asymmetric workload, allocating equal per-rank sample counts introduces critical systemic inefficiencies: it incurs (a) substantial padding waste due to fixed batch shapes; (b) severe workload imbalance across ranks when modality assignments differ; and (c) at scale NCCL collective timeouts caused by extreme step-time variance.

To address these challenges, the Cosmos 3 data loader is built around four coordinated mechanisms:
(i) \emph{token-budgeted packed sequences}, which bound each rank's per-step workload by a token budget rather than a fixed sample count;
(ii) a \emph{joint data loader}, which multiplexes per-stream loaders into a single unified training batch;
(iii) \emph{rank-synchronous stream selection}, which uses a globally seeded selector to keep all ranks aligned on the same data stream at every step;
and (iv) \emph{look-ahead packing}, which raises the average utilization of the token budget $T_{\max}$.

\paragraph{Token-budgeted packed sequences.}
Rather than fixing the per-step sample count, each rank's workload is bounded by a strict token budget $T_{\max}$. The loader greedily concatenates samples into a single packed sequence---each contributing exactly as many tokens as its serialized form requires, with no padding inserted between samples---until appending the next candidate would exceed $T_{\max}$. By eliminating cross-sample padding, this design bounds the per-step compute cost as a direct and predictable function of $T_{\max}$, isolating the hardware from execution variance induced by a fluctuating modality mix. As a defensive secondary constraint, the per-step sample count is also capped at $N_{\max}$.

\begin{figure}[t]
    \centering
    \resizebox{\linewidth}{!}{\includegraphics{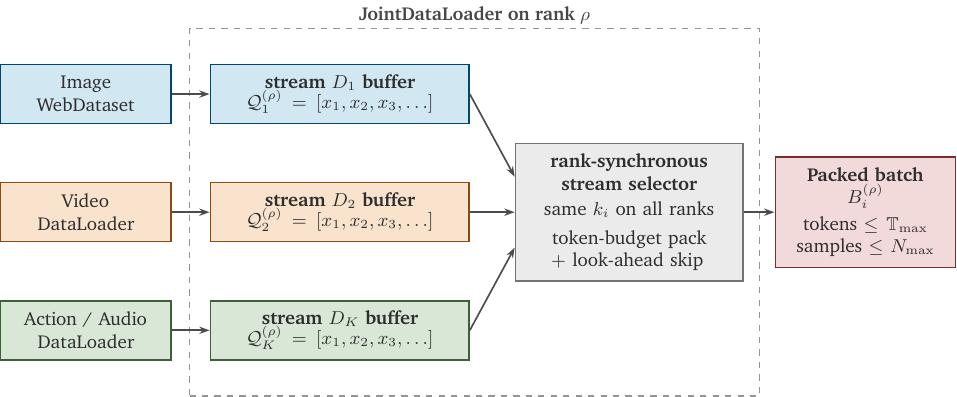}}
    \captionsetup{justification=raggedright, singlelinecheck=false}
    \caption{\textbf{Overview of the Joint Data-Loader.}
    Stream-specific data-loaders feed local per-stream buffers on each rank.
    At each global iteration, a rank-synchronous selector chooses the same
    stream $k_i$ across distributed ranks. Each rank then greedily packs
    samples from its selected local buffer into $B_i^{(\rho)}$ under token
    and sample-count budgets, using bounded look-ahead to reduce unused
    token capacity.}
    \label{fig:joint_dataloader}
\end{figure}

\paragraph{Joint data loader.}
Each modality, dataset, or finer-grained data stream is encapsulated in its own loader with a private prefetch buffer that hides storage latency. A \emph{joint data loader} (illustrated in~\cref{fig:joint_dataloader}) multiplexes across these per-stream loaders to assemble a single unified training batch per step, while fully preserving per-stream buffering, prefetching, and observability.

\paragraph{Rank-synchronous stream selection.}
Foundation-model training typically draws from datasets containing both images and videos at multiple spatial resolutions. Per-sample token counts differ by orders of magnitude across these streams. For example, video samples often carry over $100\times$ more tokens than images, and 720p videos over $10\times$ more than 256p videos. Allowing each rank to choose its stream independently would induce severe workload imbalance under FSDP, with widely divergent attention FLOPs (due to its quadratic complexity) across ranks within a single step. We mitigate this by selecting the active stream via a globally seeded selector keyed on the iteration index, ensuring that all ranks process samples drawn from the same modality and resolution bucket at every step. This eliminates cross-rank variance in compute time and activation memory, while the deterministic, seed-derived selection sequence is bit-exactly reproducible across checkpoints and restarts. Rank-synchronous stream selection improves end-to-end training throughput by $54\%$ over the unsynchronized baseline.

\paragraph{Look-ahead packing.}

Given the stream $k_i$ selected for iteration $i$, the Joint Data-Loader constructs the local batch greedily, appending samples from the head of the stream's buffer until the next candidate would exceed the token budget $\mathbb{T}_{\max}$. Pure greedy packing, however, can leave a non-trivial fraction of the budget unused whenever the next candidate is large enough to overflow but smaller candidates remain available deeper in the buffer---residual capacity that is functionally equivalent to padding and directly proportional to lost throughput. We address this with a bounded look-ahead policy. When a candidate sample exceeds the \emph{total} token budget $\mathbb{T}_{\max}$, the sample is unpackable under the current configuration and is dropped (with a logged warning). When a candidate merely exceeds the \emph{remaining} budget but the batch is already non-empty, the candidate is moved temporarily into a \emph{look-aside buffer}, and the loader continues scanning further into the stream buffer for a smaller sample that fits the residual capacity.

The mechanism is illustrated in \cref{fig:lookahead_dataloader}. In the example, samples $a_1$ and $a_2$ fit into the current batch; $a_3$ exceeds the remaining budget and is diverted to the look-aside buffer; the loader then continues scanning and packs subsequent smaller samples such as $a_4$ and $a_6$. At the end of the iteration, all samples remaining in the look-aside buffer are restored to the head of the stream buffer in their original arrival order, so that look-ahead reduces padding without permanently reordering the stream. To bound the cost of pathological cases---e.g., a stream temporarily dominated by oversized samples---the number of consecutive look-ahead attempts per iteration is capped by a configurable per-stream limit. In production, we use a cap of ten; beyond this value, we observe negligible additional reduction in unused capacity, while the size of the look-aside buffer (and its memory footprint) continues to grow. Overall, look-ahead packing increases the effective sequence length by $8\%$ over the baseline, yielding a corresponding improvement in training throughput.

\definecolor{c3OceanBlue}{HTML}{0077B6}%
\colorlet  {c3Blue}   {blue}%
\colorlet  {c3Violet} {violet}%
\colorlet  {c3Orange} {orange}%
\colorlet  {c3Rust}   {orange!60!brown}
\colorlet  {c3Sage}   {teal!70!green!70!yellow!50!gray}
%
\providecommand{\cThreeVariants}[1]{%
  \colorlet{c3#1Pale} {c3#1!4!white}%
  \colorlet{c3#1Light}{c3#1!8}%
  \colorlet{c3#1Fill} {c3#1!12}%
  \colorlet{c3#1Tok}  {c3#1!22}%
  \colorlet{c3#1Soft} {c3#1!50!gray}%
  \colorlet{c3#1Dark} {c3#1!50!black}%
}%
\cThreeVariants{Blue}%
\cThreeVariants{OceanBlue}%
\cThreeVariants{Violet}%
\cThreeVariants{Orange}%
\cThreeVariants{Rust}%
\cThreeVariants{Sage}%
\colorlet  {c3SkyFill}      {c3OceanBlue!18}%
\colorlet  {c3SkyDark}      {c3OceanBlue!60!black}%
\colorlet  {c3LavenderFill} {c3Blue!12}%
\colorlet  {c3LavenderDark} {c3Blue!45!black}%
\colorlet  {c3MintFill}     {c3Sage!25}%
\colorlet  {c3MintDark}     {c3Sage!60!black}%
\colorlet  {c3LilacFill}    {c3Violet!22}%
\colorlet  {c3LilacDark}    {c3Violet!60!black}%
\colorlet  {c3PeachFill}    {c3Orange!12}%
\colorlet  {c3PeachDark}    {c3Orange!78!black}%
\colorlet  {c3BirchFill}    {c3Orange!4!white}%
\colorlet  {c3BirchDark}    {c3Rust!70!black}%
\colorlet  {c3GrayFill}     {gray!16}%
\colorlet  {c3GrayDark}     {black!70}%
\definecolor{c3RoseFill}{HTML}{F2DADA}%
\definecolor{c3RoseDark}{HTML}{7A2E2E}%
%
\providecommand{\cPale}  [1]{#1!4!white}%
\providecommand{\cLight} [1]{#1!8}%
\providecommand{\cFill}  [1]{#1!12}%
\providecommand{\cTok}   [1]{#1!22}%
\providecommand{\cSoft}  [1]{#1!50!gray}%
\providecommand{\cBorder}[1]{#1!50!black}%
\definecolor{c3Coral} {HTML}{C95B5B}%
\colorlet  {c3Gold}   {c3Orange!55!yellow}%
\colorlet  {c3Olive}  {c3Sage!50!yellow}%
\colorlet  {c3Teal}   {c3OceanBlue!55!c3Sage}%
\colorlet  {c3Indigo} {c3Blue!65!c3Violet}%
\definecolor{c3Berry} {HTML}{B04060}%
\definecolor{c3Cocoa} {HTML}{8C5523}%
\definecolor{c3CatCoral} {HTML}{C36868}%
\colorlet  {c3CatRust}   {c3Rust!82!gray!97}%
\colorlet  {c3CatGold}   {c3Gold!78!gray!95}%
\colorlet  {c3CatOlive}  {c3Olive!85!gray!97}%
\colorlet  {c3CatSage}   {c3Sage!92!gray!97}%
\colorlet  {c3CatTeal}   {c3Teal!82!gray!95}%
\colorlet  {c3CatBlue}   {c3OceanBlue!80!gray!95}%
\colorlet  {c3CatIndigo} {c3Indigo!50!gray!92}%
\definecolor{c3CatViolet}{HTML}{9C73A2}%
\colorlet  {c3CatBerry}  {c3Berry!85!gray!95}%
\colorlet  {c3CatCocoa}  {c3Cocoa!90!gray!97}%

\newcommand{\packedbox}[1]{%
  \tikz[baseline=(b.base)]{%
    \node[
      draw=c3MintDark,
      fill=c3MintFill,
      text=c3MintDark,
      rounded corners=1.5pt,
      line width=0.7pt,
      minimum width=0.85cm,
      minimum height=0.48cm,
      inner xsep=4pt,
      inner ysep=2pt
    ] (b) {$\mathtt{x}_{#1}^{*}$};%
  }%
}

\newcommand{\skippedbox}[1]{%
  \tikz[baseline=(b.base)]{%
    \node[
      draw=c3RoseDark,
      fill=c3RoseFill,
      text=c3RoseDark,
      rounded corners=1.5pt,
      line width=0.7pt,
      minimum width=0.85cm,
      minimum height=0.48cm,
      inner xsep=4pt,
      inner ysep=2pt
    ] (b) {$\mathtt{x}_{#1}$};%
  }%
}

\newcommand{\boxgap}{\hspace{0.08cm}}

\begin{figure}[t]
\centering
\begin{adjustbox}{max width=0.98\linewidth}
\begin{tabular}{@{}p{4.4cm}p{6.9cm}p{4.1cm}p{7.0cm}@{}}
\textbf{step} &
\textbf{packed (in output\_batch)} &
\textbf{skipped (set aside)} &
\textbf{state} \\
\midrule

\texttt{fetch x1 (fits)}
& \packedbox{1}
&
& $\mathrm{cur}=\tau(x_1)$ \\[0.65em]

\texttt{fetch x2 (fits)}
& \packedbox{1}\boxgap\packedbox{2}
&
& $\mathrm{cur}=\tau(x_1)+\tau(x_2)$ \\[0.65em]

\texttt{fetch x3 (overflow)}
& \packedbox{1}\boxgap\packedbox{2}
& \skippedbox{3}
& $\mathrm{skipped}\leftarrow\{x_3\},\quad \mathrm{lookahead}=1$ \\[0.65em]

\texttt{fetch x4 (fits)}
& \packedbox{1}\boxgap\packedbox{2}\boxgap\packedbox{4}
& \skippedbox{3}
& $\mathrm{cur}\mathrel{+}= \tau(x_4)$ \\[0.65em]

\texttt{fetch x5 (overflow)}
& \packedbox{1}\boxgap\packedbox{2}\boxgap\packedbox{4}
& \skippedbox{3}\boxgap\skippedbox{5}
& $\mathrm{lookahead}=2$ \\[0.65em]

\texttt{fetch x6 (fits, last)}
& \packedbox{1}\boxgap\packedbox{2}\boxgap\packedbox{4}\boxgap\packedbox{6}
& \skippedbox{3}\boxgap\skippedbox{5}
& \texttt{batch done} \\[0.65em]

\texttt{end of iter t}
& \packedbox{1}\boxgap\packedbox{2}\boxgap\packedbox{4}\boxgap\packedbox{6}
& \skippedbox{3}\boxgap\skippedbox{5}
& \texttt{push reversed(skipped)} $\rightarrow$
  \texttt{head: [x3, x5, x7, x8, ...]} \\[0.25em]

\multicolumn{4}{@{}l}{%
  \textcolor{c3MintDark}{\footnotesize $^{*}$ fits in $\mathbb{T}_{\max}$ budget}
}
\end{tabular}
\end{adjustbox}
\captionsetup{justification=raggedright, singlelinecheck=false}
\caption{\textbf{Look-ahead packing in the JointDataLoader.}
The loader greedily scans samples from the selected stream and packs those that fit within the remaining token budget into the current mini-batch (Mint). Samples that exceed the budget are temporarily set aside in a lookaside buffer (Rose), allowing later smaller samples to fill the remaining capacity. At the end of the iteration, skipped samples are returned to the head of the stream buffer in their original arrival order, reducing padding while preserving stream order across iterations.}
\label{fig:lookahead_dataloader}
\end{figure}

\paragraph{Cold-start handling.}
Several of our data streams incur substantial first-batch latency, dominated by worker-process spawning, filesystem metadata caching for newly opened shards, and stream-specific deserialization warm-up. If this latency were paid at the first training step, it would race against the NCCL collective on that step, with a high probability of triggering a watchdog timeout on the slowest rank---particularly at scale, where the maximum over ranks is the relevant statistic. To eliminate this failure mode, the Joint Data-Loader performs an explicit pre-warm stage during construction: it fetches one batch from every stream so that worker pools, file handles, and deserialization caches are fully primed, and then issues a distributed barrier before returning control to the training loop. This guarantees that every rank has paid the cold-start cost before the first forward pass and that the first iteration runs against fully warmed streams.

\paragraph{Observability and integration.}
At every iteration, the Joint Data-Loader emits a structured record of packing statistics to a training-side monitoring callback, which aggregates the per-rank records across the distributed group and logs the result to Weights \& Biases. The reported metrics include the empirical per-stream sampling ratio (compared against the configured target mixture), the number of samples packed per iteration, per-stream buffer occupancy and wait time, the look-ahead saturation rate, and per-iteration token-budget utilization. These metrics expose the failure modes most likely to degrade large-scale training without crashing it so that such issues are surfaced in the training dashboard within minutes of onset rather than discovered later from model behavior.

\subsubsection{Attention Implementation}

As discussed in \cref{sec:cosmos3_mot}, the Mixture-of-Transformers architecture in Cosmos 3 imposes two distinct attention requirements that must coexist within a single forward pass: the Reasoner pathway employs causal attention over the Reasoner tokens only, while the Generator pathway employs bidirectional attention over the concatenation of the Reasoner and the Generator tokens, so that each Generator token can condition on the full context. Naively expressing these heterogeneous masking patterns with general-purpose operators such as FlexAttention produces correct results but underutilizes the hardware: the masking structure is opaque to the kernel, and padding-equivalent work is performed inside otherwise-skipped attention blocks. This degrades tensor-core utilization and inflates memory-bandwidth pressure.

\begin{figure*}[t]
  \centering
  \resizebox{\linewidth}{!}{\includegraphics{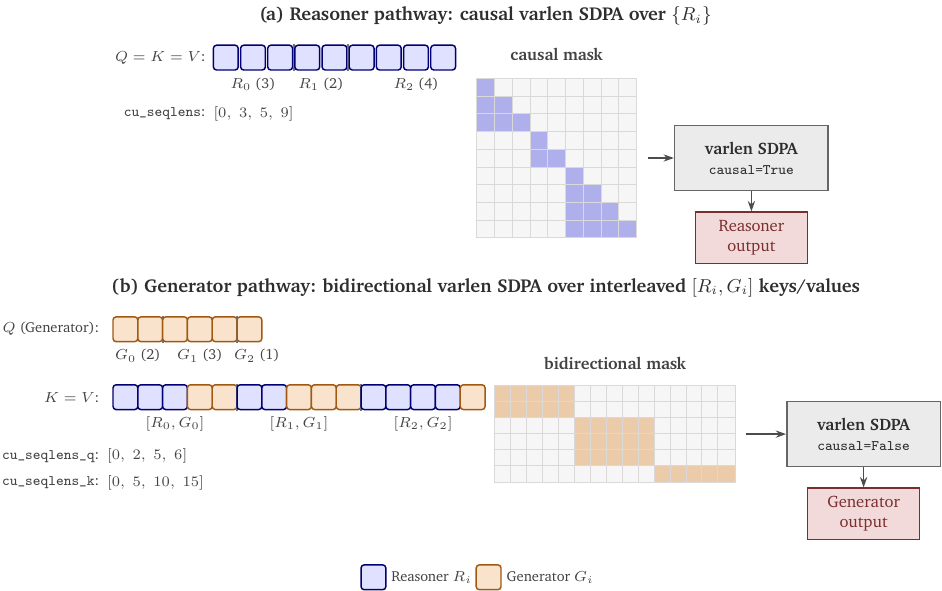}}
  \captionsetup{justification=raggedright, singlelinecheck=false}
  \caption{\textbf{Two-way flat attention.} Each pathway is implemented as a single
  variable-length SDPA call. \textbf{(a)}~The Reasoner pathway uses a standard causal
  \texttt{varlen} call on the packed Reasoner tokens, producing a block-diagonal
  causal mask. \textbf{(b)}~The Generator pathway packs Generator queries separately
  from the interleaved key/value stream $[R_0, G_0, R_1, G_1, \ldots, R_n, G_n]$. The
  resulting mask is block-diagonal but rectangular within each block, so that each
  Generator query attends bidirectionally over its own sample's $[R_i, G_i]$ context
  without crossing sample boundaries. The example uses three packed samples with
  $(|R_i|, |G_i|) = (3, 2),\,(2, 3),\,(4, 1)$.}
  \label{fig:two_way_attention}
  \vspace{-0.5\baselineskip}
\end{figure*}

To address this, we co-designed a custom two-way flat attention mechanism that exposes the cross-pathway masking structure directly to a high-performance variable-length attention kernel. \cref{fig:two_way_attention} illustrates the design. The computation is decomposed into two separate kernel invocations. The first handles the Reasoner pathway and is a standard variable-length scaled dot-product attention (SDPA)~\citep{pytorch_varlen_attn} call with a causal mask, operating only on the Reasoner queries, keys, and values. The second handles the Generator pathway and requires the Reasoner and Generator key/value streams to be visible to each Generator query within the same sample, but strictly separated across samples in a packed batch. We achieve this by flattening and interleaving the two token streams at the sample granularity in the order
\[
[R_0, G_0, R_1, G_1, \ldots, R_n, G_n]
\]
where $R_i$ and $G_i$ denote the Reasoner and Generator key/value tokens of sample $i$, respectively. Each Generator query attends bidirectionally over its own sample's $[R_i, G_i]$ block. This formulation expresses the full cross-pathway attention with two variable-length kernel launches per layer, supports both causal and bidirectional masking within one packed representation, eliminates the padding overhead inherent to fixed-length implementations, and yields $22\%$ improvement in end-to-end training throughput compared to a FlexAttention-based baseline for the Cosmos3-Nano model.

The variable-length attention backend is selected per platform to match the most performant and numerically validated implementation available on the target hardware. On Hopper-class GPUs (H100, H200), we use FlashAttention-3~\citep{flash3}, which exploits the WGMMA instructions and TMA-based asynchronous data movement of the Hopper architecture to deliver near-peak attention throughput. On Blackwell-class GPUs (GB200), we use NATTEN~\citep{natten}, whose variable-length kernels are built on the CUTLASS template library and are specifically tuned for the fifth-generation tensor cores and updated memory hierarchy of Blackwell (SM100/SM103). Both backends are accessed through a common dispatch interface, so the choice of kernel is transparent to the rest of the training stack and can be revisited as new backends mature.

\subsubsection{Distributed Training}

Cosmos 3 Reasoner and Generator are trained separately with a distributed-training stack that combines Hybrid Sharded Data Parallelism (HSDP) with Context Parallelism (CP). HSDP shards model parameters, gradients, and optimizer states within each replica group while replicating across groups, which trades a modest amount of intra-group communication for the memory headroom required to train multi-billion-parameter models on commodity per-GPU memory budgets. CP, in contrast, addresses a different bottleneck: per-sequence activation memory, which scales linearly with context length and would otherwise force a hard upper bound on the trainable sequence size. The two strategies compose orthogonally, and the (HSDP-degree, CP-degree) configuration is chosen per experiment to fit the target model size, sequence length, and cluster topology.

\paragraph{Context parallelism via the Ulysses scheme.}

For CP, we adopt the Ulysses scheme~\citep{deepspeed_ulysses}, which partitions the input sequence along the token dimension across CP-rank devices outside of attention and employs two all-to-all collectives per attention layer to transition between sharding axes. The first collective redistributes the Q/K/V activations from the sequence dimension to the attention-head dimension, so that each rank holds the complete sequence for a disjoint subset of heads and can execute attention locally without further cross-rank communication; the second collective restores the original sequence-sharded layout on the attention output.

The scheme integrates cleanly with sequence packing (described in~\cref{sec::data-loader}) and the two-way attention mechanism. The flattening and interleaving of the Reasoner and Generator key/value streams for the bidirectional attention operation is deferred until after the head-axis redistribution, at which point each rank already holds the full sequence for its assigned heads and can perform the concatenation locally. The same variable-length attention kernels are therefore reused unchanged inside CP, with no need for CP-specific kernel variants. The maximum CP degree supported by this implementation is bounded by the number of query heads in the model---32 for Cosmos3-Nano and 64 for Cosmos3-Super---which we find to be a non-restrictive limit in practice given the context lengths and per-GPU memory budgets targeted by Cosmos 3.

\paragraph{Why not ring attention?}

We considered implementing CP via ring attention as an alternative, but found it substantially less attractive in our setting. Ring attention would require materializing a single packed sequence containing the interleaved Reasoner and Generator tokens before sharding it across CP ranks, in order to expose a contiguous K/V stream to the ring schedule. This precludes the independent sharding of the two pathways that Ulysses naturally permits, and complicates the construction of the per-sample bidirectional/causal masks under the rotating ring schedule. Combined with the favorable bandwidth profile of all-to-all on NVLink-connected nodes, these factors led us to adopt Ulysses as the CP strategy for Cosmos 3.

\subsubsection{Selective Activation Checkpointing}

Computing the backward pass of a transformer requires the intermediate activations produced during the forward pass to be available, but materializing all of them simultaneously in GPU memory is prohibitive at the model sizes and context lengths targeted by Cosmos 3. The standard mitigation is activation checkpointing~\citep{sac}: the forward pass stores only a sparse set of ``anchor'' activations and discards the rest, and the discarded tensors are recomputed during the backward pass by re-running the corresponding forward subgraph from the nearest saved anchor. The default policy stores only the inputs of each transformer block and recomputes everything inside the block on demand; this minimizes activation memory but introduces an additional forward pass during backward, inflating per-step FLOPs by roughly $33\%$ and reducing end-to-end training throughput accordingly.

To reduce this recomputation overhead while staying within the activation-memory budget, we apply Selective Activation Checkpointing (SAC), in which a curated subset of intermediate tensors is additionally retained in memory rather than recomputed. The selection is guided by a simple cost-benefit heuristic: rank candidate operations by their FLOPs-to-memory ratio---i.e., the recomputation cost saved per byte of activation memory committed---and materialize those with the highest ratio first, until the activation-memory budget is exhausted. For Cosmos 3, attention outputs are by far the dominant beneficiary of this policy. Attention recomputation is expensive because its cost scales quadratically with sequence length, yet the attention output tensor itself is comparatively small (linear in sequence length and hidden size), making it the operation with the highest FLOPs-to-memory ratio. Users can additionally configure custom save sets through regular-expression patterns over operation names, which we use to retain a small number of secondary tensors when the residual activation-memory headroom permits.

In our measurements, applying SAC with attention outputs materialized yields a $13\%$ improvement in end-to-end training throughput for Cosmos3-Nano at a per-batch token budget of $74{,}000$ tokens, with no change in numerical results.

\subsubsection{Torch Compile for Transformer Blocks}

We apply \texttt{torch.compile} with \texttt{fullgraph=True} and \texttt{dynamic=True} across the training graph. The \texttt{fullgraph} mode eliminates CPU overheads and enables operator fusion, while \texttt{dynamic=True} handles the variable sequence lengths arising from mixed-modality batches---where, for example, Generator pathway tokens are substantially longer than those of image batches across iterations. Torch compile improves training throughput by $41\%$ for Cosmos3-Nano Generator training.

\subsubsection{Video Tokenizer}

Cosmos 3 training requires decoded video frames to be tokenized on-the-fly into latent representations by a video VAE: Wan2.2~\citep{wan2pt2} in our configurations. In the initial implementation, we observed that the tokenizer occupied a disproportionate fraction of each training step---dominating the forward pass for the smaller Cosmos3-Edge and Cosmos3-Nano models. In such models, the transformer compute is small enough that the VAE is not amortized by the rest of the step. Because the tokenizer sits on the critical path between the data loader and the training loop, any latency it introduces directly degrades training throughput. We therefore implemented a series of targeted optimizations that, together, reduce the tokenizer's wall-clock contribution significantly.

\paragraph{Chunked encoding.}

The Wan2.2 causal tokenizer encodes a 1-frame ``prime'' chunk followed by groups of 4 pixel frames per latent chunk; the default per-call granularity is one latent chunk (\ie, 4 frames after the prime). This default leaves the GPU substantially under-utilized for the spatial resolutions used in training, because each kernel launch operates on too little work to saturate the tensor cores. We instead invoke the encoder on a configurable number of pixel frames per call, trading additional activation memory for higher arithmetic intensity per launch. The optimal chunk size is resolution-dependent. At higher spatial resolutions, each frame already consumes substantial memory, so fewer frames per call are admissible before triggering OOMs, whereas at lower resolutions, much larger chunks are feasible and beneficial. We empirically determined the following operating points on our training hardware: 68 frames for 256p, 24 frames for 480p, and 12 frames for 720p. These configurations push the encoder onto the compute-bound side of the roofline curve while staying well under per-GPU memory budgets, yielding the bulk of the per-step speedup.

\paragraph{Ahead-of-time compilation.}

\begin{figure}[t]
\centering
\includegraphics[width=0.58719\linewidth]{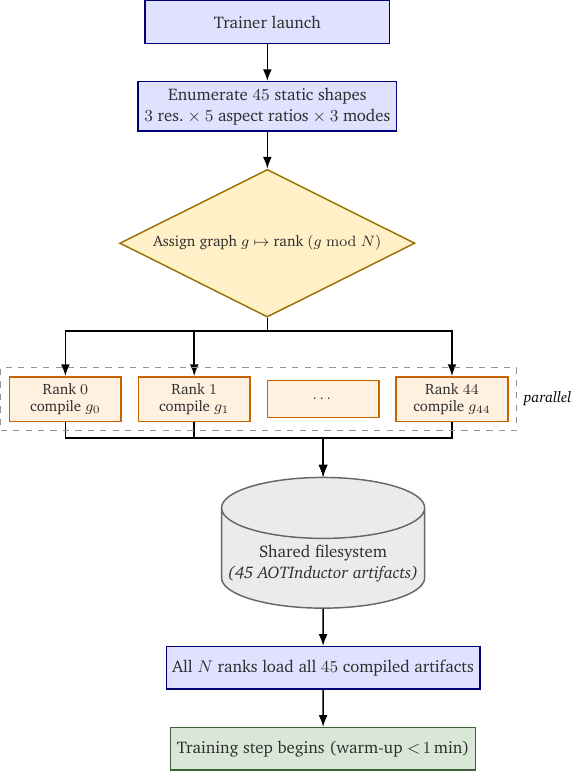}
\captionsetup{justification=raggedright, singlelinecheck=false}
\caption{\textbf{Sharded AOT compilation of the Wan2.2 tokenizer.} The $45$ static-shape graphs
arising from $\{3\text{ resolutions}\} \times \{5\text{ aspect ratios}\} \times
\{3\text{ tokenizer call modes}\}$ are partitioned across ranks; each
rank performs compilation on its assigned graph(s), writes the compiled
artifact to a shared filesystem, and loads the full set of artifacts
before training begins. Warm-up time drops from $\sim\!15$\,min (serial) to
$<\!1$\,min (sharded).}
\label{fig:aoti_sharded_compile}
\vspace{-0.5\baselineskip}
\end{figure}

On top of chunked encoding, we use \texttt{torch.compile} on the tokenizer, which delivers an additional $52\%$ reduction in encode latency by fusing pointwise operations and selecting optimized kernel schedules for the encoder's convolution and attention blocks. Maximizing throughput requires static input shapes, so the encoder is compiled separately for each shape it may be invoked with. Cosmos 3 training spans three spatial resolutions ($256\text{p}$, $480\text{p}$, $720\text{p}$) and five aspect ratios per resolution. Within each (resolution, aspect-ratio) combination, the causal tokenizer is called in two modes: prime-chunk encoding and chunked encoding with cache sizes of $1$ and $2$. This yields $3 \times 5 \times 3 = 45$ distinct graphs that must be compiled before training can begin. Compiling all $45$ graphs serially on every rank inflated trainer startup by roughly $15$ minutes.

To eliminate this overhead, we shard the compilation across data-parallel ranks using AOTInductor~\citep{pytorch_aot_inductor} as shown in \cref{fig:aoti_sharded_compile}, which performs ahead-of-time compilation and serializes the resulting kernels and host code to disk. With at least $45$ ranks (which is satisfied by all of our training configurations), each rank compiles exactly one graph; the compiled artifacts are written to a shared filesystem, after which every rank loads the full set of $45$ graphs from disk. This reduces the warm-up overhead to under one minute. To accommodate videos with arbitrary frame counts under static-shape compilation, each input clip is right-padded to the next multiple of the configured encode-chunk size prior to encoding, and the resulting latent tensor is cropped along the temporal axis to the exact expected sequence length before being consumed by the model.

\paragraph{Specialization to known frame counts.}

For datasets in which the per-clip frame count is fixed and known a priori---for example, robot action datasets, where every episode contributes a clip of identical length---we specialize the compilation to the exact tensor shapes that arise at runtime, bypassing the padding-and-crop fallback used in the general case. This eliminates the padded-tail compute, removes the corresponding latent-cropping step, and yields a small but consistent additional throughput improvement on such datasets.

\subsubsection{Checkpointing}

To eliminate save-induced stalls, checkpointing is fully overlapped with training. Following Torchtitan~\citep{torchtitan}, checkpoint writes are routed through a dedicated Gloo process group rather than the NCCL communicator carrying training collectives, isolating I/O traffic from GPU-side communication. This asynchronous design hides the highly variable object-store write latencies and removes nearly all save-time overhead, at the cost of a modest increase in host memory. \cref{tab:cosmos3_checkpointing} quantifies the resulting benefit: relative to synchronous checkpointing at a 30-minute interval, asynchronous checkpointing reduces end-to-end training time by $4\%$ for Cosmos3-Nano and $9\%$ for Cosmos3-Super.

\begin{table}[t]
    \centering
    \caption{\textbf{Benefits of asynchronous checkpointing.} Compared with synchronous checkpointing at a 30-minute interval, asynchronous checkpointing reduces end-to-end training time by $4\%$ and $9\%$ for \textbf{Cosmos3-Nano} and \textbf{Cosmos3-Super}, respectively. The larger savings on \textbf{Cosmos3-Super} reflect its longer checkpoint save times.}
    \label{tab:cosmos3_checkpointing}
    \setlength{\tabcolsep}{6pt}
    \small
    \begin{tabular}{l|rcr|c}
        \toprule
        \multirow{2}{*}{\textbf{Model}}
            & \multicolumn{3}{c|}{\textbf{Checkpoint Save Time (s)}}
            & \multirow{2}{*}{\textbf{Speedup over Synchronous}} \\
        \cmidrule(lr){2-4}
            & \textbf{Mean} & \textbf{Min} & \textbf{Max} & \\
        \midrule
        \textbf{Cosmos3-Nano}  &  72 & 43 & 250 & $4\%$ \\
        \textbf{Cosmos3-Super} & 167 & 40 & 736 & $9\%$ \\
        \bottomrule
    \end{tabular}
\end{table}

\paragraph{Asynchronous save mechanism.} At construction time, the checkpointer launches a long-lived child process using the \texttt{spawn} start method and communicates with it through multiprocessing queues. The child process joins a Gloo process group and performs CPU-side reductions needed to construct the save plan, leaving GPUs available for training. It then blocks on an inbound queue until it receives either a checkpoint save request or a termination sentinel.

\paragraph{Save plan memorization.} To further reduce overheads, checkpoint save plans are computed during the first checkpoint saving operation, and reused for subsequent saves. This is possible because the save plan is a deterministic function of the state-dict topology. Reusing the plan avoids repeated metadata communication across ranks and reduces checkpointing overhead by approximately \textbf{60\%}, further decreasing the likelihood that asynchronous checkpointing becomes a training bottleneck.

\paragraph{Optimizing for object storage.} During checkpoint saving, each rank writes its local shard of the state dictionary to object storage. Replicated tensors, which may be present on multiple ranks, are deduplicated before writing. In the default round-robin assignment, replicated tensors may be written by different ranks, requiring each rank to read all checkpoint files during loading in order to recover the replicated state. This introduces significant overhead because full files must be loaded even when only a subset of their contents is required. To reduce this overhead, we set \texttt{dedup\_to\_lowest\_rank = True}, which stores replicated tensors only on the lowest-numbered rank in the corresponding submesh, typically rank~0. During loading, each rank then reads only its own shard and the rank-0 shard. This substantially reduces checkpoint load time, particularly for optimizer state dictionaries, which contain many small replicated tensors.

\paragraph{Random state restoration.} The trainer restores random number generator (RNG) state in a rank-aware manner. Since RNG state is keyed by rank, a resumed job first checks the checkpoint metadata for the key corresponding to its own rank and requests that state only if it is present. This preserves compatibility with older checkpoints that predate the rank-keyed RNG format. If the rank-specific key is absent, the rank retains its current RNG state.

\subsubsection{Throughput Summary}
\label{sec:training_throughput_summary}

\cref{tab:training_throughput} reports steady-state, per-GPU training throughput for the Cosmos 3 dense configurations measured on NVIDIA GB200 systems. Although Cosmos 3 supports multiple training modes across heterogeneous modalities and tasks, these measurements were obtained using a joint text-to-image and text-to-video training configuration to enable a standardized throughput comparison. \cref{fig:multiresolution_seqpacking} illustrates the pre-training data mixture used. Cosmos3-Nano was benchmarked using 1024 NVIDIA GB200 GPUs, while Cosmos3-Super was benchmarked using 2048 NVIDIA GB200 GPUs.

The Nano model achieves the highest raw token throughput, processing 507 iterations per hour and reaching 4.56M image tokens and 16.23M video tokens per GPU-hour. In contrast, the larger Super model performs substantially more computation per iteration, reducing its iteration rate to 185 iterations per hour and its throughput to 1.66M image tokens and 5.91M video tokens per GPU-hour.

Despite its lower token throughput, Cosmos3-Super achieves higher arithmetic utilization, increasing per-GPU throughput from 520 to 673 TFLOPS and improving MFU from 0.23 to 0.30. This reflects the expected trade-off between model scale and training throughput: Cosmos3-Nano is optimized for maximizing token processing throughput, whereas Cosmos3-Super more effectively saturates GPU compute resources through increased model capacity and computation per token.

\begin{table}[t]
    \centering
    \caption{\textbf{Steady-state training throughput for Cosmos 3 dense model configurations.} TFLOPS and MFU are reported per GPU. Image and video token throughput are reported separately in millions of tokens per GPU-hour. The experiments were conducted with NVIDIA GB200 GPUs, where the Nano and Super runs took 2048 and 4096 GPUs, respectively.}
    \label{tab:training_throughput}
    \setlength{\tabcolsep}{5pt}
    \resizebox{0.9\linewidth}{!}{%
    \begin{tabular}{lcccccc}
        \toprule
        \textbf{Model} & \textbf{Iter (s)} & \textbf{TFLOPS} & \textbf{MFU} & \textbf{Iter/hr} & \textbf{Img Tok/hr/GPU (M)} & \textbf{Vid Tok/hr/GPU (M)} \\
        \midrule
        \textbf{Cosmos3-Nano}  &  7.1 & 520 & 0.23 & 507 & 4.56 & 16.23 \\
        \textbf{Cosmos3-Super} & 19.5 & 673 & 0.30 & 185 & 1.66 &  5.91 \\
        \bottomrule
    \end{tabular}
    }
\end{table}

\subsection{Serving Infrastructure}
\label{subsec::serving_infrastructure}

Cosmos 3 is integrated with multiple production-grade serving frameworks to support a broad range of deployment scenarios. Reasoner is supported by TensorRT-LLM~\citep{trtllm} and vLLM~\citep{vllm}, both of which provide highly optimized autoregressive decoding through paged KV-cache management, continuous batching, and fused attention kernels. Generator inference is supported by vLLM-Omni (multimodal extension of vLLM for diffusion-based generation)~\citep{vllm_omni}, which provides complementary trade-offs between peak throughput and multi-tenant scheduling efficiency. In addition to these production backends, we provide a reference implementation in native PyTorch that prioritizes readability and modifiability, serving as both a faithful specification of the inference algorithm and a starting point for downstream adaptation, research extensions, and integration into custom application pipelines.

\subsubsection{Plain PyTorch}

The plain PyTorch serving path executes the model and the surrounding inference procedure directly in eager-mode PyTorch---without dependence on specialized serving runtimes---and is designed to mirror the training-time computation as faithfully as possible. This path is responsible for the complete end-to-end inference workflow, comprising the stages described below:

\begin{itemize}

\item \textit{Input preparation.} Parses the text prompt, loads any image, video, or action conditioning, constructs the modality-specific conditioning dictionaries consumed by the model, and assembles the corresponding input tensors and metadata.
\item \textit{Autoregressive loop.} In the Cosmos 3 Reasoner, output tokens are produced autoregressively, with each step conditioned on the previously generated tokens and on the cached key/value states of the conditioning context.
\item \textit{Diffusion loop.} In the Cosmos 3 Generator, the PyTorch path constructs the timestep schedule, invokes the denoiser at each step, applies classifier-free guidance (CFG), updates the latent state according to the sampler, and manages the request-level control flow around the denoising process.
\item \textit{Decoding and post-processing.} Once denoising completes, the resulting latent representation is decoded into the target modality---text, image, video, audio, or action---post-processed as required, and returned to the caller through the serving interface.

\end{itemize}

Because the native PyTorch backend preserves the model's original PyTorch structure, it serves as the primary target for landing new model features, sampler modifications, KV-cache and activation-cache policies, and debugging instrumentation. New capabilities are validated in this backend first and only subsequently ported to the production runtimes (TensorRT-LLM and vLLM), ensuring that the reference implementation remains the authoritative specification of the inference algorithm. The PyTorch backend exposes the following features and optimizations.

\paragraph{Torch compile with CUDA graphs.}

Since the PyTorch native inference path keeps request orchestration and sampling logic in Python, a major serving bottleneck is host-side kernel launch overhead during repeated denoising steps. We therefore optimize the PyTorch path using torch.compile and CUDA graph replay. CUDA graph optimization is implemented at transformer-layer granularity, capturing repeated transformer block executions. Each block is compiled with torch.compile in \texttt{reduce-overhead} mode, which allows PyTorch Inductor to lower the block and use CUDA graph replay when the block is invoked with compatible tensor shapes and memory layouts. The outer inference loop remains in ordinary PyTorch, \ie, prompt handling, timestep scheduling, sampler updates, CFG orchestration, decoding are not part of the captured graph. The graph contains the numerically heavy and frequently repeated per-layer computation, while dynamic serving logic remains outside the graph. The benefit of CUDA graphs is most visible for T2I-style generation, where shorter generation workloads and smaller kernels make CPU launch overhead a larger fraction of end-to-end latency. CUDA Graphs on T2I generation yielded 30\% to 60\% speedups on different hardware backends.

\paragraph{Distributed inference.}

We employ context parallelism (CP) at inference time to support generations exceeding per-GPU memory capacity constraints. We retain the Ulysses scheme~\citep{deepspeed_ulysses} used during training, ensuring consistency between the two regimes. Beyond enabling long-context inference, CP also serves as a latency-reduction mechanism by distributing the forward pass across multiple GPUs; this benefit is realized even in regimes where a single device's memory is sufficient to hold the full context, making CP a general-purpose tool for accelerating inference.

In addition to context parallelism, we exploit classifier-free guidance (CFG) parallelism to further reduce end-to-end inference latency. Diffusion sampling with CFG requires, at every denoising step, two forward passes through the model---one conditioned on the input prompt and one unconditional---whose noise predictions are linearly combined to produce the guided update. Because the conditional and unconditional passes operate on independent inputs and only need to synchronize once per step to form the guided prediction, they are an ideal target for parallelization across two GPUs. In practice, we dispatch the conditional and unconditional batches concurrently, and perform a single lightweight point-to-point exchange to combine the two predictions before advancing the sampler. This nearly halves the per-step latency, and composes cleanly with context parallelism to deliver multiplicative latency reductions on multi-GPU nodes.

\paragraph{Reasoner tower caching.}

For tasks such as text-to-image (T2I), text-to-video (T2V), image-to-video (I2V), and video-to-video (V2V) generation, the conditioning inputs to the Reasoner tower---text prompts and, where applicable, conditioning images or videos---are fixed for the duration of the sampling trajectory. As a result, the Reasoner's outputs are invariant across diffusion steps and depend only on the conditioning, not on the current noise level or partially denoised sample. We exploit this property by computing the Reasoner forward pass once at the start of inference and caching its outputs for reuse across all subsequent denoising steps. Because the cached activations are mathematically identical to those that would be recomputed at each step, this optimization yields a substantial reduction in per-step latency without any impact on generation quality.

\paragraph{Batching.}

Inference throughput can be further improved by batching multiple samples into a single forward pass, amortizing per-step overheads (kernel launches, weight reads, and collective communication) across a larger volume of useful work. Our inference batcher reuses the variable-length sequence-packing mechanism developed for training: rather than padding shorter sequences to a common length---which wastes both compute and memory on padding tokens---it concatenates samples of heterogeneous shapes into a single packed tensor and supplies the corresponding cumulative sequence-length metadata to attention and other shape-sensitive operators. The user specifies one of two budgeting modes: a total token budget (the maximum number of tokens allowed within a batch) or a fixed sample count. Given the chosen budget, the batcher greedily packs incoming samples until the budget is exhausted while respecting any per-device memory constraints. This design improves GPU utilization in throughput-oriented deployments such as offline corpus generation and large-scale evaluation, but provides no benefit in latency-bound settings (\eg, robotics workloads) where a single sample must be processed in isolation; in those regimes, the batcher is configured with a sample count of one and effectively disabled.

\cref{tab:cosmos3_batching} reports the inference throughput obtained by request batching on the text-to-video (T2V) task with 189-frame outputs. At 256p, batching yields throughput gains of $8\%$ to $55\%$. The benefits diminish at 480p, where each sample already provides sufficient work to saturate the GPU and leaves limited headroom for additional parallelism. At 720p, the $74{,}000$ context window admits only $B{=}1$, precluding any batching speedup.

\begin{table}[t]
    \centering
    \caption{
    \textbf{Inference speedup from batching}. The \textbf{Cosmos3-Nano} and \textbf{Cosmos3-Super} are evaluated on the text-to-video (T2V) task with 189-frame outputs. We report results at 256p and 480p, using maximum admissible batch sizes of $B=6$ and $B=3$, respectively, under the 74k-token context limit. 720p is omitted, as it admits only $B=1$ within the same budget.}
    \label{tab:cosmos3_batching}
    \setlength{\tabcolsep}{4pt}
    \small
    \begin{tabular}{l|cc|cc}
        \toprule
        \textbf{HW Backend} &
        \multicolumn{2}{c|}{\textbf{Cosmos3-Nano}} &
        \multicolumn{2}{c}{\textbf{Cosmos3-Super}} \\
        \cmidrule(lr){2-3}
        \cmidrule(lr){4-5}
        &
        \textbf{T2V-256} & \textbf{T2V-480} &
        \textbf{T2V-256} & \textbf{T2V-480} \\
        \midrule
        H100 80GB &
        $8\%$ & $2\%$ &
        $55\%$ & $5\%$ \\
        GB200 &
        $40\%$ & $2\%$ &
        $9\%$ & $1\%$ \\
        \bottomrule
    \end{tabular}
\end{table}

\paragraph{Generation with prompt upsampling.}

Inference additionally supports a prompt-upsampling mode (see more details in \cref{sec::prompt_upsampling}) in which a short, free-form user prompt is expanded by the Reasoner into a richer, structured JSON description of the desired output (covering, for example, scene composition, subject attributes, camera and motion specifications, lighting, and style). The upsampled description is generated autoregressively by the Reasoner and is then supplied to the Generator as an additional conditioning input, alongside any image, video, or action conditioning provided by the user, to synthesize the final output. This pipeline serves two purposes: it offloads the burden of detailed prompt engineering from the end user to the model itself, consistently improving downstream generation quality, and it exercises the full omnimodal end-to-end across both Reasoner and Generator pathways within a single inference invocation---demonstrating that two pathways can be composed seamlessly to produce high-fidelity multimodal outputs from a minimal user specification.

\paragraph{Throughput vs serving considerations.}

Inference workloads generally optimize for one of two competing objectives: throughput or latency. Batch inference, in which a large set of outputs is generated for a fixed corpus of inputs, is typically optimized for throughput, since end-to-end wall-clock time and aggregate cost are the metrics of interest. In contrast, latency-sensitive applications such as robotics---where actions must be produced from a specific starting context with tight per-step deadlines---prioritize responsiveness, and the relevant metric is time-to-first-token (or time-to-first-action) and per-step latency. The optimizations described above target these regimes differentially: distributed inference (context parallelism and CFG parallelism) primarily reduces latency by parallelizing a single request across multiple devices, whereas batching primarily improves throughput by amortizing per-step overheads across multiple concurrent requests. The remaining optimizations---torch compile, and reasoner-output caching---benefit both regimes simultaneously.

\subsubsection{Inference Frameworks for Reasoner: vLLM and TensorRT-LLM}

Because the Nano and Super Reasoners are built on the Qwen3-VL~\citep{qwen3vl2025} backbone, their integration into vLLM and TensorRT-LLM reuses the upstream Qwen3-VL support already present in both frameworks. This allows us to inherit, out of the box, the optimized attention kernels, paged KV-cache management, continuous batching, and multimodal input handling that the two backends already provided for the Qwen3-VL family, requiring only minimal configuration changes to bind the Cosmos 3 model weights and tokenizer to the existing execution paths.

The Edge Reasoner, in contrast, is built on a custom Nemotron backbone~\citep{nemotron3-family}, which is not natively supported in vLLM. We therefore implemented a dedicated integration that follows the vLLM model-contributor conventions. This integration is structured to be upstreamable, easing future maintenance and enabling community contributions.

\subsubsection{Inference Frameworks for Generator: vLLM-Omni}

Cosmos 3 Generator is integrated into vLLM-Omni to leverage its optimized serving stack for diffusion-based multimodal generation. The integration implements Cosmos 3 Generator as a first-class vLLM-Omni model and supports the full set of Generator modalities, including image, video, audio, and action-conditioned generation. It follows the model-contributor conventions and coding style of the vLLM-Omni framework, making the implementation compatible with the framework's existing scheduling, distributed execution, memory-reduction, and quantization features. 

This integration is particularly important for Generator serving because diffusion-based generation requires many repeated transformer evaluations over large image, video, or multimodal token sequences. The vLLM-Omni backend therefore focuses on reducing per-step latency, lowering peak memory usage, and improving throughput while preserving generation quality. The Cosmos 3 Generator integration supports the following vLLM-Omni features:

\begin{itemize}

    \item \textit{Cache-DiT.} A training-free acceleration method that reuses cached transformer-block outputs across adjacent denoising steps, allowing redundant computation to be skipped with negligible impact on generation quality.

    \item \textit{Ulysses context parallelism.} A context-parallel execution scheme that shards long image and video token sequences across multiple GPUs and uses all-to-all attention communication to reduce per-device memory usage and improve latency.

    \item \textit{CFG-Parallel.} A parallelization strategy for classifier-free guidance that dispatches the conditional and unconditional forward passes to separate GPU ranks. The two predictions are synchronized once per denoising step to form the guided update, reducing the latency of CFG-based sampling.

    \item \textit{HSDP.} A memory-efficient distributed inference mode that shards transformer weights across GPUs using FSDP2 and gathers parameters on demand during the forward pass, reducing peak GPU memory requirements for large Generator models.

    \item \textit{CPU offload.} A layer-wise offloading mechanism that moves model parameters between CPU and GPU memory during inference, trading additional data-transfer overhead for substantially lower peak GPU memory usage.

    \item \textit{VAE-Patch-Parallel.} A parallel VAE execution mode that partitions latent or pixel tensors into spatial tiles and encodes or decodes them across multiple ranks, reducing both per-device memory consumption and VAE latency.

    \item \textit{Quantization.} A dynamic FP8 quantization path that lowers the precision of dominant compute operations to reduce inference latency and peak GPU memory usage while maintaining acceptable generation quality.

\end{itemize}

\begin{figure}[t]
    \centering
    \includegraphics[width=0.98\textwidth]{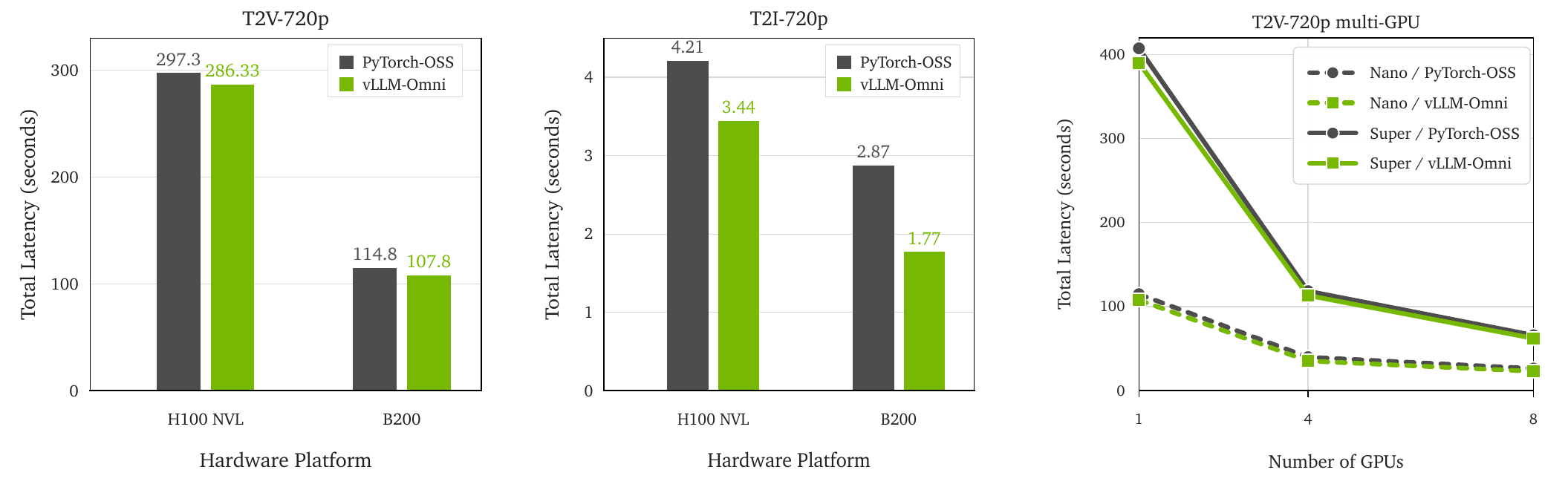}\\[-0.2em]
    \vspace{0.5em}
    \makebox[0.98\textwidth]{\makebox[0.90\textwidth]{%
        \footnotesize
        \hspace*{0.4em}\makebox[0.30\textwidth][c]{\shortstack[c]{(a) Nano T2V, 1-GPU\\H100 NVL/B200 backend latency}}\hfill
        \hspace*{0.4em}\makebox[0.30\textwidth][c]{\shortstack[c]{(b) Nano T2I, 1-GPU\\H100 NVL/B200 backend latency}}\hfill
        \hspace*{2.6em}\makebox[0.30\textwidth][c]{\shortstack[c]{(c) T2V on B200, 1--8 GPUs\\Nano/Super scaling}}\hspace*{-2.6em}%
    }}
    \caption{\textbf{Cosmos 3 serving performance.}
    (a) Cosmos3-Nano 720p T2V 1-GPU latency on H100 NVL and B200, to observe performance on different hardware backends.
    (b) Cosmos3-Nano 720p T2I 1-GPU latency on H100 NVL and B200, to observe performance on different hardware backends.
    (c) 720p T2V latency scaling on B200 from 1 to 8 GPUs for Cosmos3-Nano and Cosmos3-Super. Lower is better throughout.}
    \label{fig:cosmos3_serving_latency_combined}
\end{figure}

Together, these features allow Cosmos 3 Generator to scale from memory-constrained single-GPU deployments to high-throughput multi-GPU serving. Cache-DiT and quantization reduce the cost of repeated denoising computation, context parallelism and CFG-Parallel improve latency by distributing a single request across GPUs, and HSDP, CPU offload, and VAE-Patch-Parallel reduce memory pressure for large-resolution or long-duration generation tasks. Figure~\ref{fig:cosmos3_serving_latency_combined} summarizes Cosmos3 serving performance by comparing single-GPU runs on different hardware backends as well as comparing multi-GPU runs on B200s. The evaluation is done for PyTorch-OSS and vLLM-Omni frameworks. 

\subsection{Benchmark Infrastructure}
\label{subsec::benchmark_infra}

The Cosmos benchmark system manages evaluation jobs for Cosmos models and stores both generated artifacts and evaluation results. An orchestration layer schedules generation, scoring, and endpoint evaluation jobs on Lepton or Slurm clusters and tracks the execution status of each stage. For every run, the system records metadata including the model checkpoint, code version, selected benchmarks, generation settings, benchmark-specific parameters, and associated datasets. Together, these records establish full traceability between each reported score and the exact model weights, inputs, parameter configurations, and evaluation code used to produce it.

The system supports the heterogeneous benchmark suite described in \cref{sec::results} without requiring all benchmarks to share a common implementation. Benchmarks evaluate generated video, audio, action trajectories, and text responses from reasoner models across a diverse set of criteria, including visual fidelity, audio quality, audio-visual synchronization, prompt and control adherence, action or trajectory accuracy, task completion, physical plausibility, and reasoning correctness. Evaluators include integrations with open-source libraries and public benchmark suites, as well as custom evaluators developed specifically for Cosmos. Scoring methods span reference-based error metrics, perceptual and temporal consistency measures, audio-video alignment metrics, VLM-based judges, human annotations, and exact-match or numeric-answer evaluation.

For Generator evaluation, benchmarking is separated into generation and scoring stages. Generation jobs execute models using either the PyTorch inference pipeline or one of the serving frameworks described in \cref{subsec::serving_infrastructure}, and write generated outputs to object storage. Scoring jobs subsequently consume these stored artifacts, compute per-sample and aggregate metrics, and record results together with run metadata. This decoupled design allows outputs to be rescored with new metrics or evaluators without rerunning generation.

For Reasoner evaluation, we use the VLMEvalKit~\citep{duan2024vlmevalkit} framework together with vLLM. These jobs send prompts and multimodal inputs to deployed model endpoints, process model responses, and record benchmark scores and associated metadata.

Evaluation scores and run metadata are stored in a relational database, while generated artifacts are stored in object storage. Human-evaluation annotations are stored together with the evaluated artifacts and question sets, and aggregate human-evaluation results are tracked alongside automated metrics. A benchmark portal provides access to these records through dashboards, leaderboards, example-level inspection tools, and model-to-model or checkpoint-to-checkpoint comparisons.

\section{Results}
\label{sec::results}

We evaluate Cosmos 3 across a broad spectrum of understanding and generation tasks that are central to Physical AI. Unlike prior systems that focus on a single modality or capability, Cosmos 3 is designed as a unified omnimodal world model that jointly supports reasoning, perception, simulation, and action generation. Our evaluation therefore spans both the Reasoner and Generator components, covering multimodal understanding, spatial and temporal reasoning, image and video generation, audio-visual generation, transfer generation, forward and inverse dynamics, and robot policy learning. Across these diverse benchmarks, Cosmos 3 consistently demonstrates strong results relative to both specialized open-source models and leading proprietary systems, highlighting the benefits of a unified world-model architecture for Physical AI. The following sections present detailed results for the Reasoner and Generator, together with analyses of their capabilities across robotics, autonomous driving, smart infrastructure, and general multimodal domains. 

\subsection{Reasoner Evaluation}

\newcommand{\reasonerModelHeader}[1]{\rotatebox[origin=c]{90}{\makecell[c]{\tiny #1}}}
\newcommand{\reasonerGroupCell}[4][1]{\multirow{#2}{*}{\rotatebox[origin=c]{90}{\scalebox{#1}{\makecell[c]{\textbf{#3}\\#4 benchmarks}}}}}
\newcommand{\reasonerAvgCell}[1]{\cellcolor{black!4}#1}
\newcommand{\reasonerAvgLastCell}[1]{\multicolumn{1}{>{\columncolor{black!4}[\tabcolsep][0pt]}c@{}}{#1}}
\begin{table*}[p]
    \centering
    \tiny
    \caption{ \textbf{Reasoner benchmark results} for Cosmos 3 variants and comparison models across general multimodal understanding, robotics, smart-infrastructure, and autonomous-driving benchmarks. Rows report individual benchmarks or group averages, and columns report model scores. Within each separated model block, best and second-best scores per row are shown in \textbf{bold} and \underline{underlined}. $^\dagger$ denotes a closed model.
    }
    
    \label{tab:reasoner_benchmark_group}
    \setlength{\tabcolsep}{1.55pt}
    \renewcommand{\arraystretch}{1.05}
    \begin{adjustbox}{width=\textwidth}
    \begin{tabular}{@{}c l|>{\columncolor{columnours}}ccccc|>{\columncolor{columnours}}cccccc|>{\columncolor{columnours}}ccccc@{}}
        \toprule
        & \textbf{Benchmark} & \reasonerModelHeader{\textbf{Cosmos 3}\\\textbf{Super}} & \reasonerModelHeader{Qwen3-VL\\32B} & \reasonerModelHeader{Cosmos-Reason2\\32B} & \reasonerModelHeader{Gemma-4\\31B} & \reasonerModelHeader{Gemini 3.1\\Pro$^\dagger$} & \reasonerModelHeader{\textbf{Cosmos 3}\\\textbf{Nano}} & \reasonerModelHeader{Qwen3-VL\\8B} & \reasonerModelHeader{Cosmos-Reason2\\8B} & \reasonerModelHeader{Gemma-4\\E4B} & \reasonerModelHeader{RynnBrain\\8B} & \reasonerModelHeader{MiMo-Embodied\\7B} & \reasonerModelHeader{\textbf{Cosmos 3}\\\textbf{Edge}} & \reasonerModelHeader{Qwen3-VL\\2B} & \reasonerModelHeader{Cosmos-Reason2\\2B} & \reasonerModelHeader{Gemma-4\\E2B} & \reasonerModelHeader{RynnBrain\\2B} \\
        \midrule
        \reasonerGroupCell{20}{General}{19} & MMBench-Dev & 87.4 & \underline{87.8} & 86.3 & 86.7 & \textbf{93.2} & 85.1 & \underline{85.3} & 82.2 & 66.2 & \textbf{85.5} & 81.4 & 76.6 & \underline{77.2} & 73.6 & 60.2 & \textbf{81.9} \\
        & RealWorldQA & 79.2 & \underline{80.3} & 76.1 & 71.8 & \textbf{81.8} & 72.2 & \textbf{73.2} & 68.2 & 61.0 & \underline{72.5} & 72.0 & \textbf{73.3} & \underline{67.3} & 61.4 & 57.1 & 65.9 \\
        & CVBench & 88.0 & 86.8 & \underline{88.1} & 84.1 & \textbf{88.6} & 86.5 & 85.2 & 85.6 & 68.1 & \underline{87.5} & \textbf{87.8} & \underline{84.9} & 78.7 & 78.7 & 56.1 & \textbf{85.7} \\
        & VideoPhy2 & \textbf{47.4} & 36.8 & \underline{43.3} & 33.1 & 28.7 & \textbf{45.6} & 28.2 & \underline{37.1} & 13.8 & 10.5 & 20.8 & \textbf{40.3} & 7.9 & \underline{12.8} & 8.4 & 7.3 \\
        & CausalVQA & 77.0 & \underline{81.0} & 74.5 & 76.5 & \textbf{92.0} & 70.0 & \textbf{72.0} & \underline{71.5} & 38.0 & 68.5 & 63.0 & 37.0 & \textbf{57.0} & \underline{53.5} & 29.5 & 51.0 \\
        & MVPBench & \textbf{70.3} & 58.6 & \underline{62.2} & 27.0 & 59.4 & \textbf{66.9} & 51.6 & \underline{54.2} & 32.8 & 50.1 & 43.3 & \textbf{53.3} & 43.6 & 43.7 & 31.9 & \underline{44.9} \\
        & CountBenchQA & 89.1 & \underline{93.6} & 87.5 & 79.1 & \textbf{95.3} & 84.8 & \underline{89.5} & 79.9 & 55.4 & \textbf{90.5} & 84.6 & \textbf{89.9} & \underline{87.5} & 79.7 & 56.9 & 86.0 \\
        & AI2D & 87.8 & 88.2 & 87.5 & \underline{88.9} & \textbf{93.8} & \underline{85.0} & 84.8 & 83.6 & 78.2 & \textbf{85.7} & 83.2 & 75.4 & \underline{76.7} & 75.4 & 73.0 & \textbf{79.8} \\
        & DocVQA & 90.4 & \textbf{96.0} & 95.1 & 89.6 & \underline{95.8} & 94.2 & \textbf{95.6} & 94.3 & 78.1 & \underline{95.4} & 93.9 & 86.8 & \textbf{92.8} & 89.9 & 73.4 & \underline{91.8} \\
        & InfoVQA & 82.4 & \textbf{87.6} & \underline{85.1} & 66.5 & 85.0 & 81.8 & \underline{83.4} & 79.8 & 46.1 & 81.8 & \textbf{84.2} & 60.1 & \textbf{71.7} & 65.0 & 37.8 & \underline{70.5} \\
        & OCRBench-v2 & \textbf{66.7} & \underline{65.9} & 57.4 & 61.8 & 64.5 & \underline{60.1} & \textbf{64.1} & 56.6 & 42.4 & 58.6 & 41.2 & 43.7 & \textbf{54.2} & \underline{50.1} & 37.8 & 41.0 \\
        & LogicVista & 55.9 & 47.9 & 46.1 & \underline{57.0} & \textbf{81.9} & \textbf{43.2} & \underline{41.8} & 37.4 & 31.5 & 40.3 & 39.6 & 34.7 & \textbf{39.4} & 34.0 & 29.5 & \underline{34.9} \\
        & MMMU-Pro & 48.1 & 49.0 & 45.6 & \underline{70.6} & \textbf{76.7} & 41.1 & 41.1 & 38.6 & \textbf{46.9} & \underline{42.0} & 37.3 & 26.4 & \underline{32.3} & 26.9 & \textbf{39.9} & 30.3 \\
        & MVBench & \textbf{74.5} & 72.6 & 72.5 & 64.4 & \underline{72.8} & \textbf{73.2} & 69.1 & \underline{70.1} & 48.9 & 69.5 & 56.1 & 58.2 & 60.3 & \underline{60.5} & 41.2 & \textbf{64.7} \\
        & BlinkSpatial & 88.8 & 88.1 & 87.4 & \underline{90.9} & \textbf{92.3} & 81.8 & \textbf{87.4} & \underline{83.9} & 76.9 & 76.9 & 83.2 & 72.0 & \underline{77.6} & 75.5 & 62.9 & \textbf{79.7} \\
        & BlinkDepth & \textbf{91.9} & 82.3 & 85.5 & \underline{87.1} & 79.8 & \textbf{92.7} & 87.1 & 87.9 & 77.4 & \underline{91.1} & 81.5 & 79.8 & 74.2 & \underline{83.1} & 70.2 & \textbf{87.9} \\
        & RefCOCO & \underline{89.5} & \textbf{90.6} & 70.7 & 81.8 & 84.3 & \underline{84.3} & \textbf{87.3} & 81.1 & 71.5 & 75.9 & 74.3 & 80.1 & \textbf{84.5} & \underline{80.8} & 66.3 & 71.2 \\
        & HallusionBench & 49.0 & 52.8 & 50.8 & \underline{57.8} & \textbf{64.2} & 45.3 & \textbf{50.5} & 42.0 & 42.0 & \underline{45.7} & 40.2 & 40.7 & \underline{42.6} & 28.6 & 36.0 & \textbf{45.0} \\
        & IFBench & 37.0 & 37.0 & 28.2 & \textbf{52.3} & \underline{42.5} & 28.5 & \underline{32.0} & 26.0 & \textbf{34.2} & 22.8 & 25.8 & \underline{20.8} & 20.0 & 19.8 & \textbf{29.0} & 17.8 \\
        \cmidrule{2-18}
        & \reasonerAvgCell{\textbf{General Avg.}} & \underline{73.7} & \reasonerAvgCell{72.8} & 70.0 & \reasonerAvgCell{69.8} & \reasonerAvgCell{\textbf{77.5}} & \textbf{69.6} & \reasonerAvgCell{\underline{68.9}} & 66.3 & \reasonerAvgCell{53.1} & \reasonerAvgCell{65.8} & \reasonerAvgCell{62.8} & 59.7 & \reasonerAvgCell{\textbf{60.3}} & 57.5 & \reasonerAvgCell{47.2} & \reasonerAvgLastCell{\underline{59.9}} \\
        \midrule
        \reasonerGroupCell{18}{Robotics}{17} & Cosmos-ER & \underline{74.1} & 61.3 & \textbf{74.9} & 54.3 & 61.1 & \underline{69.7} & 56.9 & \textbf{71.2} & 44.3 & 54.9 & 55.6 & \underline{56.6} & 48.9 & \textbf{59.0} & 38.0 & 48.0 \\
        & Cosmos-CS & \underline{66.4} & 63.4 & 65.2 & 61.1 & \textbf{69.5} & \textbf{63.9} & 58.4 & \underline{63.1} & 43.2 & 54.1 & 53.8 & \underline{51.5} & 49.7 & \textbf{54.8} & 35.3 & 49.0 \\
        & RefSpatial & \underline{57.0} & 52.7 & 48.0 & 46.6 & \textbf{70.0} & \textbf{53.1} & \underline{47.6} & 41.1 & 24.6 & 46.6 & 41.3 & \textbf{48.4} & 27.1 & 30.7 & 16.2 & \underline{39.0} \\
        & VSI-Bench & \textbf{60.9} & \underline{59.5} & 58.0 & 47.6 & 47.5 & 54.9 & \underline{55.1} & 52.0 & 28.4 & \textbf{63.0} & 46.7 & \underline{59.2} & 49.8 & 45.0 & 27.3 & \textbf{62.5} \\
        & SparBench & \textbf{54.9} & 48.0 & 42.9 & 46.6 & \underline{51.5} & \textbf{54.8} & 40.1 & 38.0 & 28.5 & \underline{49.5} & 41.2 & \textbf{52.8} & 34.5 & 35.5 & 30.6 & \underline{47.8} \\
        & RynnBrain-Area & 53.0 & 53.1 & 50.4 & \underline{58.7} & \textbf{65.4} & \underline{52.0} & 33.2 & 43.4 & 35.7 & \textbf{56.6} & 47.1 & \underline{39.1} & 24.4 & 31.9 & 31.2 & \textbf{58.1} \\
        & RynnBrain-Spatial & 34.5 & 16.8 & \textbf{42.2} & 32.4 & \underline{36.8} & 26.1 & \underline{37.5} & 35.5 & 33.9 & \textbf{59.0} & 37.2 & 22.6 & 29.7 & \underline{33.9} & 19.7 & \textbf{55.7} \\
        & RynnBrain-Trajectory & \underline{69.3} & 61.6 & 64.6 & 64.4 & \textbf{71.3} & \textbf{67.9} & 54.8 & \underline{64.4} & 63.5 & 61.1 & 61.1 & 58.7 & 54.7 & \textbf{61.1} & \underline{60.9} & 53.5 \\
        & RynnBrain-Affordance & 84.6 & 84.2 & \underline{86.8} & \textbf{87.8} & \underline{86.8} & 85.1 & 82.6 & 84.6 & 84.4 & \underline{85.3} & \textbf{85.7} & 77.6 & 70.5 & \underline{80.0} & 77.0 & \textbf{90.4} \\
        & RynnBrain-Object & 48.3 & \textbf{57.0} & 44.9 & 47.2 & \underline{51.5} & 39.8 & \underline{49.2} & 42.0 & 35.2 & \textbf{71.6} & 33.5 & 24.0 & \underline{41.2} & 30.1 & 26.7 & \textbf{70.7} \\
        & RynnBrain-Grounding & 74.4 & \underline{77.0} & 76.7 & 72.4 & \textbf{82.8} & \underline{72.9} & 68.8 & 72.5 & 59.3 & \textbf{74.2} & 57.8 & \underline{51.6} & 33.1 & \textbf{54.3} & 31.1 & 45.5 \\
        & MMSIBench & \textbf{41.8} & 33.0 & 31.4 & 32.3 & \underline{40.8} & 36.2 & 26.3 & 29.5 & \underline{37.6} & \textbf{38.4} & 30.1 & 32.3 & 28.5 & 28.9 & \textbf{33.7} & \underline{33.2} \\
        & MMSIVideoBench & 26.1 & \underline{33.8} & 29.6 & 33.6 & \textbf{38.6} & 27.2 & 28.8 & 29.4 & \textbf{39.4} & 28.2 & \underline{30.0} & 24.9 & \underline{25.5} & 23.9 & \textbf{32.5} & 24.5 \\
        & HealthSurgiBench & \underline{44.5} & 24.4 & \textbf{53.9} & 19.1 & 24.6 & \textbf{56.1} & 23.0 & \underline{46.9} & 20.7 & 23.3 & 24.0 & \textbf{62.1} & 26.0 & \underline{31.3} & 17.3 & 18.3 \\
        & ERQA & \underline{51.2} & 46.5 & 42.8 & 47.8 & \textbf{65.2} & \textbf{46.0} & 44.0 & \underline{44.2} & 30.2 & 43.0 & 42.0 & \textbf{42.0} & 37.8 & 37.2 & 32.5 & \underline{38.8} \\
        & RoboSpatialHome & \textbf{70.0} & \underline{65.1} & 64.3 & 63.0 & \underline{65.1} & \underline{66.3} & 64.8 & 64.5 & 42.0 & \textbf{71.4} & 66.1 & \textbf{63.4} & 44.6 & 52.0 & 36.3 & \underline{62.9} \\
        & Where2Place & \textbf{71.0} & 56.0 & 59.0 & 52.0 & \underline{61.0} & \textbf{64.0} & 53.0 & 50.0 & 17.0 & 11.0 & \underline{58.0} & \textbf{55.0} & 32.0 & \underline{33.0} & 15.0 & 11.0 \\
        \cmidrule{2-18}
        & \reasonerAvgCell{\textbf{Robotics Avg.}} & \underline{57.8} & \reasonerAvgCell{52.6} & 55.0 & \reasonerAvgCell{51.0} & \reasonerAvgCell{\textbf{58.2}} & \textbf{55.1} & \reasonerAvgCell{48.5} & 51.3 & \reasonerAvgCell{39.3} & \reasonerAvgCell{\underline{52.4}} & \reasonerAvgCell{47.7} & \textbf{48.3} & \reasonerAvgCell{38.7} & 42.5 & \reasonerAvgCell{33.0} & \reasonerAvgLastCell{\underline{47.6}} \\
        \midrule
        \reasonerGroupCell{10}{Smart Infrastructure}{9} & VANTAGE-2DGrounding & \textbf{76.2} & \underline{72.4} & 45.7 & 45.1 & 46.9 & \textbf{75.6} & \underline{73.3} & 66.9 & 10.1 & 67.6 & 59.4 & 49.8 & \textbf{65.1} & \underline{56.3} & 5.5 & 49.2 \\
        & VANTAGE-Astro2D & \textbf{81.5} & 76.8 & 22.6 & 68.7 & \underline{77.7} & \underline{78.3} & 69.2 & \textbf{81.1} & 58.8 & 10.5 & 57.8 & \textbf{76.7} & 56.6 & \underline{72.5} & 48.4 & 0.0 \\
        & VANTAGE-2DPointing & 72.9 & 75.6 & 74.0 & \underline{76.8} & \textbf{85.6} & \textbf{74.8} & 68.6 & \underline{68.7} & 43.4 & 64.9 & 55.0 & \textbf{63.0} & 53.1 & 59.7 & 31.7 & \underline{60.3} \\
        & VANTAGE-DVC & 29.5 & 29.4 & \underline{30.1} & 29.6 & \textbf{30.2} & \underline{31.4} & 29.6 & \textbf{32.5} & 14.3 & 28.4 & 2.2 & \underline{20.9} & 0.8 & \textbf{28.5} & 9.3 & 0.0 \\
        & VANTAGE-EventVerif & \underline{71.3} & 60.0 & \textbf{73.6} & 55.2 & 67.5 & \textbf{68.9} & 59.4 & \underline{64.1} & 40.6 & 58.9 & 58.0 & \textbf{64.8} & 44.7 & \underline{55.3} & 27.6 & 41.6 \\
        & VANTAGE-SOT & \underline{62.2} & 44.2 & 33.1 & 54.7 & \textbf{72.7} & \textbf{59.2} & 33.1 & \underline{37.7} & 16.0 & 4.8 & 9.2 & 18.7 & \textbf{29.8} & \underline{26.7} & 11.5 & 4.8 \\
        & VANTAGE-Temporal & \textbf{51.9} & 46.8 & \underline{50.5} & 33.0 & 41.7 & \textbf{48.0} & 43.3 & \underline{47.3} & 16.9 & 20.1 & 5.9 & \textbf{39.1} & 35.2 & \underline{39.0} & 9.0 & 25.9 \\
        & VANTAGE-VQA & 69.5 & \textbf{71.3} & 70.3 & 67.0 & \underline{71.2} & \textbf{69.0} & 66.4 & \underline{68.0} & 51.5 & 65.4 & 67.5 & \underline{64.4} & 63.9 & \textbf{64.7} & 46.2 & 62.9 \\
        & TARBench & \textbf{48.4} & 28.1 & \underline{36.6} & 31.9 & 33.6 & \textbf{43.6} & 31.5 & 34.1 & 12.8 & \underline{34.8} & 32.6 & \underline{34.1} & 32.7 & 26.6 & 8.5 & \textbf{34.2} \\
        \cmidrule{2-18}
        & \reasonerAvgCell{\textbf{Smart Infra. Avg.}} & \textbf{62.6} & \reasonerAvgCell{56.1} & 48.5 & \reasonerAvgCell{51.3} & \reasonerAvgCell{\underline{58.6}} & \textbf{61.0} & \reasonerAvgCell{52.7} & \underline{55.6} & \reasonerAvgCell{29.4} & \reasonerAvgCell{39.5} & \reasonerAvgCell{38.6} & \textbf{47.9} & \reasonerAvgCell{42.4} & \underline{47.7} & \reasonerAvgCell{22.0} & \reasonerAvgLastCell{31.0} \\
        \midrule
        \reasonerGroupCell{4}{Driving}{3} & LingoQA & \textbf{76.8} & 66.4 & 70.0 & 57.2 & \underline{71.2} & \underline{71.4} & 68.4 & \textbf{71.6} & 24.2 & 60.4 & 70.4 & \underline{58.4} & \textbf{59.2} & 58.2 & 19.0 & 50.2 \\
        & AVSpecialCollision & \textbf{79.3} & 37.3 & \underline{77.3} & 32.3 & 54.0 & \textbf{79.0} & 34.0 & \underline{74.0} & 33.3 & 33.7 & 37.3 & \underline{66.7} & 36.3 & \textbf{74.3} & 33.7 & 33.3 \\
        & AVSpecialStopBehavior & \textbf{81.6} & 18.4 & \underline{69.4} & 20.4 & 16.3 & \textbf{77.5} & 36.7 & \underline{59.2} & 20.4 & 36.7 & 42.9 & \textbf{53.1} & 32.6 & 34.7 & 20.4 & \underline{36.7} \\
        \cmidrule{2-18}
        & \reasonerAvgCell{\textbf{Driving Avg.}} & \textbf{79.3} & \reasonerAvgCell{40.7} & \underline{72.2} & \reasonerAvgCell{36.6} & \reasonerAvgCell{47.2} & \textbf{76.0} & \reasonerAvgCell{46.4} & \underline{68.3} & \reasonerAvgCell{26.0} & \reasonerAvgCell{43.6} & \reasonerAvgCell{50.2} & \textbf{59.4} & \reasonerAvgCell{42.7} & \underline{55.7} & \reasonerAvgCell{24.4} & \reasonerAvgLastCell{40.1} \\
        \bottomrule
    \end{tabular}
    \end{adjustbox}
\end{table*}

Cosmos 3 Reasoner is evaluated on a total of \textbf{48} benchmarks. The results are aggregated into four categories: general, robotics, smart infrastructure, and driving. \Cref{tab:reasoner_benchmark_group} reports results for the Edge, Nano, and Super models and their comparisons to existing open-source and closed-source models. The numbers are evaluated using VLMEvalKit~\cite{duan2024vlmevalkit}, which is integrated into our benchmark infrastructure.

\paragraph{General.} We select 19 benchmarks, listed below, to assess the model's general capabilities.
\begin{itemize}
  \item \textit{Broad visual question answering and multimodal understanding} is measured by MMBench\_DEV~\citep{liu2024mmbench}, RealWorldQA~\citep{xai2024realworldqa}, and AI2D~\citep{kembhavi2016ai2d}, which test whether the model can interpret natural images, diagrams, and real-world scenes while answering diverse semantic and compositional questions.
  \item \textit{Spatial, grounding, and quantitative reasoning.} is evaluated by CVBench~\citep{tong2024cambrian}, BlinkSpatial, BlinkDepth~\citep{fu2024blink}, RefCOCO~\citep{yu2016refcoco}, and CountBenchQA~\citep{paiss2023countbench}, covering 2D/3D spatial relations, depth perception,
  referring-expression localization, and object counting.
  \item \textit{Text-rich visual understanding} is covered by DocVQA~\citep{mathew2021docvqa}, InfoVQA~\citep{mathew2022infographicvqa}, and OCRBench-v2~\citep{fu2026ocrbench}, which require reading and reasoning over documents, infographics, scene text, and structured
  visual layouts.
  \item \textit{Video, physical, and causal reasoning} is assessed by MVBench~\citep{li2024mvbench}, VideoPhy2~\citep{bansal2025videophy}, MVPBench~\citep{krojer2025shortcut}, and CausalVQA~\citep{foss2025causalvqa}, testing temporal event understanding, physical plausibility, and cause--effect
  reasoning across frames.
  \item \textit{Advanced reasoning, robustness, and instruction following} is measured by LogicVista~\citep{xiao2024logicvista}, MMMU\_Pro~\citep{yue2024mmmupro}, HallusionBench~\citep{guan2024hallusionbench}, and IFBench~\citep{pyatkin2025ifbench}, which probe visual logic, expert-level multimodal problem solving, hallucination resistance, and adherence to user instructions.
\end{itemize}
Together, these benchmarks provide a broad view of model's general reasoning ability across perception, localization, text recognition, temporal understanding, and reliable instruction-conditioned response generation.

\paragraph{Robotics.} 
We group 17 robotics and embodied reasoning benchmarks into several capability families.
\begin{itemize}
  \item \textit{Embodied commonsense and task reasoning} is measured by Embodied Reasoning, CommonSense in Cosmos Reason~\citep{azzolini2025cosmos}, ERQA~\citep{gemini2025robotics}, and Where2Place~\citep{yuan2024robopoint}, which test whether the model can reason about
  object affordances, feasible actions,  placement decisions, and physical commonsense in embodied environments.
  \item \textit{Spatial grounding and scene geometry} is evaluated by RefSpatial~\citep{zhou2025roborefer}, VSI-Bench~\citep{yang2024vsibench}, SparBench~\citep{zhang2025spar}, and RoboSpatialHome\citep{song2025robospatial}, covering referring-expression grounding, metric and relational
  spatial understanding, indoor layout reasoning, free-space awareness, and robot-relevant localization.
  \item \textit{Robotics-oriented perception and action understanding} is captured by RynnBrain~\citep{dang2026rynnbrain} and ERQA~\citep{gemini2025robotics}, which probe manipulation-relevant perception, action feasibility, trajectory reasoning, and object-centric decision making over real robot episodes.
  And MMSIVideoBench~\citep{lin2025mmsivideobench} requires reasoning across multiple views or frames to infer object
  correspondences, spatial relations, temporal changes, and scene-level structure. We curate HealthSurgiBench, an in-house benchmark for operating-room understanding with 23 question types. It combines rule-based scoring for structured outputs such as counts, tools, roles,
  boxes, coordinates, time/distance estimates, ordered lists, scene graphs, and monitor text, with a local Qwen3-4B judge for six free-form answer types. The final score uses the applicable evaluator for each sample and then reports the unweighted mean across all samples.
\end{itemize}
Together, these benchmarks evaluate whether the model can move beyond static visual recognition toward embodied reasoning: understanding where objects are, how they relate, what actions are possible, and how spatial evidence evolves across views and time.

\paragraph{Smart infrastructure.} We evaluate on \textbf{VANTAGE-Bench}~\citep{nvidia2026vantage} and \textbf{Traffic Anomaly Reasoning (TAR)}~\citep{nvidia2026tar}, covering warehouse logistics, transportation, and smart-infrastructure fixed-camera settings. VANTAGE-Bench measures semantic, spatial, temporal, and spatiotemporal understanding through event verification, VQA, referring, pointing, localization, temporal localization, dense captioning, and VLM-native single-object tracking, with $3{,}346$ assets and $35{,}027$ expert annotations, including synthetic anomaly footage. TAR, the AI City Challenge 2026 Track~3 suite, evaluates anomaly verification, temporal localization, scene description, causality, summarization, and out-of-domain generalization with heterogeneous QA, temporal IoU, and text-generation metrics.

\paragraph{Driving.} We evaluate the driving capabilities with \textbf{LingoQA}~\citep{marcu2024lingoqa} and two in-house benchmarks for safety-critical driving event classification. \textbf{AVSpecialCollisionBench} measures whether the model correctly classifies each video into one of three event categories: collision, near collision, or no collision. The benchmark contains 100 videos per category and the per-category accuracy is computed. Finally, the mean accuracy across the three categories is reported as the final score. \textbf{AVSpecialStopBehaviorBench} evaluates stop-sign behavior classification across five categories: full stop, rolling stop, no stop, not relevant, and false sign. This benchmark contains 10 videos per category, and its final score is the mean of the five per-category accuracies.

Cosmos 3 is competitive with open-source models on general benchmarks, while still trailing Gemini 3.1 Pro~\citep{gemini3}. Compared with Cosmos-Reason2, Cosmos 3 shows stronger general capabilities, benefiting from the additional 20\% pre-training data that increases data diversity. In the robotics, smart infrastructure, and driving domains, Cosmos 3 outperforms both open-source and closed-source models including RynnBrain~\citep{dang2026rynnbrain}, Mimo-Embodied~\citep{hao2025mimo}, and Gemma-4~\citep{gemma4modelcard}, with the exception of a small gap to Gemini 3.1 Pro in robotics. Overall, Cosmos 3 demonstrates strong domain-specific reasoning across robotics, smart infrastructure, and autonomous driving, supporting a broad range of Physical AI applications.

\subsection{Generator Evaluation}

The Generator component of Cosmos 3 is evaluated across a diverse set of tasks that collectively measure its ability to simulate and generate multimodal worlds for Physical AI. Unlike conventional generative models that focus on a single modality, Cosmos 3 jointly models images, videos, audio, and actions within a unified framework, enabling evaluation across image generation, video generation, audio-visual generation, transfer generation, forward and inverse dynamics, and robot policy learning. We further evaluate specialized post-trained variants, including Cosmos3-Super-Text2Image, Cosmos3-Super-Image2Video, and Cosmos3-Nano-Policy-DROID, to assess the effectiveness of downstream adaptation from a shared omnimodal foundation model. Our benchmark suite combines automated metrics, human evaluation, domain-specific Physical AI benchmarks, and real-world robotics evaluations, covering critical capabilities such as prompt following, physical plausibility, temporal consistency, audio-video synchronization, controllability, action prediction, and task completion.

\subsubsection{Image Generation Evaluation}
\label{subsec::image_eval}

\par
\begin{figure}[t]
    \centering
    \includegraphics[width=0.95\linewidth]{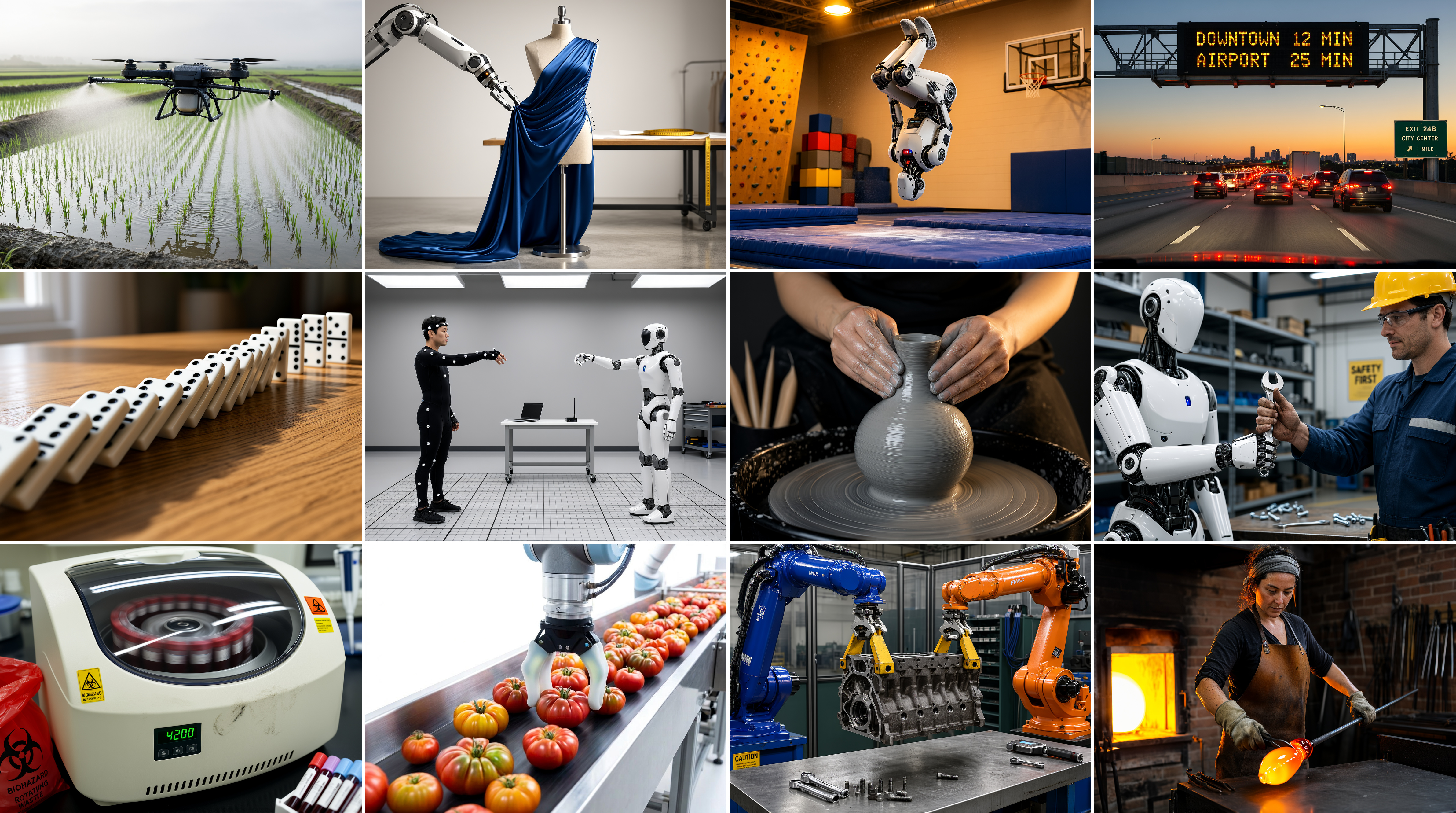}
    \caption{\textbf{Example images generated by Cosmos3-Super-Text2Image.} Our model generates images that are both physically plausible and photorealistic, exhibiting coherent object geometry, consistent object environment interactions, \etc. All images are generated from single-shot upsampled JSON prompts using shift=3.0, guidance=4.0, and 50 diffusion steps, as also described in Table~\ref{tab:sampling_configs}. These results highlight the model's potential as an effective real-world image simulator for robotics, autonomous driving, and other scenarios where adherence to physical laws is essential.}
    \label{fig:sft_t2i_demo}
\end{figure}

 \noindent
We evaluate Cosmos 3 image generation as single-frame visual generation, focusing on four complementary axes: broad semantic prompt following, exact scene-text rendering, human-preference alignment, and visual aesthetics.
UniGenBench is the primary prompt-following metric because it exposes failures at the testpoint level. We enhance UniGenBench by adding a Physical-AI subset.
CVTG isolates a frequent failure mode of image generators---misspelled, omitted, blurred, or duplicated scene text---through OCR-based GNED and PNED scores.
HPSv3 and LAION aesthetic complement those targeted checks by measuring overall human preference and visual appeal, giving a more actionable view of T2I quality than any single aggregate score.
For all benchmarks, we use Claude-Opus-4.7 as a prompt rewriter to convert evaluation text prompts into a structured format described in \cref{sec::gen_guide} and \cref{sec::prompt_upsampling}, ensuring they match the training prompt style preferred by the generator. The instructions template can be found in Appendix~\ref{appendix:sft_t2i_upsampler_prompt}. We generate and compare all models at 1024x1024 resolution. For open-source models, we follow the recommended hyper-parameters and negative prompts provided in their documentation. Qualitative examples can also be found in \cref{fig:sft_t2i_demo}.

\paragraph{UniGenBench.} UniGenBench~\citep{wang2025unigenbench} is a unified semantic evaluation benchmark for text-to-image generation.
It comprises 600 prompts spanning 5 main themes and 20 subthemes, each assessed across 10 primary and 27 sub-evaluation criteria using an MLLM-as-Judge framework.
To better assess the capabilities of Cosmos 3 in Physical AI scenarios, we augment the benchmark with 570 additional prompts (UniGenBench-Phys) targeting photorealistic physical-world scenes.
These prompts cover six physical-world sub-domains: (a) robotics and industrial, (b) physical signage and text, (c) autonomous driving, (d) construction sites, (e) fluid dynamics, and (f) medical/clinical.
Gemini 3.1 Pro evaluates each generated image against each testpoint with binary pass/fail decisions; the primary score is mean testpoint accuracy, reported for the full 1{,}170-prompt (\ie "All") with additional breakdowns over original and physics subsets (\ie "Orig" and "Phys").

\paragraph{CVTG.} CVTG~\citep{du2025textcrafter} (Complex Visual Text Generation) evaluates a model's ability to accurately render text in visually complex scenes. This capability is particularly important for real-world environments, where signage, labels, and instructional text must be legible and correctly spelled. Generated images are first processed by an OCR system to extract visible text regions, and the predicted strings are then matched to the target strings using Hungarian matching based on normalized edit distance (\ie NED).

We evaluate on two prompt sets:
(a) \textit{CVTG-500L}, an English-focused subset of 500 prompts randomly sampled from CVTG-2K while preserving comparable coverage of the number of text regions. We further upsample each short prompt into a long, dense description to better reflect the requirements of modern world-model generation.
(b) \textit{CVTG-102ch}, a Chinese-focused set of 102 prompts designed to evaluate accurate Chinese character rendering. These prompts are drawn from both common and physically grounded domains, including public spaces, outdoor and scenic environments, digital displays, and other real-world settings.

\textbf{HPSv3.} HPSv3~\citep{ma2025hpsv3} evaluates prompt-aware text-to-image quality with a learned human-preference reward model. It is built on HPDv3, a preference dataset spanning both synthetic and real images across a wide range of quality levels. HPSv3 adopts a VLM-based architecture trained with an uncertainty-aware ranking loss, producing a prompt-aware score that reflects human judgments of semantic alignment, realism, and aesthetic quality. We directly apply HPSv3 on Cosmos 3 results to obtain this complementary human-preference metric.

\textbf{Aesthetic V2.} Aesthetic V2~\citep{laionaesthetics} measures prompt-independent visual appeal using the LAION aesthetic predictor. The predictor estimates how much people would like an image on a 1--10 scale, using a lightweight model trained on top of CLIP image embeddings. Unlike HPSv3, this score does not condition on the input prompt and therefore does not directly measure instruction following or semantic alignment. We use it as a standalone image-quality signal to capture general visual attractiveness and composition quality.

\begin{table*}[t]
    \centering
    \footnotesize
    \caption{\textbf{Text-to-Image benchmark results.} UniGenBench scores are fractions of evaluation criteria satisfied; ``All (1170)'' aggregates the original 600 prompts (Orig) and 570 Physical AI prompts (Phys). PNED and GNED are character-level accuracy metrics (higher is better); CVTG-102ch tests Chinese character rendering, CVTG-500L tests English long-prompt rendering. Aesthetic v2 and HPSv3 are higher-is-better image-level metrics.$^\dagger$}
    \label{tab:t2i_results}
    \setlength{\tabcolsep}{4pt}
    \resizebox{\textwidth}{!}{
    \begin{tabular}{ll|ccc|cc|cc|cc}
        \toprule
        & & \multicolumn{3}{c|}{\textbf{UniGenBench}} & \multicolumn{2}{c|}{\textbf{CVTG-500L}} & \multicolumn{2}{c|}{\textbf{CVTG-102ch}} & \multicolumn{2}{c}{\textbf{Image-Level Metrics}} \\
        \cmidrule(lr){3-5} \cmidrule(lr){6-7} \cmidrule(lr){8-9} \cmidrule(lr){10-11}
        \textbf{Model} & \textbf{Type} & \textbf{All} ($\uparrow$) & \textbf{Orig} ($\uparrow$) & \textbf{Phys} ($\uparrow$) & \textbf{GNED} ($\uparrow$) & \textbf{PNED} ($\uparrow$) & \textbf{GNED} ($\uparrow$) & \textbf{PNED} ($\uparrow$) & \textbf{Aesv2} ($\uparrow$) & \textbf{HPSv3} ($\uparrow$) \\
        \midrule
        \rowcolor{rowours}
        \textbf{Cosmos3-Super-Text2Image} & Open-source & \textbf{91.36} & \textbf{93.34} & \textbf{89.54} & \textbf{80.88} & \underline{89.08} & 32.02 & 41.22 & \underline{5.91} & 11.60 \\
        \rowcolor{rowours}
        \textbf{Cosmos3-Super} & Open-source & 87.33 & 85.21 & 89.64 & 66.77 & 70.97 & 8.48 & 16.31 & 5.76 & 9.49 \\
        \rowcolor{rowours}
        \textbf{Cosmos3-Nano} & Open-source & 84.61 & 87.32 & 82.12 & 24.23 & 26.53 & 4.63 & 9.70 & 5.76 & 8.99 \\
        \midrule
        Gemini 3 Pro Image & Closed-source & \underline{90.69} & \underline{92.81} & \underline{89.74} & 59.24$^\dagger$ & 71.79$^\dagger$ & 46.00 & \textbf{76.40} & 5.70 & \underline{11.78} \\
        FLUX.2-dev         & Open-source & 87.60 & 89.77 & 85.61 & 74.71 & 84.98 & 44.33 & 68.74 & 5.75 & 11.38 \\
        Qwen-Image-2512    & Open-source & 84.25 & 87.32 & 81.44 & \underline{79.68} & \textbf{90.86} & 46.33 & 71.26 & \textbf{5.92} & 11.03 \\
        Hunyuan 3.0        & Open-source & 84.02 & 87.68 & 80.67 & 71.40 & 87.68 & \underline{49.05} & 71.31 & 5.90 & \textbf{11.93} \\
        Z-Image-Turbo      & Open-source & 78.14 & 81.53 & 75.03 & 75.20 & 86.95 & \textbf{49.18} & \underline{73.32} & 5.70 & 11.36 \\
        \bottomrule
    \end{tabular}
    }
    \vspace{0.5ex}
    \par\noindent\scriptsize $^\dagger$ Gemini 3 Pro Image has a high probability of generating case-insensitive scene text (e.g., ``Adventure'' $\rightarrow$ ``ADVENTURE''). When this error is ruled out, the CVTG-500L scores increase to GNED = 75.97 and PNED = 91.45.
\end{table*}

\begin{figure}[t]
    \centering
    \includegraphics[width=\linewidth]{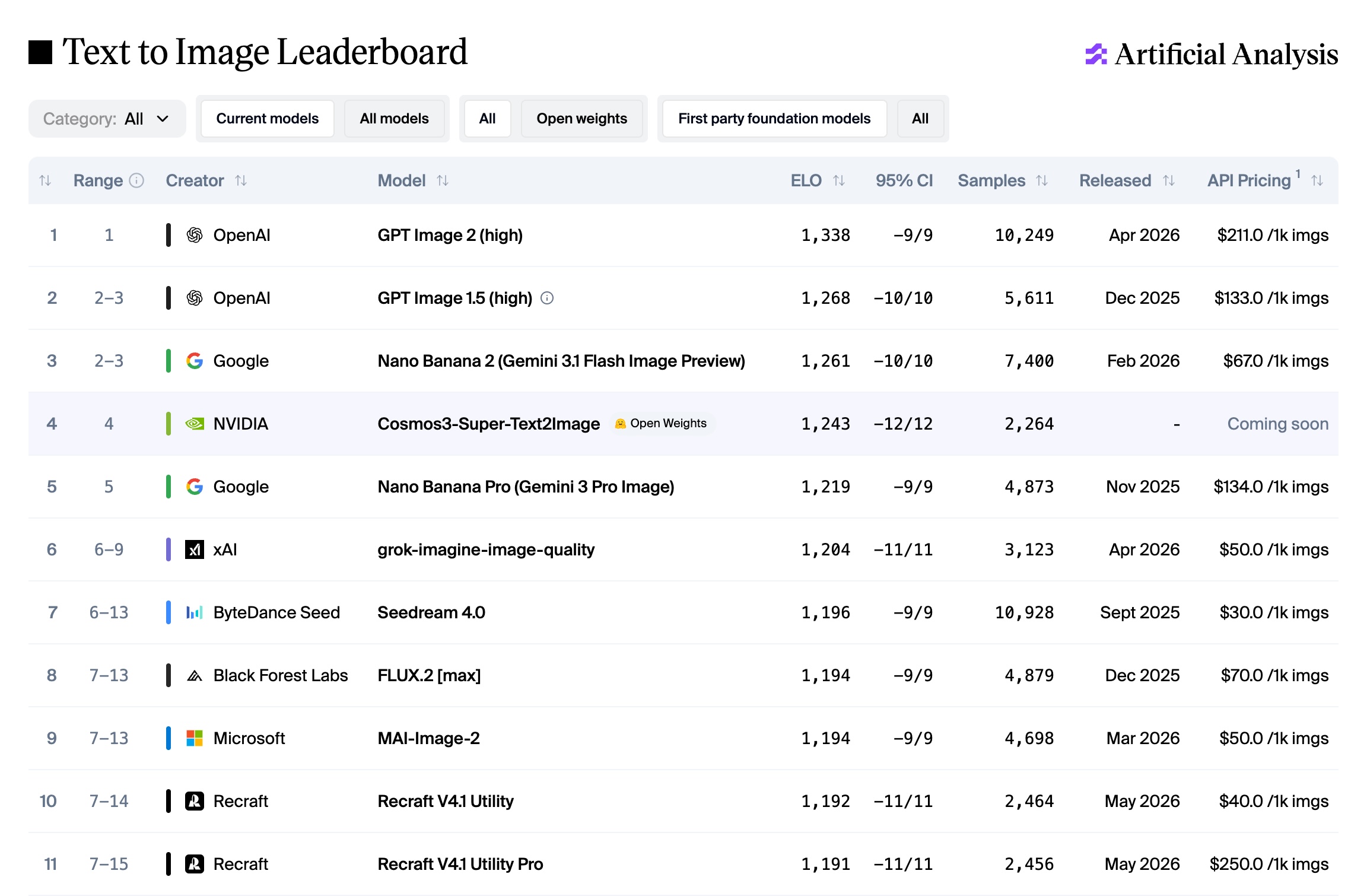}
    \caption{\textbf{Cosmos3-Super-Text2Image is the \#1 open-weight model on crowdsourced arena rankings.} \textbf{Cosmos3-Super-Text2Image} ranked \#1 among open-weight models (\#4 including proprietary models) on the Artificial Analysis Text to Image Leaderboard (Date: 2026-05-28).}
    \label{fig:sft_t2i_aa_leaderboard}
    \vspace{-0.5\baselineskip}
\end{figure}

\paragraph{Artificial Analysis Text-to-Image Leaderboard.} To evaluate Cosmos 3 under real-world scenarios, we submitted our specialized T2I model, Cosmos3-Super-Text2Image, alongside an agentic harness, to the Artificial Analysis Text-to-Image leaderboard for crowdsourced public voting. The model ranked \#1 among all open-weight models and \#4 among all models (open $+$ closed). For more details on the submission, please see Appendix ~\ref{appendix:t2i_agentic_upsampling}.

\FloatBarrier
\subsubsection{Video Generation Evaluation}
\label{subsec::video_eval}

We evaluate the video generation capabilities of Cosmos 3 through complementary automated and human benchmarks. Automated benchmarks offer scalable, reproducible comparisons and capture a broad range of quality and domain-specific signals, but their discriminative power diminishes as models improve (especially in domain-specific Physical-AI regimes, where such metrics are usually myopic to temporal and physics-based failures). Human evaluation addresses these gaps by catching long-tail physics, embodiment, and semantic failures that automated protocols systematically miss, while spreading scores across a meaningfully wider range so that small but real differences between state-of-the-art models remain detectable. We report results on three automated benchmarks---PAIBench-G, RBench, and Physics-IQ---followed by Cosmos HUE, a dedicated human evaluation protocol for Physical AI video generation, and Human World Bench (HWB), targeted human evaluation for realistic human motion from task-level instructions. For all benchmarks (automated or human-evaluation), we use Claude-Opus-4.6 as a prompt rewriter to convert text prompts into a structured format described in~\cref{sec::gen_guide} and~\cref{sec::prompt_upsampling}, ensuring they match the training prompt distribution.

\phantomsection\label{subsubsec::paibench_g}
\paragraph{PAIBench-G.} PAIBench-G is the video generation track of the Physical AI Benchmark~\citep{zhou2025paibench}, comprising 1,044 image–text prompt pairs across six Physical AI domains: Human (299), Autonomous Vehicle (239), Common Sense (174), Robotics (107), Physics (107), and Industry (107). We adopt it for its broad domain coverage and balanced scoring: each model receives a Quality Score (aggregating metrics for frame consistency, motion smoothness, aesthetic quality, and video–text alignment) and a Domain Score (VLM-as-Judge binary verification of physical and semantic accuracy across domains). The overall score weights both equally: $\text{Overall} = 0.5 \times \text{Quality} + 0.5 \times \text{Domain}$. With a Pearson correlation of
r=0.918 against human ELO rankings, PAIBench-G is well-suited for reliably measuring progress in early training phases, before models converge to the frontier where automated scores begin to saturate. For every prompt, we generate videos across 5 seeds and evaluate at 720p resolution, 16:9 aspect ratio for 189 frames. While PaiBench-G natively supports only Image-to-Video evaluation, we extend it to compute Text-to-Video scores as well\footnote{We found that the public PAIBench-G Image-to-Video leaderboard results (judged by Qwen3-VL-235B-A22B) were not reproducible. We therefore use Qwen2.5-VL-72B-Instruct as the domain-score judge for our internal evaluation (\cref{tab:paibench_rbench_results_combined}) and independently submit to the public leaderboard; on both, Cosmos3-Super and Cosmos3-Nano rank first and second overall.}. \cref{tab:paibench_rbench_results_combined} reports both text-to-video and image-to-video PAIBench-G results using Qwen2.5-VL-72B-Instruct as the VLM judge; Cosmos3-Super achieves state-of-the-art overall score on both tracks, emerging as the best open-source model, outperforming strong closed-source models such as Veo-3.1.

\begin{table*}[t]
    \centering
    \captionsetup{justification=raggedright, singlelinecheck=false}
    \caption{\textbf{PAIBench-G and RBench results, across Text-to-Video and Image-to-Video settings.} PAIBench-G evaluates visual quality and domain-specific accuracy across six Physical AI domains while RBench evaluates task correctness and physical plausibility in embodied robotics scenarios. \textbf{Cosmos3-Super} achieves the highest overall scores on both PAIBench-G T2V and I2V among open-source models, while \textbf{Cosmos3-Nano} leads on RBench. \textbf{Bold} indicates the winner in the column and \underline{underline} indicates the second best.}
    \label{tab:paibench_rbench_results_combined}
    \footnotesize
    \setlength{\tabcolsep}{4pt}
    \renewcommand{\arraystretch}{1.15}
    \begin{tabular}{l l | ccc | ccc | c}
        \toprule
        & & \multicolumn{3}{c}{\textbf{PAIBench-G Text2Video ($\uparrow$)}} & \multicolumn{3}{c}{\textbf{PAIBench-G Image2Video ($\uparrow$)}} & \textbf{RBench Image2Video ($\uparrow$)} \\
        \cmidrule(lr){3-5} \cmidrule(lr){6-8} \cmidrule{9-9}
        \textbf{Model} & \textbf{Type} & \textbf{Overall} & \textbf{Domain} & \textbf{Quality} & \textbf{Overall} & \textbf{Domain} & \textbf{Quality} & \textbf{Score} \\
        \midrule
        \rowcolor{rowours}
        \textbf{Cosmos3-Super} & Open-source & \textbf{80.0} & \textbf{86.8} & \underline{73.1} & \textbf{82.8} & \underline{87.3} & \textbf{78.2} & 58.1\% \\
        \rowcolor{rowours}
        \textbf{Cosmos3-Nano}  & Open-source & \underline{79.4} & \underline{85.8} & 73.0 & \underline{82.7} & 87.2 & \underline{78.1} & \underline{58.4}\% \\
        \midrule
        Wan2.2-A14B                   & Open-source & 78.0 & 83.2 & 72.8 & 81.3 & 85.3 & 77.3 & 50.7\% \\
        HunyuanVideo-1.5              & Open-source & 76.5 & 80.9 & 72.0 & 81.7 & 85.9 & 77.6 & 46.0\% \\
        Cosmos-Predict2.5-2B   & Open-source & 76.5 & 79.9 & \textbf{73.2} & 81.2 & 84.6 & 77.9 & 46.4\% \\
        Cosmos-Predict2.5-14B  & Open-source & 76.4 & 79.5 & \textbf{73.2} & 81.1 & 84.0 & \underline{78.1} & ---  \\
        Wan2.1-14B                    & Open-source & 76.4 & 80.1 & 72.7 & 80.2 & 83.6 & 76.8 & ---  \\
        Wan2.2-5B                     & Open-source & 76.3 & 79.6 & 73.0 & 81.0 & 84.6 & 77.4 & ---  \\
        \midrule
        Veo-3.1                       & Closed-source   & 79.1 & 85.2 & 72.9 & 82.6 & \textbf{87.6} & 77.6 & 56.3\% \\
        Seedance-1.5-Pro              & Closed-source   & 76.9 & 82.1 & 71.6 & 80.8  & 84.7  & 76.9  & \underline{58.4}\% \\
        Wan 2.6                       & Closed-source & 78.6 & 85.2 & 72.0 & 81.9 & 85.9 & 77.8  & \textbf{60.7\%}\\
        \bottomrule
    \end{tabular}
    \vspace{-0.5\baselineskip}
\end{table*}

\paragraph{RBench.} RBench~\citep{deng2026rbench} evaluates video generation in embodied task scenarios, emphasizing task correctness and physical plausibility in robot–object interactions over purely perceptual realism. While PAIBench-G covers Physical AI domains broadly, RBench delves deeper into robotics as a critical Physical AI use case, stress-testing models on generating physically coherent manipulation sequences across diverse robot morphologies. The benchmark comprises 650 image–text evaluation cases from two complementary splits: 250 task-oriented pairs spanning five categories (Common Manipulation, Long-horizon Planning, Multi-entity Collaboration, Spatial Relationship, and Visual Reasoning) and 400 embodiment-specific pairs covering four robot morphologies (Dual-arm, Humanoid, Single-arm, and Quadruped). Prompts are sourced from RoVid-X~\citep{deng2026rbench}, a curated corpus of 4M robotics clips spanning over 1,300 skills. Each video is scored on Task Completion (TC, averaging physical-semantic plausibility and task-adherence consistency) and Visual Quality (VQ, a penalized combination of robot–subject stability and motion smoothness), and the final score is the mean of both across all 650 cases. With a Spearman correlation of $\rho = 0.96$ with human judgments across 25 evaluated models, RBench provides a reliable signal for embodied generation quality. For every prompt, we generate videos with a single seed and evaluate at 720p resolution, 16:9 aspect ratio for 121 frames. \cref{tab:paibench_rbench_results_combined} reports Cosmos 3 image-to-video results on RBench -- Cosmos 3 emerges as the best open-source model.

As shown in \cref{tab:paibench_rbench_results_combined}, Cosmos3-Super achieves the highest overall PAIBench-G scores on both T2V and I2V across all models, including closed-source, while Cosmos3-Nano matches the second best result on RBench. Both Cosmos 3 variants also significantly outperform their Cosmos-Predict2.5 predecessors, reflecting the gains from the omnimodal architecture. While both benchmarks incorporate physical plausibility into their scoring, they do not probe it in depth. We therefore turn to Physics-IQ for a targeted evaluation of physics adherence.

\paragraph{Physics-IQ.} Physics-IQ~\citep{motamed2025generative} evaluates whether video generation models capture physical principles by conditioning on a real-world starting context and scoring how closely the generated continuation matches an actual physical outcome. Unlike PAIBench-G and RBench, which incorporate physical plausibility as one scoring component among many, Physics-IQ isolates it as the sole evaluation axis, measuring whether generated motion matches ground-truth physical outcomes along precise spatial and temporal dimensions.
Physics-IQ covers five physics categories---solid mechanics, fluid dynamics, optics, thermodynamics, and magnetism---across 396 real-world scenes captured from three fixed viewpoints, with paired textual descriptions for text-conditioned models.
Two evaluation modes are supported.
In \emph{image-to-video} (I2V), the model is conditioned on a single \emph{switch frame} plus an optional text prompt and predicts the subsequent motion.
In \emph{video-to-video} (V2V) continuation, the model is conditioned on a 3\,s conditioning video plus an optional text prompt and predicts the next 5\,s of motion.
Generated videos are compared against ground-truth physical continuations along four complementary axes (spatial overlap, temporal alignment, magnitude-weighted spatial agreement, and pixel-level error), normalized by a real-vs-real upper bound into a single 0–100 score directly comparable across models and conditioning modes.

\begin{table*}[t]
    \centering
    \captionsetup{justification=raggedright, singlelinecheck=false}
    \caption{\textbf{Physics-IQ benchmark results, separated by conditioning modes}. WMReward (BoN) denotes best-of-$N$ reranking with a WMReward~\citep{yuan2026wmreward}. Higher Physics-IQ score is better. \textbf{Cosmos3-Super} achieved state-of-the-art results for both I2V and V2V, with and without using the WMReward+BoN. \textbf{Bold} indicates best in the column group and \underline{underline} indicates the second best.}
    \label{tab:physics_iq_results}
    \footnotesize
    \setlength{\tabcolsep}{4pt}
    \renewcommand{\arraystretch}{1.15}
    \begin{subtable}{0.48\linewidth}
        \centering
        \caption{Image-to-Video (I2V).}
        \label{tab:physics_iq_i2v}
        \resizebox{\linewidth}{!}{%
        \begin{tabular}{l | l c}
            \toprule
            \textbf{Model} & \textbf{Mode} & \textbf{Score ($\uparrow$)} \\
            \midrule
            \rowcolor{rowours}
            \textbf{Cosmos3-Super} & I2V + WMReward (BoN) & \textbf{48.9} \\
            \rowcolor{rowours}
            \textbf{Cosmos3-Nano}  & I2V + WMReward (BoN) & 43.8 \\
            Sora2 (Closed-sourced)                         & I2V + WMReward (BoN) & \underline{46.4} \\
            Wan2.2-A14B (Open-sourced)                       & I2V + WMReward (BoN) & 44.4 \\
            \midrule
            \rowcolor{rowours}
            \textbf{Cosmos3-Super} & I2V                  & \textbf{43.8} \\
            \rowcolor{rowours}
            \textbf{Cosmos3-Nano}  & I2V                  & 40.2 \\
            Sora2 (Closed-sourced)                        & I2V                  & \underline{42.3} \\
            Wan2.2-A14B (Open-sourced)                       & I2V                  & 38.3 \\
            \bottomrule
        \end{tabular}
        }
    \end{subtable}
    \hfill
    \begin{subtable}{0.48\linewidth}
        \centering
        \caption{Video-to-Video (V2V).}
        \label{tab:physics_iq_v2v}
        \resizebox{\linewidth}{!}{%
        \begin{tabular}{l | l c}
            \toprule
            \textbf{Model} & \textbf{Mode} & \textbf{Score ($\uparrow$)} \\
            \midrule
            \rowcolor{rowours}
            \textbf{Cosmos3-Super} & V2V + WMReward (BoN) & \textbf{63.4} \\
            \rowcolor{rowours}
            \textbf{Cosmos3-Nano}  & V2V + WMReward (BoN) & 57.7 \\
            Magi-1 (Open-sourced)                       & V2V + WMReward (BoN) & \underline{62.6} \\
            \midrule
            \rowcolor{rowours}
            \textbf{Cosmos3-Super} & V2V                  & \textbf{59.7} \\
            \rowcolor{rowours}
            \textbf{Cosmos3-Nano}  & V2V                  & 50.2 \\
            Magi-1 (Open-sourced)                       & V2V                  & \underline{56.0} \\
            Video-GPT (Open-sourced)                    & V2V                  & 35.0 \\
            VideoPoet (Closed-sourced)                     & V2V                  & 29.5 \\
            \bottomrule
        \end{tabular}
        }
    \end{subtable}
\end{table*}

We evaluate Cosmos3-Super in both I2V and V2V modes in \cref{tab:physics_iq_results} alongside leading open-source and commercial baselines in both modes.
For V2V, we use the full 3\,s conditioning video as input.
We also use a prompt upsampler, following a similar iterative-refinement strategy to PhyT2V~\citep{xue2025phyt2v}, to generate text prompts from the switching frame in the I2V setting and from the 3\,s conditioning video in the V2V setting.
Following the public Physics-IQ leaderboard and the WMReward inference-time alignment protocol~\citep{yuan2026wmreward}, we additionally run WMReward + best-of-$N$ (BoN) scores for Cosmos3-Super in both I2V and V2V.
From \cref{tab:physics_iq_results}, we find that Cosmos3-Super achieves state-of-the-art in I2V: its direct score of $43.8$ exceeds the direct I2V baselines, and WMReward+BoN further improves the score to $48.9$, above the strongest listed I2V baseline with WMReward+BoN.
In V2V, Cosmos3-Super also achieves the state-of-the-art, reaching $59.7$ directly and $63.4$ with WMReward + BoN, above the strongest listed V2V baseline with WMReward+BoN.

While automated evaluation metrics are valuable for measuring quality, prompt alignment, and physical plausibility, even the strongest VLM-based metrics miss certain artifacts and failure cases. Human evaluation complements automated metrics on two fronts: coverage of long-tail failures that automated protocols systematically miss, and discriminability across a wider scoring range. For instance, on the same prompt set, the evaluated T2V generators span $\sim$10 points on human evaluation versus $\sim$4 points on the comparable automated PAIBench-G overall score (see Appendix~\ref{sec::cosmos_hue_bench}). We therefore complement our automated results with two human evaluation protocols: Cosmos HUE, which targets broad Physical AI video generation, and Human World Bench (HWB), which focuses specifically on realistic human motion under task-level instructions.

\paragraph{Cosmos HUE.} Cosmos HUE (HUman Evaluation) is a human scoring protocol grounded in the same PAIBench-G prompt set (\cref{subsubsec::paibench_g}). HUE departs from prior
human-eval protocols in two ways. First, it replaces subjective Likert-scale grading with \emph{atomic binary} verification: each video is decomposed into a set of single-fact \emph{Yes / No / Unclear} questions phrased so that ``Yes'' always denotes the desirable outcome, shifting the annotator's task from holistic judgment to objective fact-verification. Each (video, question) pair is independently rated by two annotators, with disagreements escalated to a quality-control reviewer.   Second, the per-prompt question set is \emph{automatically generated by a three-layer VLM pipeline}: a Domain Strategist classifies the prompt into one of seven Physical AI domains, a Scene Parser produces a structured scene manifest from the prompt (T2V) or from sampled frames of the paired ground-truth reference video (I2V), and an Auditor emits the final atomic questions. This ensures that questions track the actual prompt and reference scene rather than a static rubric. All three layers run on GPT~5.2. The VLM-generated question set is then supplemented by a human review pass in which reviewers inspect generated videos from top-ranked models and propose additional questions targeting failure modes the automated pipeline does not yet cover; the proposed questions are folded back into the question bank. The evaluation set consists of 100 fixed prompts sampled from PAIBench-G preserving its native domain distribution, with each model producing 5 random-seed generations per prompt for 500 videos per model; the per-prompt question set (up to 20 binary questions) is applied to all 5 generated videos. Each question is classified into one of four dimensions: ``Semantic Alignment'', ``Physical Laws'', ``Geometric Reasoning'', and ``Visual Integrity''. Per-dimension and per domain T2V and I2V leaderboards, the formal scoring scheme, and reliability estimates are detailed in Appendix~\ref{sec::cosmos_hue_bench}.

\begin{table*}[t]
  \centering
  \captionsetup{justification=raggedright, singlelinecheck=false}
  \caption{\textbf{Human evaluation results on Cosmos HUE and Human World Bench (HWB).} Cosmos HUE evaluates broad Physical AI video generation quality via atomic binary verification
across four dimensions; HWB evaluates realistic human motion under task-level instructions via instruction-following and physics pass rates. \emph{Ground Truth} scores real videos
paired with the same prompts and serves as an upper reference for Cosmos-HUE. \textbf{Cosmos3-Super} achieves the highest HUE T2V score and the highest HWB score among open-source models. Full
per-dimension HUE leaderboards in Appendix~\ref{sec::cosmos_hue_bench}. \textbf{Bold} indicates the best in the column; \underline{underline} indicates the second best.}
  \label{tab::human_eval_results}
  \footnotesize
  \setlength{\tabcolsep}{4pt}
  \renewcommand{\arraystretch}{1.15}
  \resizebox{0.85\linewidth}{!}{%
  \begin{tabular}{l l | cc|c}
      \toprule
      & & \multicolumn{2}{c}{\textbf{Cosmos HUE}} & \textbf{Human World Bench} \\
  \cmidrule(lr){3-4} \cmidrule(lr){5-5}
      \textbf{Model} & \textbf{Type} & \textbf{Text-to-Video ($\uparrow$)} & \textbf{Image-to-Video ($\uparrow$)} & \textbf{Image-to-Video ($\uparrow$)} \\
      \midrule
      \rowcolor{black!8}
      \emph{Ground Truth}              & ---          &    93.6       &   94.4        &    ---    \\
      \midrule
      \rowcolor{rowours}
      \textbf{Cosmos3-Super}    & Open-sourced & 89.3  &       \underline{89.6}          & \textbf{71.9} \\
      \rowcolor{rowours}
      \textbf{Cosmos3-Nano}     & Open-sourced & 87.6          & 88.6          & 66.9 \\
      \midrule
      Wan2.2-A14B                      & Open-sourced &  88.2         &  88.4         & 60.7 \\
      HunyuanVideo-1.5                 & Open-sourced &  86.5         &  85.6         & 54.7 \\
      Wan2.1-14B                       & Open-sourced &  84.0         &  83.9         & 33.1 \\
      Wan2.2-5B                        & Open-sourced &  80.8         &  80.4         & 25.4 \\
      Cosmos-Predict2.5-14B     & Open-sourced &  82.1         &    83.0       & 38.7 \\
      Cosmos-Predict2.5-2B      & Open-sourced &     81.8      &   82.6        & 32.8 \\
      \midrule
      Veo-3.1                          & Closed-sourced   &   \textbf{91.3}       & \textbf{89.7}  & \underline{67.8} \\
      Seedance-1.5-Pro                 & Closed-sourced   &    \underline{90.0}       & 87.6          &  ---     \\
      \bottomrule
  \end{tabular}
  }
\end{table*}

\paragraph{Human World Bench (HWB).} While Cosmos HUE evaluates broad Physical AI video generation, Human World Bench (HWB) focuses on a more targeted and challenging setting: egocentric image-to-video generation for human manipulation tasks under task-level instructions.
The benchmark videos in HWB are sourced from EgoVerse~\citep{punamiya2026egoverse} and comprise 180 samples.
HWB uses an absolute failure-mode protocol: annotators judge each generated video independently for instruction following and physical plausibility. \label{subsec::hwb}
We report two top-level pass rates.
\emph{Instruction following} measures whether the video depicts the requested actions and objects.
\emph{Physical plausibility} measures temporal coherence and physical plausibility, including object dynamics, contact, and hand anatomy.
The HWB score is the average of the instruction-following and physics pass rates.

\cref{tab::human_eval_results} reports Cosmos-HUE scores on both T2V and I2V, alongside Human World Bench (HWB) results. On HUE T2V, Cosmos3-Super is the best open-source model at 89.3, behind the closed-source Veo-3.1 (91.3) and Seedance-1.5-Pro (90.0). On HUE I2V, Cosmos3-Super is again the best open-source model and is essentially tied with the leading closed-source generator, trailing Veo-3.1 by only 0.1 points (89.6 vs. 89.7). Cosmos3-Nano is also competitive, finishing second among open-source models on I2V (88.5) and third on T2V(87.6).

On HWB, Cosmos3-Super achieves 71.9, the state-of-the-art score among all evaluated models in \cref{tab::human_eval_results}, outperforming the strongest listed closed-source baseline, Veo-3.1 (67.8), by 4.1 points and the strongest non-Cosmos open-source baseline, Wan2.2-A14B (60.7), by 11.2 points.
Cosmos3-Nano also performs strongly at 66.9, ranking second among open-source models and outperforming every non-Cosmos open-source baseline.
Together, the two Cosmos 3 variants take the top two open-source positions on HWB, demonstrating strong egocentric human-motion generation across both model scales.

\begin{figure}[t]
    \centering
    \includegraphics[width=\linewidth]{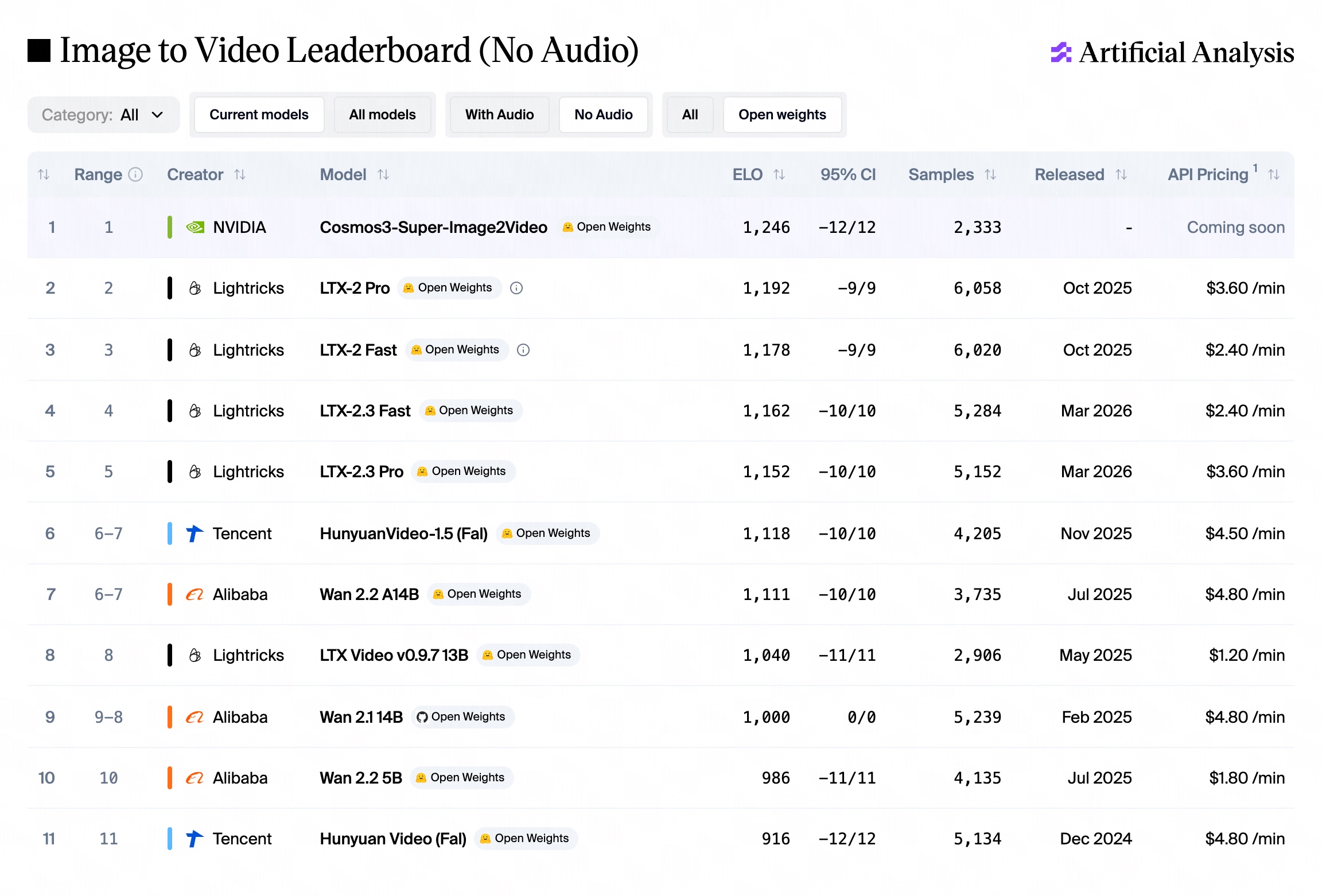}
    \caption{\textbf{Cosmos3-Super-Image2Video is the best open-weight model on crowdsourced arena rankings.} \textbf{Cosmos3-Super-Image2Video} ranked \#1 among open-weight models (\#22 including proprietary models) on the Artificial Analysis Image to Video Leaderboard (No Audio) (Date: 2026-05-28).}
    \label{fig:sft_aa_leaderboard}
\end{figure}

\paragraph{Artificial Analysis Image-to-Video Leaderboard.} To evaluate Cosmos 3 under real-world scenarios, we submitted our specialized I2V model, Cosmos3-Super-Image2Video, to the Artificial Analysis Image-to-Video leaderboard (No Audio) for crowdsourced public voting. The model ranked the top among all open-weight models, and achieved the \#22 position globally including proprietary models. In particular, the model is on par with proprietary offerings such as Veo 3.1 \citep{veo31} and Wan 2.5 \citep{wan2pt2}, demonstrating its exceptional temporal dynamics and visual fidelity.

\FloatBarrier
\subsubsection{Audio Generation Evaluation}
\label{subsec::audio_eval}
Text-to-audiovisual generation must satisfy two requirements that are not captured by video-only benchmarks. The generated audio should contain the sound events requested by the prompt, and those events should be attributable to the correct visual sources with plausible timing. We therefore evaluate audio generation with Cosmos-SoundBench, a targeted benchmark for audio-visual prompt following and synchronization. The metric separates semantic audio-visual correctness from prompt-blind audio fidelity. A sample can contain the right sound at the right moment but still suffer from audio artifacts, or it can be acoustically clean while missing the requested event.

\paragraph{Cosmos-SoundBench.}
Cosmos-SoundBench contains 144 evaluation prompts drawn from FoleyBench~\citep{foleybench}. The prompts cover non-speech sound categories such as ambient scenes, impacts, object interactions, tools, vehicles, water, and other environmental effects. We focus on non-speech audio because these signals are especially important for Physical AI. Contact sounds reveal material and collision properties, tool sounds indicate ongoing actions, and ambient cues provide scene context that may be only partially visible.

\paragraph{Audiovisual quality metric (AVQ).}
We evaluate each generated sample with a structured MLLM-as-judge protocol. The judge sees the prompt, video, and audio only in the stages where that information is needed, keeping each judge stage focused on the intended evidence rather than unrelated cross-modal cues.

\begin{enumerate}
    \item \textit{Prompt checklist construction.} Before inspecting any generated media, an ensemble of Claude Opus~4.7, Gemini~3.1 Pro, and GPT~5.5 extracts the required foreground sounds, ambient audio conditions, and prompt-critical visual evidence. We use majority voting to fix the checklist used by downstream scoring.
    \item \textit{Prompt-blind visual observation.} Gemini~3.1 Pro Preview describes visible entities, actions, scene context, and plausible sound sources without seeing the prompt. These observations provide evidence for source attribution and temporal alignment, rather than a standalone visual-quality score.
    \item \textit{Semantic audiovisual scoring.} The judge evaluates whether the requested sounds are present, source-specific, dynamically appropriate, and aligned with visible or plausible actions. These checks are mapped to semantic audio correctness (\textbf{SA}), audiovisual alignment (\textbf{AVAlign}), and prompt-critical visual support (\textbf{VisualSupport}), combined as:
    \[
        \mathrm{SAV}=0.60\,\mathrm{SA}+0.30\,\mathrm{AVAlign}+0.10\,\mathrm{VisualSupport}.
    \]
    We run this stage three times and average the scores to account for judge variability.
\end{enumerate}

We additionally measure audiobox-aesthetics Production Quality (PQ) as a proxy for perceptual audio quality, as it tries to quantify sample-level clarity \& fidelity, dynamics, frequency bandwidth and spatialization~\citep{tjandra2025audioboxaesthetics}

The final audiovisual quality score is:
\[
    \mathrm{AVQ}=0.5\,\mathrm{SAV}+0.5\,\mathrm{AQ}.
\]
For each model, we evaluate five random-seed generations per prompt. For Cosmos 3 models, we use Claude Opus~4.6 to rewrite SoundBench prompts into the structured generation format described in \cref{sec::gen_guide} and \cref{sec::prompt_upsampling}, matching the prompt distribution used by the generator.

\begin{table*}[t]
    \centering
    \captionsetup{justification=raggedright, singlelinecheck=false}
    \caption{\textbf{Cosmos-SoundBench Audiovisual Quality.}
    We report the overall Audiovisual Quality (AVQ), Semantic Audiovisual quality (SAV), its sub-scores defined in~\cref{subsec::audio_eval}, and audiobox-aesthetics Production Quality (PQ).
    \textbf{Bold} marks the best score in each column and \underline{underline} marks the second best.
    Seedance-1.5-Pro achieves the highest AVQ through stronger PQ, while Cosmos 3 achieves the strongest semantic audio-visual grounding and alignment.}
    \label{tab:cosmos_soundbench_results}
    \footnotesize
    \setlength{\tabcolsep}{4pt}
    \renewcommand{\arraystretch}{1.15}
    \begin{tabular}{l l | cccccc}
        \toprule
        & & \multicolumn{6}{c}{\textbf{SoundBench Audiovisual Quality ($\uparrow$)}} \\
        \cmidrule(lr){3-8}
        \textbf{Model} & \textbf{Type} & \textbf{AVQ} & \textbf{SAV} & \textbf{SA} & \textbf{AVAlign} & \textbf{Visual Sup.} & \textbf{PQ}\\
        \midrule
        \rowcolor{rowours}
        \textbf{Cosmos3-Super} & Open-sourced & 7.31 & \underline{8.34} & \underline{8.30} &
  \underline{8.14} & \textbf{9.18} & 6.28 \\
          \rowcolor{rowours}
          \textbf{Cosmos3-Nano} & Open-sourced & 7.34 & \textbf{8.35} & \textbf{8.33} &
  \textbf{8.16} & \underline{9.10} & 6.32 \\
          \midrule
          LTX-2.3 & Open-sourced & 7.10 & 7.80 & 7.86 & 7.58 & 8.12 & 6.39\\
          \midrule
          Seedance-1.5-Pro & Closed-sourced & \textbf{7.64} & 8.21 & 8.22 & 8.06 & 8.61  & \textbf{7.06} \\
          Veo-3.1 & Closed-sourced & \underline{7.45} & 8.21 & 8.21 & 8.01 & 8.85 & 6.68 \\
          LTX-2.3 Pro & Closed-sourced & 7.32 & 7.93 & 7.96 & 7.74 & 8.35 & \underline{6.70} \\
          Wan2.6 & Closed-sourced & 7.23 & 7.90 & 7.99 & 7.54 & 8.45 & 6.55 \\
          Sora 2 & Closed-sourced & 6.90 & 7.94 & 7.97 & 7.70 & 8.49 & 5.85 \\
        \bottomrule
    \end{tabular}
\end{table*}

\begin{figure}[t]
    \centering
    \includegraphics[width=0.99\linewidth]{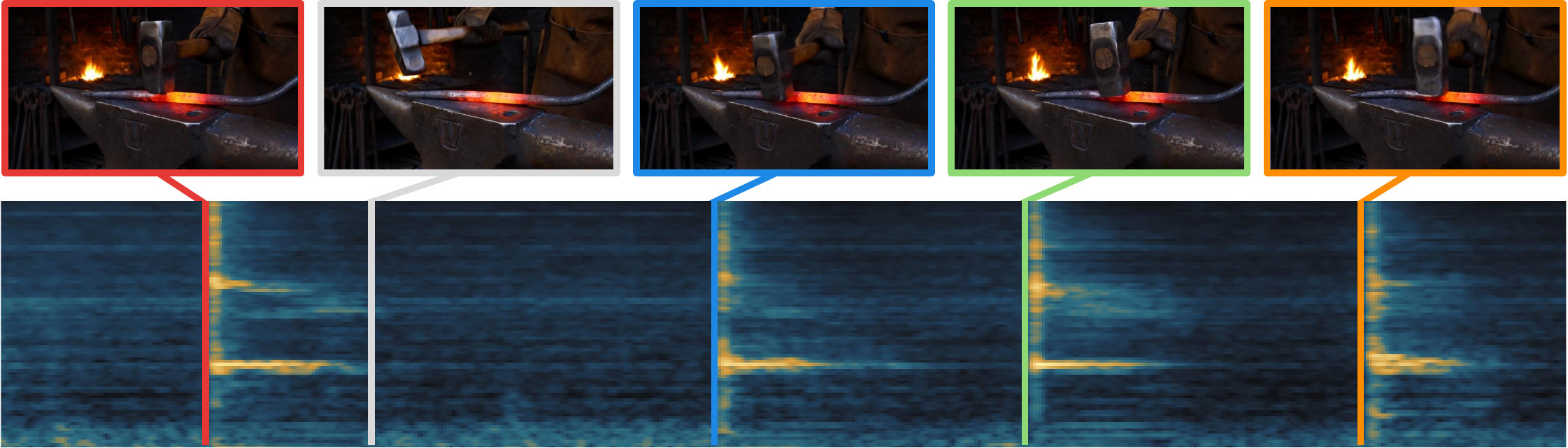}
    \caption{\textbf{Audio-video event alignment.} Selected frames from a \textbf{Cosmos3-Nano} generation are paired with the spectrogram of the generated audio. Colored frames denote hammer-strike moments, and their temporal markers coincide with sharp spectral transients. The gray frame shows a non-contact moment between strikes, where no comparable acoustic transient is observed. This contrast provides qualitative evidence that the acoustic transients are aligned with visual impact moments rather than intervening motion.}
    \label{fig:av_alignment_example}
\end{figure}

\cref{tab:cosmos_soundbench_results} shows that the strongest closed-source systems retain an advantage in overall AVQ, primarily through higher perceptual audio quality. Seedance-1.5-Pro achieves the best AVQ ($7.64$) and PQ ($7.06$), followed by Veo-3.1 ($7.45$ AVQ, $6.68$ AQ). In contrast, Cosmos 3 is strongest on the semantic and alignment components that measure whether the sound matches the visual event. Cosmos3-Nano obtains the best SAV, SA, and AVAlign scores, while Cosmos3-Super achieves the best visual-support score. This pattern suggests that Cosmos 3 mid-training is effective at grounding sound events in the generated video, with remaining headroom concentrated in low-level audio fidelity. \cref{fig:av_alignment_example} demonstrates an example of strong temporal alignment between visual and acoustic events in Cosmos3-Nano generations.

\FloatBarrier
\subsubsection{Transfer Generation Evaluation}
\label{subsec::transfer_eval}

\paragraph{Two-weight classifier-free guidance.}
Video transfer is conditioned on both a structured text prompt and a control video, and the optimal balance between caption fidelity and structural adherence varies across modalities. To control these two factors independently, we use a two-weight classifier-free guidance scheme with separate weights, one for the control video and one for the text prompt.

At each denoising step, we evaluate the denoiser three times---with both conditions, with the prompt only (dropping the control video), and with the control video kept but the prompt replaced by a fixed negative caption. We then combine the predictions so that the control weight extrapolates from the prompt-only prediction toward the fully conditional one (strengthening structural control), while the text weight extrapolates away from the negative-prompt prediction (strengthening caption fidelity). Exposing the two weights as separate knobs lets us tune each modality to its own quality--fidelity sweet spot, and we find this factored guidance to be more effective than the standard single-guidance formulation.

\paragraph{General video transfer.}
We evaluate video transfer on PAIBench-C~\citep{cosmos_transfer1}, a control-conditioned video-to-video benchmark covering four spatial control modalities: blur, edge, segmentation, and depth. The benchmark contains $600$ clips spanning three Physical AI domains: $200$ robotic-arm manipulation clips from AgiBot World~\citep{bu2025agibot}, $200$ driving clips from OpenDV~\citep{yang2024genad}, and $200$ egocentric everyday-life clips from Ego-Exo-4D~\citep{grauman2024egoexo4d}.

For each example, the model receives a text prompt and a control video and must generate a photorealistic video that follows the control while preserving the prompted scene semantics. We report the single-control setting (exactly one modality per example) so each modality's contribution can be measured in isolation. Quality is assessed by re-extracting the relevant control signal from the generated and reference videos and comparing them in the corresponding space (blur SSIM after bilateral filtering, Canny edge F1, scale-invariant RMSE on estimated depth, mIoU on open-vocabulary segmentation masks), plus DOVER~\citep{dover} as a content-agnostic measure of perceptual realism.

We compare Cosmos 3 against the Cosmos-Transfer2.5 baseline, which handles different modalities with a dedicated ControlNet~\citep{Zhang_2023_ICCV} branch per modality. Note that Cosmos 3 is instead a unified model that consumes the control video alongside the text prompt in its input sequence, natively supporting any combination of modalities without per-modality ControlNet adapters.

\cref{tab:paibenchc_results} shows that, despite the collapse of four dedicated ControlNet branches into a single unified backbone, Cosmos 3 matches or surpasses Cosmos-Transfer2.5 in every modality
through one of its two variants: Cosmos3-Nano leads on perceptual
quality (DOVER) and segmentation, while Cosmos3-Super takes the
geometrically demanding edge and depth tasks. All three models are effectively on par on blur SSIM, which is already saturated near the metric's upper bound. These results
indicate that per-modality ControlNet adapters are not a
prerequisite for strong control fidelity---a single unified
backbone natively supports all four spatial controls and beats the
dedicated-adapter baseline at one of its two scales on every metric.

\begin{table}[t]
    \centering
    \captionsetup{justification=justified, singlelinecheck=false}
    \caption{\textbf{PAIBench-C single-control results.} Each generation is conditioned on one of four control modalities (depth, segmentation, blur, or edge) and scored by the corresponding ground-truth-based metric: \textbf{Depth si-RMSE}, the scale-invariant root-mean-squared error between predicted and reference depth; \textbf{Seg.\ mIoU}, the mean intersection-over-union between predicted and reference segmentation masks; \textbf{Blur SSIM}, the structural similarity between predicted and reference blur maps; and \textbf{Edge F1}, the F1 score between predicted and reference edge maps. \textbf{DOVER} is a no-reference perceptual video-quality score, averaged across the four per-modality generations. Depth si-RMSE: lower is better; DOVER, Seg.\ mIoU, Blur SSIM, and Edge F1: higher is better. \textbf{Bold} indicates the best result per column.}
    \label{tab:paibenchc_results}
    \setlength{\tabcolsep}{6pt}
    \resizebox{0.8\linewidth}{!}{%
    \begin{tabular}{l|ccccc}
        \toprule
        \textbf{Model} & \textbf{DOVER} $\uparrow$ & \textbf{Seg.\ mIoU} $\uparrow$ & \textbf{Blur SSIM} $\uparrow$ & \textbf{Edge F1} $\uparrow$ & \textbf{Depth si-RMSE} $\downarrow$ \\
        \midrule
        \rowcolor{rowours}
        \textbf{Cosmos3-Super} & 10.14 & 0.71 & \textbf{0.91} & \textbf{0.50} & \textbf{0.58} \\
        \rowcolor{rowours}
        \textbf{Cosmos3-Nano} & \textbf{10.39} & \textbf{0.72} & \textbf{0.91} & 0.49 & 0.62 \\
        Cosmos-Transfer2.5 & 9.49 & 0.68 & 0.90 & 0.45 & {0.68} \\
        \bottomrule
    \end{tabular}
    }
\end{table}
\begin{figure}[t]
    \centering
    \includegraphics[width=0.99\linewidth]{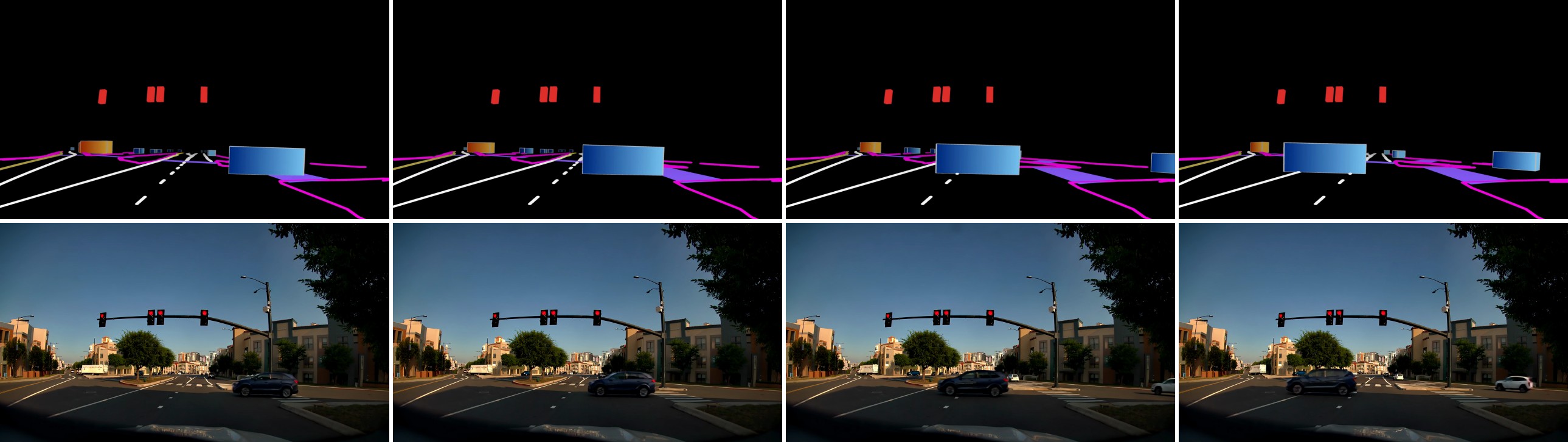}
    \caption{\textbf{Driving scene Video Transfer results.} \textbf{Cosmos3-Nano} generates frames (bottom) from the corresponding 720p control video (top). The control video encodes HD map elements---lanes, road markings, poles, and traffic lights (with or without state)---which together represent complex road topologies (including overpasses), as well as actors represented as cuboids. Each cuboid is color-coded by a coarse class ontology (\eg, truck, vehicle, pedestrian) and shaded to differentiate front from back.}
    \label{fig:av_transfer_example}
\end{figure}

\paragraph{Autonomous driving.}
PAIBench-C is complemented by an AV-specific benchmark of $486$ single-view driving clips~\citep{cosmos_predict2p5} for world-scenario-map-conditioned video transfer, called AVBench-C. The control input
is a camera-view rendering of the driving scene that combines static map structure (lane lines, road boundaries, traffic signals, and traffic signs) with all scene objects (vehicles, pedestrians, and other agents), both stationary and moving. Given this rendered scenario together with a text prompt, the model must produce a photorealistic driving video that is consistent with both inputs. We evaluate the results with both automatic and human evaluations:
\begin{itemize}
\item \textit{AVBench-C automatic evaluation.} We evaluate each generation with three ground-truth-based automatic checkers. The \emph{egomotion} checker uses visual odometry to compute trajectory drift per meter travelled. The \emph{object-correspondence} checker matches scene entities to the control input, comparing dynamic actors against rendered trajectories and static structure against the projected map. The \emph{environment} checker uses a VLM judge to score prompt-level driving conditions, including weather, time of day, region, and road-surface state.
\item \textit{AVBench-C human evaluation.} Trained annotators rate each generated video on a 1--3 scale along two axes. The \emph{video quality} axis measures overall realism, with 3 indicating realistic textures and temporally consistent agent behavior, 2 indicating mild deformities or brief temporal stutter, and 1 indicating clearly unrealistic dynamics or low-quality textures. The \emph{lane line} axis measures fidelity to the world-scenario map: annotators view the generation alongside a reference overlay of the expected road layout and assign 3 if lanes, crossings, and entities are mostly in the correct locations, 2 if a few lanes are missing but most entities are correctly placed, and 1 if most lanes or crossings are missing or entities are misplaced.
\end{itemize}

\begin{table}[t]
    \centering
    \captionsetup{justification=justified, singlelinecheck=false}
    \caption{\textbf{AVBench-C evaluation scores}. We report both automatic and human evaluation results. \textit{Auto scores} are computed by three ground-truth-based checkers: \textbf{Ego drift} (lower is better), the visual-odometry trajectory drift; \textbf{Dyn.\ Obj.} and \textbf{Static Obj.} (higher is better), how closely scene entities match the control input; and \textbf{Environment} (higher is better), a VLM-judge score for prompt-level driving conditions. \textit{Human scores} are 1--3 ratings from trained annotators along two axes: \textbf{Video quality}, which measures overall realism (textures, agent behavior, and temporal consistency), and \textbf{Lane line}, which measures fidelity to the world-scenario map (correct placement of lanes, crossings, and entities). Higher is better for both human metrics. \textbf{Bold} indicates the best result per column.}
    \label{tab:avbenchc_transfer}
    \setlength{\tabcolsep}{6pt}
    \resizebox{\linewidth}{!}{%
    \begin{tabular}{l|cccc|cc}
        \toprule
        & \multicolumn{4}{c}{\textbf{Automatic evaluation}} & \multicolumn{2}{c}{\textbf{Human evaluation}} \\
        \cmidrule(lr){2-5} \cmidrule(lr){6-7}
        \textbf{Model} & \textbf{Ego drift} $\downarrow$ & \textbf{Dyn.\ Obj.} $\uparrow$ & \textbf{Static Obj.} $\uparrow$ & \textbf{Environment} $\uparrow$ & \textbf{Video quality} $\uparrow$ & \textbf{Lane line} $\uparrow$ \\
        \midrule
        \rowcolor{rowours}
        \textbf{Cosmos3-Super} & \textbf{0.003} & 0.64 & 0.41 & \textbf{0.90} & \textbf{2.86} & 2.45 \\
        \rowcolor{rowours}
        \textbf{Cosmos3-Nano}  & \textbf{0.003} & \textbf{0.67} & 0.41 & \textbf{0.90} & 2.82 & \textbf{2.50} \\
        Cosmos-Transfer2.5-AV-Singleview & 0.008 & 0.62 & \textbf{0.42} & \textbf{0.90} & 2.59 & 2.47 \\
        \bottomrule
    \end{tabular}
    }
\end{table}

\cref{tab:avbenchc_transfer} shows that Cosmos 3 matches or surpasses Cosmos-Transfer2.5 on every driving-scene metric through one of its two variants. On the automatic side, ego-trajectory drift is uniformly small across all three models, and Cosmos3-Nano further leads on dynamic-object correspondence, while static structure and environment consistency are effectively on par and already saturated near the metric's upper bound. Human evaluation reinforces this picture: Cosmos3-Super and Cosmos3-Nano deliver markedly higher video quality ($2.86$ and $2.82$ vs.\ $2.59$), while lane-line fidelity is comparable across all three models (within $\pm 0.05$). Taken together, these results indicate that Cosmos 3 preserves geometric and structural fidelity while delivering visibly higher-quality driving-scene generations, beating the baseline at one of its two scales on every metric. \cref{fig:av_transfer_example} shows a qualitative example of a driving-scene video generated by Cosmos3-Nano from its corresponding control input.

\FloatBarrier
\subsubsection{Action Generation Evaluation}
\label{subsec::action_benchmarks}

We evaluate whether action mid-training endows Cosmos 3 with a reusable world-action prior across domains and inference modes.
The central question is whether unified action mid-training accelerates adaptation to a domain-specific action interface, and whether a short post-training stage can turn the shared base model into a specialized model that is competitive with or stronger than state-of-the-art domain baselines.
\Cref{tab:action_posttraining_summary_wide} reports results on camera motion, autonomous driving, robotics, and egocentric motion domains, covering forward- and inverse-dynamics settings.
\Cref{tab:action_posttraining_policy} reports results on robot policy settings, evaluating Cosmos 3's ability to complete language-specified tasks.

\paragraph{Setup.} We compare two initialization protocols for each downstream domain. The first, pre-training initialization (\textit{PT-init}), starts from the Cosmos 3 pre-trained checkpoint, which has not been trained on action-domain data. The second, mid-training initialization (\textit{MT-init}), starts from our mid-trained checkpoint, which has seen action data spanning multiple domains and prediction modes, including forward dynamics (FD), inverse dynamics (ID), and policy. For each comparison, we use the same training recipe and keep the model size, data, and compute budget fixed. We report results for both Cosmos3-Nano and Cosmos3-Super.

\paragraph{Metrics.} We report metrics for each action interface. For autonomous-vehicle inverse dynamics, relative rotation error (RRE) and relative translation error (RTE) measure frame-to-frame pose accuracy, while absolute trajectory error (ATE) measures global trajectory consistency. We also use these metrics to measure camera-following accuracy in camera-motion forward dynamics, comparing the ground-truth camera poses with the estimated camera trajectories from generated videos. For robotics and egocentric forward dynamics, PSNR measures the reconstruction quality of action-conditioned future observations. Although a plausible generative rollout need not be pixel-identical to the single recorded ground-truth future, we find that PSNR is as a useful proxy for temporal alignment, motion consistency, and reconstruction fidelity under a relatively short temporal horizon. For robotics policy, success rate measures the percentage of evaluation tasks completed successfully.

\begin{table}[t]
    \centering
    \caption{\textbf{Post-training comparisons for forward and inverse dynamics across domains.} FD denotes forward dynamics and ID denotes
  inverse dynamics. \textit{PT-init} initializes from the generic Cosmos 3 pre-trained checkpoint, while \textit{MT-init} initializes from
  the mid-trained checkpoint. Domain-specific baselines are reported only for their corresponding application, while Cosmos 3 variants are compared across all evaluated action settings. \textbf{Bold} indicates the best in the column and
  \underline{underline} indicates the second best.}
  \label{tab:action_posttraining_summary_wide}
    \small
    \setlength{\tabcolsep}{5pt}
    \resizebox{\linewidth}{!}{%
    \begin{tabular}{@{}l|ccc|ccc|c|c@{}}
        \toprule
        & \multicolumn{3}{c|}{\textbf{Autonomous Vehicle (ID)}} & \multicolumn{3}{c|}{\textbf{Camera Motion (FD)}} & \multicolumn{1}{c|}{\textbf{Egocentric Motion (FD)}} & \textbf{Robotics (FD)} \\
        \cmidrule(lr){2-4} \cmidrule(lr){5-7} \cmidrule(lr){8-8} \cmidrule(lr){9-9}
        \textbf{Model} & RRE ($^\circ$, $\downarrow$) & RTE (m, $\downarrow$) & ATE (m, $\downarrow$) & RRE ($^\circ$, $\downarrow$) & RTE (m, $\downarrow$) & ATE (m, $\downarrow$) & PSNR ($\uparrow$) & PSNR ($\uparrow$) \\
        \midrule
        \rowcolor{rowours}
        \textbf{Cosmos3-Super} (\textit{MT-init}) & \underline{0.232} & \textbf{0.014} & \textbf{0.90} & \textbf{0.142} & \textbf{0.026} & \textbf{0.99} & \textbf{16.19} & \textbf{26.04} \\
        \rowcolor{rowours}
        \textbf{Cosmos3-Nano} (\textit{MT-init})  & \textbf{0.211} & \textbf{0.014} & \underline{0.98} & \underline{0.147} & \underline{0.029} & \underline{1.24} & \underline{16.12} & \underline{25.52} \\
        \midrule
        \rowcolor{rowours}
        \textbf{Cosmos3-Super} (\textit{PT-init}) & 0.284 & 0.018 & 1.32 & 0.293 & 0.036 & 1.82 & 15.34 & 22.69 \\
        \rowcolor{rowours}
        \textbf{Cosmos3-Nano} (\textit{PT-init})  & 0.249 & \underline{0.017} & 1.20 & 0.172 & 0.034 & 1.61 & 15.22 & 23.24 \\
        \midrule
        Lingbot-World  & --    & --    & --    & 0.299 & 0.057 & 2.88 & --    & --    \\
        HY-World1.5 & --    & --    & --    & 0.377 & 0.042 & 1.39 & --    & --    \\
        \midrule
        VGGT           & 0.596 & 0.768 & 23.46 & --    & --    & --   & --    & --    \\
        DepthAnything3 & 0.312 & 0.354 & 9.29  & --    & --    & --   & --    & --    \\
        \midrule
        LOME       & -- & -- & -- & -- & -- & -- & 9.36 & --    \\
        \midrule
        Ctrl-World & -- & -- & -- & -- & -- & -- & --   & 22.99 \\
        \bottomrule
    \end{tabular}
    }
\end{table}

\begin{figure}[t]
    \centering
    \includegraphics[width=0.99007\linewidth]{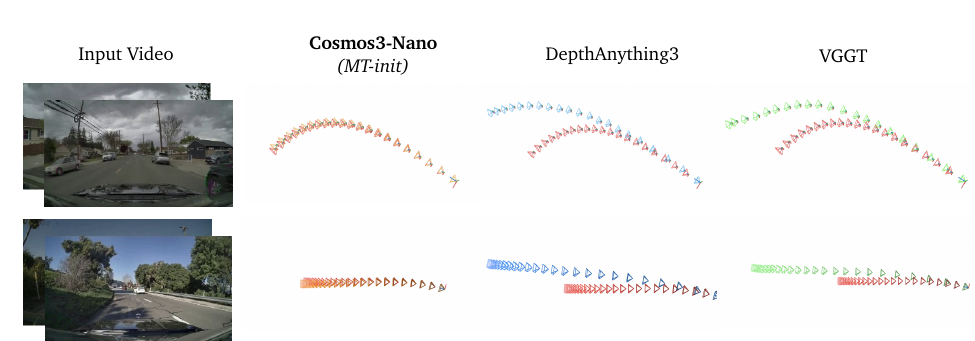}
    \caption{\textbf{Comparison for autonomous vehicle inverse dynamics}. We qualitatively compare ego-vehicle trajectories estimated from input videos by different methods, with the red trajectory representing the ground truth. \textbf{Cosmos3-Nano} \textit{(MT-init)} demonstrates the ability to estimate accurate, metric-scale ego poses.}
    \label{fig:av_id}
\end{figure}

\paragraph{Autonomous vehicle (ID).} 
We evaluate the inverse-dynamics capabilities of Cosmos 3 using an in-house driving dataset, which consists of 6-second video clips with accurate ego-vehicle trajectories at 10 FPS. 
Under this setup, the model directly predicts the ego-trajectory from video inputs. 
We compare our approach against two state-of-the-art general-domain baselines: VGGT~\citep{wang2025vggt} and DepthAnything3~\citep{lin2025depthanything3}. 
As reported in~\cref{tab:action_posttraining_summary_wide}, both Cosmos3-Nano and Cosmos3-Super initialized with \textit{PT-init} outperform the baselines. 
Applying \textit{MT-init} yields further improvements for Cosmos3-Nano. 
Notably, our method trained with specialized driving data achieves much better metric-scale translation estimation, whereas the general-domain baselines suffer from drifting errors. 
Overall, these results demonstrate that physical-AI-oriented video models can excel in specific embodied applications with minimal adaptation. 
Qualitative results are visualized in~\cref{fig:av_id}.

\begin{figure}[hbt]
    \centering
    \includegraphics[width=0.98415\linewidth]{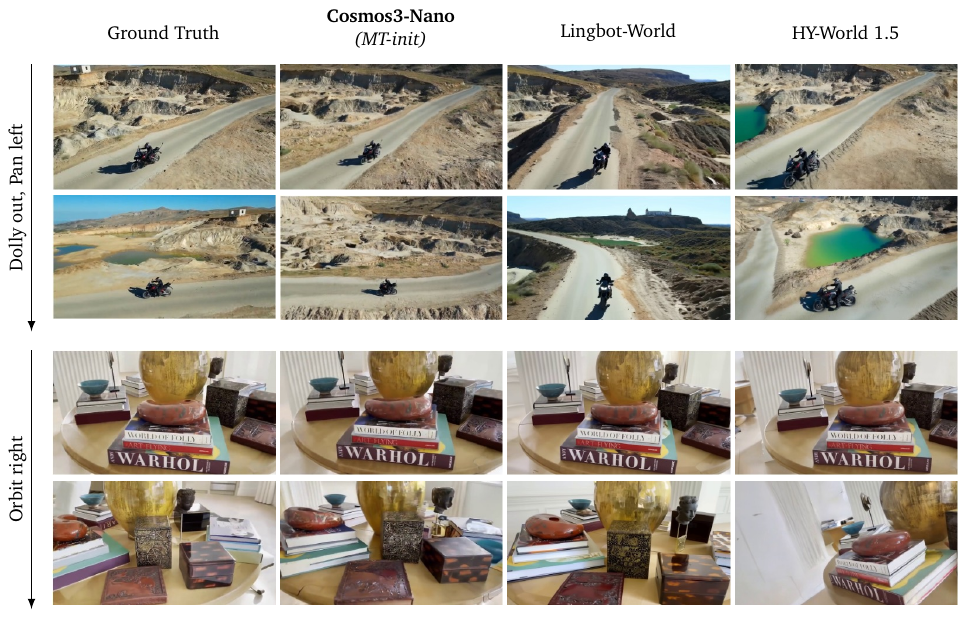}
    \caption{\textbf{Camera forward dynamics comparison}. Given complex realistic trajectories, \textbf{Cosmos3-Nano} \textit{(MT-init)} faithfully reproduces the same camera motion in the generated video. For each motion example, the first row shows frames near the start of the sequence and the second row shows frames near the end. The downward arrow indicates temporal progression from start to finish, while the text beside it specifies the commanded camera motion.}
    \label{fig:camera_fd}
\end{figure}

\paragraph{Camera motion (FD).}
\label{sec:camera_fdm}
Camera-conditioned video generation can serves as world simulation for many applications. 
We task the model with predicting future frames given an initial image and a specified camera trajectory.
To quantify camera-following accuracy, we utilize DepthAnything3~\citep{lin2025depthanything3} to estimate metric-scale camera trajectories from the generated videos and compare them against the conditioning input.
If the generated videos are highly consistent with the conditioning trajectory, the estimated poses from the predicted sequence should closely match the input poses.
We benchmark our approach against the bidirectional models of Lingbot-World~\citep{lingbot-world} and HY-World 1.5~\citep{hyworld2025} on an internal dataset of one hundred 5-second realistic video clips with camera motions estimated by DepthAnything3~\citep{lin2025depthanything3}. Qualitative results are reported in~\cref{fig:camera_fd}. 
The results show that Cosmos3-Nano and Cosmos3-Super with \textit{PT-init} already provide robust camera control. 
\textit{MT-init} improves the result further: Cosmos3-Super achieves 0.142$^\circ$ RRE, 0.026 m RTE, and 0.99 m ATE, outperforming Lingbot-World (0.299$^\circ$ RRE, 0.057 m RTE, 2.88 m ATE) and HY-World 1.5 (0.377$^\circ$ RRE, 0.042 m RTE, 1.39 m ATE) across all three metrics.

\paragraph{Egocentric motion (FD).}
\label{sec:egocentric_fdm}
We evaluate videos from Human World Bench (HWB, see \cref{subsec::hwb}) in the forward-dynamics
setting. We use the HWB action annotations, which include both camera ego-motion and hand motion tracking, and report the PSNR of the generated
videos. We adapt both \textit{PT-init} and \textit{MT-init} checkpoints to the egocentric domain and compare against LOME~\citep{gao2026lome}, a recent action-conditioned video generation method specialized for egocentric hand manipulation data. 
Although LOME is one of the closest available baselines for this setting, it performs poorly on HWB, achieving a PSNR value of 9.36dB, likely due to distribution shift between its training data and the HWB benchmark.
In contrast, Cosmos 3 achieves substantially higher PSNR across both model scales and initialization protocols. With \textit{PT-init},
Cosmos3-Nano reaches a PSNR value of 15.22dB and Cosmos3-Super reaches a PSNR value of 15.34dB. 
Starting from \textit{MT-init}
further improves the scores to PSNR values of 16.12dB for Cosmos3-Nano and 16.19dB for Cosmos3-Super. 
These results show that \textit{MT-init} provides a consistent gain, indicating that unified action mid-training
provides a substantially stronger starting point for egocentric forward dynamics.
Overall, these results support the central hypothesis of unified action mid-training: co-training across robot embodiments, camera
motion, autonomous-vehicle motion, and egocentric motion induces a transferable action-domain prior, enabling faster convergence and
stronger downstream adaptation in the egocentric domain.

\begin{figure*}[t]
    \centering
    \includegraphics[width=0.98810\linewidth]{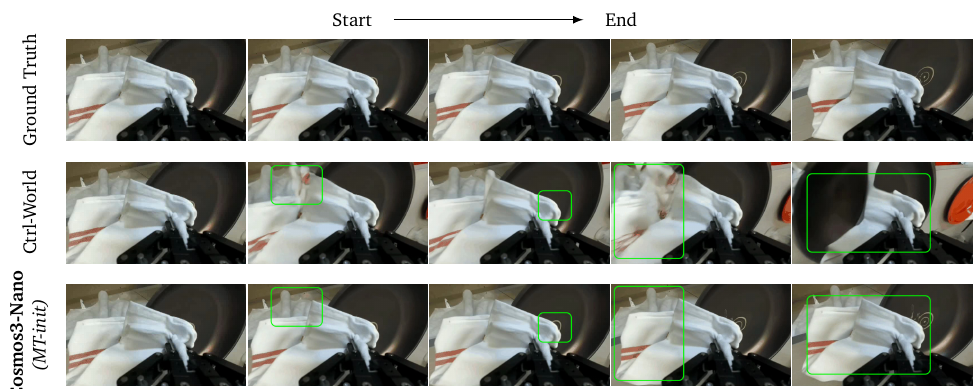}
    \caption{\textbf{Qualitative comparison for robotics forward dynamics.} The generated frames closely follow the action commands, and the interactions between the robot arm and the fabric are more realistic than those from the baseline. Green boxes highlight regions with visible distortions or artifacts in the baseline outputs, and the corresponding regions in \textbf{Cosmos3-Nano} \textit{(MT-init)}, where these artifacts are absent.}
    \label{fig:robotics_fd_qualitative}
\end{figure*}

\paragraph{Robotics (FD).}
\label{sec:robot_fdm}
Robotics forward dynamics unlocks applications such as policy evaluations with video models.
We conduct experiments on the DROID dataset~\citep{khazatsky2024droid}, a large-scale real-robot manipulation dataset featuring diverse scenes and objects. 
Given an initial frame and the robot end-effector action chunk of size 16, the model predicts the subsequent 16 frames.
We post-train both the \textit{PT-init} and \textit{MT-init} checkpoints on the DROID dataset.
We compare against Ctrl-World~\citep{guo2026ctrlworld} as our primary baseline.
We present the qualitative results in \cref{fig:robotics_fd_qualitative} and quantitative results in~\cref{tab:action_posttraining_summary_wide}.
As shown in \cref{tab:action_posttraining_summary_wide}, post-training from the pre-trained checkpoint already matches the baseline at Nano scale, reaching a PSNR value of 23.24dB compared with 22.99dB for Ctrl-World. 
Post-training from the mid-trained checkpoint yields substantially stronger results, reaching PSNR values of 25.52dB with Cosmos3-Nano and 26.04dB with Cosmos3-Super. 
These results demonstrate that Cosmos 3 is a strong and flexible backbone for forward dynamics prediction.

\begin{figure}[t]
    \centering
    \includegraphics[width=0.85\linewidth]{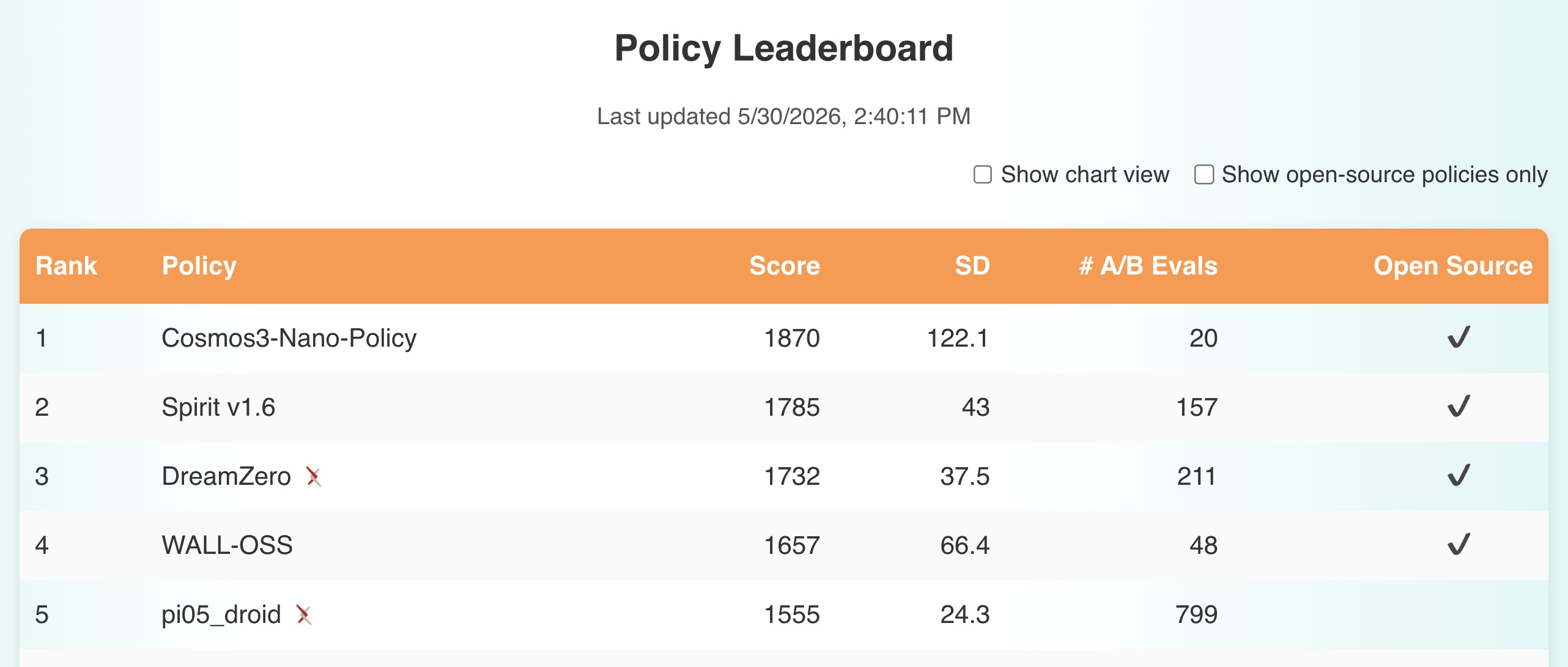}
    \vspace{0.5em}
    \caption{\textbf{Cosmos3-Nano-Policy-DROID held the top position on RoboArena.}
    Cosmos3-Nano-Policy-DROID ranked \#1 on the RoboArena real-world benchmark leaderboard
    (Date: 2026-05-30).}
    \label{fig:robotarena_leaderboard}
\end{figure}

\begin{figure}[t]
 \centering
 \includegraphics[width=1.00\linewidth]{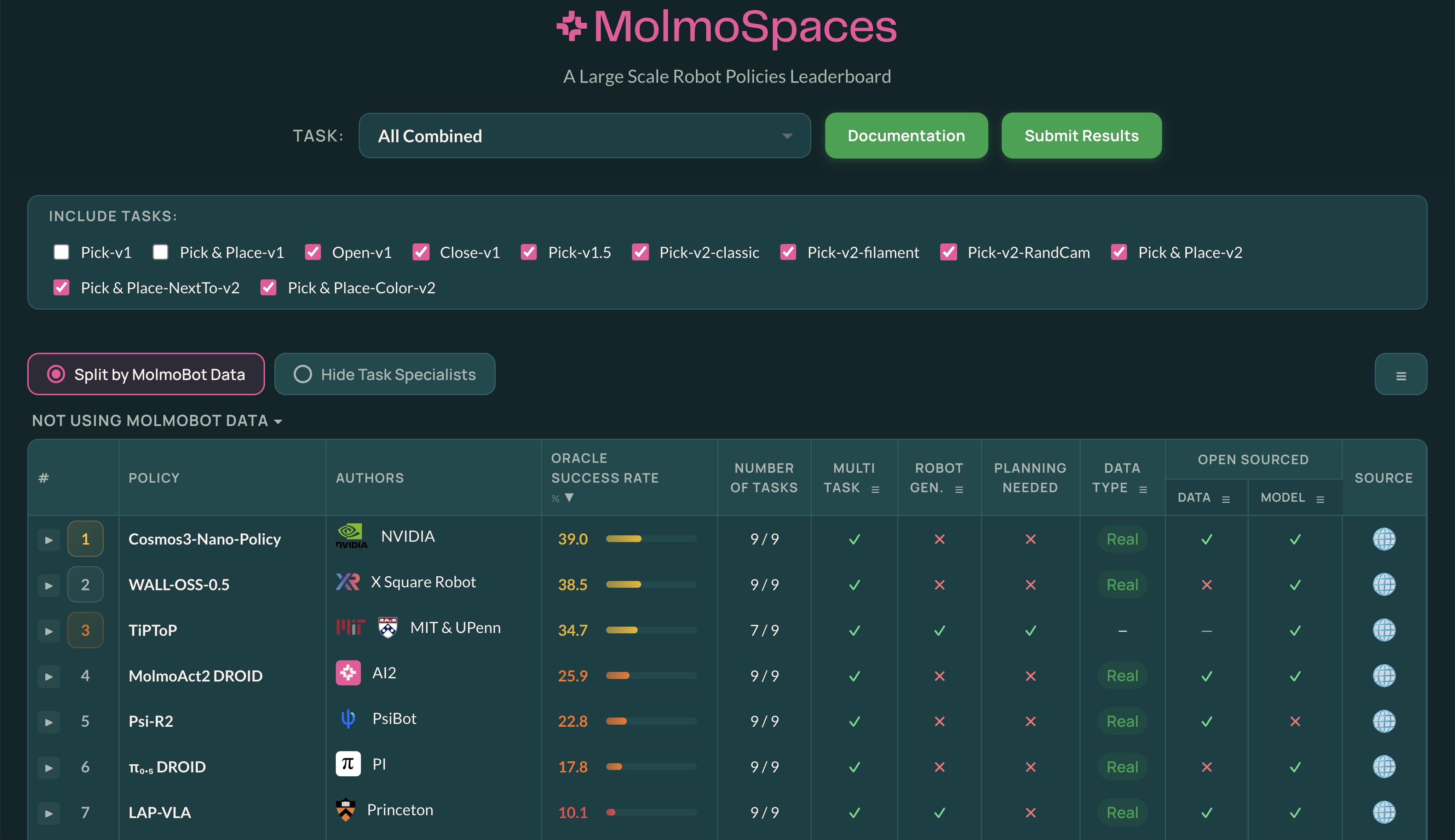}
 \caption{\textbf{Cosmos3-Nano-Policy-DROID ranked \#1 on MolmoSpaces.} Cosmos3-Nano-Policy-DROID was the top-ranked model on the MolmoSpaces simulation benchmark leaderboard (Date: 2026-06-20).}
 \label{fig:molmospaces_leaderboard}
\end{figure}

\paragraph{Robot manipulation (policy).}
\label{sec:robot_manip_policy}
To evaluate the policy mode, we post-train and submit our Cosmos3-Nano-Policy-DROID to the RoboLab simulation benchmark~\citep{yang2026robolab}, the RoboArena real-world benchmark~\citep{atreya2025roboarena}, and the MolmoSpaces simulation benchmark~\citep{kim:arxiv2026}. 
As shown in~\cref{tab:action_posttraining_policy}, \cref{fig:robotarena_leaderboard}, and \cref{fig:molmospaces_leaderboard}, our model achieves new state-of-the-art results across all three benchmarks, demonstrating that Cosmos3 omnimodal world models can be effectively post-trained into strong robot manipulation policies.

RoboLab~\citep{yang2026robolab} is a high-fidelity, robot- and policy-agnostic simulation benchmark comprising 120 language-conditioned tasks designed to evaluate task-generalist robot manipulation policies across visual, relational, and procedural competencies.
RoboLab evaluates each task under three instruction specificity levels, across vague prompts, default prompts and specific prompts.
This split tests robustness to language phrasing rather than a single canonical command.
On RoboLab, our post-trained policy surpasses all prior models on task success rate by clear margins (\cref{tab:action_posttraining_policy}), including strong vision-language-action (VLA) models such as $\pi_{0.5}$~\citep{pi05} and world-action models (WAM) such as DreamZero~\citep{ye2026dreamzero},
across all task instruction granularities and task difficulty levels.
For example, under \textit{specific} task instructions, our policy achieves a 39.7\% average success rate across 120 tasks with 10 rollouts per task, outperforming $\pi_{0.5}$ at 28.1\% and DreamZero at 25.2\%.
Additionally, compared with a model post-trained directly from the pre-trained checkpoint, Cosmos3-Nano (\textit{PT-init}), our final policy performs better, demonstrating the effectiveness of incorporating action-modality data from diverse sources into mid-training.

\begin{table}[t]
    \centering
    \caption{\textbf{Cosmos3-Nano-Policy-DROID establishes a new state of the art on RoboLab.}
    Task success rates (\%) are reported on RoboLab-120 across language specificity levels and task difficulty levels.
    All policies use off-the-shelf checkpoints fine-tuned on the DROID dataset.
    \textbf{Bold} indicates the best in the column.}
    \label{tab:action_posttraining_policy}
    \small
    \setlength{\tabcolsep}{4pt}
    \resizebox{\linewidth}{!}{%
    \begin{tabular}{@{}l|ccc|ccc|ccc|ccc@{}}
        \toprule
        & \multicolumn{3}{c}{\textbf{Overall}} & \multicolumn{3}{c}{\textbf{Simple}} & \multicolumn{3}{c}{\textbf{Moderate}} & \multicolumn{3}{c}{\textbf{Complex}} \\
        \cmidrule(lr){2-4} \cmidrule(lr){5-7} \cmidrule(lr){8-10} \cmidrule(lr){11-13}
        \textbf{Model} & Vague & Default & Specific & Vague & Default & Specific & Vague & Default & Specific & Vague & Default & Specific \\
        \midrule
        \rowcolor{rowours}
        \textbf{Cosmos3-Nano-Policy-DROID} & \textbf{20.6} & \textbf{36.8} & \textbf{39.7} & \textbf{23.3} & \textbf{40.6} & \textbf{42.0} & \textbf{23.3} & \textbf{35.4} & \textbf{40.3} & 4.1  & \textbf{25.3} & \textbf{29.4} \\
        \rowcolor{rowours}
        \textbf{Cosmos3-Nano} (\textit{PT-init})  & 16.7 & 28.1 & 30.2 & 17.8 & 30.3 & 32.8 & 19.0 & 28.7 & 29.5 & \textbf{7.1} & 18.2 & 21.8 \\
        \midrule
        $\pi_{0.5}$       &  15.2 &  28.0 &  28.1 &  16.2 &  29.7 &  29.8 &  17.9 &  31.5 &  31.0 & 5.3 &  13.5 &  14.7 \\
        DreamZero         &  14.9 &  25.7 &  23.9 &  15.0 &  26.1 &  25.8 &  19.5 &  30.0 &  26.7 & 4.1 &  14.1 &  10.6 \\
        $\pi_{0}$-FAST    & ~~9.2 &  15.5 &  14.9 & ~~9.5 &  20.2 &  19.4 &  12.8 &  13.3 &  12.6 & 0.0 & ~~2.9 & ~~3.5 \\
        paligemma-binning & ~~3.1 & ~~3.4 & ~~5.5 & ~~2.2 & ~~3.4 & ~~4.1 & ~~5.9 & ~~4.9 &  10.3 & 0.0 & ~~0.0 & ~~0.0 \\
        GR00T N1.6        & ~~5.4 & ~~7.2 & ~~5.3 & ~~7.2 & ~~8.8 & ~~7.5 & ~~4.9 & ~~7.9 & ~~4.1 & 0.0 & ~~0.0 & ~~0.0 \\
        $\pi_{0}$         & ~~2.8 & ~~5.0 & ~~3.5 & ~~2.8 & ~~7.2 & ~~5.3 & ~~3.8 & ~~3.6 & ~~2.1 & 0.0 & ~~0.0 & ~~0.0 \\
        \bottomrule
    \end{tabular}
    }
\end{table}
\begin{figure}[ht]
    \centering
    \includegraphics[width=0.99\linewidth]{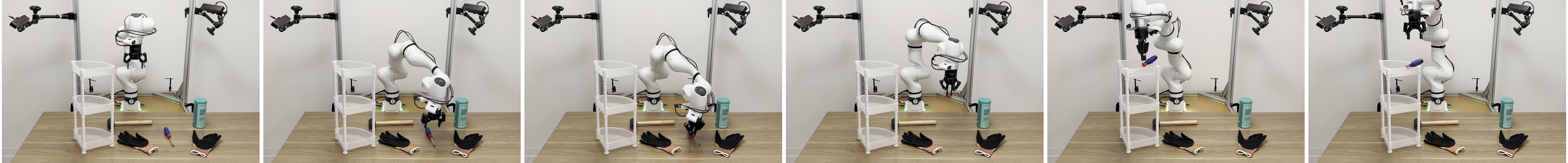}
    {\small Task: put the screwdriver on the top shelf\par}
    \vspace{4pt}
    \includegraphics[width=0.99\linewidth]{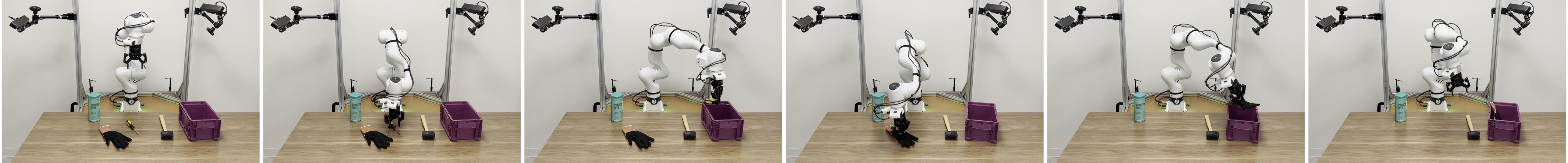}
    {\small Task: put the screwdriver and the glove in the purple container\par}
    \caption{\textbf{Real-world evaluation of Cosmos3-Nano-Policy-DROID.} We show snapshot frames from physical-robot rollouts. These examples demonstrate successful real-world deployment of our policy for language-conditioned manipulation tasks on physical robots.}
    \label{fig:robot_policy_example}
\end{figure}

RoboArena~\citep{atreya2025roboarena} is a distributed, real-world benchmark that uses crowdsourced pairwise comparisons to evaluate and rank generalist robot policies across diverse tasks and real world environments.
Anyone with a DROID platform can evaluate a pair of policies in any environment and on any task by conducting double-blind A/B comparisons, and the final scores are aggregated from these pairwise preferences to produce an overall rating for each policy.
As of 2:40 p.m. on May 30, 2026, our robot manipulation policy tops the leaderboard (\cref{fig:robotarena_leaderboard}), outperforming many prior strong policy models. 
We expect to receive more evaluations from the community, which will further validate our policy. 
Across several tested tasks, the policy accomplishes tasks reliably and follows language instructions well.
The policy can perform simple pick-and-place tasks, as well as long-horizon tasks requiring multiple sequential steps.
We observe strong generalization to a variety of unseen objects and tasks.
The policy also tolerates failures, often retrying when necessary, and remains robust to human interventions during execution.
Some qualitative results are shown in~\cref{fig:robot_policy_example}.

MolmoSpaces~\citep{kim:arxiv2026} is a simulation benchmark for evaluating generalist policies with a focus on generalization under systematic and controlled variations. We focus on the manipulation tasks, which consist of two task sets: (1) \textit{MolmoSpaces Combined} and (2) \textit{MolmoBot Combined}. The \textit{MolmoSpaces Combined} set comprises four benchmark tasks (\textit{Pick-v1}, \textit{Pick \& Place-v1}, \textit{Open-v1}, and \textit{Close-v1}), while the \textit{MolmoBot Combined} set comprises seven benchmark tasks (\textit{Pick-v1.5}, \textit{Pick-v2-classic}, \textit{Pick-v2-filament}, \textit{Pick-v2-RandCam}, \textit{Pick \& Place-v2}, \textit{Pick \& Place-NextTo-v2}, and \textit{Pick \& Place-Color-v2}). As of June 20, 2026, Cosmos3-Nano-Policy-DROID ranked first on the leaderboard under the \textit{All Combined} setting, achieving a 39.0\% oracle success rate (\cref{fig:molmospaces_leaderboard}). The model outperformed several strong prior policy models, such as WALL-OSS-0.5~\citep{zhai:arxiv2025} and TiPToP~\citep{shen:arxiv2026}. Notably, we submitted the exact same model and hyperparameters for the RoboLab and RoboArena evaluations, without any benchmark-specific tuning for MolmoSpaces.

\paragraph{Adaptation to new embodiments.}
We evaluate whether \textit{MT-init} enables Cosmos3-Nano to adapt more quickly to an unseen embodiment and environment.
We use the LIBERO-10 environment~\citep{liu2023libero} with multi-view observations from a third-person camera and a wrist camera.
We report progress across post-training iterations by running 50 trials per validation task, giving 500 rollouts per checkpoint.

As shown in \cref{tab:libero_posttraining_success_rate}, post-training rapidly improves closed-loop manipulation success rate for both initializations, but \textit{MT-init} has a clear advantage early in post-training.
At 500 iterations, the Cosmos3-Nano \textit{MT-init} reaches 24.6\% success, while the Cosmos3-Nano \textit{PT-init} remains at 0.0\%.
By 2000 iterations, the Cosmos3-Nano \textit{MT-init} can achieve a 97.4\% success rate.
These results show that \textit{MT-init} enables faster adaptation to new embodiments and environments.

\begin{table}[t]
    \centering
    \small
    \caption{
        \textbf{Fast adaptation to a new embodiment using LIBERO-10.} Closed-loop success rates for \textbf{Cosmos3-Nano} from \textit{MT-init} and \textit{PT-init}, evaluated with 500 rollouts per checkpoint. The results show that \textit{MT-init} adapts faster than \textit{PT-init}.
    }
    \label{tab:libero_posttraining_success_rate}
    \begin{tabular}{@{}lrrrr@{}}
        \toprule
        \multirow{2}{*}{\textbf{Model}} & \multicolumn{4}{c}{\textbf{Iteration}} \\
        \cmidrule(lr){2-5}
            & \textbf{500} & \textbf{1000} & \textbf{1500} & \textbf{2000} \\
        \midrule
        \textbf{Cosmos3-Nano} (\textit{MT-init}) & 24.6\% & 91.4\% & 95.8\% & \textbf{97.4\%} \\
        \textbf{Cosmos3-Nano} (\textit{PT-init}) & 0.0\% & 73.8\% & 93.4\% & 95.2\% \\
        \bottomrule
    \end{tabular}
\end{table}

\begin{figure}[t]
    \centering
    \captionsetup[subfigure]{font=small,skip=3pt,justification=centering,singlelinecheck=false}

    \begin{subfigure}[t]{0.43\linewidth}
        \centering
        \resizebox{\linewidth}{!}{%
            \includegraphics{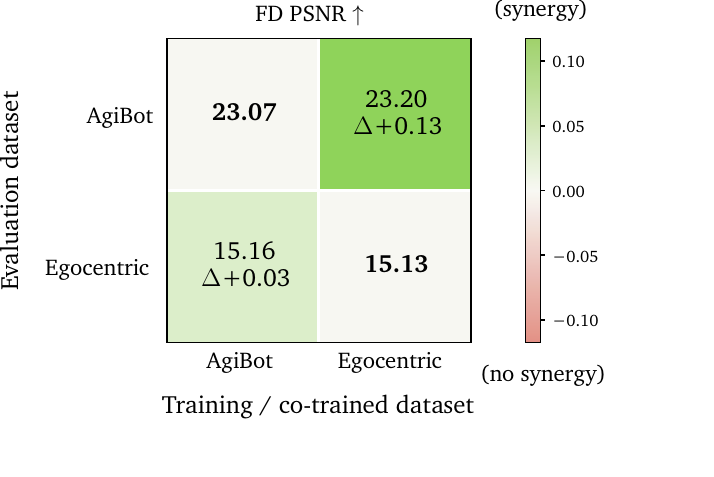}%
        }
        \caption{Synergy between AgiBot and Egocentric motion.}
        \label{fig:action_synergy_agibot_handpose}
    \end{subfigure}
    \hfill
    \begin{subfigure}[t]{0.53\linewidth}
        \centering
        \resizebox{\linewidth}{!}{%
            \includegraphics{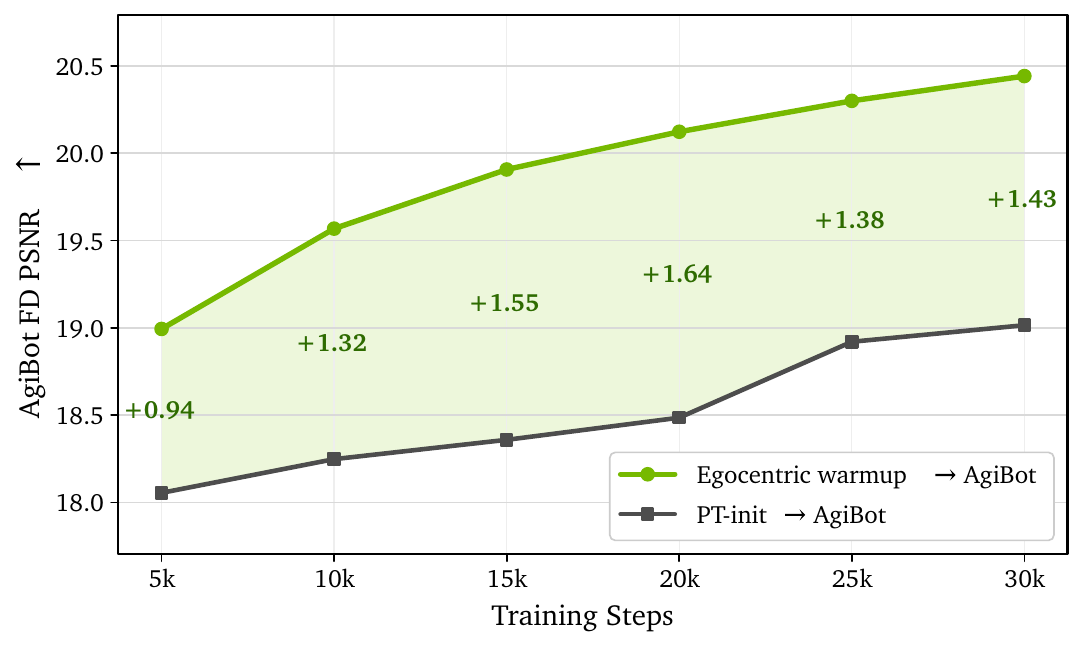}%
        }
        \caption{AgiBot benefits from Egocentric warmup.}
        \label{fig:action_synergy_agibot_curve}
    \end{subfigure}

    \caption{\textbf{Egocentric motion as a robot-adaptation prior.}
    (a) The matrix measures pairwise transfer between AgiBot robot manipulation and Egocentric motion. (b)
    The curve compares AgiBot adaptation from an Egocentric-warmed checkpoint against direct adaptation from the \textit{PT-init} checkpoint.}
    \label{fig:action_synergy_agibot}
\end{figure}

\paragraph{Action data synergy.}
\label{app:action_data_synergy}
Choosing which action domains to train together is a practical mixture-design problem: an added domain can provide a useful visual-action prior, but it can also dilute supervision for the target domain.
Inspired by recent work on language-domain transfer~\citep{longpre2026atlas}, which measures how training on one language benefits or interferes with another, we build analogous transfer maps for action mid-training across ego-motion domains (Camera Motion, Autonomous Vehicle), robot manipulation domains (Google Robot, WidowX-250, Franka Panda Single, Franka Panda Dual, AgiBot), and human egocentric motion (Egocentric).
The quantitative synergy matrices are summarized in~\Cref{fig:action_synergy_av_camera_robots} and~\Cref{fig:action_synergy_agibot}.
We report PSNR for FD and MSE for ID.
For policy mode, we report the minimum PSNR and MSE across 4 rollouts because the rollout distribution is multimodal.
In each synergy matrix, an off-diagonal cell uses a 50/50 mixture of the row and column domains for 4{,}000 iterations, while a diagonal cell is the corresponding single-domain baseline at 2{,}000 iterations.
This setup matches the per-domain training budget between single-domain and paired runs.
Each row fixes the evaluation domain, and columns vary the single-domain or paired training setting.
Positive deltas indicate cross-domain transfer, while negative deltas indicate interference.

We summarize our key findings below:

\begin{figure}[t]
    \centering
    \resizebox{\linewidth}{!}{%
        \includegraphics{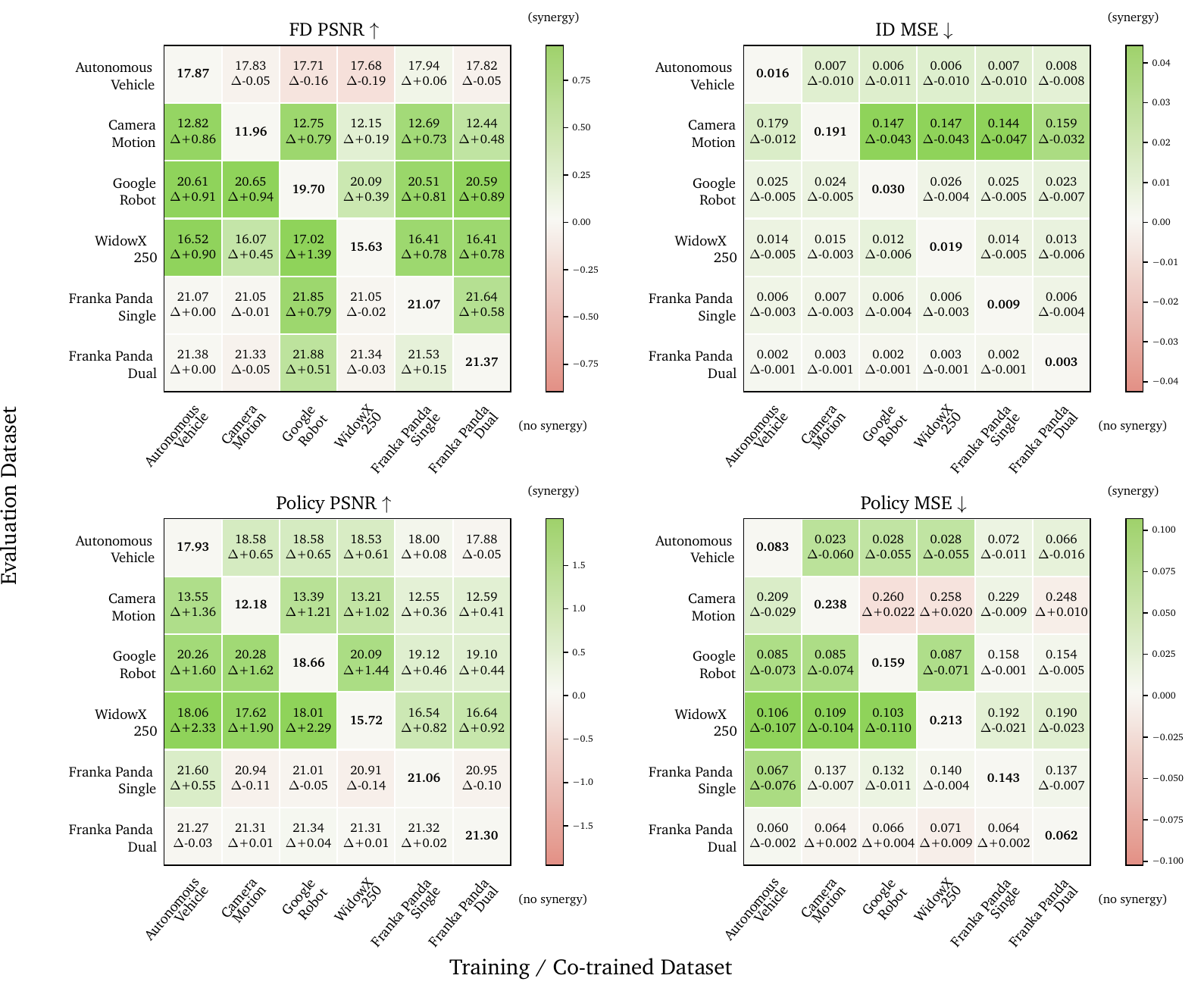}%
    }
    \caption{\textbf{Synergy study across ego-motion and robot manipulation domains.}
    Rows denote evaluation domains; columns denote the added co-training domain.
    Diagonal cells are single-domain baselines, while off-diagonal cells use a 50/50 row--column mixture with matched row-domain training exposure.
    Each cell reports the score and delta from the row diagonal.
    Green indicates positive transfer, with signs adjusted so that higher PSNR ($\uparrow$) and lower MSE ($\downarrow$) are both better.}

    \label{fig:action_synergy_av_camera_robots}
\end{figure}

\begin{itemize}[leftmargin=*]
    \item \textit{Camera motion benefits from broad action co-training.}
    \Cref{fig:action_synergy_av_camera_robots} shows that camera motion, one of the lower-performing domains, benefits from several co-training partners rather than only from another ego-motion source.
    AV data improves camera FD PSNR from 11.96 to 12.82 (+0.86), while robot domains also provide positive FD gains, including Google Robot (+0.79), Franka Panda Single (+0.73), and Franka Panda Dual (+0.48).
    ID MSE also improves for every shown co-training partner in the row.
    This suggests that camera-motion learning benefits from general action-conditioned visual priors, such as object persistence, scene geometry, and motion-correspondence cues, even when the added domain has a different embodiment.

    \item \textit{Robot manipulation domains share early-training priors.}
    The robot-domain panels in \Cref{fig:action_synergy_av_camera_robots} show broad early transfer among manipulation datasets, especially for domains with lower-performing single-domain baselines or more heterogeneous data.
    WidowX-250 benefits strongly from Google Robot, gaining +1.39 FD PSNR and +2.29 policy PSNR, with matching improvements in ID and policy MSE.
    Google Robot also gains from several robot co-training partners, with FD PSNR improvements up to +0.89 and policy PSNR improvements up to +1.44.
    By contrast, the Franka Panda single-arm and dual-arm entries in this synergy study correspond to small RoboMIND Franka subsets~\citep{wu2024robomind}, with only 23 and 4 hours of data, respectively. Compared with the much larger action-data sources summarized in~\Cref{fig:action_data_distribution}, these small target domains saturate quickly under single-domain training, so adding larger co-training domains yields smaller or mixed gains on their own evaluations. However, they still provide useful transfer to other domains when used as co-trained sources.

    \item \textit{Human egocentric motion helps robot adaptation.}
    \Cref{fig:action_synergy_agibot_handpose} shows modest but consistent transfer between egocentric motion and AgiBot.
    Egocentric data improves AgiBot FD PSNR from 23.07 to 23.20 (+0.13), while AgiBot slightly improves egocentric FD PSNR from 15.13 to 15.16 (+0.03).
    The warmup curve in \Cref{fig:action_synergy_agibot_curve} further tests this transfer as an initialization effect: we first warm up the base PT-init checkpoint on egocentric motion and then adapt it to AgiBot, comparing against direct AgiBot adaptation from the same PT-init checkpoint.
    Starting from the egocentric-warmed checkpoint improves AgiBot FD PSNR at every measured step, with gains of +0.94 at 5K and roughly +1.3--1.6 later in training.
    Together, these results suggest that egocentric human action data provides a useful manipulation prior for robot adaptation, while domain-aware sampling or specialization remains important as training progresses.
\end{itemize}

\paragraph{Conclusion.}
\label{sec:action_exp_summary}
We show that Cosmos 3 is a strong action foundation model across diverse domains, embodiments, and inference modes.
A single foundation model can be adapted to camera-controlled video generation, autonomous-vehicle inverse dynamics, egocentric hand forward dynamics, robot forward dynamics, and robot policy learning, while remaining competitive with or stronger than specialized domain baselines.
The comparison between \textit{PT-init} and \textit{MT-init} further shows that unified action mid-training does more than improve isolated domains: it produces a reusable action-domain prior that accelerates convergence across downstream settings, including adaptation to a new embodiment and environment.
This is especially notable because the evaluated domains differ substantially in embodiment and supervision format, ranging from camera and vehicle motion to articulated human hands and robot end-effectors.
The synergy study provides evidence that action domains share transferable structure: co-training across ego-motion, robot manipulation, and egocentric motion can improve early adaptation, particularly for domains with weaker single-domain baselines.
The additional action-specific ablations in Appendix~\ref{app:pusht_fd_id_policy_synergy} and \ref{app:robolab_video_action_consistency} reinforce the same conclusion: FD, ID, and policy objectives share useful structure under joint training, and videos predicted alongside policy actions remain aligned with simulator rollouts driven by those same actions.
Overall, these findings support Cosmos 3 as a general-purpose world-action foundation model that can be efficiently specialized to a broad range of action generation and control tasks.

\FloatBarrier

\subsection{Generator User Guide}

While benchmark results provide a quantitative assessment of generation quality, effectively using an omnimodal world model also requires understanding its inference interface, prompting methodology, and sampling configurations. Because Cosmos 3 supports a diverse set of modalities and generation modes—including image, video, audio-visual, transfer, forward-dynamics, inverse-dynamics, and policy generation—the quality of the outputs depends not only on the model weights but also on how generation requests are specified and conditioned. In this section, we provide practical guidance for using Cosmos 3 Generator, including recommended prompting strategies, structured caption formats, sampling hyperparameters, and the role of the Cosmos 3 Reasoner as a prompt upsampler. These guidelines reflect the configurations used throughout our evaluations and serve as a reference for obtaining high-quality, physically plausible generations across a wide range of Physical AI applications.

\subsubsection{Generation Guide}
\label{sec::gen_guide}

The Cosmos 3 Generator supports flexible visual (and audio-visual) generation across a broad inference envelope: frame rates from 10–30 FPS, 5 to 400 frames, resolutions spanning 256p, 480p, and 720p, and common aspect ratios (1:1, 3:4, 4:3, 9:16, 16:9). This allows a single model to serve use cases ranging from short preview clips to longer, higher-resolution landscape or portrait video without changing the sampling interface. The Cosmos 3 Generator is trained for forward dynamics, inverse dynamics and policy modes to support different action-related applications. For action generation, the base Cosmos3-Nano and Cosmos3-Super models support action prediction at native control frequencies ranging from 10–30 FPS, with prediction horizons spanning 16–400 frames across inverse dynamics and policy modes. Post-trained models specialize to a single mode and frequency---for example, Cosmos3-Nano-Policy-DROID operates at 15 FPS with a 32-step prediction horizon. Below, we detail the key components for successful generation using Cosmos 3 Generator. These details are also summarized in \cref{tab:sampling_configs}.

\begin{table*}[!t]
    \centering
    \captionsetup{justification=raggedright, singlelinecheck=false}
    \caption{\textbf{Default sampling configurations and negative prompts for each generator and generation modality.} We summarize the generation settings for different Cosmos 3 Generator modes. Negative prompts are provided in full for each setting in the Appendix; ``Null'' indicates that the null string was the best-performing variant.}
    \label{tab:sampling_configs}
    \setlength{\tabcolsep}{8pt}
    \renewcommand{\arraystretch}{1.25}
    \resizebox{\textwidth}{!}{%
    \begin{tabular}{l l l l}
        \toprule
         & \textbf{Generation Modality} & \textbf{Sampling Hyperparameters} & \textbf{Negative Prompt} \\
        \midrule
        \textbf{Cosmos3-Nano}             & Audio-Visual   & steps=50, guidance=6, shift=10, full-range CFG            & Appendix~\ref{appendix:generator_negative_prompt} \\
        \textbf{Cosmos3-Super}            & Audio-Visual   & steps=50, guidance=6, shift=10, full-range CFG            & Appendix~\ref{appendix:generator_negative_prompt} \\
        \textbf{Cosmos3-Super-Text2Image}        & Visual   & steps=50, guidance=4, shift=3, full-range CFG             & Null \\
        \textbf{Cosmos3-Super-Image2Video}        & Visual   & steps=50, guidance=6, shift=5, full-range CFG & Appendix~\ref{appendix:sft_i2v_upsampler_prompt} \\
        \midrule
        \textbf{Cosmos3-Nano}             & Forward/Inverse Dynamics   & steps=50, guidance=1, shift=5, full-range CFG                            & Null \\
        \textbf{Cosmos3-Super}            & Forward/Inverse Dynamics   & steps=50, guidance=1, shift=5, full-range CFG                             & Null \\
        \textbf{Cosmos3-Nano-Policy-DROID} & Policy & steps=4, guidance=3, shift=5, full-range CFG & Null \\
        \midrule
        \textbf{Cosmos3-Nano}             & Transfer & steps=50, guidance=3, control guidance=1.5,  shift=10                           & Appendix~\ref{appendix:generator_negative_prompt} \\
        \textbf{Cosmos3-Super}            & Transfer & steps=50, guidance=3, control guidance=1.5, shift=10                           & Appendix~\ref{appendix:generator_negative_prompt} \\
        \bottomrule
    \end{tabular}
    }
\end{table*}

\paragraph{Media specifications and prompting guide.}
The Cosmos 3 Generator is trained on structured JSON captions that provide fine-grained control over scene composition, covering subjects, background, lighting, aesthetics, cinematography, and for video, temporal fields such as actions, state changes, camera motion, and segment-level descriptions (full schema in Appendix~\ref{appendix:captioning_details}). At inference time, a prompt upsampler---served either by Claude Opus 4.6 or by the Cosmos 3 Reasoner---converts user requests into this same structured format, ensuring that generation prompts match the distribution seen during training (see Appendix~\ref{appendix:upsampler_prompt_template} for template instructions). The upsampler is instructed to first describe the scene layout and world state, then specify the temporal progression of events, and finally add any audio descriptions. The specific upsampler instruction varies slightly across generation modes---for instance, action and transfer generation impose additional task constraints tailored to their conditioning inputs (see Appendix~\ref{appendix:generator_transfer_prompt_prefix} for the transfer generation prompt prefix, and Appendix~\ref{appendix:action_upsampler_prompt} for the action prompt guide). The JSON specification additionally includes explicit media controls (duration, FPS, spatial height and width, and aspect ratio), keeping prompt interpretation and sampling configuration inspectable and reproducible.

\paragraph{Negative prompt.}
We tune negative prompts separately for each model and generation mode through automated benchmark iteration. For each configuration, we ablate over candidate templates spanning natural-language descriptions, keyword lists, instruction-style directives, compositional extensions targeting physical consistency and identity preservation, and the null string. The best-performing variant is selected based on automated benchmark scores. For the base Cosmos3-Nano and Cosmos3-Super generators, the explicit negative prompt can be found in Appendix~\ref{appendix:generator_negative_prompt}. For the post-trained variants, we found that the null string negative prompt works best for Cosmos3-Super-Text2Image, while using negative prompts automatically derived from the user prompts yields the best result for Cosmos3-Super-Image2Video (Appendix~\ref{appendix:sft_i2v_upsampler_prompt}). For action generation modes, we found the null string negative prompt to work best.

\paragraph{Generation sampling hyperparameters.} We adopt the following sampling parameters for different modalities:
\begin{enumerate}
    \item \textit{Audio-visual generation.} For audio-visual generation, we tune sampling hyperparameters on automated benchmarks for image and video generation (\cref{subsec::image_eval} and \cref{subsec::video_eval}). For the base Cosmos3-Nano and Cosmos3-Super generators, we use 50 denoising steps, a guidance scale of 6, a time shift of 10, and full-range classifier-free guidance. For the post-trained Cosmos3-Super-Text2Image model, we use a guidance scale of 4 and a time shift of 3. For the post-trained Cosmos3-Super-Image2Video model, we use a shift of 5.
    \item \textit{Action generation.} Action generation has three supported modes. For forward and inverse dynamics, we use 50 denoising steps, a guidance scale of 1, a time shift of 5, and full-range classifier-free guidance. For policy mode, we switch to 4 denoising steps and a guidance scale of 3.
    \item \textit{Transfer generation.} We tune sampling hyperparameters for video transfer generation on the automated PAIBench-C benchmark (\cref{subsec::transfer_eval}). We use 50 denoising steps, a text guidance scale of 3, a control guidance scale of 1.5, a time shift of 10, and full-range classifier-free guidance.
\end{enumerate}

\subsubsection{Cosmos 3 Reasoner as Prompt Upsampler}
\label{sec::prompt_upsampling}

Prompt upsampling is the pre-generation reasoning step in Cosmos 3 that translates compact user intent into a structured spatiotemporal scene specification. Rather than merely rewriting the prompt, it expands sparse user input into a physically grounded control language that captures scene layout, temporal evolution, and audio cues when applicable. Users typically provide a short natural-language request, while high-quality image and video generation depends on many details that are rarely specified explicitly such as subject attributes, spatial layout, camera behavior, lighting, temporal ordering, audio cues, and generation controls such as resolution, aspect ratio, duration, and frame rate. The upsampler fills this gap by acting as an instruction-following LLM module between the user interface and the generator. It takes the brief request, the output's structure template, plus the optional conditioning input signals such as a starting image for image-to-video, and produces a dense typed JSON scene specification that the downstream image or video model can render from.

The prior work falls into two threads. The first thread maps short user queries into richer text prompts for image generation. DALL-E 3 uses descriptive re-captioning and caption upsampling to better match the generator's caption distribution \citep{betker2023improving}; Prompt Expansion explicitly learns to expand a query into multiple optimized text-to-image prompts \citep{datta-etal-2024-prompt}; and prompt-adaptation systems such as Promptist, BeautifulPrompt, and RePrompt train language models to rewrite user inputs into model-preferred prompts while optimizing image-level objectives \citep{hao2023optimizing,cao-etal-2023-beautifulprompt,wu2025reprompt}. The second thread uses LLMs as planners for downstream visual generators, producing object layouts, grounded scene descriptions, frame-level prompts, or multi-scene video plans \citep{lian2024llmgrounded,feng2023layoutgpt,hong2023direct2v,
lin2023videodirectorgpt}. Cosmos 3 takes the same basic idea of inserting a reasoning module before generation, but further changes the interface. The upsampler emits one typed multimodal scene program, rather than only a rewritten prompt or a generator-specific layout plan, and the program spans image, video, and audio-conditioned generation. In a sense, the upsampler proceeds by first imagining the scene, considering the temporal/spatial aspects, and adding further multimodal content (\eg, audio descriptors) in an abstract language description state, then translating them into the structured output that matches the provided template.

We treat upsampling as complex \emph{multimodal} instruction following with \emph{physical priors} over \emph{structured data}. The input may be text-only, image-plus-text, or video-plus-text. The output is a schema-constrained JSON control program that captures the semantic content, visual attributes, task-specific caption, temporal plan for video tasks, audio-description field for video tasks, and generation parameters needed by the renderer. The structured output is not only a formatting choice, it exposes the latent variables that the generator must condition on, including entities, spatial relations, actions, timing, camera behavior, audio consequences, and generation controls. This connects prompt upsampling to the body of work on structured LLM reasoning and verification, where natural-language instructions are converted into explicit, checkable programs over structured fields \citep{chegini-etal-2025-repanda}. At the same time, filling a dense scene schema requires reasoning under uncertainty in which the upsampler must infer likely object states, temporal transitions, causal interactions, and acoustic outcomes from incomplete user input \citep{pournemat2025reasoning} in order to expand along the creativity axis without contradicting physical constraints of imposed reality. In Cosmos 3, these probabilistic and physical priors are expressed through the schema itself. The model first imagines a coherent world state, then maps it into a temporal rollout, and finally derives audio cues that remain synchronized with visible events. This separation is useful because it exposes prompt understanding as an independently inspectable component rather than entangling it with the rendering model.

\paragraph{Prompt contract.}
At inference time, the upsampler receives the user description together with the selected generation controls. For image-to-video generation, it also receives the conditioning image, which is treated as definitive visual evidence for the first frame. The request is formatted as four tagged blocks: 1) \texttt{instructions} block: introduces all blocks, defines the task-level behavior such as producing a single fenced JSON object, avoiding extra commentary, and treating the template as mandatory. 2) \texttt{image} or \texttt{video} \texttt{\_description} block: carries the raw semantic request; for conditioned requests such as I2V, the attached condition content is provided alongside this text and the constraints specify how visual facts should be anchored to it. 3) \texttt{task\_constraints} block: carries most of the
operational detail. It specifies field ordering, timing format, duration bounds, copied controls such as resolution, aspect ratio, duration, and frame rate, preservation requirements for user-provided entities and actions, and task-specific grounding rules such as matching the I2V first frame at \(t=0\). 4) \texttt{output\_json\_template} block: defines the schema surface: which keys must be populated, which nested fields are expected, and where scene, temporal, audio, subject, camera, lighting, style, and control information should be placed. In other words, the constraints describe how the upsampler should reason and copy values, while the JSON template defines the shape that the final answer
must take. The system prompt is a minimal ``You are a helpful assistant''. The full template can be found in appendix~\ref{appendix:upsampler_prompt_template}. Using this template, certain placeholder arguments such as description, FPS, duration, aspect ratio, resolution, \etc, have to be filled accordingly before passing to the Upsampler.

\section{Related Work}
\label{sec::related_work}

Recent advances in Physical AI have been driven by progress in several previously distinct research areas, including world models, multimodal understanding, video generation, action modeling, audio-visual generation, and omnimodal foundation models. While each of these directions has independently contributed important capabilities for perception, reasoning, simulation, and control, Physical AI systems ultimately require these capabilities to operate together within a unified framework. In this section, we review the literature most relevant to Cosmos 3, focusing on prior work in world simulation, multimodal reasoning, generative modeling, and embodied intelligence. We highlight how these research threads have evolved toward increasingly unified representations of the physical world and discuss how Cosmos 3 extends this trajectory by jointly modeling language, image, video, audio, and action for both understanding and generation within a single omnimodal world model.

\subsection{World Models for Physical AI}
World models describe how the world evolves and how an agent's actions change future observations.
A useful distinction is between predictive latent world models, which learn compact internal dynamics for planning and control, and generative world models, which expose predicted futures as inspectable multimodal simulations.

Predictive latent world models learn compact hidden states whose value is that they make planning and control cheaper: a controller can search or optimize in a representation that abstracts away pixel-level detail.
Early latent-dynamics systems such as World Models \citep{ha2018world}, Embed-to-Control \citep{watter2015embed}, PlaNet \citep{hafner2018planet}, and Dreamer \citep{hafner2019dream} showed that compact predictive states can support control and planning from pixels.
More recent JEPA-style models shift prediction from pixels to latent abstractions that are better aligned with perception, forecasting, and planning \citep{lecun2022path,assran2023ijepa,bardes2024vjepa,assran2025v,leworldmodel}.
These approaches establish internal dynamics modeling as a core ingredient of intelligent agents, but they are usually optimized for compact prediction or downstream control rather than high-fidelity multimodal simulation.

Generative world models make simulation itself the modeling interface: the model predicts possible future observations as images, videos, audio, or other grounded tokens.
By exposing the predicted future directly, they make errors in geometry, contact, timing, and sound observable rather than only inferred from a task loss.
Sora \citep{sora} made this view prominent for video generation and implicit world simulation \citep{he2025pre}.
The Cosmos world model series develops this direction directly for Physical AI \citep{cosmos_predict2,cosmos_predict2p5,cosmos_transfer1,cosmos_v1}.
Other domain-specific systems study manipulation, driving, and embodied navigation and interaction \citep{wang2023drivedreamer,zhao2025drivedreamer,liang2024dreamitate,jang2025dreamgen,ren2025cosmos,robodreamer,gao2026dreamdojo}.
Recent interactive systems extend the same direction toward controllable environments and agent-facing rollout interfaces \citep{bruce2024genie,genie2,genie3,gaia1,unisim,worldlabs2025marble,bahmani2025lyra,shen2026lyra2,waymo_world_model,li2025wonderplay,wang2025embodiedgen}.
Cosmos 3 builds on this direction but treats world modeling as a problem of both understanding and generation rather than as video synthesis alone: its reasoner tower interprets multimodal context and infers structured world state, while its generator tower synthesizes future image, video, audio, and action tokens from that context.
This lets one framework connect scene understanding, future synthesis, action inference, and multimodal rollout, so a generated trajectory can be conditioned on text, observations, actions, and audio rather than on prompts alone.

\subsection{Multimodal Understanding and Embodied Reasoning}
Multimodal understanding models provide the perception and reasoning layer needed for physical intelligence.
Flamingo \citep{alayrac2022flamingo} and BLIP-2 \citep{li2023blip2} helped establish the modern pattern of coupling large language models with visual inputs, while LLaVA \citep{liu2023visual,llavaonevision} made instruction-tuned open vision-language assistants broadly influential \citep{zhu2023minigpt4}.
Recent model families improve scale, grounding, temporal reasoning, OCR, and instruction following across images and videos \citep{chen2024internvl,wang2025internvl35,dai2024nvlm,bai2023qwen,qwen3vl2025,qwen2p5vl,deitke2024molmo,kosmos2,you2024ferret,zhang2024ferret,mm1,zhang2025mm1,tschannen2025siglip,xiao2024unified,beyer2024versatile,steiner2024family}.
GPT-4o \citep{openai2024gpt4o} reflects the move toward omnimodal interaction.
These capabilities make vision-language models useful front-ends for embodied systems, but physical intelligence needs more than recognition over isolated images or clips.
It requires temporally persistent state, spatial grounding tied to objects and agents, affordance reasoning over possible interactions, and task-progress tracking as conditions change.
This shifts the unit of reasoning from a caption or answer to a maintained, actionable scene estimate.

For Physical AI, the relevant challenge is therefore not only visual question answering, but maintaining grounded scene state while reasoning about space, time, affordance, object state, and task progress.
Cosmos-Reason1 \citep{azzolini2025cosmos} addresses this setting through physical common sense and embodied chain-of-thought reasoning.
Related robotics and egocentric work stresses spatial grounding, embodied planning, and human-object or robot-object interaction reasoning, using both model development and task-oriented benchmarks \citep{sermanet2024robovqa,HoloAssist2023,chen2025robo2vlm,yang2025magma,chen2024spatialvlm,song2025robospatial,zhou2025roborefer,yang2025embodiedbench}.
Despite this progress, most understanding models remain primarily discriminative or language-output systems: they can describe scenes, answer questions, or infer next-step intent, but they usually do not generate future observations or executable actions in the same representation space.
Cosmos 3 addresses this separation by allowing structured multimodal understanding to condition world generation directly.

\subsection{Video Generation and Visual World Simulation}
Video generation has progressed from short text-to-video clips to high-resolution, temporally coherent, instruction-following synthesis.
Early diffusion and transformer systems established the basic recipe for text-conditioned video \citep{ho2022video,singer2022makeavideo,ho2022imagen,villegas2022phenaki,blattmann2023stablevideo,bartal2024lumiere,hacohen2024ltx,videopoet,emuvideo,yang2024cogvideox,opensora2,mochi1,waver}.
Sora \citep{sora,sora2}, Movie Gen \citep{polyak2024movie}, and Veo 3 \citep{veo3,veo31} reflect the recent shift toward long-horizon realism, stronger prompt adherence, and higher-fidelity visual dynamics \citep{kong2024hunyuanvideo,wan2025,gao2025seedance,arkhipkin2025kandinsky}.
Industrial systems have also made controllability, creator workflows, and API deployment central parts of the video-generation landscape \citep{gen3,runway_gen4,kling,kling2,dreammachine,ray2,minimax,hailuo02,pika,firefly_video,nova_reel}.

The resulting landscape largely frames progress around perceptual video generation: plausible frames, smooth motion, and outputs that match textual or visual conditions.
Work on video world simulation and physical-consistency evaluation shows why these objectives are necessary for world modeling, but not sufficient for Physical AI \citep{he2025pre,bansal2024videophy,guo2025t2vphysbench,motamed2025generative}.
World simulation imposes a stronger contract: rollouts must preserve object identity and permanence, respect temporal causality, expose controllable factors such as actions or camera motion, and remain consistent when the same scene is queried under different interventions.
For an agent, a video is useful only if changes can be traced back to state, actions, and contacts; otherwise realism does not imply a reliable simulator.
The gap is especially visible under repeated prompting or closed-loop use, where small inconsistencies compound into incorrect downstream decisions.
Simulation-oriented systems must therefore couple synthesis with grounded state and intervention semantics.
Against this backdrop, Cosmos 3 places video inside a broader world modeling framework.
Video is not only an output modality, but also an input for reasoning, an observation stream for action inference, and a state trajectory coupled with actions and control.

\subsection{Action Modeling, VLAs, and World-Action Models}
Actions provide the causal link between agents and changing world states, and prior work is commonly organized into three settings: forward dynamics predicts future observations from state and action histories; inverse dynamics infers the action behind an observed transition; and policy mode maps observations, goals, and instructions directly to actions.
This taxonomy spans heterogeneous action spaces, from robot joint commands and vehicle controls to camera, human body, and egocentric wearer motion, each with different units, rates, and causal scopes.

Forward dynamics models are most explicit when the action is treated as a condition on future observations.
In driving, GAIA-1 \citep{gaia1}, DriveDreamer \citep{wang2023drivedreamer,zhao2025drivedreamer}, and Cosmos-Drive-Dreams \citep{ren2025cosmos} illustrate how future scenes can be generated under ego-motion, trajectories, or other controls.
In robotics and camera-controlled generation, related systems explore action-conditioned manipulation videos, robot-aware simulators, camera trajectories, depth, segmentation, and 3D-consistent controls \citep{liang2024dreamitate,jang2025dreamgen,robodreamer,zhu2024irasim,hma,gao2026dreamdojo,he2024cameractrl,wang2024motionctrl,xu2024camco,ren2025gen3c}.
These systems show that generative models can learn useful action-conditioned priors, but many are specialized to a particular domain, action space, or embodiment.

In practice, inverse dynamics often asks which action explains an observed transition.
VPT \citep{baker2022vpt} studies action labeling from video, and later work extends this idea through imitation from observation, latent action discovery, and learning from videos without explicit action annotations \citep{zhang2022learningdrive,torabi2018bco,yang2019iddm,pavse2019ridm,schmidt2023lapo,ye2024latentaction,latent_action_wild}.
Large embodied datasets such as Open X-Embodiment \citep{vuong2023open} and DROID \citep{khazatsky2024droid} provide the data substrate for learning across embodiments, tasks, viewpoints, and manipulation regimes \citep{walke2023bridgedata,liu2023libero,nasiriany2024robocasa,contributors2024agibotworldrepo,bu2025agibot,grauman2024egoexo4d,li2023behavior1k}.

Policy-mode systems predict actions from observations, goals, and instructions.
RT-1/RT-2 \citep{brohan2022rt,brohan2023rt}, PaLM-E \citep{driess2023palme}, OpenVLA \citep{kim2024openvla}, $\pi_0$ \citep{black2024pi0}, Gemini Robotics \citep{gemini2025robotics,gemini_robotics15}, and GR00T N1 \citep{bjorck2025gr00t} trace the progression from vision-language policies to general robot foundation models \citep{roboflamingo,octo2024,pi05,liu2024rdt,zhou2025chatvla,zheng2026xvla,yang2025egovla,shi2025hi,lee2025molmoact,gr1,gr2}.
Related embodied-agent systems use language or vision-language models for spatial value maps, action reasoning, and interaction-aware planning \citep{voxposer,zawalski2024ecot,yang2025magma}.
World-action models bring policy learning closer to world modeling by using video dynamics and action representations as priors \citep{li2025uva,ye2026dreamzero,univla,rynnvla,world_action_survey,acwmphys}.
Cosmos 3 treats forward dynamics, inverse dynamics, and policy mode as conditioning patterns of one multimodal sequence model over video, audio, text, and action tokens.

\subsection{Audio and Audio-Visual Generation}
Audio is an important part of physical-world modeling because many events are defined not only by how they look, but also by how they sound.
AudioLDM 2 \citep{audioldm2}, AudioCraft \citep{audiocraft}, and MusicGen \citep{musicgen} are representative of the rapid progress in text-conditioned audio and music generation, with later systems extending the space toward speech, sound effects, and creator-facing products \citep{voicebox,audiobox,stableaudioopen,lee2024etta,suno,udio,elevenlabs_sfx}.
These models improve the acoustic realism of generated media, but physical-world simulation also requires synchronizing sound with visible dynamics.

Audio-visual learning studies whether sound and vision correspond to the same event, source, or motion.
Classical threads include cross-modal correspondence, source separation, synchronization, and speech-lip alignment, while Diff-Foley \citep{luo2023difffoley} and related video-to-audio systems focus on generating sounds that follow visible dynamics \citep{owens2018audiovisual,zhao2018sound,ephrat2018looking,chung2016out,prajwal2020wav2lip,comunita2023syncfusion,zhang2024foleycrafter,ren2024stav2a,gramaccioni2024stablev2a,mmaudio,klingfoley}.
Talking-head, human-animation, and joint audio-video generators further connect speech, body motion, scene dynamics, and sound, including synchronized audio-visual media generation in Movie Gen \citep{polyak2024movie} and Veo 3 \citep{veo3,vasa1,emo,hallo,omnihuman,xing2024seeing,ruan2022mmdiffusion,wang2024avdit,liu2024syncflow,li2025threemdit,hacohen2026ltx2,liu2025javisditt}.
For physical simulation, the timing is as important as the identity of the sound: a collision should produce an impact at the contact frame, footsteps should match gait and surface, and a tool or engine should change sound when its visible state changes.
Speech similarly needs mouth motion, speaker identity, and turn-taking to remain aligned, while alarms, motors, scraping, and other environmental sounds can provide causal evidence for events that are partly occluded.
For Physical AI, audio should be modeled as synchronized evidence about events, contacts, speech, tools, engines, and environments rather than as post-hoc decoration.
Cosmos 3 treats audio as another modality in the same world modeling interface, enabling joint video-audio generation, video-conditioned audio generation, and audio-conditioned video generation.

\subsection{Omnimodels for Understanding and Generation}
A growing line of work studies omnimodels that combine multimodal understanding and generation in a single framework.
It is useful to separate this goal from two neighboring families.
Vision-language models (VLMs) usually map multimodal inputs to text, while media generators map prompts or conditions to generated images, video, or audio.
Omnimodels aim to support both directions in one framework, so perception, language, and synthesis can share representations instead of being connected only by external pipelines.

Unified-IO 2 \citep{lu2023unifiedio2} and Chameleon \citep{chameleon2024} study unified multimodal modeling, while GPT-4o \citep{openai2024gpt4o} and Gemini \citep{gemini2,gemini3} show how frontier systems are moving toward native multimodal input and output.
Other unified or any-to-any systems explore different mixtures of text, image, audio, video, and discrete tokenization or diffusion interfaces, with Qwen-Omni \citep{xu2025qwen3omni,qwen35omni} representing a recent omnimodal family \citep{tang2023codi,wu2023nextgpt,anygpt,seedx,fourm,xie2024showo,chen2025januspro,emu3}.

Within this broader omnimodal family, recent MoT-style work is especially relevant because it asks how one model can share sequence-level context while reserving specialized capacity for different modalities or functions.
Transfusion \citep{zhou2024transfusion} connects this thread to one-transformer training over mixed-modality sequences by combining next-token prediction for text with diffusion-based image generation.
Mixture-of-Transformers \citep{liang2024mot} makes the specialization pattern explicit for multimodal foundation models, with related work adapting pre-trained language models for multimodal generation and studying asymmetric bridges between heterogeneous understanding and generation experts \citep{shi2025lmfusion,wang2025hbridge}.
BAGEL \citep{deng2025bagel} extends this direction to unified multimodal pre-training for both understanding and generation with a decoder-only architecture trained on large-scale interleaved multimodal data.
Recent embodied extensions make the same architectural question physical by specializing pathways for scene understanding, visual foresight, latent action, and control in VLA and world-action models \citep{lv2025f1,cai2026internvlaa1,shou2026halo,bi2025motus,motubrain2026,li2026causalworld,tencent2026hyembodied}.

Taken together, these omnimodels show that understanding and generation can be handled within shared multimodal frameworks rather than by separate systems.
However, existing work still emphasizes text-image modeling or general multimodal media generation, with less focus on physical-world dynamics, action-conditioned generation, inverse dynamics, and embodied control.
Cosmos 3 extends the omnimodal idea to Physical AI by pairing an autoregressive reasoner tower for multimodal understanding with a diffusion generator tower conditioned on the reasoner's representations.
This interface supports video/text-to-text understanding, text/video-to-video generation, audio-video generation, and world-action modeling for action understanding and prediction.
Cosmos 3 is not only multimodal but also omni-functional: the same model can interpret the world, simulate how it evolves, infer the actions behind observed changes, and generate future observations and actions.

\section{Conclusion}
\label{sec::conclusion}

We presented Cosmos 3, a family of omnimodal world models for Physical AI. Cosmos 3 unifies multimodal understanding and generation across language, image, video, audio, and action within a single architecture, reducing the need to compose separate vision-language models, video generation models, world models, and action models. This unified formulation is enabled by modality-specific encoders, structured token arrangements, and a Mixture-of-Transformers backbone that couples autoregressive reasoning with diffusion-based generation. Together with scalable data, training, serving, and evaluation infrastructure, Cosmos 3 provides a solid foundation for developing Physical AI agents. Across diverse understanding and generation benchmarks, Cosmos 3 demonstrates strong results and broad capability coverage, while remaining amenable to downstream specialization through post-training. We expect Cosmos 3 to serve as a bridge between the synthetic world and the real world, providing better synthetic data, a better starting point for specialized models, and better closed-loop training environments for Physical AI. By releasing source code, pre-trained checkpoints, and curated benchmarks, we aim to accelerate research toward general-purpose Physical AI agents that can perceive, reason, simulate, and act in the real world.

\clearpage
\appendix

\section{Caption Details}
\label{appendix:captioning_details}
This appendix details the design of our in-house captioning models and full structured caption schemas used for image and video training data. We train our own models rather than relying on off-the-shelf VLMs to ensure maximum control over the output structure and to prevent hallucinated or missing schema fields. Instead of relying solely on free-form natural-language captions, we organize annotations as JSON objects with predefined semantic fields. This design encourages comprehensive coverage of visual details while keeping the representation consistent and interpretable. The image schema captures static scene properties, including subjects, layout, lighting, aesthetics, style, and spatial composition, while the video schema extends this representation with temporal information such as actions, state changes, transitions, camera motion, and audio cues.

\subsection{Captioner Models}
To maximize detail and spatial coverage in our image training data, we employ a quadrant-scan annotation strategy. Each image is divided into four quadrants, and the content within each quadrant---along with the center region---is described independently. This approach is highly effective for capturing multiple distinct subjects and complex layouts, which frequently occur in our dataset but are typically under-described by standard captioning methods.

For video data, we introduce a second annotation pass dedicated exclusively to temporal dynamics. Because motion and state transitions are fundamental to video understanding, this secondary pass comprehensively labels temporal changes across all explicitly tracked visual attributes.

For both our final image and video captioning models, we determined that LoRA fine-tuning of \mbox{Qwen3-VL-8B} provided the optimal balance between benchmark metrics and inference efficiency. For video input, we sample frames at $8$ FPS and set the generation temperature to $0.7$. We also use the default \mbox{Qwen3-VL} minimum and maximum pixel bounds, $131,072$ and $25,165,824$, respectively.

\subsection{Image Schema}

We use a structured caption representation for text-to-image training data that captures a broad range of visual attributes, including foreground subjects, scene composition, background elements, lighting, aesthetics, artistic style, camera viewpoint, and cinematographic properties. Rather than relying on dense free-form natural language, captions are represented as JSON objects with predefined semantic fields. This structured representation improves detail recall and annotation consistency while maintaining high precision.

To improve spatial coverage, the captioning pipeline incorporates a quadrant-based scanning mechanism that partitions each image into four spatial quadrants together with a central region. Each region is described independently before being merged into the final caption. This design is particularly effective for images containing multiple distinct subjects, localized interactions, or complex spatial layouts that are often under-described in conventional captioning approaches.

We summarize the schema by semantic category and provide a compact JSON skeleton below.

\begin{tcolorbox}[
    colback=black!2,
    colframe=nvidiagreen!75!white,
    boxrule=0.4pt,
    arc=1pt,
    left=4pt,
    right=4pt,
    top=4pt,
    bottom=4pt,
    title=Top-level Fields in the Image Caption Schema
]
\small
\renewcommand{\arraystretch}{1.1}
\begin{tabularx}{\linewidth}{@{}>{\ttfamily}lX@{}}
\toprule
Field & Description \\
\midrule
subjects[] & Per-subject records capturing identity, appearance, spatial placement, pose, clothing, expression, and count-sensitive anatomy fields when applicable. Human or human-like subjects additionally include demographic and facial-attribute fields when visible. \\
subject\_details & Open-ended attributes or clarifications that do not fit cleanly into a single subject slot, including free-form fine-grained facial details when useful. \\
background\_setting & Global scene context, environment, and background elements. \\
lighting & Illumination conditions, directionality, shadows, and notable lighting effects. \\
aesthetics & Composition, color palette, mood, and repeated visual patterns. \\
cinematography & Framing, camera angle, depth of field, focus behavior, and lens characteristics. \\
style\_medium / artistic\_style & Medium, rendering style, and artistic treatment. \\
context & High-level narrative or situational context for the scene. \\
text\_and\_signage\_elements[] & Any visible text, its appearance, placement, category, and scene relevance. \\
quadrant\_scan & Region-wise scan of the top-left, top-right, bottom-left, bottom-right, and absolute center. \\
comprehensive\_t2i\_caption & A natural-language summary distilled from the structured fields. \\
resolution / aspect\_ratio & Image size metadata used to preserve scale and layout cues. \\
\bottomrule
\end{tabularx}
\end{tcolorbox}

The number of entries in \texttt{subjects} and the contents of \texttt{subject\_details} vary according to scene complexity.

\paragraph{Human and human-like attributes.}
For human or human-like subjects, the schema may additionally include:
\begin{itemize}
    \item \texttt{clothing}
    \item \texttt{expression}
    \item \texttt{gender}
    \item \texttt{age}
    \item \texttt{skin\_tone\_and\_texture}
    \item fine-grained facial attributes, such as eye shape, eye color, lip shape, hair color, wrinkles, or moles, typically represented through \texttt{appearance\_details} or \texttt{subject\_details}
\end{itemize}

\paragraph{Count-sensitive attributes.}
If a subject corresponds to a group or cluster of similar entities, the schema may additionally include:
\begin{itemize}
    \item \texttt{number\_of\_subjects}
    \item \texttt{number\_of\_arms}
    \item \texttt{number\_of\_hands}
    \item \texttt{number\_of\_fingers}
    \item \texttt{number\_of\_legs}
\end{itemize}

\paragraph{Compact JSON skeleton.}
To standardize our image annotations and prevent the omission of critical visual details, we enforce a strict, predefined JSON structure. The schema detailed below outlines the specific semantic fields---ranging from background settings to aesthetic style---used to comprehensively capture the static properties of each image.
\begin{tcolorbox}[
    breakable,
    colback=black!2,
    colframe=nvidiagreen!75!white,
    boxrule=0.4pt,
    arc=1pt,
    left=4pt,
    right=4pt,
    top=4pt,
    bottom=4pt,
    title=Compact Image-specific JSON Skeleton
]
\begin{Verbatim}[breaklines=true,breakanywhere=true,fontsize=\scriptsize]
{
  "subjects": [
    {
      "description": "...",
      "appearance_details": "...",
      "relationship": "...",
      "location": "...",
      "relative_size": "...",
      "orientation": "...",
      "pose": "...",
      // human or human-like only
      "clothing": "...",
      "expression": "...",
      "gender": "...",
      "age": "...",
      "skin_tone_and_texture": "...",
      // group- or count-sensitive attributes
      "number_of_subjects": 0,
      "number_of_arms": 0,
      "number_of_hands": 0,
      "number_of_fingers": 0,
      "number_of_legs": 0
    },
    ...
  ],
  "background_setting": "...",
  "lighting": {
    "conditions": "...",
    "direction": "...",
    "shadows": "...",
    "illumination_effect": "..."
  },
  "aesthetics": {
    "composition": "...",
    "color_scheme": "...",
    "mood_atmosphere": "...",
    "patterns": "..."
  },
  "cinematography": {
    "framing": "...",
    "camera_angle": "...",
    "depth_of_field": "...",
    "focus": "...",
    "lens_focal_length": "..."
  },
  "style_medium": "...",
  "artistic_style": "...",
  "context": "...",
  "text_and_signage_elements": [
    {
      "text": "...",
      "category": "...",
      "appearance": "...",
      "spatial": "...",
      "context": "..."
    }
  ],
  "quadrant_scan": {
    "top_left": "...",
    "top_right": "...",
    "bottom_left": "...",
    "bottom_right": "...",
    "absolute_center": "..."
  },
  "comprehensive_t2i_caption": "...",
  "resolution": {
    "H": xx,
    "W": xx
  },
  "aspect_ratio": xx
}
\end{Verbatim}
\end{tcolorbox}

\subsection{Video Schema}

The video caption schema extends the image caption schema with fields that capture temporal information. Shared static attributes, such as subjects, background, lighting, aesthetics, style, and camera viewpoint, follow the image caption schema described above. Here, we focus only on video-specific additions.

The video schema explicitly records how the scene evolves over time. It captures subject-level actions and state changes, global action timelines, temporal segments, transitions between segments, camera motion, and optional audio descriptions. These fields improve the representation of motion, interactions, and temporal continuity, which are not captured by image-only captions.

We summarize the video-specific schema components by semantic category and provide a compact JSON skeleton below.

\begin{tcolorbox}[
    colback=black!2,
    colframe=nvidiagreen!75!white,
    boxrule=0.4pt,
    arc=1pt,
    left=4pt,
    right=4pt,
    top=4pt,
    bottom=4pt,
    title=Additional Top-level Fields in the Video Caption Schema
]
\small
\renewcommand{\arraystretch}{1.1}
\begin{tabularx}{\linewidth}{@{}>{\ttfamily}lX@{}}
\toprule
Field & Description \\
\midrule
actions[] & Key visual actions in chronological order. \\
segments & Distinct temporal segments by shot, scene, or meaningful change within the video. \\
transitions & Notable transitions between segments or temporal changes in the video. \\
temporal\_caption & Dense description of all temporal changes in the video. \\
resolution / aspect\_ratio / duration / fps & Video metadata. \\
audio\_description & Description of the audio content in the video. \\
\bottomrule
\end{tabularx}
\end{tcolorbox}

The number of entries in \texttt{actions}, \texttt{segments}, and \texttt{transitions} varies according to the temporal complexity of the video. Short or static videos may contain only a few temporal events, while longer videos with multiple scene changes, interactions, or camera movements may require more detailed temporal segmentation.

\paragraph{Video-specific fields.}
Compared with the image schema, the video schema introduces the following additional components:
\begin{itemize}
    \item \texttt{action}: subject-level motion or activity.
    \item \texttt{state\_changes}: changes in subject appearance, pose, position, or condition over time.
    \item \texttt{camera\_motion}: temporal camera behavior, such as panning, tilting, zooming, tracking, or handheld motion.
    \item \texttt{actions}: a time-indexed description of important events in the video.
    \item \texttt{segments}: temporally localized descriptions of major video intervals.
    \item \texttt{transitions}: changes between segments, shots, scenes, or subject states.
    \item \texttt{temporal\_caption}: an overall summary of the video’s temporal progression.
    \item \texttt{duration} and \texttt{fps}: basic video metadata.
    \item \texttt{audio\_description}: optional description of speech, music, sound effects, or ambient audio.
\end{itemize}

\paragraph{Compact video-specific JSON skeleton.}
While the image schema captures static scene properties, video annotations require tracking temporal dynamics over time. To avoid redundancy, we use a compact JSON skeleton for video data that intentionally omits static visual attributes covered previously. Instead, the schema detailed below focuses strictly on video-specific semantic fields, such as actions, transitions, camera motion, and audio cues.
\begin{tcolorbox}[
    breakable,
    colback=black!2,
    colframe=nvidiagreen!75!white,
    boxrule=0.4pt,
    arc=1pt,
    left=4pt,
    right=4pt,
    top=4pt,
    bottom=4pt,
    title=Compact Video-specific JSON Skeleton
]
\begin{Verbatim}[breaklines=true,breakanywhere=true,fontsize=\scriptsize]
{
  "subjects": [
    {
      // fields inherited from the image schema are omitted here
      "action": "",
      "state_changes": ""
    },
    ...
  ],
  "cinematography": {
    // fields inherited from the image schema are omitted here
    "camera_motion": ""
  },
  "actions": [
    {
      "time": "",
      "description": ""
    },
    ...
  ],
  "text_and_signage_elements": [
    {
      // image-level text fields are inherited from the image schema
      "spatial_temporal": ""
    },
    ...
  ],
  "segments": [
    {
      "segment_index": 0,
      "time_range": "",
      "description": "",
      "key_changes": "",
      "camera": ""
    },
    ...
  ],
  "transitions": [
    "",
    ...
  ],
  "temporal_caption": "",
  "duration": xx,
  "fps": xx,
  "audio_description": ""
}
\end{Verbatim}
\end{tcolorbox}
\section{Default Prompts and Prompt Upsampling Templates for Generator}
This appendix provides the prompt upsampler instruction templates and  full negative prompts used at inference time for Cosmos 3 Generator under different conditions. We specify the upsampler in each subsection and the corresponding use case. The sampling configurations that accompany these templates are summarized in ~\cref{tab:sampling_configs}.

\subsection{Upsampler Prompt Template for Cosmos 3 Reasoner}
\label{appendix:upsampler_prompt_template}

The following snippet shows the instruction template used to invoke the Cosmos 3 Reasoner as a prompt upsampler, converting user prompts into the structured JSON schema expected by the Cosmos 3 Generator at inference time (see \cref{sec::prompt_upsampling} for details).

\begin{tcolorbox}[
    breakable,
    colback=black!2,
    colframe=nvidiagreen!75!white,
    boxrule=0.4pt,
    arc=1pt,
    left=4pt,
    right=4pt,
    top=4pt,
    bottom=4pt,
    title=Abridged Canonical Prompt-Template Examples
]
\begin{Verbatim}[breaklines=true,breakanywhere=true,fontsize=\scriptsize]
T2V user message:
<instructions>
Prompt upsampler for a text-to-video model. Produce exactly one fenced JSON
object that fully populates the template and satisfies all constraints.
</instructions>
<video_description>{description}</video_description>
<task_constraints>
1. Write scene_imagination first.
2. Write temporal_caption second as the timestamped M:SS playback timeline.
3. Write audio_description third, aligned with the visual beats when possible.
4. Copy exactly: duration="0:06", fps=24, aspect_ratio="1,1",
   resolution={"W":640,"H":640}.
5. Keep all timed fields within duration and mutually consistent.
6. Preserve the user's described subjects, actions, props, and setting.
</task_constraints>
<output_json_template>
{"scene_imagination":"...", "temporal_caption":"...", "audio_description":"...",
 "subjects":[...], "background_setting":"...", "lighting":{...},
 "aesthetics":{...}, "cinematography":{...}, "style_medium":"...",
 "artistic_style":"...", "context":"...", "actions":[...],
 "text_and_signage_elements":[...], "segments":[...], "transitions":[...],
 "resolution":"Per task constraints", "aspect_ratio":"Per task constraints",
 "duration":"Per task constraints", "fps":"Per task constraints"}
</output_json_template>

T2I user message:
<instructions>
Prompt upsampler for a text-to-image model. Produce exactly one fenced JSON
object that fully populates the template and satisfies all constraints.
</instructions>
<image_description>{description}</image_description>
<task_constraints>
1. Write scene_imagination first.
2. Write comprehensive_t2i_caption second as a dense 80-200 word image prompt.
3. Copy exactly: aspect_ratio="1,1", resolution={"W":960,"H":960}.
4. Keep subjects, background, lighting, aesthetics, and camera fields consistent.
5. Populate subject_details with 2-5 image-specific attributes.
</task_constraints>
<output_json_template>
{"scene_imagination":"...", "comprehensive_t2i_caption":"...",
 "subjects":[...], "subject_details":{...}, "background_setting":"...",
 "lighting":{...}, "aesthetics":{...}, "cinematography":{...},
 "style_medium":"...", "artistic_style":"...", "context":"...",
 "text_and_signage_elements":[...], "quadrant_scan":{...},
 "resolution":"Per task constraints",
 "aspect_ratio":"Per task constraints"}
</output_json_template>

I2V user message, with attached starting frame:
<instructions>
Prompt upsampler for an image-to-video model. Treat the attached starting frame
as definitive visual ground truth and the text as temporal/action intent.
</instructions>
<video_description>{description}</video_description>
<task_constraints>
1. Write scene_imagination first, anchoring visual facts to the image and
   temporal facts to the description.
2. Copy exactly: duration="0:20", fps=20, aspect_ratio="9,16",
   resolution={"W":480,"H":832}.
3. Use only M:SS timing and keep all timed fields within duration.
4. Ensure the first segment and earliest actions match the image at t=0.
5. Preserve concrete facts from both the image and the description.
</task_constraints>
<output_json_template>
{"scene_imagination":"...", "temporal_caption":"...", "audio_description":"...",
 "subjects":[...], "background_setting":"...", "lighting":{...},
 "aesthetics":{...}, "cinematography":{...}, "style_medium":"...",
 "artistic_style":"...", "context":"...", "actions":[...],
 "text_and_signage_elements":[...], "segments":[...], "transitions":[...],
 "resolution":"Per task constraints", "aspect_ratio":"Per task constraints",
 "duration":"Per task constraints", "fps":"Per task constraints"}
</output_json_template>

V2V user message, with attached conditioning video:
<instructions>
Prompt upsampler for a video-to-video continuation model. Treat the attached
conditioning video as definitive visual and temporal ground truth for the
observed prefix, and the text as future/action intent.
</instructions>
<video_description>{description}</video_description>
<task_constraints>
1. Write scene_imagination first, summarizing the conditioning video's state,
   subjects, motion history, and final visible configuration.
2. Write temporal_caption second as the future M:SS playback timeline after the
   conditioning video, preserving continuity with the observed prefix.
3. Write audio_description third, aligned with visible future events when possible.
4. Copy exactly: duration="0:05", fps=24, aspect_ratio="16,9",
   resolution={"W":1280,"H":720}.
5. Preserve concrete facts from the conditioning video and the description.
</task_constraints>
<output_json_template>
{"scene_imagination":"...", "temporal_caption":"...", "audio_description":"...",
 "subjects":[...], "background_setting":"...", "lighting":{...},
 "aesthetics":{...}, "cinematography":{...}, "style_medium":"...",
 "artistic_style":"...", "context":"...", "actions":[...],
 "text_and_signage_elements":[...], "segments":[...], "transitions":[...],
 "resolution":"Per task constraints", "aspect_ratio":"Per task constraints",
 "duration":"Per task constraints", "fps":"Per task constraints"}
</output_json_template>
\end{Verbatim}
\end{tcolorbox}

\subsection{Upsampler Prompt Template for Cosmos3-Super-Text2Image}
\label{appendix:sft_t2i_upsampler_prompt}

We use the following instruction template to produce upsampled JSON prompts for post-trained Cosmos3-Super-Text2Image. See ~\ref{sec::t2i_post_train} for more details.

\begin{tcolorbox}[
    breakable,
    colback=black!2,
    colframe=nvidiagreen!75!white,
    boxrule=0.4pt,
    arc=1pt,
    left=4pt,
    right=4pt,
    top=4pt,
    bottom=4pt,
    title=\textbf{Cosmos3-Super-Text2Image} Prompt Upsampler Template,
]
\begin{Verbatim}[breaklines=true,breakanywhere=true,fontsize=\scriptsize]
Given the user's natural-language request below, generate a dense structured JSON that fully describes the image to be produced. The JSON must strictly follow the template provided after the request, including every top-level key and every nested sub-field.

The output is always DENSE. Even when the request is brief, you must infer plausible, scene-consistent details for every field. Do not leave fields empty merely because the request did not mention them - the purpose of this task is to upsample a sparse request into a rich, complete annotation. Be creative but stay grounded: your additions must be physically plausible and internally consistent with the request.

Requirements:
- For every visual field, write rich, specific content inferred from the request's scene, subjects, mood, and context.
- Empty values ("", 0, [], {}) are permitted ONLY for truly inapplicable fields:
    * Human-only subject fields (clothing, expression, gender, age, skin_tone_and_texture, facial_features, number_of_arms, number_of_legs, number_of_hands, number_of_fingers) when the subject is non-human.
    * text_and_signage_elements = [] when no visible text or signage is present.
    * aesthetics.patterns = "" when there are no notable repeating patterns.
    * subject_details = {} when no image-specific structured attributes apply.
- Do not add keys beyond the template. Do not omit keys required by the template.

Return only the JSON object wrapped in a ```json code fence.

USER VISUAL REQUEST:
{caption_dense}

Lists (subjects, text_and_signage_elements) may contain zero or more items of the shape shown. All top-level keys must always be present in the output; fill unused fields with "", 0, {}, or [] as appropriate.

{
  "subjects": [
    {
      "description": "full visual description of the subject",
      "appearance_details": "additional visual details (accessories, texture, distinguishing features)",
      "relationship": "how this subject relates to others or to the scene",
      "location": "where in frame (e.g., 'Center foreground', 'Top right')",
      "relative_size": "size within frame",
      "orientation": "direction subject faces relative to camera",
      "pose": "body position and posture",
      "clothing": "clothing and accessories; '' if non-human or N/A",
      "expression": "facial expression; '' if non-human or N/A",
      "gender": "one of 'Male', 'Female', 'Unknown', 'N/A'",
      "age": "age category",
      "skin_tone_and_texture": "skin tone description; '' if non-human",
      "facial_features": "notable facial features, including eye shape/color, hair color/style, lip shape, wrinkles, moles, scars, freckles, facial hair, and other visible fine-grained facial attributes; '' if non-human or not visible",
      "number_of_subjects": "int; total in this subject group, 0 if N/A",
      "number_of_arms": "int; 2 for humans, 0 if non-human",
      "number_of_legs": "int; 2 for humans, 0 if non-human",
      "number_of_hands": "int; 2 for humans, 0 if non-human",
      "number_of_fingers": "int; 10 for humans, 0 if non-human"
    }
  ],
  "subject_details": {
    "key_name_1": "free-form image-specific attribute (keys vary by image content; {} if N/A)"
  },
  "background_setting": "full prose description of the environment and setting",
  "lighting": {
    "conditions": "type and quality of light",
    "direction": "where light comes from; 'None' for flat digital images",
    "shadows": "shadow description; 'None' for flat digital images",
    "illumination_effect": "overall effect of the lighting"
  },
  "aesthetics": {
    "composition": "framing and compositional choices",
    "color_scheme": "dominant colors and palette",
    "mood_atmosphere": "emotional atmosphere in short phrases",
    "patterns": "notable repeating visual patterns; 'None' if none"
  },
  "cinematography": {
    "framing": "shot type",
    "camera_angle": "angle (e.g., 'Eye-level', 'Low angle', 'High angle')",
    "depth_of_field": "'Shallow', 'Deep', 'Uniform focus', or 'N/A'",
    "focus": "what is in sharp focus",
    "lens_focal_length": "descriptive focal length"
  },
  "style_medium": "visual medium (e.g., 'Photography', 'Digital presentation slide', 'Screenshot')",
  "artistic_style": "genre or approach",
  "context": "scene context or use case (brief)",
  "text_and_signage_elements": [
    {
      "text": "the visible text content",
      "category": "one of 'physical_in_scene', 'ui_text', 'body_text', 'scene_sign', 'logo', 'label'",
      "appearance": "font, color, size, style",
      "spatial": "position in image",
      "context": "purpose or meaning of the text"
    }
  ],
  "quadrant_scan": {
    "top_left": "description of what appears in the top-left region",
    "top_right": "description of what appears in the top-right region",
    "bottom_left": "description of what appears in the bottom-left region",
    "bottom_right": "description of what appears in the bottom-right region",
    "absolute_center": "description of what appears at the center"
  },
  "comprehensive_t2i_caption": "a comprehensive, full-scene natural-language prose description of the image"
}
\end{Verbatim}
\end{tcolorbox}

\subsection{Upsampler Prompt Template for Cosmos3-Super-Image2Video}
\label{appendix:sft_i2v_upsampler_prompt}

The following snippet shows the instruction template used with Claude Opus 4.7 to produce upsampled JSON prompts and per-sample contextual negative prompts for the post-trained Cosmos3-Super-Image2Video model (see \cref{sec::i2v_post_train} and \cref{sec::gen_guide} for more details).

\begin{tcolorbox}[
    breakable,
    colback=black!2,
    colframe=nvidiagreen!75!white,
    boxrule=0.4pt,
    arc=1pt,
    left=4pt,
    right=4pt,
    top=4pt,
    bottom=4pt,
    title=\textbf{Cosmos3-Super-Image2Video} Prompt Upsampler Template,
]
\begin{Verbatim}[breaklines=true,breakanywhere=true,fontsize=\scriptsize]
You are an expert prompt engineer for an image-to-video generative model. You are given a STARTING FRAME image (the first frame of the video) and a USER INSTRUCTION describing the desired motion or changes to animate. Your task is to produce a dense, cinematic video description that the model will use to generate the full video, together with a customized negative prompt.

Complete this task in two phases.

---
### PHASE 1: VIDEO DESCRIPTION
Write a dense, narrative caption inside `<final_prompt>` XML tags, formatted as a JSON object using this exact template:

<final_prompt>{"temporal_caption": "..."}</final_prompt>

Rules for the caption:
- The provided image is the exact starting frame - all described motion must be consistent with the starting frame.
- Opening: Establish the scene — subjects, environment, lighting — describing what is directly visible in the starting frame accurately and faithfully, noting essential elements that the motion will directly involve, the subject's orientation (e.g., "facing away", "in three-quarter profile"), and any implied ongoing motion (e.g., a cyclist leaning into a curve, water already splashing) so the video continues smoothly. Phrase it naturally as a scene description (do not say "in the starting frame", "initially shown", or similar meta-references).
- Motion: Describe the changes and actions in chronological order. Flow naturally from one action to the next. Advance time using natural conjunctions (e.g., "while," "as," "and").
- Physical Accuracy: All motion must obey gravity and reflect realistic material behavior (e.g., cloth ripples, water splashes, rigid objects resist deformation).
- Cause-and-effect: Always describe causes before their effects. Reflections, shadows, and secondary effects cannot appear on their own — the source object must first enter the frame or move into the relevant position before any reflection or shadow is described. E.g., a person must walk to the water's edge before their reflection appears on the surface; an object must strike the water before a splash erupts.
- Object Permanence: Every subject must persist throughout or have a clear reason for entering or exiting. When a new subject not present in the starting frame is introduced (e.g., an opposing team, an arriving vehicle), briefly describe their appearance (e.g., uniform color, vehicle type and color) so the generator can render them consistently, and describe a logical way for them to come into the frame (e.g., entering from a specific side of the frame, walking in through a door, or emerging from behind an existing object) rather than having them appear out of nowhere.
- Taboo Phrases: NEVER refer to the video medium itself. Avoid "the video shows...", "the scene...", "the clip...", "the frame...", "the camera shows...", "we see...".
- Perspective: Describe human body sides from the subject's own perspective (e.g., "her right hand" = the subject's right hand) to avoid ambiguity. This applies whenever a body part enters or moves in the frame: always specify whether it is the left or right (e.g., "his right hand reaches in from the lower edge"), never a bare "a hand enters the frame".
- Pronouns: Use singular pronouns ("he", "she", "him", "her", "it") or a singular noun phrase ("the person", "the rider", "the child") for single subjects. Never use "they"/"them"/"their" to refer to one person, as this can cause the model to render multiple subjects.
- Spatial Phrasing: Use spatial relationships for motion (e.g., "enters from the left", "rises above the horizon") rather than camera-centric descriptions.
- Camera: Include camera motion only if specified in the instruction; otherwise describe from a static viewpoint. Keep any described camera movement subtle and gradual — do not exaggerate altitude loss, tilt angle, or speed beyond what is minimally implied by the instruction. Do not use the word "transition" when describing camera motion.
- Cinematography Terms: When the instruction references a lens, camera, or filming technique (e.g., "probe lens", "macro lens", "fisheye", "drone shot", "GoPro"), treat it as a cinematographic style describing how the footage is captured — never as a physical object visible in the scene. Mention the style (e.g., for a probe lens: extreme close shot; for a fisheye lens: extreme wide angle fisheye view) rather than mentioning the lens or camera apparatus itself.
- Timelapse: If the instruction implies timelapse, explicitly use the word "timelapse" in the caption and avoid exaggerating its effects.
- Cuts & Montages: Always describe a single continuous shot with no hard cuts unless the user instruction explicitly used words like "cut", "hard cut", "jump cut", "shot change", or "montage". When multiple shots are requested without specifying an exact number, describe at most 3 shots, and dedicate the majority of the description to the opening action before any cut. Never use phrases like "the first shot", "the opening shot", or number shots as "first", "second", etc. — simply describe the action directly.
- Tone: Neutral, objective, descriptive. No opinions, value judgments, or inferred emotions unless physically observable.
- Length & Format: Write exactly ONE coherent paragraph of 5-8 sentences. No bullet points or lists.

USER INSTRUCTION:
"{description}"

---
### PHASE 2: NEGATIVE PROMPT
Using your final video description from Phase 1, create a customized negative prompt.

HOW IT WORKS:
A negative prompt describes exactly what a bad video looks like. Use declarative statements (e.g., "blurry faces"). Never use negative instructions like "avoid" or "do not".

---
DEFAULT NEGATIVE PROMPT:
The video captures a series of frames showing macroblocking artifacts, chromatic aberration, high-frequency noise, and rolling shutter distortion. It includes static with no motion, motion blur, over-saturation, shaky footage, low resolution, grainy texture, pixelated images, poorly lit areas, underexposed and overexposed scenes, poor color balance, washed out colors, choppy sequences, jerky movements, low frame rate, bit-depth compression artifacts, color banding, unnatural transitions, outdated special effects, fake elements, unconvincing visuals, poorly edited content, jump cuts, hard cut, visual noise, and flickering. It features moiré patterns, edge halos, and temporal aliasing. Furthermore, the content defies common sense, generating illogical scenarios, nonsensical entities, absurd character behaviors, and conceptual paradoxes that violate basic human reasoning and everyday reality. The video looks like a surreal or glitchy hallucination. Overall, the video is of poor quality.
---

INSTRUCTIONS:
Delete any words from the default negative prompt that contradict your intended video. Keep most of the original wording and structure intact, and do not add new items. Examples:
* If you want scene cuts/montages -> REMOVE "jump cuts" and "hard cut".
* If you want a motionless/static scene -> REMOVE "static with no motion".
* If you want fantasy, sci-fi, or surrealism -> REMOVE "defies common sense", "illogical scenarios", "nonsensical entities", "surreal", and related logic-violation terms.
* If the scene has flickering light -> REMOVE "flickering".
* If it is a night-time timelapse -> REMOVE "motion blur".

Output only the final negative prompt as a single paragraph, wrapped in <negative_prompt> tags. Do not output any explanation or preamble.
\end{Verbatim}
\end{tcolorbox}

\subsection{Prompt Prefix for Video Transfer}
\label{appendix:generator_transfer_prompt_prefix}
For video transfer tasks, we prefix the user caption with a system prompt that instructs the model to condition generation on visual control signals. Specifically, the following system prompt is prepended via the chat template before the user-provided caption during both training and inference:

\begin{tcolorbox}[
    breakable,
    colback=black!2,
    colframe=nvidiagreen!75!white,
    boxrule=0.4pt,
    arc=1pt,
    left=4pt,
    right=4pt,
    top=4pt,
    bottom=4pt,
    title=Cosmos 3 Base Generator Video-to-Video Transfer Prompt Prefix
]
\begin{Verbatim}[breaklines=true,breakanywhere=true,fontsize=\scriptsize]
You are a helpful assistant that generates images or videos following the user's instructions and control signals (edge maps, blur, depth, or segmentation).
\end{Verbatim}
\end{tcolorbox}

\subsection{Prompt Template for Action Generation}
\label{appendix:action_upsampler_prompt}
For action-related tasks, the model accepts a natural-language description of the intended behavior. The metadata information related to the action and video is populated in the JSON format. This format follows the same structured-caption convention used for video generation, while exposing action-specific fields for camera framing, temporal extent, conditioning frame rate, output resolution, and aspect ratio. The resulting prompt has the following form:

\begin{tcolorbox}[
    breakable,
    colback=black!2,
    colframe=nvidiagreen!75!white,
    boxrule=0.4pt,
    arc=1pt,
    left=4pt,
    right=4pt,
    top=4pt,
    bottom=4pt,
    title=Action Generation JSON Prompt Template
]
\begin{Verbatim}[breaklines=true,breakanywhere=true,fontsize=\scriptsize]
{
  "cinematography": {
    "framing": "<viewpoint description>"
  },
  "actions": [
    {
      "time": "0:00-<end time>",
      "description": "<action caption as a sentence>",
      "idle_frame": "<idle frames out of total frames>"
    }
  ],
  "duration": "<integer seconds>s",
  "fps": <conditioning fps>,
  "resolution": {"H": <height>, "W": <width>},
  "aspect_ratio": "<width,height>"
}
\end{Verbatim}
\end{tcolorbox}

\begin{figure}[H]
    \centering
    \begin{minipage}[t]{0.44\linewidth}
        \vspace{0pt}
        \includegraphics[width=\linewidth]{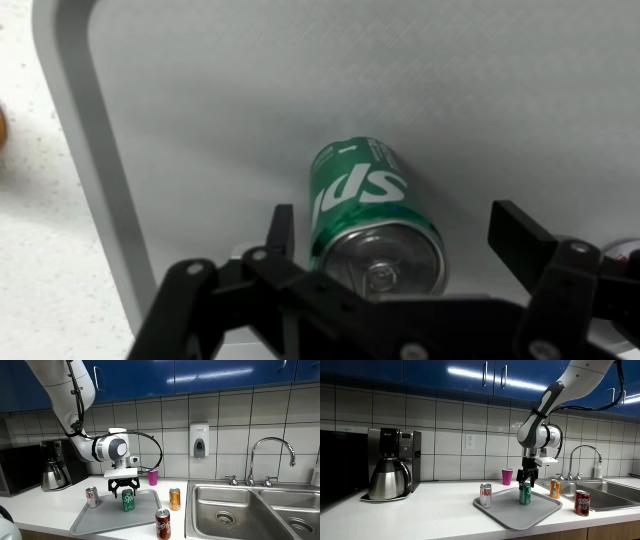}
    \end{minipage}
    \hfill
    \begin{minipage}[t]{0.52\linewidth}
    \begin{tcolorbox}[
        breakable,
        colback=black!2,
        colframe=nvidiagreen!75!white,
        boxrule=0.4pt,
        arc=1pt,
        left=4pt,
        right=4pt,
        top=3pt,
        bottom=3pt,
        title=Prompt
    ]
    \begin{Verbatim}[breaklines=true,breakanywhere=true,fontsize=\tiny]
    {
      "cinematography": {
        "framing": "This video contains concatenated views from multiple camera perspectives. The top row is from the wrist-mounted camera. The bottom row contains two horizontally concatenated third-person perspective views of the scene from opposite sides, with the robot visible."
      },
      "actions": [
        {
          "time": "0:00-0:02",
          "description": "Remove the cans from the tray and put them on the countertop.",
          "idle_frame": "4 out of 16."
        }
      ],
      "duration": "2s",
      "fps": 10.0,
      "resolution": {
        "H": 544,
        "W": 736
      },
      "aspect_ratio": "4,3"
    }
    \end{Verbatim}
    \end{tcolorbox}
    \end{minipage}
    \captionsetup{justification=raggedright, singlelinecheck=false}
    \caption{\textbf{Multiview action prompt formatting.} When multiple viewpoints are available, we concatenate them into a single canvas and attach view-layout metadata in the structured JSON prompt so the model can associate each pixel region with its camera stream.}
    \label{fig:action_multiview_packaging}
\end{figure}

The \texttt{cinematography.framing} field is filled to describe the input or output viewpoints, covering first-person, third-person, wrist-mounted, or concatenated camera views.
The action entry spans the full generated clip, with the end time rounded from the measured duration; the top-level \texttt{duration} is truncated to integer seconds to remain consistent with the video JSON-caption format.
Aspect ratios are mapped from a fixed set of generator resolutions.
The optional \texttt{idle\_frame} field indicates how many frames contain no action (idle frames).
We omit idle-frame metadata for inverse-dynamics samples as action prediction should strictly follow the video instead of user preferences.
\cref{fig:action_multiview_packaging} shows a multiview example in which the concatenated camera layout is encoded directly in the JSON prompt. We use Claude-Opus-4.6 for prompt upsampling.

\subsection{Cosmos 3 Generator Negative Prompt}
\label{appendix:generator_negative_prompt}

We use the following negative prompt for the base Cosmos3-Nano and Cosmos3-Super generators.

\begin{tcolorbox}[
    breakable,
    colback=black!2,
    colframe=nvidiagreen!75!white,
    boxrule=0.4pt,
    arc=1pt,
    left=4pt,
    right=4pt,
    top=4pt,
    bottom=4pt,
    title=Cosmos 3 Base Generator Negative Prompt,
]
\begin{Verbatim}[breaklines=true,breakanywhere=true,fontsize=\scriptsize]
"subjects": [
  {
    "description": "Blurry, poorly defined subjects with inconsistent shapes and unrealistic proportions.",
    "appearance_details": "Distorted features, visible compression artifacts, muddy textures lacking fine detail, color bleeding between elements, and unnatural skin tones or surface textures that appear artificial or computer-generated.",
    "relationship": "Subjects appear disconnected from the environment, floating or improperly grounded in the scene without proper occlusion or spatial coherence.",
    "location": "Subjects are poorly placed within the frame, appearing at awkward positions that violate basic compositional rules.",
    "relative_size": "Inconsistent scale relationships between subjects and the environment, with objects appearing too large or too small relative to their surroundings.",
    "orientation": "Unnatural orientations that defy physics and spatial logic.",
    "pose": "Stiff, mannequin-like poses with unnatural joint angles and impossible limb positions that look computer-generated.",
    "action": "Incoherent motion with visible frame-to-frame discontinuities. Movement appears as a slideshow rather than smooth animation. Limbs and appendages pop between positions without interpolation.",
    "state_changes": "Visual state transitions are abrupt and jarring. Colors shift without motivation. Surface textures flicker between different materials randomly. Outlines shimmer and vibrate.",
    "clothing": "Clothing appears painted on with no sense of material weight or drape. Fabric textures are flat and repeat visibly.",
    "expression": "Frozen, uncanny valley expressions or expressions that change abruptly without natural transition.",
    "gender": "",
    "age": "",
    "skin_tone_and_texture": "Waxy, plastic-looking skin with visible artifacts and inconsistent texture resolution across the frame.",
    "facial_features": "Asymmetric facial features, extra fingers or limbs, teeth that appear blurry or malformed.",
    "number_of_subjects": 0,
    "number_of_arms": 0,
    "number_of_legs": 0
  },
  {
    "description": "Extremely low-quality subjects with visible rendering artifacts, broken mesh geometry, and completely unrealistic proportions throughout.",
    "appearance_details": "Distorted features, visible compression artifacts, muddy textures lacking fine detail, color bleeding between elements, and unnatural skin tones or surface textures that appear artificial or computer-generated.",
    "relationship": "Subjects appear disconnected from the environment, floating or improperly grounded in the scene without proper occlusion or spatial coherence.",
    "location": "Subjects are poorly placed within the frame, appearing at awkward positions that violate basic compositional rules.",
    "relative_size": "Inconsistent scale relationships between subjects and the environment, with objects appearing too large or too small relative to their surroundings.",
    "orientation": "Unnatural orientations that defy physics and spatial logic.",
    "pose": "Stiff, mannequin-like poses with unnatural joint angles and impossible limb positions that look computer-generated.",
    "action": "Incoherent motion with visible frame-to-frame discontinuities. Movement appears as a slideshow rather than smooth animation. Limbs and appendages pop between positions without interpolation.",
    "state_changes": "Visual state transitions are abrupt and jarring. Colors shift without motivation. Surface textures flicker between different materials randomly. Outlines shimmer and vibrate.",
    "clothing": "Clothing appears painted on with no sense of material weight or drape. Fabric textures are flat and repeat visibly.",
    "expression": "Frozen, uncanny valley expressions or expressions that change abruptly without natural transition.",
    "gender": "",
    "age": "",
    "skin_tone_and_texture": "Waxy, plastic-looking skin with visible artifacts and inconsistent texture resolution across the frame.",
    "facial_features": "Asymmetric facial features, extra fingers or limbs, teeth that appear blurry or malformed.",
    "number_of_subjects": 0,
    "number_of_arms": 0,
    "number_of_legs": 0
  },
  {
    "description": "Poorly generated subjects exhibiting all hallmarks of failed neural rendering -- flickering edges, inconsistent depth, and uncanny spatial relationships.",
    "appearance_details": "Distorted features, visible compression artifacts, muddy textures lacking fine detail, color bleeding between elements, and unnatural skin tones or surface textures that appear artificial or computer-generated.",
    "relationship": "Subjects appear disconnected from the environment, floating or improperly grounded in the scene without proper occlusion or spatial coherence.",
    "location": "Subjects are poorly placed within the frame, appearing at awkward positions that violate basic compositional rules.",
    "relative_size": "Inconsistent scale relationships between subjects and the environment, with objects appearing too large or too small relative to their surroundings.",
    "orientation": "Unnatural orientations that defy physics and spatial logic.",
    "pose": "Stiff, mannequin-like poses with unnatural joint angles and impossible limb positions that look computer-generated.",
    "action": "Incoherent motion with visible frame-to-frame discontinuities. Movement appears as a slideshow rather than smooth animation. Limbs and appendages pop between positions without interpolation.",
    "state_changes": "Visual state transitions are abrupt and jarring. Colors shift without motivation. Surface textures flicker between different materials randomly. Outlines shimmer and vibrate.",
    "clothing": "Clothing appears painted on with no sense of material weight or drape. Fabric textures are flat and repeat visibly.",
    "expression": "Frozen, uncanny valley expressions or expressions that change abruptly without natural transition.",
    "gender": "",
    "age": "",
    "skin_tone_and_texture": "Waxy, plastic-looking skin with visible artifacts and inconsistent texture resolution across the frame.",
    "facial_features": "Asymmetric facial features, extra fingers or limbs, teeth that appear blurry or malformed.",
    "number_of_subjects": 0,
    "number_of_arms": 0,
    "number_of_legs": 0
  }
],
"background_setting": "A poorly rendered, flat background with visible seams, repeated textures, and inconsistent depth cues. The environment lacks volumetric depth and appears as a painted backdrop rather than a three-dimensional space. Vegetation looks like flat cutouts with no volumetric depth. The background appears to have been composited from multiple source materials at different resolutions, creating visible seams and edge artifacts where elements meet. Textures swim and shift across surfaces in a way that breaks the illusion of solidity -- patterns drift laterally rather than staying anchored to the geometry they belong to. Background elements flicker in and out of existence between frames, particularly at the edges of the field of view. The rendering resolution is visibly lower for distant elements, creating a jarring transition between near and far objects. Cloud textures repeat obviously in the sky with visible tiling. Water surfaces lack proper reflection and refraction, appearing as flat animated textures. Fog and atmospheric effects pop in and out rather than smoothly transitioning. Trees and vegetation exhibit obvious LOD (level-of-detail) switching. Building facades have inconsistent window spacing and pattern repetition. The overall scene feels like a poorly assembled collage of individually rendered elements rather than a coherent whole.",
"lighting": {
  "conditions": "Harsh, flat lighting with no natural variation. The scene appears uniformly lit as if by a single overhead fluorescent light, removing all sense of depth and atmosphere.",
  "direction": "Inconsistent light sources -- shadows point in multiple contradictory directions, breaking physical plausibility.",
  "shadows": "Hard-edged, unrealistic shadows that pop in and out of existence between frames. Some objects cast no shadows while others have impossibly dark ones that don't animate smoothly with the object's motion. Shadow edges exhibit visible staircase aliasing artifacts. Shadow maps appear to have been rendered at extremely low resolution, creating blocky patterns. Self-shadowing on characters shows visible peter-panning artifacts where shadows detach from their source. Contact shadows between objects and the ground appear and disappear as objects move slightly. Shadow color is pure black with no ambient contribution, creating an unnaturally harsh contrast that flattens the image. Multiple shadow cascades have visible boundaries where resolution changes. The shadow rendering appears to be temporally unstable -- even static objects have shadows that shimmer and crawl frame to frame, breaking the illusion of a stable light source.",
  "illumination_effect": "No bounce light, no ambient occlusion, no subtle color interactions between surfaces. The scene looks like a poorly lit 3D render from the early 2000s."
},
"aesthetics": {
  "composition": "Cluttered, poorly framed composition with no clear focal point. Important elements are cut off by the frame edges. The rule of thirds is completely ignored, leading to an unbalanced and visually unpleasant arrangement.",
  "color_scheme": "Oversaturated, garish colors that clash violently. Color banding is visible in gradient areas. The overall palette feels artificial and digitally processed rather than natural.",
  "mood_atmosphere": "Unsettling, uncanny atmosphere that fails to evoke any intended emotional response. The scene feels lifeless and sterile despite attempting to portray dynamic action.",
  "patterns": "Visible tiling artifacts in textures, moir\u00e9 patterns, and aliasing on edges."
},
"cinematography": {
  "camera_motion": "Extremely shaky, unstable camera with visible rolling shutter artifacts. The motion is jerky and discontinuous, causing motion sickness and making the scene impossible to follow.",
  "framing": "Poorly framed shots that cut off important elements and include unnecessary empty space.",
  "camera_angle": "Awkward, disorienting camera angles that provide no useful spatial information about the scene. The camera path exhibits visible mathematical artifacts suggesting simple interpolation between keyframes rather than natural camera operation. Camera motion is completely disconnected from the scene content -- panning away from action, dollying during dialogue, and shaking during still moments. The camera appears to pass through solid objects occasionally. Zoom is applied digitally rather than optically, revealing progressively worse resolution. Camera motion exhibits non-physical acceleration profiles -- instant starts and stops rather than smooth ease-in/ease-out. Rolling shutter simulation is applied inconsistently, present in some frames but not others. The camera occasionally exhibits impossible motion like teleporting between positions. Virtual camera stabilization creates an uncanny floating sensation disconnected from any physical camera rig.",
  "depth_of_field": "Uniform focus throughout, creating a flat, documentary-like appearance with no cinematic depth separation.",
  "focus": "Soft, out-of-focus imagery with visible chromatic aberration and lens distortion that was not corrected in post-processing.",
  "lens_focal_length": "Inappropriate focal length causing barrel distortion and unnatural perspective compression."
},
"style_medium": "Low quality compressed digital video with visible encoding artifacts",
"artistic_style": "Amateur, unpolished with inconsistent visual style",
"context": "A poorly produced video with numerous technical and artistic flaws that detract from any intended narrative or visual impact.",
"actions": [
  {
    "time": "0:00-0:08",
    "description": "Subjects attempt to move but their motion is jerky, temporally inconsistent, and physically implausible. Background elements flicker and shift between frames."
  }
],
"text_and_signage_elements": [],
"segments": [
  {
    "segment_index": 0,
    "time_range": "0:00-0:08",
    "description": "A single continuous shot suffering from severe temporal inconsistencies -- subjects that morph and deform between frames, backgrounds that shift and wobble, and rendering quality that fluctuates visibly over time. Motion blur is applied incorrectly, smearing in directions that don't match actual movement. Frame-to-frame coherence breaks down with individual pixels changing color randomly in flat areas. Texture detail level fluctuates between frames as if the rendering budget varied shot to shot. Color grading drifts over the duration with no creative motivation. Noise patterns change between frames in ways that draw attention rather than being invisible. Overall visual quality degrades progressively from start to finish.",
    "key_changes": "No meaningful progression or narrative development. Visual quality degrades over time.",
    "camera": "Unstable, poorly controlled camera work with visible mathematical interpolation artifacts."
  }
],
"transitions": [],
"temporal_caption": "The scene opens at 0.0 seconds with a poorly rendered establishing shot that immediately reveals low production quality. At 1.0 seconds, subjects begin to move but their motion is jerky and inconsistent, with limbs bending at unnatural angles and objects clipping through each other. From 2.0 to 4.0 seconds, the camera shakes violently while the scene exhibits visible compression artifacts, color banding in the sky, and flickering in the shadows. Between 4.0 and 6.0 seconds, temporal coherence breaks down as elements appear and disappear between frames, textures swim and morph unnaturally, and the lighting shifts abruptly without physical cause. In the final 2 seconds, the overall visual quality deteriorates further with increasing noise, blur, and a general loss of spatial coherence that makes the scene nearly unwatchable. Additionally, the frame rate appears inconsistent with visible judder and stuttering throughout. Color temperature shifts randomly between warm and cool tones with no motivation. The encode quality degrades in complex regions showing macro-blocking and mosquito noise around moving edges. Temporal noise patterns are spatially correlated, creating swimming artifacts on flat surfaces.",
"audio_description": "",
"physical_realism": "No adherence to physical laws. Objects defy gravity, pass through solid surfaces, and change mass and momentum without cause. Fluid dynamics, cloth simulation, and rigid body physics are all fundamentally broken. Furthermore, conservation of energy is violated as objects gain or lose kinetic energy spontaneously. Elastic collisions produce inelastic results and vice versa. Surface friction is inconsistent -- objects slide on rough surfaces while sticking to smooth ones. Air resistance appears to affect only some objects while others move through the atmosphere unimpeded."
}
\end{Verbatim}
\end{tcolorbox}

\subsection{Agentic Upsampling for Cosmos3-Super-Text2Image}
\label{appendix:t2i_agentic_upsampling}

The majority of top (closed-source) text-to-image generation models do some variation of in-the-loop iterative refinement and/or multi-modal reasoning \citep{gpt_image_2}.
The ability for Cosmos 3 models to accept JSON-structured prompts opens a variety of opportunities for fine-grained iterative agentic refinement. We put this to the test in our Artificial Analysis submission. At test-time, we infer Cosmos3-Super-Text2Image via an agentic harness that iteratively upsamples the user prompt, scores the generated image by outlining its flaws/issues (if any) and giving a score between 1 to 10, and re-writes both the positive and negative prompt. The output result is simply the best image of this loop, as per our critic's overall score. We use at most 2 re-write iterations, and do early stopping if the score is at least 9 and if there are no severe issues highlighted by the critic.

For the first iteration, we use the LLM upsampler template found in Appendix~\ref{appendix:sft_t2i_upsampler_prompt}, and an empty negative prompt.
The VLM critic prompt template is found in Box~\ref{box:agentic_t2i_critic_prompt}. The LLM positive/negative rewriter prompt template is found in Box~\ref{box:agentic_t2i_rewriter_prompt}.
For our submission we use GPT-5.5 for caption upsampling and rewriting, and Gemini3.1-Pro for the critic. The harness is, by design, flexible and can accept any LLM/VLM, including Cosmos 3 (reasoning tower) itself. For more information and scripts, please visit the \href{https://huggingface.co/nvidia/Cosmos3-Super-Text2Image/blob/main/AGENTIC_UPSAMPLING.md}{nvidia/Cosmos3-Super-Text2Image} HuggingFace repository.

\begin{samepage}
The loop structure is:
\begin{center}
\small
\setlength{\fboxsep}{4pt}
\includegraphics[width=0.93936\linewidth]{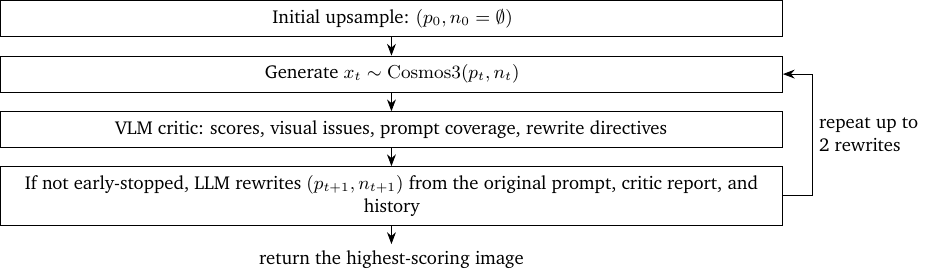}
\end{center}
\end{samepage}

\newcounter{agenticpromptbox}
\refstepcounter{agenticpromptbox}\label{box:agentic_t2i_critic_prompt}
\begin{tcolorbox}[
    breakable,
    colback=black!2,
    colframe=nvidiagreen!75!white,
    boxrule=0.4pt,
    arc=1pt,
    left=4pt,
    right=4pt,
    top=4pt,
    bottom=4pt,
    title=Box~\theagenticpromptbox: Agentic T2I Critic Prompt
]
\begin{Verbatim}[breaklines=true,breakanywhere=true,fontsize=\scriptsize]
You are an expert image quality analyst specialising in AI-generated image evaluation.
Your job is to produce an exhaustive defect report. Be meticulous: go beyond the obvious
problems and look carefully for subtle, fleeting, or background issues too.

The following image was generated by an AI image model.
{generated_image}

The attached image was generated from this prompt:
{user_text_prompt}

Analyze this image carefully and list EVERY quality issue you observe.
For each issue give an approximate location and name the specific object or
region involved. Report each distinct occurrence separately.

To make sure you don't miss anything, mentally step through these areas before finalising
your list — but only report issues you actually see:
• Physics: gravity violations, impossible collisions, implausible trajectories
• Object deformation: morphing, melting, stretching of solid objects
• Anatomy: distorted hands, faces, fingers, limbs; wrong body proportions
• Lighting & shadows: missing shadows, inconsistent illumination
• Depth & scale: wrong spatial relationships, perspective issues, scale inconsistencies
• Text / numbers: garbled, floating, or shifting text and digits
• Visual quality: blur patches, noise, compression blocking, visual artefacts, low-resolution regions
• Colour: inconsistent coloration, bleeding, banding
• Action correctness: prompted actions are correctly displayed
• Prompt following: missing subjects, wrong objects, wrong setting, wrong action

Depending on the category of the prompt, you may also need to apply additional checks from the following list:
- Text/commercial/UI/logo checks: readable text for logos, labels, posters, billboards, product packaging, or UI. Verify exact quoted strings, spelling, legibility, typography, placement, layout, and whether commercial/UI intent is visually clear.
- People/anatomy checks: if humans, human-like characters, body parts, portraits, or poses are present or required by the prompt, inspect faces, eyes, hands, fingers, limbs, pose, proportions, expression, clothing coherence, and physically possible interactions.
- Fantasy/cartoon/vector/pixel-art checks: if a stylized medium is requested, judge whether stylization is intentional and clean. Penalize messy geometry, inconsistent line language, broken vector shapes, muddy palettes, and unwanted photorealistic texture.
- Photorealistic/physical checks: if realism, physical objects, geometry, camera behavior, reflections, transparent materials, shadows, perspective, scale, or contact matter, judge material realism, lighting physics, lens plausibility, and whether objects obey real-world physical constraints.

Return exactly one JSON object, no markdown fences and no prose outside JSON:
{
  "prompt_adherence_score": <number 0-10>,
  "visual_quality_score": <number 0-10>,
  "aesthetics_score": <number 0-10>,
  "physical_plausibility_score": <number 0-10>,
  "category_score": <number 0-10>,
  "text_rendering_score": <number 0-10 or null>,
  "photorealism_score": <number 0-10 or null>,
  "overall_score": <number 0-10>,
  "issues": [
    {
      "category": "<concise label of your choosing>",
      "description": "<what, where in frame, at what timestamp>",
      "severity": "minor" | "moderate" | "severe"
    }
  ],
  "prompt_elements": {
    "<key noun or action from the prompt>": "present" | "absent" | "partial"
  },
  "category_findings": {{"<check area>": "<concise finding>"}},
  "improvement_directives": ["<specific prompt rewrite instruction>"],
  "rationale": "<2-4 concise sentences>"
}
\end{Verbatim}
\end{tcolorbox}

\refstepcounter{agenticpromptbox}\label{box:agentic_t2i_rewriter_prompt}
\begin{tcolorbox}[
    breakable,
    colback=black!2,
    colframe=nvidiagreen!75!white,
    boxrule=0.4pt,
    arc=1pt,
    left=4pt,
    right=4pt,
    top=4pt,
    bottom=4pt,
    title=Box~\theagenticpromptbox: Agentic T2I Joint Positive/Negative Rewriter Prompt
]
\begin{Verbatim}[breaklines=true,breakanywhere=true,fontsize=\scriptsize]
You are a precise text-to-image prompt engineer. Return valid JSON only, no markdown.
Jointly coordinate the positive structured prompt and generator-side negative prompt so they do not contradict each other.

Original user prompt:
{user_text_prompt}

Application-specific guidance:
Apply the following sections as one checklist program. Do not first classify the prompt.
Apply each section only when relevant to the original user prompt, previous JSON, or VLM failures.
- Text/commercial/UI/logo checks: readable text for logos, labels, posters, billboards, product packaging, or UI. Verify exact quoted strings, spelling, legibility, typography, placement, layout, and whether commercial/UI intent is visually clear.
- People/anatomy checks: if humans, human-like characters, body parts, portraits, or poses are present or required by the prompt, inspect faces, eyes, hands, fingers, limbs, pose, proportions, expression, clothing coherence, and physically possible interactions.
- Fantasy/cartoon/vector/pixel-art checks: if a stylized medium is requested, judge whether stylization is intentional and clean. Penalize messy geometry, inconsistent line language, broken vector shapes, muddy palettes, and unwanted photorealistic texture.
- Photorealistic/physical checks: if realism, physical objects, geometry, camera behavior, reflections, transparent materials, shadows, perspective, scale, or contact matter, judge material realism, lighting physics, lens plausibility, and whether objects obey real-world physical constraints.
- General scene checks: always judge object completeness, layout clarity, subject relationships, background coherence, visual appeal, and absence of obvious AI artifacts.

Previous generated image failed or scored according to this VLM analysis:
{
  "overall_score": <number 0-10>,
  "prompt_adherence_score": <number 0-10>,
  "visual_quality_score": <number 0-10>,
  "aesthetics_score": <number 0-10>,
  "physical_plausibility_score": <number 0-10>,
  "category_score": <number 0-10>,
  "text_rendering_score": <number 0-10 or null>,
  "photorealism_score": <number 0-10 or null>,
  "issues": [
    {
      "category": "<concise issue label>",
      "description": "<what failed and where>",
      "severity": "minor" | "moderate" | "severe"
    }
  ],
  "prompt_elements": {"<prompt element>": "present" | "absent" | "partial"},
  "category_findings": {"<check area>": "<concise finding>"},
  "improvement_directives": ["<specific prompt rewrite instruction>"],
  "rationale": "<brief rationale>"
}

Iteration history summary:
[
  {
    "iteration": <integer>,
    "overall_score": <number 0-10>,
    "prompt_adherence_score": <number 0-10>,
    "category_score": <number 0-10>,
    "threshold_cleared": <boolean>
  }
]

Previous positive JSON prompt:
{previous_t2i_json_prompt}

Previous negative prompt:
{previous_negative_prompt}

Joint rewrite task:
Return a JSON object with exactly two top-level keys: "positive_prompt" and "negative_prompt".
"positive_prompt" must be a complete JSON object with exactly these top-level keys, preserving their names and types:
{schema_keys}

"positive_prompt" must keep resolution previous "resolution" and "aspect_ratio".
"negative_prompt" must be a concise generator-side negative prompt string.
Coordinate both fields: strengthen required positive constraints while using the negative prompt only to suppress concrete wrong alternatives or artifacts.
Do not put positive instructions in negative_prompt. Do not negate content required by the original user prompt.
For exact counts, grids, text, geometry, or anatomy, explicitly block wrong alternatives when useful.
The positive "comprehensive_t2i_caption" should be direct generation guidance, not an explanation of this rewrite process.
\end{Verbatim}
\end{tcolorbox}

\section{Synthetic Dataset for Generator Training}
\label{appendix:sdg_datasets}

This appendix provides detailed descriptions of each synthetic data generation (SDG) dataset used in the generator mid-training, the distribution of the SDG datasets with respect to the pre-training video dataset, and a detailed ablation study with these SDG datasets in Cosmos 3 training. \Cref{tab:sdg_overview} summarizes the scale and provided modalities across the five datasets, with URL links to Hugging Face datasets; per-dataset cards follow.

\begin{table}[h]
\centering
\captionsetup{justification=raggedright, singlelinecheck=false}
\caption{\textbf{Overview of SDG datasets}. RGB, depth, instance segmentation (Seg), bounding boxes (BBox), physics state (Phys), camera parameters (Cam), and captions (Cap) indicate whether the modality is provided. ``part.'' denotes partial coverage (subset of clips or specific generators only).}
\label{tab:sdg_overview}
\small
\setlength{\tabcolsep}{4pt}
\begin{tabular}{lrlccccccc}
\toprule
Dataset & Clips & Resolution / FPS & RGB & Depth & Seg & BBox & Phys & Cam & Cap \\
\midrule
\href{https://huggingface.co/datasets/nvidia/PhysicalAI-WorldModel-Synthetic-Physical-Interaction-Scenes}{SDG-PhyxSim}   & 76{,}489  & $1920{\times}1080$ / 30 & \checkmark & \checkmark & \checkmark & ---        & \checkmark & \checkmark & \checkmark \\
\href{https://huggingface.co/datasets/nvidia/PhysicalAI-WorldModel-Synthetic-Embodied-Robot-Scenes}{SDG-RobotSim}  & 208{,}022 & varies                  & \checkmark & part.      & part.      & ---        & part.      & \checkmark & \checkmark \\
\href{https://huggingface.co/datasets/nvidia/PhysicalAI-WorldModel-Synthetic-Autonomous-Driving-Scenarios}{SDG-DriveSim}  & 264{,}000 & $3840{\times}2160$ / 24 & \checkmark & ---        & ---        & ---        & ---        & ---        & \checkmark \\
\href{https://huggingface.co/datasets/nvidia/PhysicalAI-WorldModel-Synthetic-Digital-Human-Scenes}{SDG-SynHuman}  & 236{,}937 & $1920{\times}1080$ / 30 & \checkmark & \checkmark & ---        & ---        & ---        & \checkmark & ---        \\
\href{https://huggingface.co/datasets/nvidia/PhysicalAI-WorldModel-Synthetic-Warehouse-Operations-Scenes}{SDG-Warehouse} & 122{,}952 & $1920{\times}1080$ / 30 & \checkmark & \checkmark & \checkmark & \checkmark & ---        & \checkmark & ---        \\
\bottomrule
\end{tabular}
\end{table}

\subsection{SDG-PhyxSim}
\label{appendix:sdg_phyxsim}

\begin{figure}[htbp]
    \centering
    \captionsetup{justification=raggedright, singlelinecheck=false}
    \footnotesize
    \setlength{\tabcolsep}{2pt}
    \renewcommand{\arraystretch}{1.0}
    \begin{tabular}{@{}*{5}{c@{\hspace{2pt}}}@{}}
        RGB & Center of mass & Rotation & Linear velocity & Angular velocity \\[2pt]
        \includegraphics[width=0.19\textwidth]{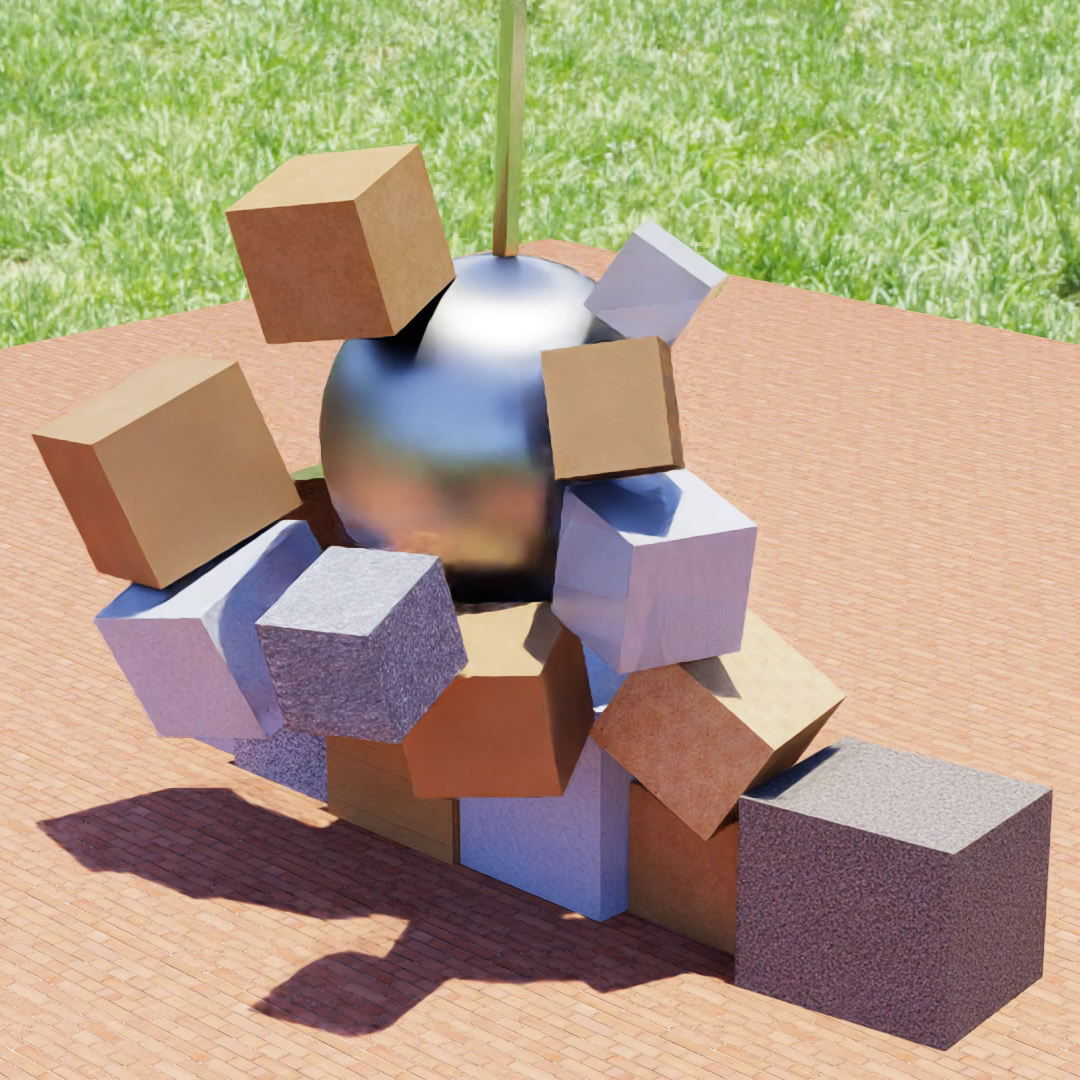} &
        \includegraphics[width=0.19\textwidth]{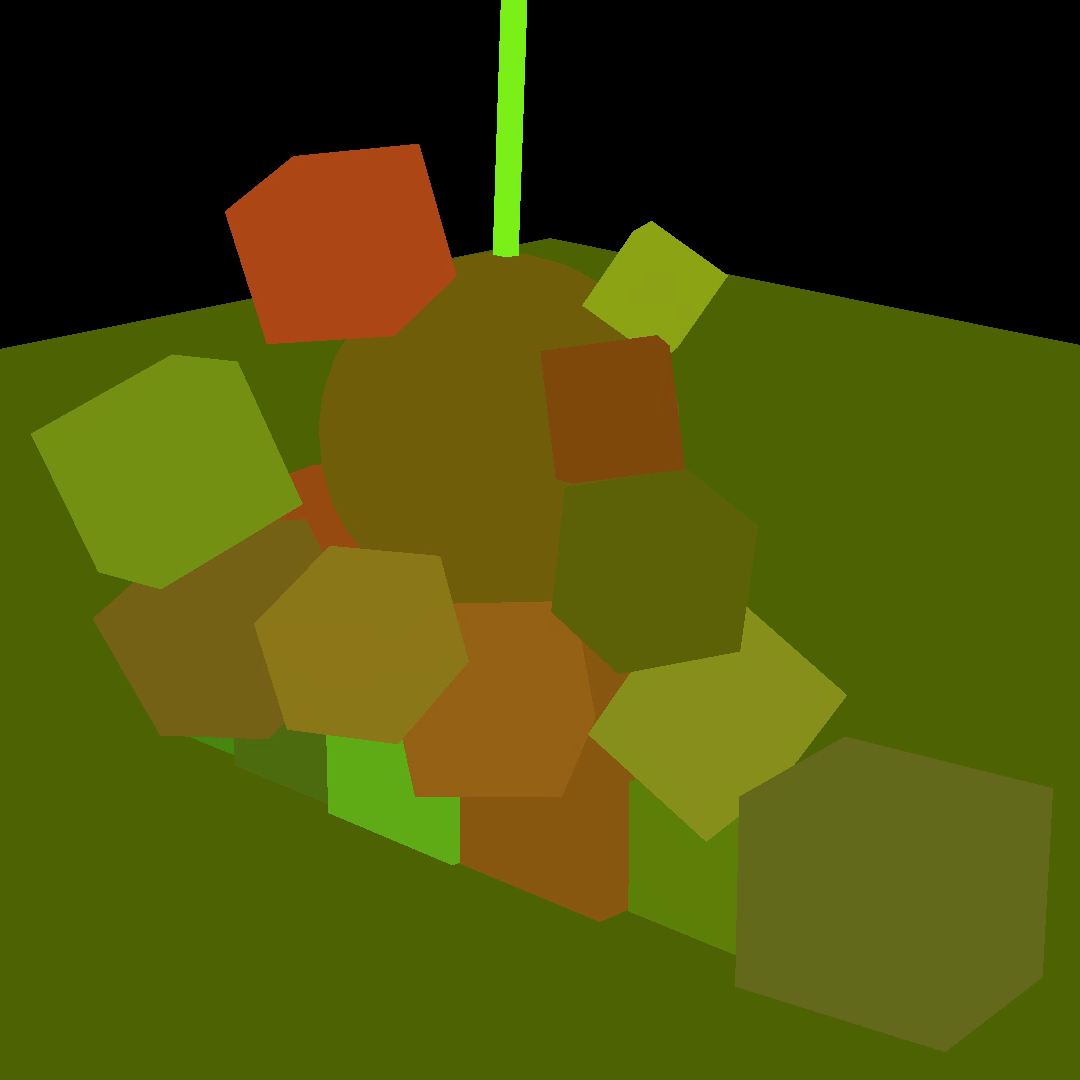} &
        \includegraphics[width=0.19\textwidth]{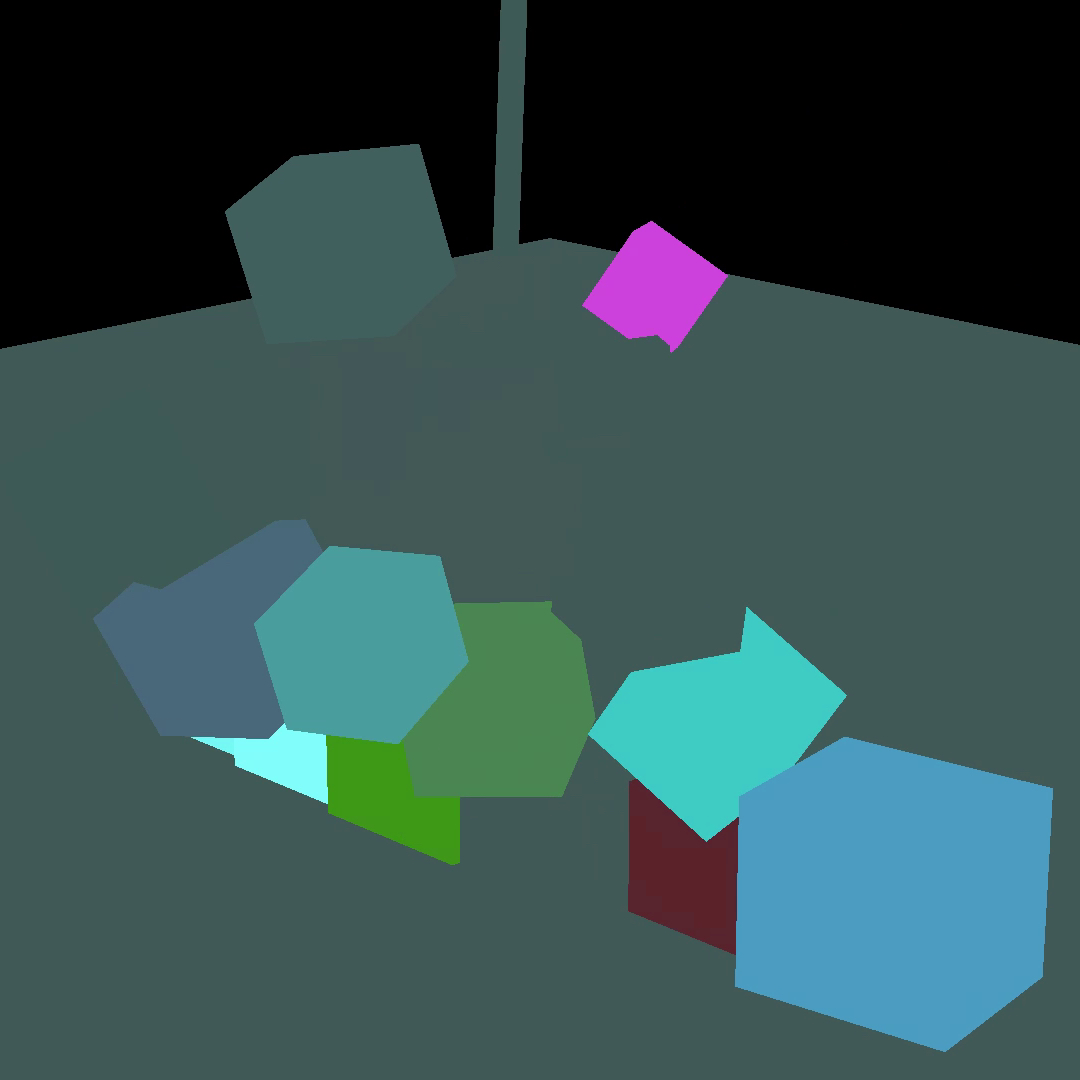} &
        \includegraphics[width=0.19\textwidth]{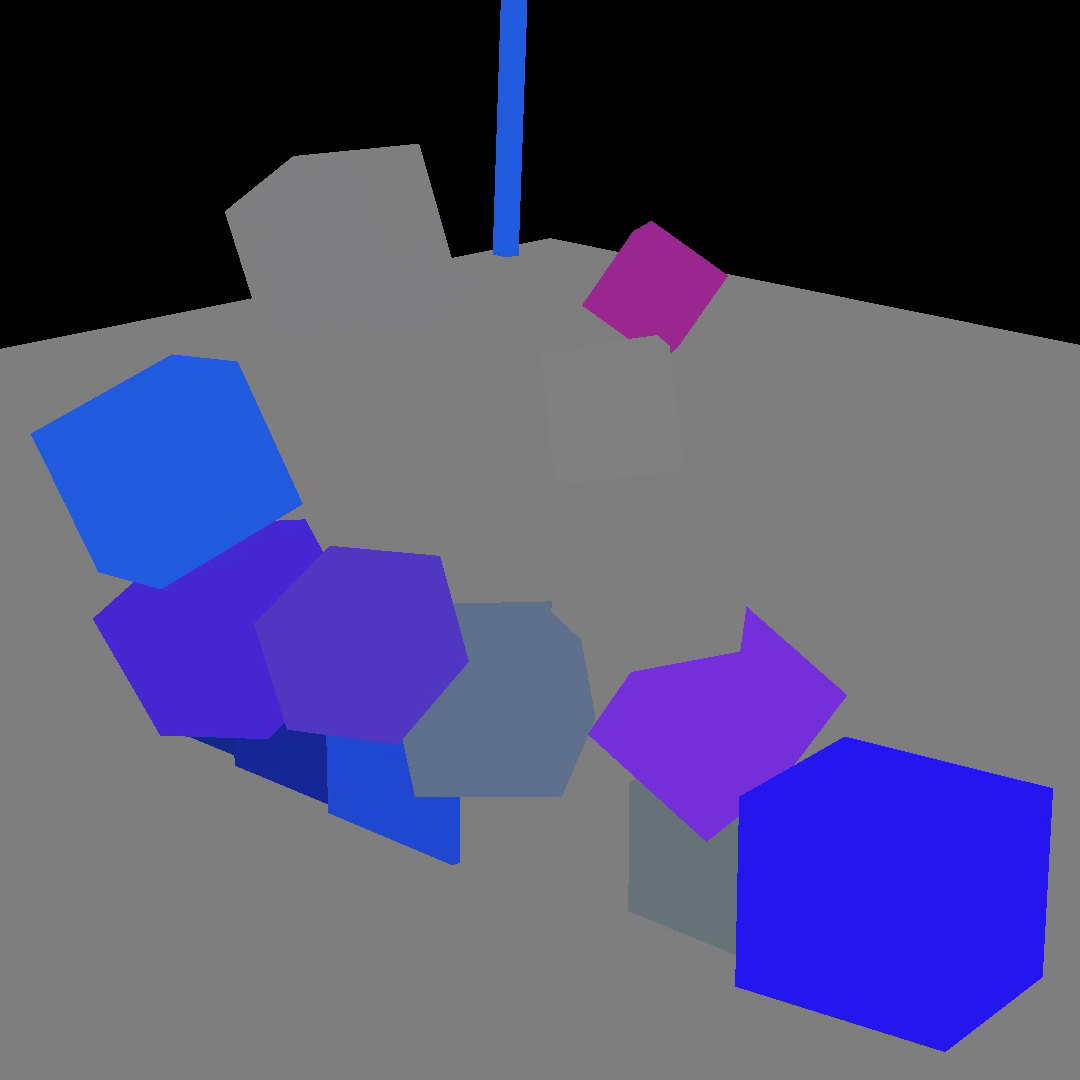} &
        \includegraphics[width=0.19\textwidth]{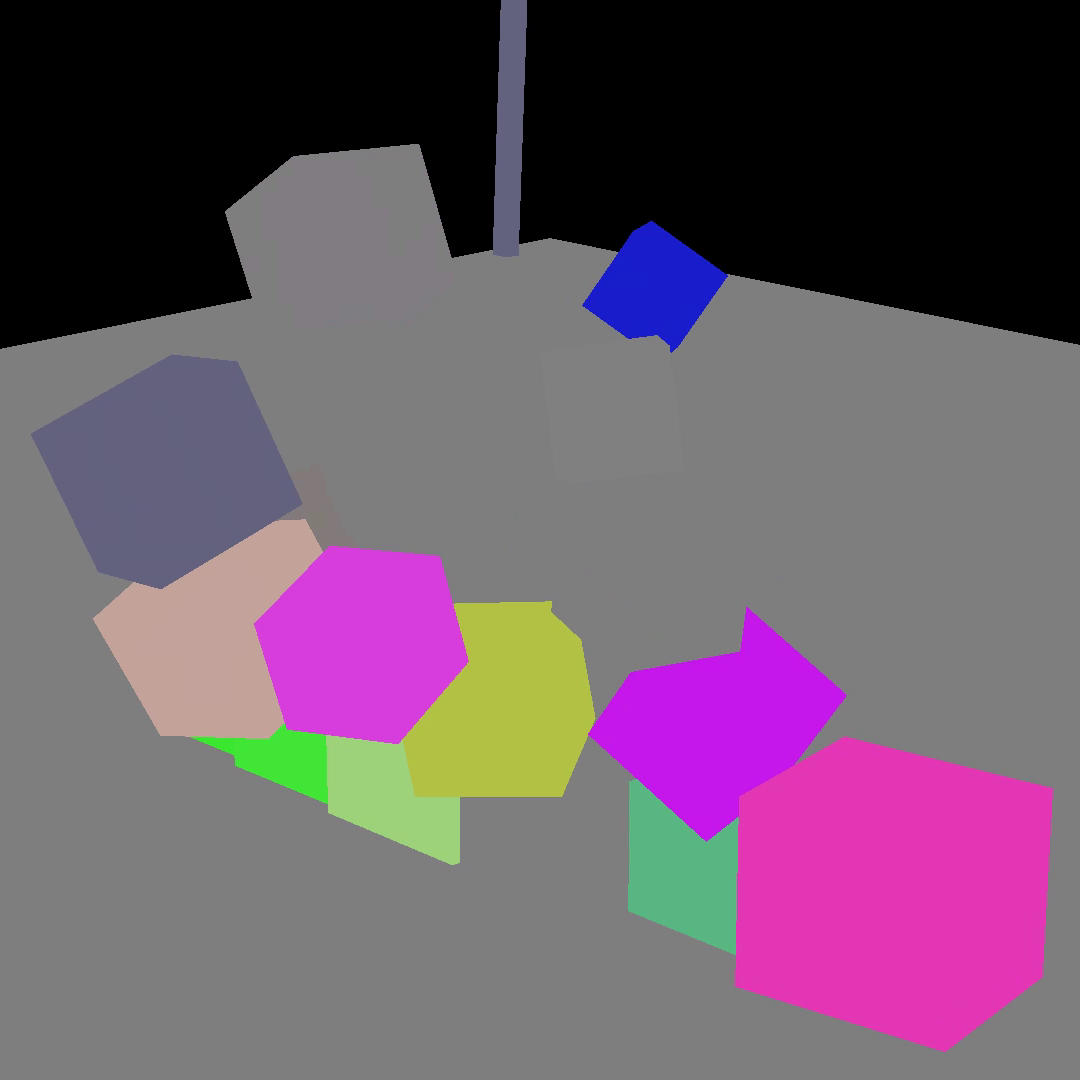} \\
    \end{tabular}
    \caption{\textbf{SDG-PhyxSim}. A single frame of the \texttt{wrecking\_ball} scene at the
    moment of impact (\texttt{Corner} camera). From left to right: RGB,
    center-of-mass displacement, cumulative rotation, linear velocity, and
    angular velocity.}
    \label{fig:sdg_phyxsim_modalities}
\end{figure}

\paragraph{Overview.}
\label{appendix:sdg_phyxsim_overview}
SDG-PhyxSim (PhysicsAI-WorldModel-Synthetic-Physical-Interaction-Scenes) is a large-scale synthetic video dataset of physically simulated
multi-object interaction scenes, designed to expose Cosmos 3 to dense,
ground-truthed rigid-body dynamics that are difficult to obtain from real
video. Every simulation run is captured from four fixed camera viewpoints
simultaneously, yielding four synchronized 5--8\,s, $1920{\times}1080$,
30\, FPS clips per run, each paired with per-frame instance segmentation,
lossless metric depth, and structured per-object physics annotations
(linear velocity, angular velocity, center-of-mass displacement, cumulative
rotation) read directly from the simulator at render time. The release
covers ten procedurally parameterized scene families
(\texttt{dominoes}, \texttt{ball\_mixer}, \texttt{bowling}, \texttt{billiards},
\texttt{towers}, \texttt{wrecking\_ball}, \texttt{objects\_falling},
\texttt{rolling\_ramp\_objects}, \texttt{rolling\_ramp\_obstruct}, and
\texttt{obstruction}), each chosen to exercise a distinct class of physical
phenomena---cascading impact chains, multi-body mixing, ballistic
trajectories, constrained-pendulum dynamics, freefall and settling,
gravity-driven rolling, mid-path deflection, object permanence, and
concurrent multi-directional collisions.

\paragraph{Simulation setup.}
\label{appendix:sdg_phyxsim_simulation_setup}
SDG-PhyxSim is generated with NVIDIA Isaac Sim~\citep{nvidia_isaac_sim_2026} using the PhysX rigid-body
engine, and captured with NVIDIA Omniverse
Replicator. Each scene is authored as a self-contained USD asset whose
geometry, material bindings (with physical properties), initial poses,
kinematic constraints, and randomization parameters are fully captured in a
single \texttt{.usda} file, so any clip can be replayed exactly in Isaac Sim
from its scene file alone. Simulation accuracy is set high---16 PhysX
substeps per rendered frame for stable collision resolution---and a short
warmup phase ($0.01$\,s at $0.001$\,s/step) is run before capture to settle
objects into a quiescent initial state. Every clip is keyed by a
\texttt{(scene\_name, scene\_hash, seed)} triple; the integer seed
deterministically controls all randomized scene parameters (object counts,
sizes, masses, materials, spacings, initial velocities).

\paragraph{Scenes.}
\label{appendix:sdg_phyxsim_scenes}
The ten scene families and their target phenomena are:
\begin{itemize}
    \item \texttt{dominoes}: sequential momentum transfer in a curved chain,
    \item \texttt{ball\_mixer}: persistent multi-body mixing under a rotating paddle, $8$\,s clips,
    \item \texttt{bowling}: directed rolling impact into a pin formation,
    \item \texttt{billiards}: elastic multi-ball collisions with spin transfer on a bumpered table,
    \item \texttt{towers}: structural collapse of stacked block arches under a projectile,
    \item \texttt{wrecking\_ball}: constrained pendulum dynamics and high-impulse demolition of cubes,
    \item \texttt{objects\_falling}: freefall, impact, bounce, settling of mixed props,
    \item \texttt{rolling\_ramp\_objects}: rolling and rotational inertia down an angled ramp,
    \item \texttt{rolling\_ramp\_obstruct}: the ramp scene with static obstacles for mid-path deflection and object permanence,
    and
    \item \texttt{obstruction}: multi-directional rolling balls through a field of static pins, with ricochets and pin-occluded object permanence.
\end{itemize}
Per seed, ball diameter, initial velocity, paddle RPM, enclosure size, object count, material assignments, ramp angle, and obstacle layout are randomized.

\paragraph{Dataset statistics.}
\label{appendix:sdg_phyxsim_dataset_statistics}
The SDG-PhyxSim release comprises 76{,}489 independent simulation
runs. Most scenes produce $5$\,s ($150$-frame) clips; \texttt{ball\_mixer}
produces $8$\,s ($240$-frame) clips. All clips are rendered at
$1920{\times}1080$ and $30$\,fps, yielding approximately $57$\,M RGB frames.
Each run is captured from four fixed cameras whose names vary by scene
family---for example, \texttt{Front}, \texttt{Side}, \texttt{TopDown},
and \texttt{Corner} for the \texttt{wrecking\_ball} scene. The release totals 1{,}529{,}752 rendered MP4 files (RGB plus
physics-colorized variants), 346{,}147 depth videos, and
47{,}073{,}033 per-frame segmentation PNGs, with aggregate storage of
${\sim}14.9$\,TiB dominated by lossless depth video.

\paragraph{Metadata and annotations.}
\label{appendix:sdg_phyxsim_metadata_annotations}
In addition to the modalities listed in~\cref{tab:sdg_overview}, every camera of every run provides:
\begin{itemize}
    \item \textit{Dynamic per-object physics (NPZ).} Four files per camera---linear velocity (m/s), angular velocity (deg/s), cumulative rotation (rad), and center-of-mass displacement (m)---each an (\texttt{objects}\,$\times$\,\texttt{frames}\,$\times$\,$3$) tensor indexed by segmentation color, with per-axis bounds used to normalize the physics-colorized videos.
    \item \textit{Static per-object physics (JSON).} World gravity, and per-rigid-body mass, diagonal inertia tensor, body-frame center of mass, principal-axes quaternion, static and dynamic friction, restitution, density, and collision flag, keyed by USD prim path and segmentation color.
    \item \textit{Physics-colorized videos.} Per-camera, per-quantity MP4s in which each object's color encodes its instantaneous physics state (red\,=\,X, green\,=\,Y, blue\,=\,Z; stationary objects gray; saturation grows with magnitude), normalized to the NPZ bounds so the same color maps to the same physical magnitude within a clip (see~\cref{fig:sdg_phyxsim_modalities}).
    \item \textit{Scene file (USD).} The self-contained \texttt{.usda} used to produce the run, replayable exactly in Isaac Sim.
\end{itemize}
Metric depth is encoded as a 16-bit FFV1 MKV per camera, with the per-camera quantization ceiling $d_{\max}$ and observed range recorded in \texttt{depth\_metadata.json}, so metric depth recovers as $d = (v/65535)\,d_{\max}$ (the value $65535$ marks invalid depth). Instance segmentation PNGs ship with a color$\rightarrow$USD-prim-path mapping for cross-frame identity, and the per-frame camera JSON records intrinsics, extrinsics, FOV, elevation, and gravity orientation. All annotations are produced deterministically from the simulator and USD scene graph.

\subsection{SDG-RobotSim}
\label{appendix:sdg_robotsim}

\begin{figure}[htbp]
    \centering
    \captionsetup{justification=raggedright, singlelinecheck=false}
    \includegraphics[width=0.95\textwidth]{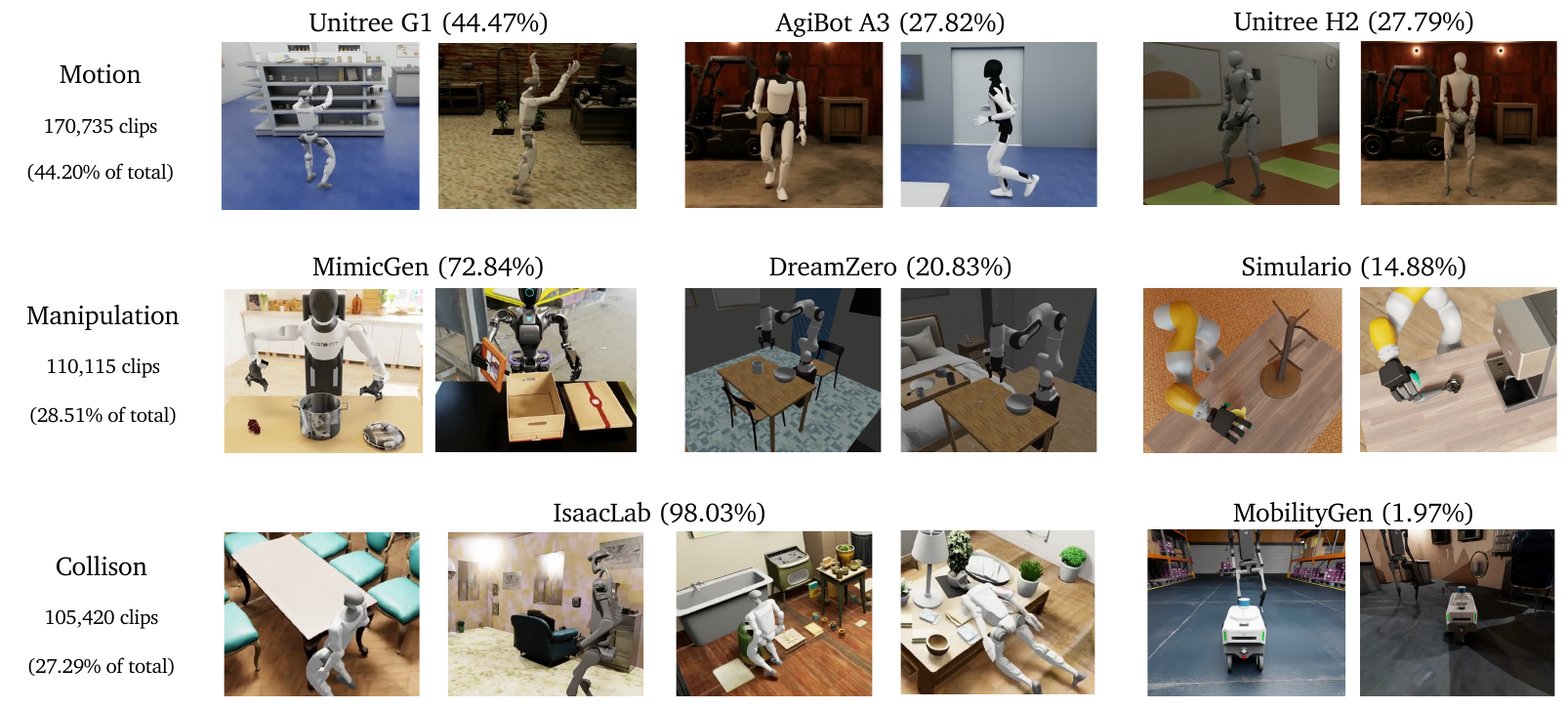}
    \caption{\textbf{Overview of the SDG-RobotSim dataset.} Clips are partitioned into three task categories---Motion, Manipulation, and Collision---with each category further broken down by its dominant robot embodiment.}
    \label{fig:sdg_robotsim_dataset_overview}
\end{figure}

\paragraph{Overview.}
\label{appendix:sdg_robotsim_overview}
PhysicalAI-WorldModel-Synthetic-Embodied-Robot-Scenes, abbreviated here as SDG-RobotSim, is a fully synthetic robotics video corpus for Cosmos training. It is designed to improve physical plausibility, embodiment persistence, contact understanding, long-horizon robot video modeling, and action-conditioned reasoning. The public v1.0 release contains 386{,}270 RGB MP4 clips across collision, manipulation, and humanoid motion. Rather than modeling one platform exhaustively, the release covers mobile robots, quadrupeds, humanoids, fixed-base manipulators, bimanual systems, and dexterous hand-arm embodiments. An overview of the dataset composition is shown in~\cref{fig:sdg_robotsim_dataset_overview}.

\paragraph{Generation pipelines.}
\label{appendix:sdg_robotsim_generation_pipelines}
\label{appendix:sdg_robotsim_simulation_setup}
SDG-RobotSim is generated from USD-based simulation and rendering pipelines built around NVIDIA Isaac Sim, Omniverse, Isaac Lab, and related robot data-generation systems~\citep{nvidia_isaac_sim_2026,nvidia_gr00t_mimic_blueprint_2026}. The release combines collision clips from IsaacLab and MobilityGen~\citep{nvidia_mobilitygen_2026}, manipulation clips from DreamZero~\citep{ye2026dreamzero}, MimicGen/DexMimicGen~\citep{mandlekar2023mimicgen,jiang2025dexmimicgen}, and Simulario/DextrAH~\citep{nvidia_dextrah_2026}, and SOMA humanoid motion clips across SAGE \citep{xia2026sage} and SceneSmith scene sources. The generation loop defines physics-grounded scenarios, randomizes assets and environments, renders synchronized camera views, attaches simulator-derived metadata, and scales through distributed rendering and indexing; curation uses rule-based filters, metadata checks, duplicate motion-sequence removal, and VLM-assisted critique for visible simulation artifacts.

\paragraph{Dataset statistics.}
\label{appendix:sdg_robotsim_dataset_statistics}
The public v1.0 release contains 386{,}270 RGB MP4 clips in 389 WebDataset shards. Motion is the largest family, with 170{,}735 SOMA clips (44.20\%); manipulation contains 110{,}115 clips (28.51\%); and collision contains 105{,}420 clips (27.29\%). By source group, the release contains 170{,}735 SOMA motion clips, 103{,}340 IsaacLab collision clips, 80{,}162 MimicGen manipulation clips, 22{,}926 DreamZero manipulation clips, 16{,}384 Simulario manipulation clips, and 2{,}080 MobilityGen clips. The SOMA subset contains 37{,}709 curated motion groups and is organized under SAGE and SceneSmith branches, each with AgiBot A3, Unitree G1, and Unitree H2 robot families.

\paragraph{Metadata and annotations.}
\label{appendix:sdg_robotsim_metadata_annotations}
Each clip includes RGB video and release metadata covering task family, generator family, embodiment, scene identifier, task text or motion name, camera setup, frame rate, clip length, and available simulator state. Generator-specific metadata may additionally include robot pose, joint state, end-effector state, object pose, contact tags, and task success flags. Captions are generated from RGB clips and simulator metadata where available.

\subsection{SDG-DriveSim}
\label{appendix:sdg_drivesim}

\paragraph{Overview.}
\label{appendix:sdg_drivesim_overview}
Real-world driving video is abundant but structurally biased: it oversamples nominal cruising and undersamples the safety-critical, long-tail interactions that matter most for autonomy and for stress-testing world models. SDG-DriveSim (PhysicsAI-WorldModel-Synthetic-Autonomous-Driving-Scenarios) is a large-scale synthetic video dataset of autonomous-driving scenes generated with NVIDIA Omniverse simulation platform, designed to fill this gap along two axes that real fleet data cannot easily provide. Each clip is a temporally consistent multi-camera surround capture of one ego vehicle and surrounding traffic participants, paired with per-camera VLM captions.

\paragraph{Targeted long-tail coverage.} The dataset is built around scenario families that are explicitly rare or hard to capture in real data---emergency-vehicle interactions, nudging around parked obstacles, cut-ins from adjacent lanes, weather-degraded visibility, and pedestrian crossings with non-standard trajectories. Because scenarios are authored declaratively from natural-language prompts via the Scenario Agent rather than mined post-hoc from driving logs, we can produce many permutations of the same corner case at controllable density.

\paragraph{Environment variation.} Each authored scenario is expanded into deterministic permutations over time of day, cloud coverage, visibility, road material, and vehicle and pedestrian asset choices. The same underlying interaction is thus observed under varied environmental conditions, helping models separate scene content from environment.

\begin{figure}[t]
    \centering
    \captionsetup[subfigure]{justification=centering, singlelinecheck=false}
    \begin{subfigure}[b]{0.24\linewidth}
        \includegraphics[width=\linewidth]{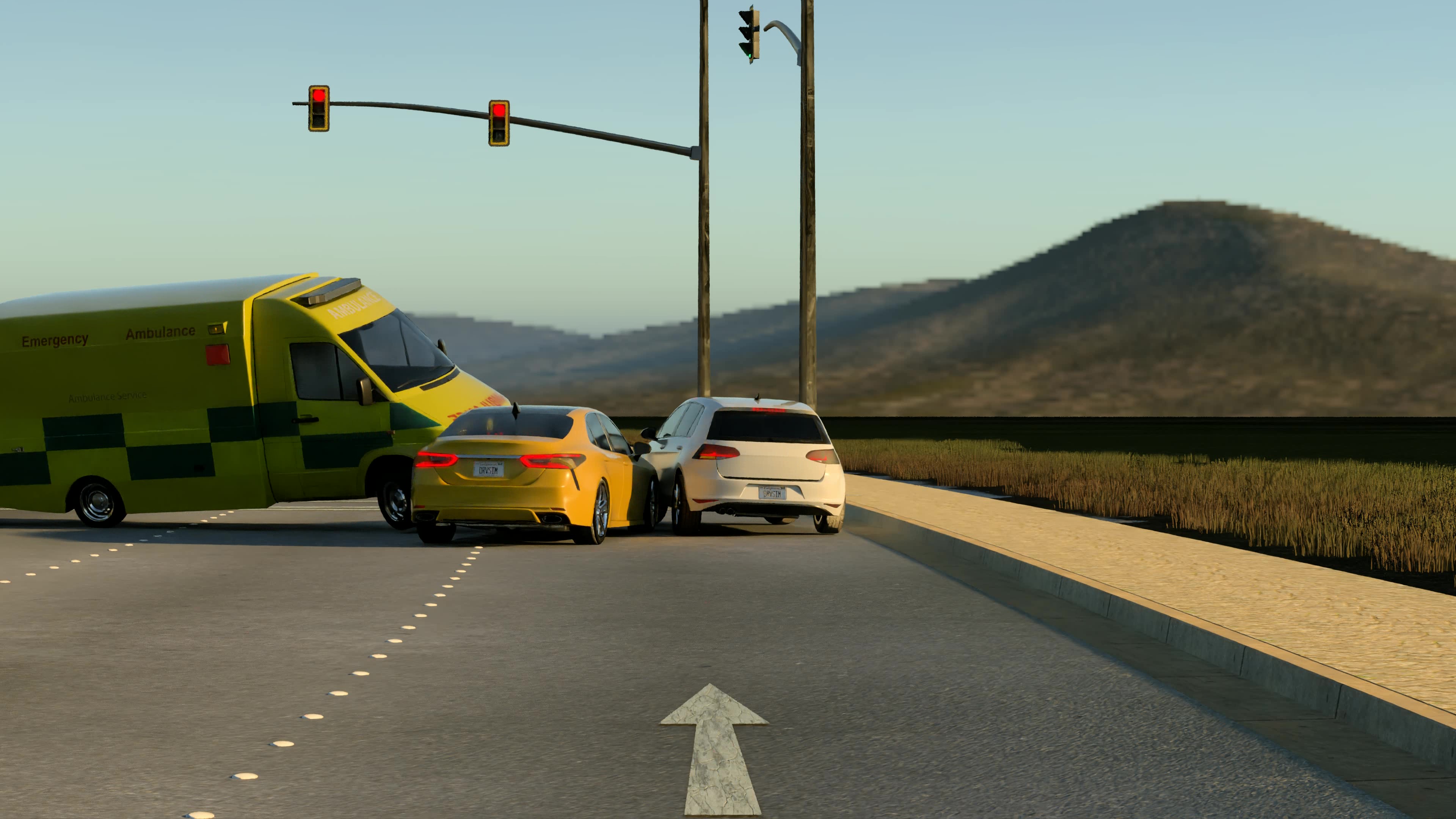}
        \caption*{Cars Collision}
    \end{subfigure}\hfill
    \begin{subfigure}[b]{0.24\linewidth}
        \includegraphics[width=\linewidth]{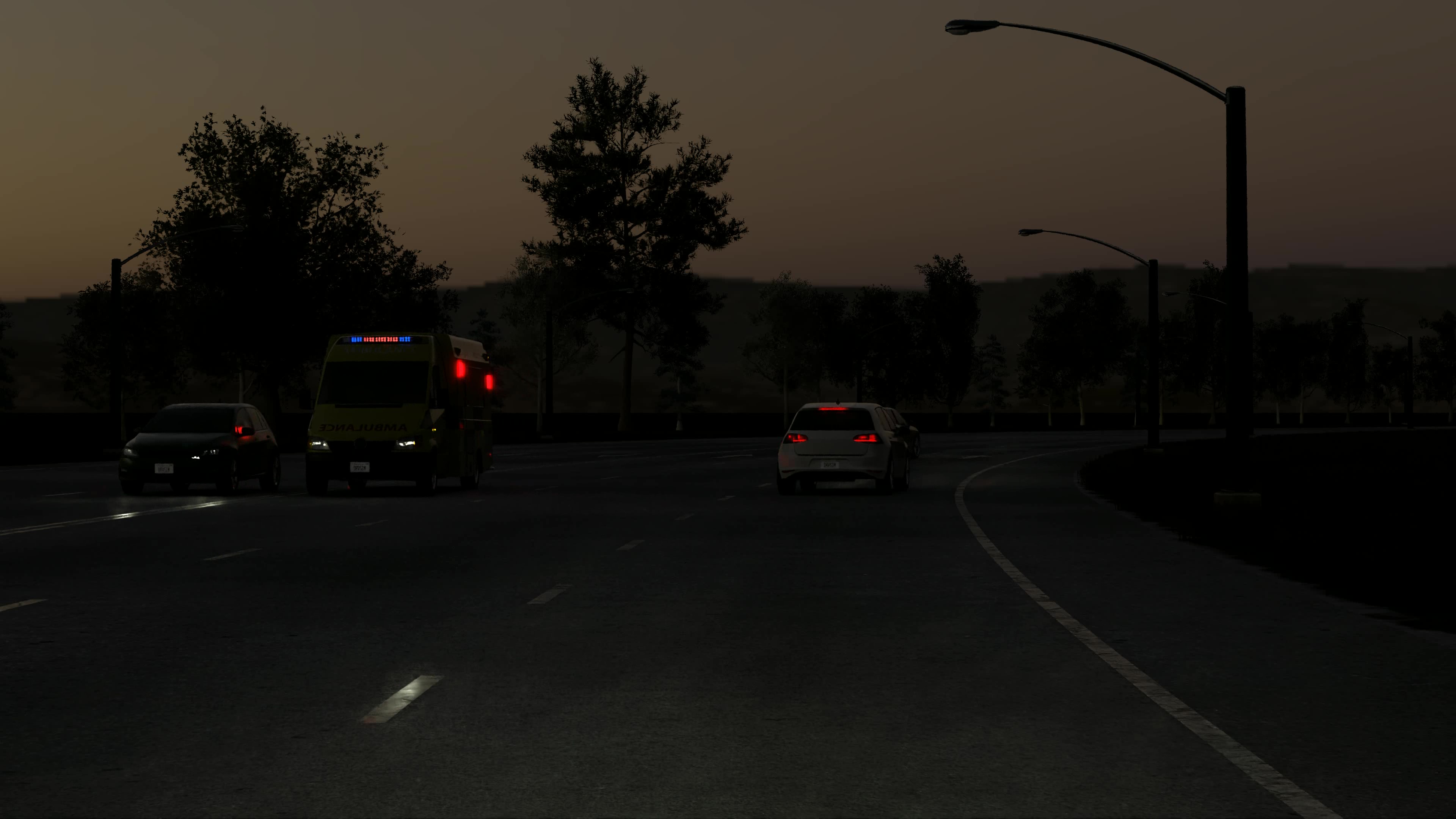}
        \caption*{Emergency Vehicle + Night}
    \end{subfigure}\hfill
    \begin{subfigure}[b]{0.24\linewidth}
        \includegraphics[width=\linewidth]{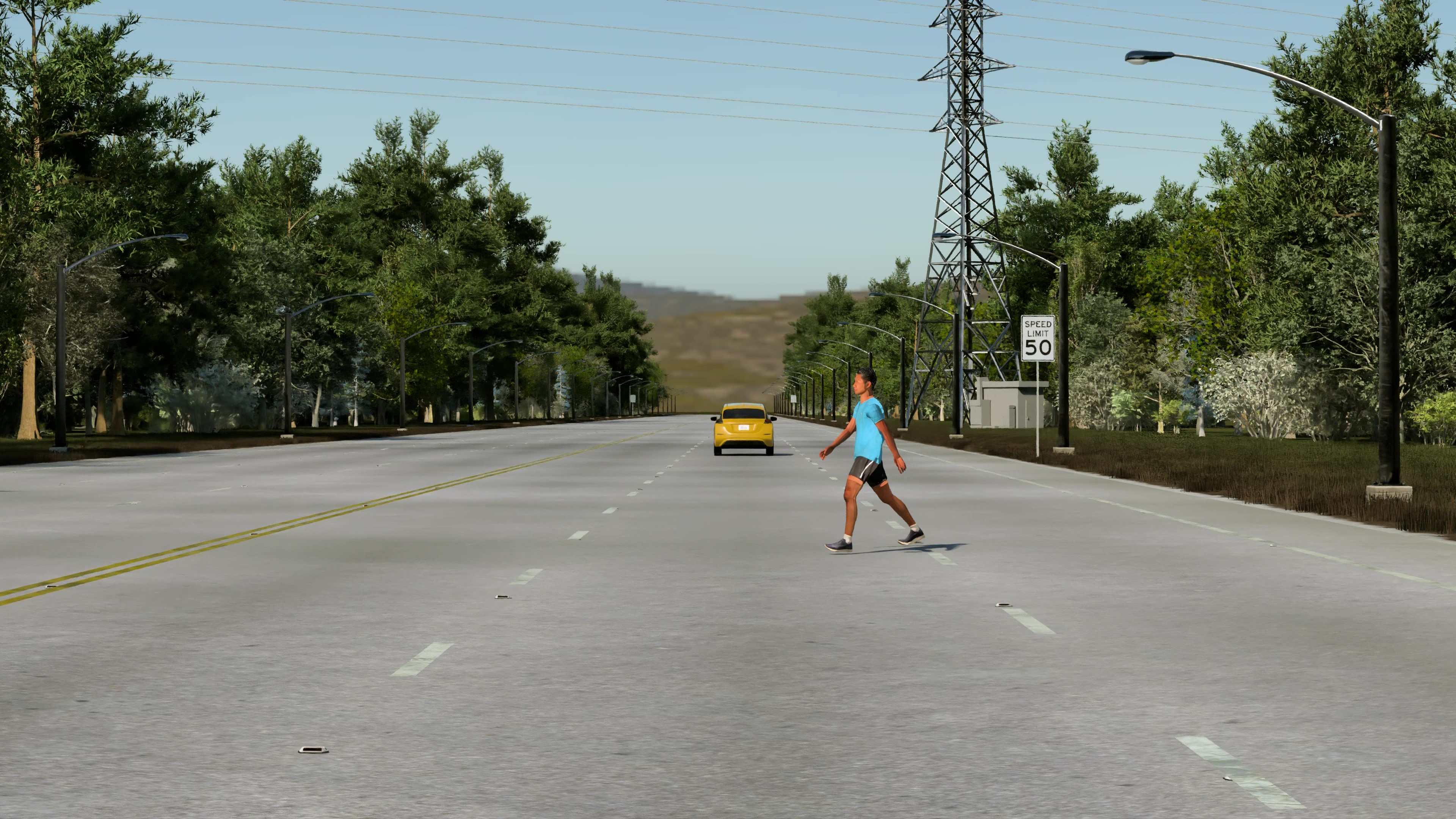}
        \caption*{Jaywalking}
    \end{subfigure}\hfill
    \begin{subfigure}[b]{0.24\linewidth}
        \includegraphics[width=\linewidth]{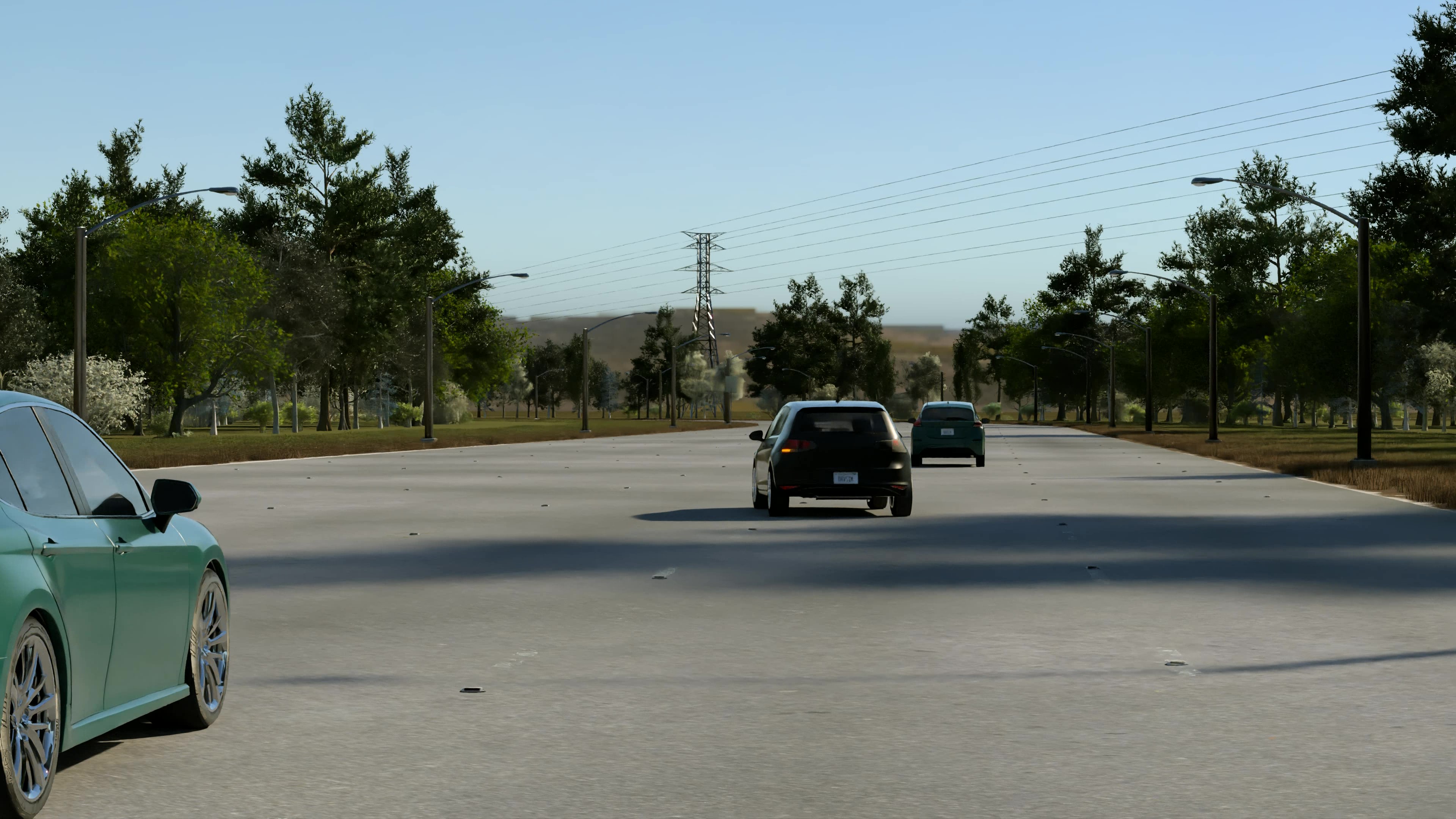}
        \caption*{Lane Change}
    \end{subfigure}

    \vspace{4pt}

    \begin{subfigure}[b]{0.24\linewidth}
        \includegraphics[width=\linewidth]{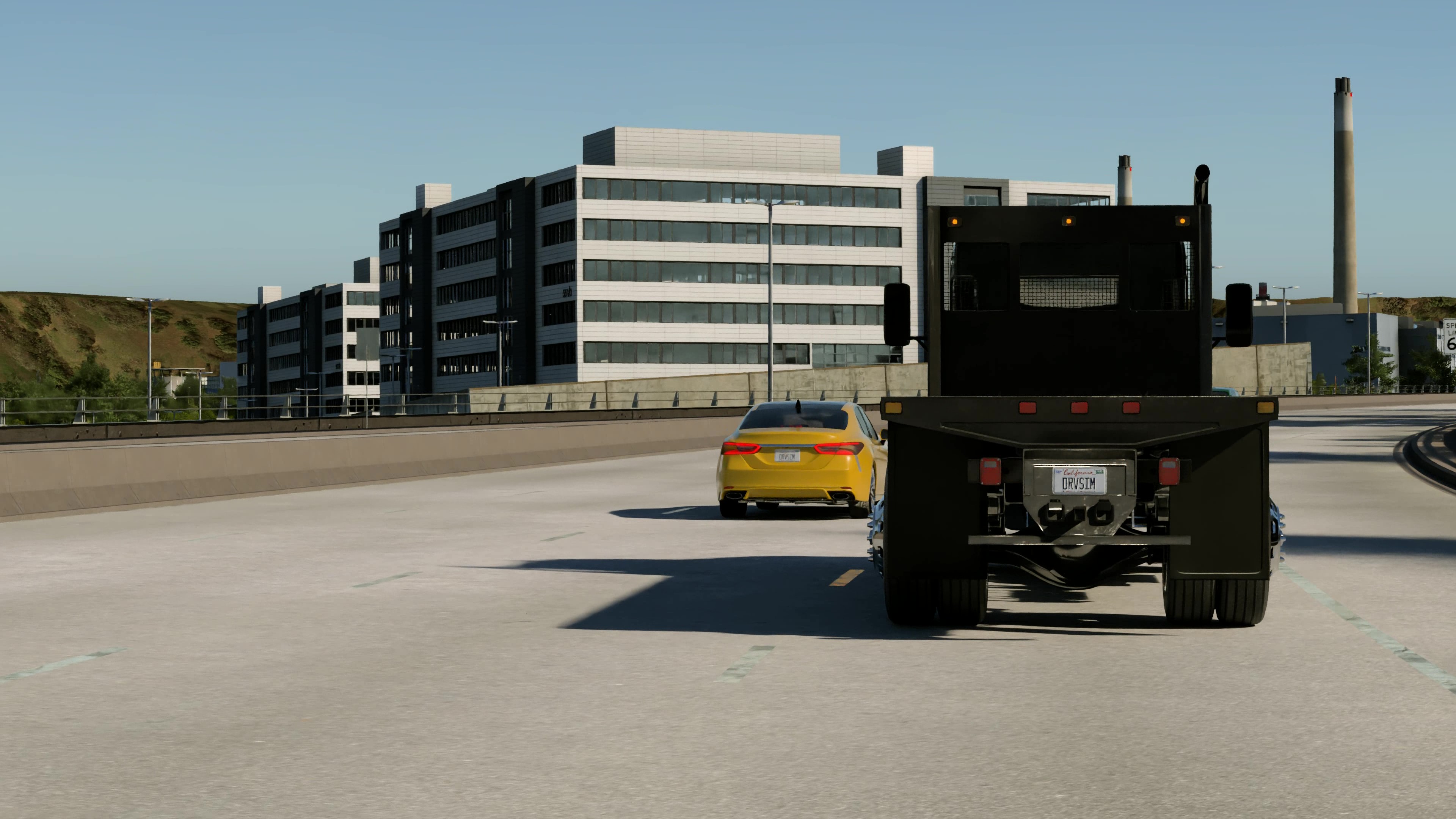}
        \caption*{Cars Nudging}
    \end{subfigure}\hfill
    \begin{subfigure}[b]{0.24\linewidth}
        \includegraphics[width=\linewidth]{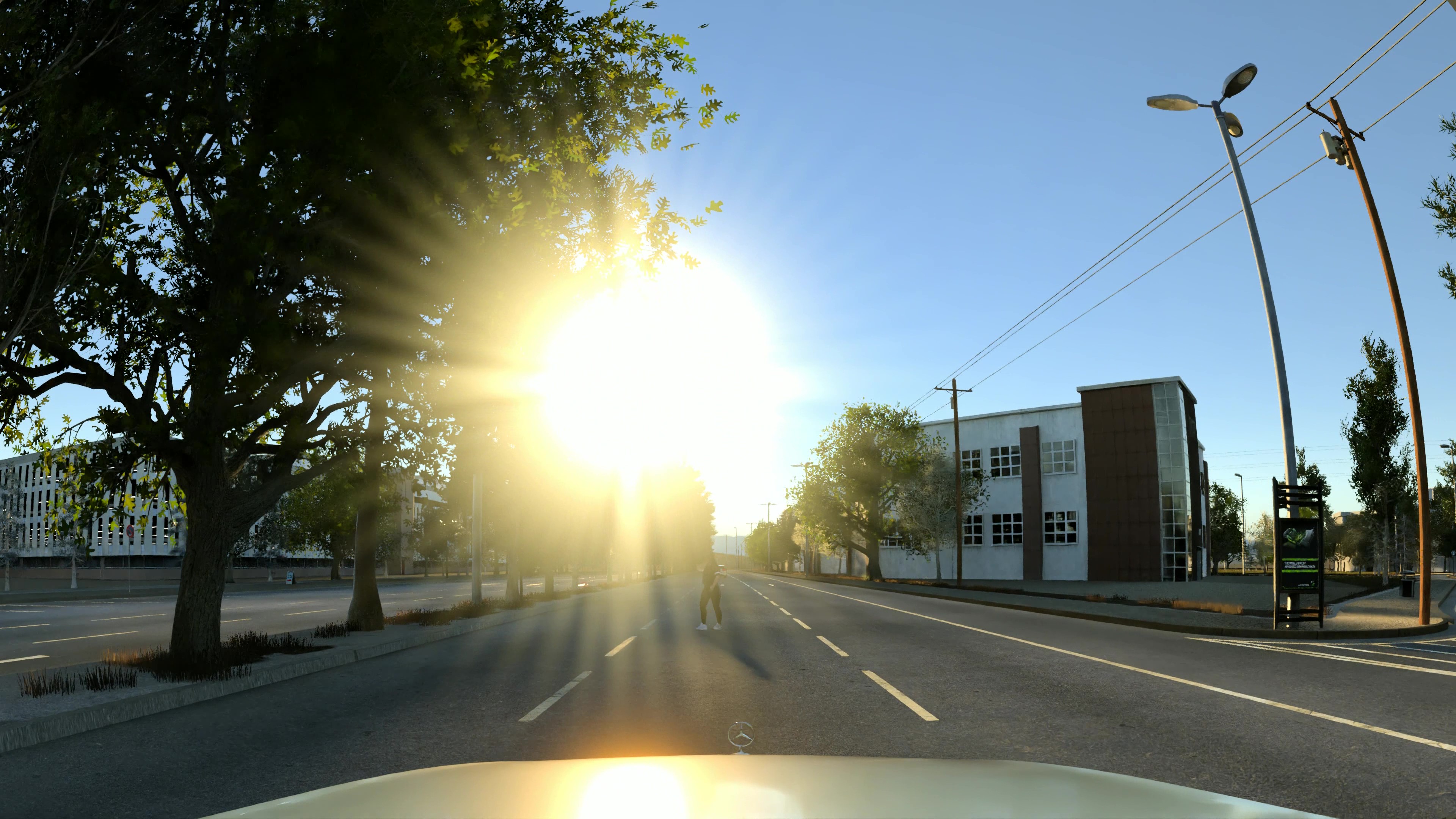}
        \caption*{Pedestrian + Glare}
    \end{subfigure}\hfill
    \begin{subfigure}[b]{0.24\linewidth}
        \includegraphics[width=\linewidth]{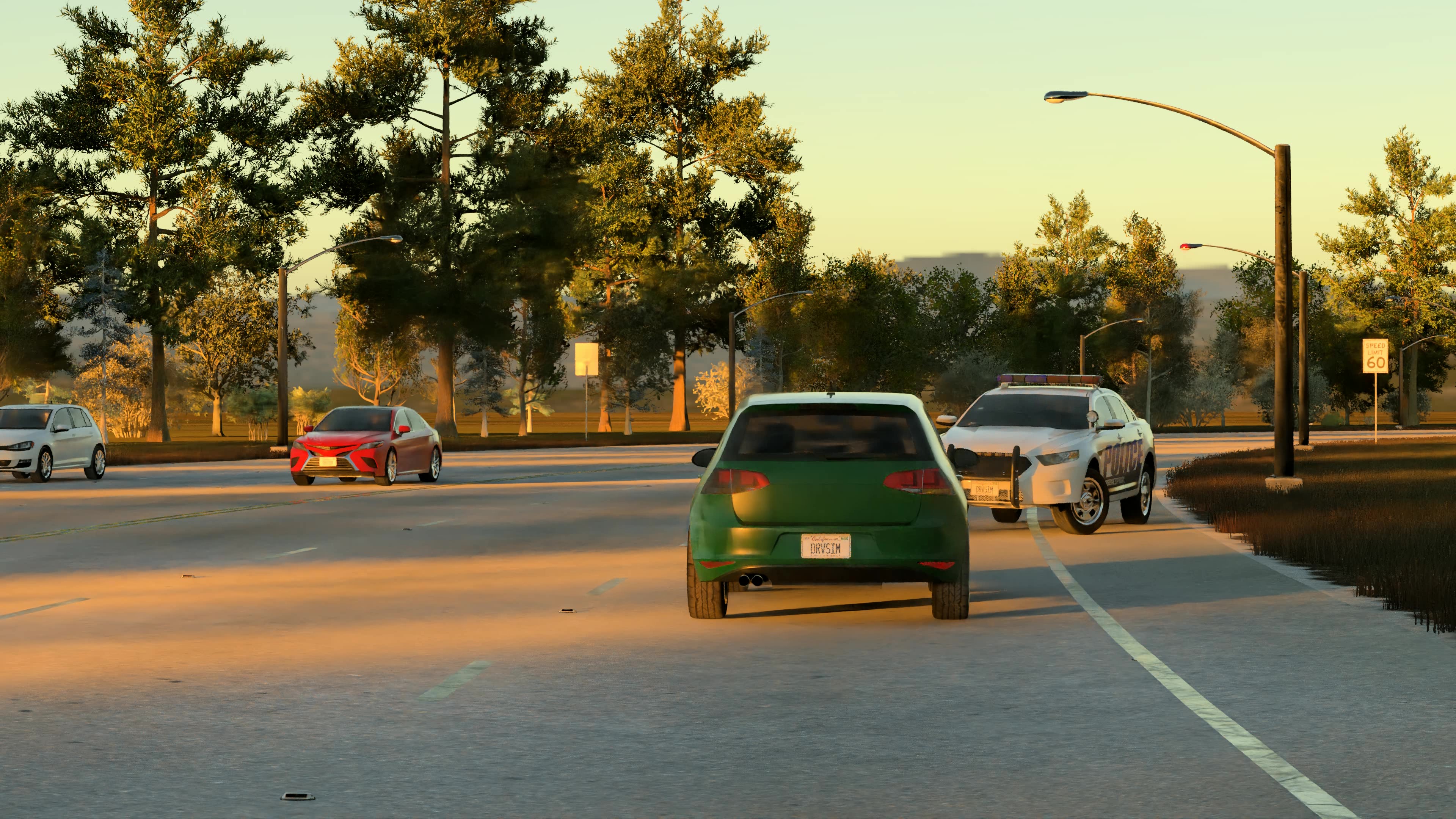}
        \caption*{Police Cars}
    \end{subfigure}\hfill
    \begin{subfigure}[b]{0.24\linewidth}
        \includegraphics[width=\linewidth]{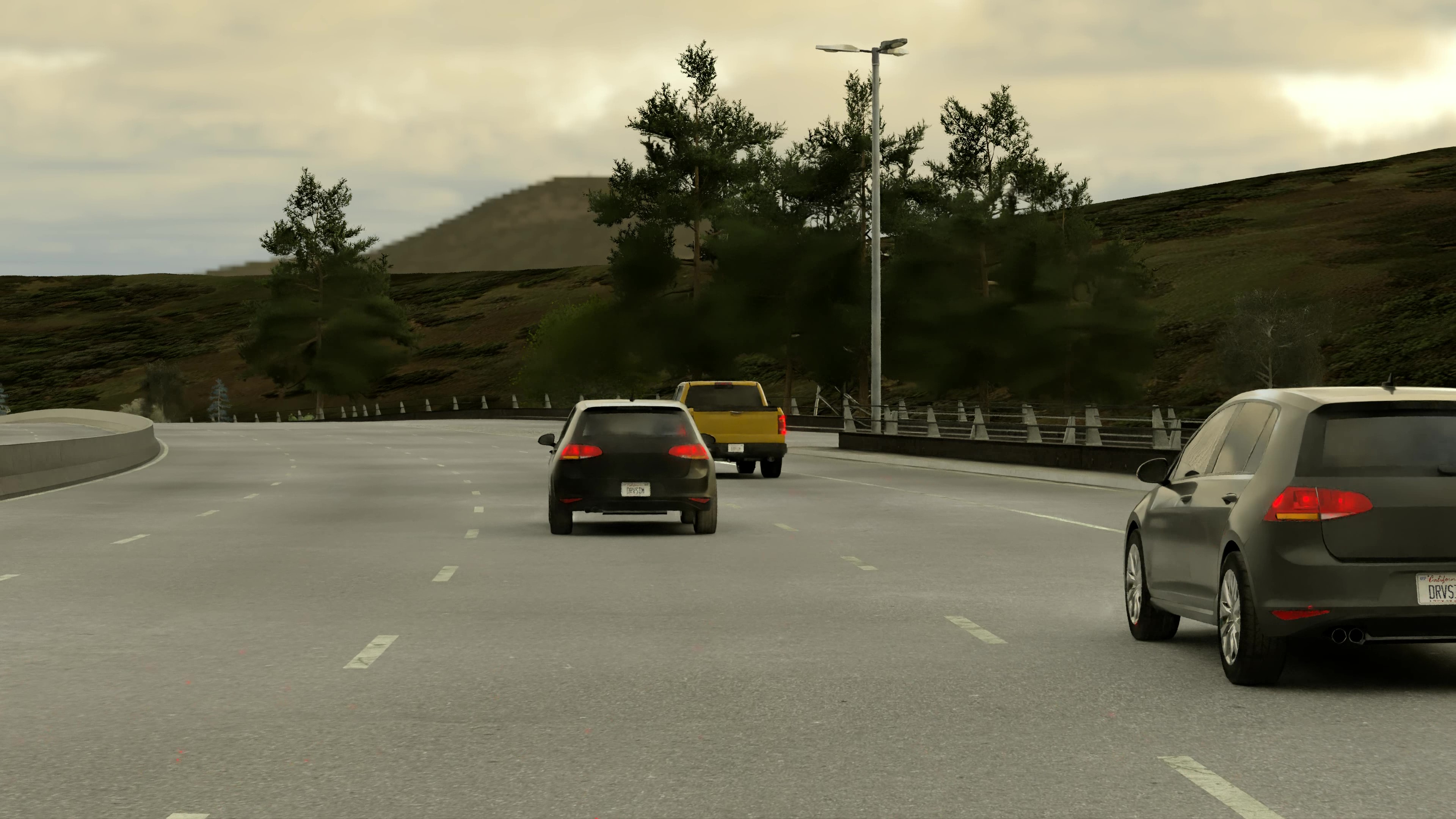}
        \caption*{Weather Change}
    \end{subfigure}

    \caption{\textbf{SDG-DriveSim dataset.} SDG-DriveSim is built to cover long-tail, rare scenarios which are hard to capture in real world. Eight representative driving scenarios are provided from the dataset.}
    \label{fig:drivesim_scenarios}
\end{figure}

\paragraph{Simulation setup.}
\label{appendix:sdg_drive_simulation_setup}
SDG-DriveSim is generated with an agentic scenario-generation pipeline built on OpenUSD: an LLM-backed scenario agent converts a natural-language prompt into a runnable USD world configuration, which the simulator then renders and re-runs across deterministic permutations to produce a batch of clips per prompt. Each scenario instantiates a map (urban, highway, intersection, oval, or test-track), weather and time-of-day condition, camera rig, ego vehicle, and a configurable set of traffic agents and pedestrians with assigned behaviors (drive, follow trajectory, lane change, cut-in, nudge, pull over, pedestrian animation). Two camera rigs are used: a forward-biased 4-camera rig (120\textdegree~front-wide, 30\textdegree~front-tele, plus two 70\textdegree~rear-corner cameras) and a 7-camera rig extending the same set with three 200\textdegree~fisheye cameras (left, right, rear) for 360\textdegree~wraparound coverage. Each authored scenario is expanded into up to ten deterministic permutations over time of day, cloud coverage and visibility, road material, and vehicle and pedestrian asset choices.

\paragraph{Dataset statistics.}
\label{appendix:sdg_drive_dataset_statistics}
The current SDG-DriveSim release contains 264{,}000 clips totaling
approximately 1{,}467 hours of video, rendered at 4K (3840$\times$2160) and
24\,fps with per-clip durations of approximately 20\,s, corresponding to
roughly 127 million RGB frames. Clips are distributed across seven scenario families: vehicle cut-in (32.9\%),
vehicle--pedestrian (21.1\%), vehicle lane change (12.9\%), pedestrian (12.4\%),
vehicle weather degradation (9.2\%), vehicle nudging (8.8\%), and emergency
vehicle (2.7\%). The release draws on
9 unique driving maps, 10 vehicle asset categories,
8 pedestrian assets, and 3 pedestrian animation variations,
with one to nine traffic vehicles and pedestrians per scene.

\paragraph{Metadata and annotations.}
\label{appendix:sdg_drive_metadata_annotations}
The dataset is partitioned by scenario category, with separate \path{video/} and \path{description/} subfolders. Each clip yields one (video, caption) pair per camera of the surround rig: H.264-encoded RGB at 24\,fps, plus a per-camera caption file containing frame rate, frame count, and a list of time-windowed natural-language captions (\texttt{t2w\_windows} entries as \texttt{(start\_frame, end\_frame, caption)}). Clip-level scene metadata records \texttt{weather}, \texttt{time\_of\_day}, \texttt{surface\_type}, and \texttt{region}.

\subsection{SDG-SynHuman}
\label{appendix:sdg_synhuman}

\paragraph{Overview.}
\label{appendix:sdg_synhuman_overview}
SDG-SynHuman (PhysicsAI-WorldModel-Synthetic-Digital-Human-Scenes) is a large-scale synthetic video dataset of digital humans rendered in diverse 3D environments, supplying the dense geometric supervision (per-frame metric depth and camera intrinsics/extrinsics) that real-world human video rarely provides. Clips are temporally consistent at 60--120\,s with 1--9 humans per scene, sampled across diverse human appearances, animations, indoor and outdoor environments, lighting conditions, and camera trajectories. Per-frame camera parameters are produced deterministically from the underlying scene graph, so camera pose can serve as both an input conditioning signal and a prediction target, supporting world model pre-training, camera-motion generalization, depth-aware learning, and human-scene interaction modeling.
\begin{figure}[htbp]
    \centering
    \captionsetup{justification=raggedright, singlelinecheck=false}
    \includegraphics[width=0.24\textwidth]{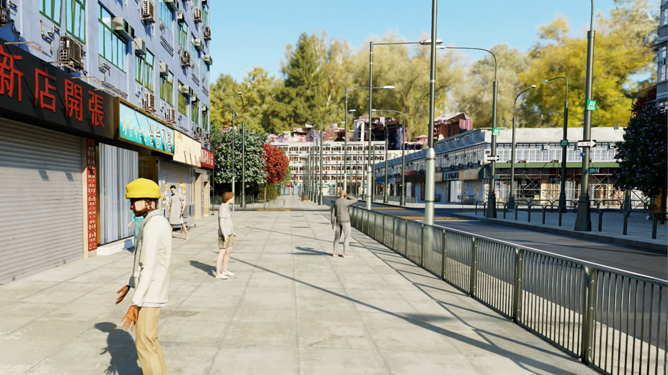}
    \hfill
    \includegraphics[width=0.24\textwidth]{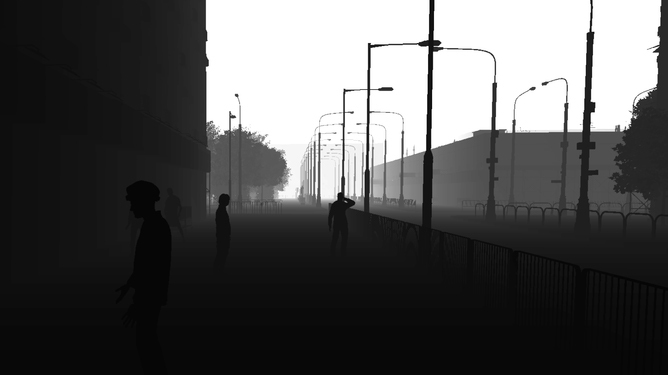}
    \hfill
    \includegraphics[width=0.24\textwidth]{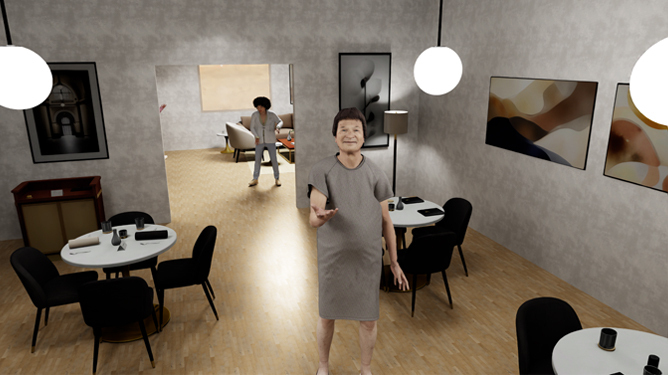}
    \hfill
    \includegraphics[width=0.24\textwidth]{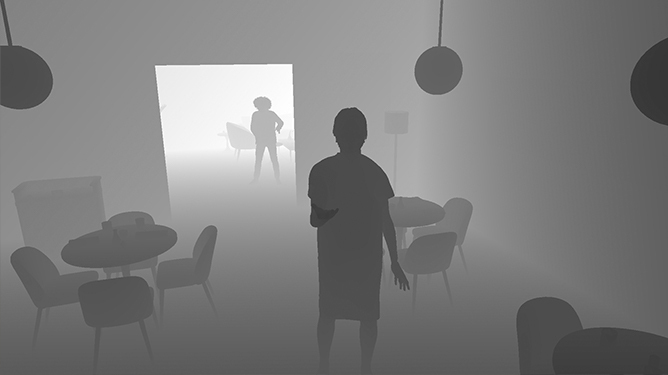}
    \caption{\textbf{SDG-SynHuman samples.} From left to right: RGB, depth; exterior and interior views.}
    \label{fig:synhuman_samples1}
\end{figure}

\paragraph{Simulation setup.}
\label{appendix:sdg_synhuman_simulation_setup}
SDG-SynHuman is generated with NVIDIA's internal SDG pipeline, built on the NeMo Agent Toolkit, Omniverse, OpenUSD, and an internal orchestration backend. The pipeline converts high-level scenario specifications into structured world configurations that define the environment, lighting, digital humans, animations, camera model, camera trajectory, and requested ground-truth outputs. Each scenario instantiates one 3D environment, one lighting condition, one camera, and multiple digital humans, with each character assigned an animation sequence and scene placement.

Camera behavior is controlled through a motion configuration that combines one primary camera motion, such as static, tracking, fly-through, arc/orbit, egocentric, zig-zag, or bird's-eye motion, with optional secondary motion layers such as shake, drift, breathing sway, dutch angle, zoom, crab, or tilt. Temporal remapping and velocity profiles, including ease-in-out and procedurally sampled custom curves, are used to vary motion pacing and acceleration while maintaining temporally coherent clips. Digital human selection, animation, camera behavior, and scene composition are sampled per scenario from the world configuration to produce broad variation across human appearance, motion, scene context, and camera trajectories.

Environment assets include internal NVIDIA scenes, indoor scenes adapted from the SceneSmith example-scenes dataset \citep{pfaff2026scenesmith}, and outdoor city environments generated with The City Generator Blender plugin \citep{durr2026citygenerator}. During rendering, the simulator produces RGB video together with metric depth and per-frame camera calibration directly from the underlying USD scene, enabling deterministic camera and geometry supervision without manual labeling.

\paragraph{Dataset statistics.}
\label{appendix:sdg_synhuman_dataset_statistics}
The final SDG-SynHuman release contains 236{,}937 clips totaling 5{,}841 hours of video. Clips are rendered at 1080p and 30 fps, with durations between 60 and 120 seconds and an average duration of approximately 88.8 seconds. This corresponds to roughly 631 million RGB frames, with paired metric depth frames and camera parameters generated at the same temporal resolution.

The dataset spans 4{,}050 unique digital human assets, 8{,}184 unique animations, 198 indoor environments, 200 outdoor city environments, and 14 camera-motion scenarios including static, flythrough, tracking, arc, egocentric, zig-zag, bird's eye, tilt, shake, drift, breathing sway, dutch angle, zoom, and crab. Each scenario contains one 3D environment, one lighting condition, one camera trajectory, and one to nine digital humans, providing broad variation across human appearance, animation, scene context, and camera motion. The statistics of camera motion in this dataset are summarized in ~\cref{tab:synhuman_primary_motion} and ~\cref{tab:synhuman_secondary_motion}.

\begin{table}[t]
\centering
\captionsetup{justification=raggedright, singlelinecheck=false}
\caption{\textbf{Primary camera-motion distribution in SDG-SynHuman.} Share of 190{,}670 scenes across seven primary types; each scene has exactly one primary camera motion type. See \cref{tab:synhuman_secondary_motion} for overlapping secondary camera-motion types.}
\label{tab:synhuman_primary_motion}
\small
\begin{tabular}{p{0.16\linewidth}ccp{0.48\linewidth}}
\toprule
Motion Type & \# scenes & Percentage & Description \\
\midrule
static     & 71{,}006  & 29.97\%  & Camera remains fixed in position and orientation throughout the sequence. \\
egocentric & 33{,}070  & 13.96\%  & Camera behaves as the viewpoint of a character or agent. \\
tracking   & 36{,}978  & 15.61\%  & Camera follows a subject while maintaining framing. \\
flythrough & 36{,}852  & 15.55\%  & Camera travels forward through the scene along a path. \\
arc        & 37{,}166  & 15.69\%  & Camera moves in a curved arc around a target. \\
zigzag     & 15{,}294  &  6.45\%  & Camera advances with alternating lateral motion. \\
birdseye   &  6{,}571  &  2.77\%  & Top-down overhead camera perspective. \\
\midrule
Total      & 190{,}670 & 100.00\% & --- \\
\bottomrule
\end{tabular}
\end{table}
 
\begin{table}[t]
\centering
\captionsetup{justification=raggedright, singlelinecheck=false}
\caption{\textbf{Secondary camera-motion activity durations in SDG-SynHuman.} Secondary motions are compositional layers that may overlap temporally, therefore reported durations are independent per-motion totals rather than shares of a fixed duration budget.}
\label{tab:synhuman_secondary_motion}
\small
\begin{tabular}{p{0.18\linewidth}cp{0.55\linewidth}}
\toprule
Motion Type & Active Time (hours) & Description \\
\midrule
breathing   & 1{,}221.77 & Gentle breathing-like motion to mimic human respiration. \\
drift       & 1{,}211.96 & Slow subtle positional motion over time. \\
dutch\_angle & 1{,}205.72 & Rotation around the forward axis producing a tilted horizon. \\
shake       & 1{,}197.87 & Rapid positional and rotational handheld jitter. \\
sway        & 1{,}191.28 & Pendulum-like oscillatory motion with horizon rocking. \\
zoom        & 1{,}184.09 & Forward/backward motion without lens-property changes. \\
crab        &   373.72   & Lateral translation while maintaining viewing direction. \\
tilt        &   195.33   & Up/down rotational motion around the horizontal axis. \\
\bottomrule
\end{tabular}
\end{table}

\paragraph{Metadata and annotations.}
\label{appendix:sdg_synhuman_metadata_annotations}
The dataset is delivered in 1{,}215 tar shards under \path{shards/}, each bundling 200 samples. Per sample (keyed by UUID): RGB as H.264 in \path{video/uuid.mp4}; FFV1 lossless 1080p depth in \path{depth/uuid.mkv} with a companion JSON recording depth range, resolution, frame rate, and dtype for metric reconstruction; per-frame camera data in \path{meta/uuid_camera.json}; scene-level metadata in \path{metas/uuid.json} (environment, lighting, agents, camera configuration, animation tasks); and asset-level metadata in \path{description/uuid.json} (spawned inventory, motion assets, placements, provenance). All annotations are generated deterministically from the USD scene graph. ~\cref{fig:synhuman_samples1} shows sample RGB and depth outputs, for both the exterior and interior views.

\subsection{SDG-Warehouse}
\label{appendix:sdg_warehouse}

\paragraph{Overview.}
SDG-Warehouse (PhysicsAI-WorldModel-Synthetic-Warehouse-Operation-Scenes) is a synthetic video dataset of indoor industrial-safety events rendered in fully simulated warehouse environments. Real surveillance footage of these events is rare, hard to release at scale, and seldom carries the kind of dense ground truth that physical-AI training benefits from, so we generate it in simulation, where the event is guaranteed to happen, every parameter is controllable, and every frame is paired with deterministic per-pixel and per-object annotations. The release covers four representative scenarios---a forklift--human near-miss (23\%), a warehouse fire with worker evacuation (36\%), a forklift--shelf collision (20\%), and a warehouse box-pickup action (21\%)---totaling ${\sim}123$K clips ($\sim$412 hours of video) at $1920{\times}1080$ and $30$\,fps, with multiple synchronized camera viewpoints per simulation run.

\paragraph{Common simulation infrastructure.}
All four scenarios are built on NVIDIA Isaac Sim. Procedural scene composition---warehouse layout, shelf placement, prop variation, and per-light randomization of color temperature, intensity, exposure, and color---is handled by Isaac Sim Replicator Object (IRO). Agent and sensor population---worker spawning and behavior, forklift placement and navigation, and the camera rigs that define the dataset's multi-view viewpoints---is handled by Isaac Sim Replicator Agent (IRA). Camera placement is parametric, with height, distance, and look-down angle sampled per run, and worker assets and motions are sampled from Isaac Sim's character library to diversify human appearance and gait. Each simulation run is seeded with a unique random seed that controls all randomized variables (scene composition, lighting, agent identity and motion, camera pose, and event timing), so runs are independent and reproducible.

\begin{figure}[htbp]
    \centering
    \footnotesize
    \setlength{\tabcolsep}{2pt}
    \renewcommand{\arraystretch}{0}
    \begin{tabular}{@{}c@{\hspace{2pt}}*{5}{c}@{}}
        & RGB & Depth & Segmentation & Shaded Segmentation & Edges \\[2pt]
        \rotatebox{90}{Near-miss}
            & \includegraphics[width=0.17\textwidth]{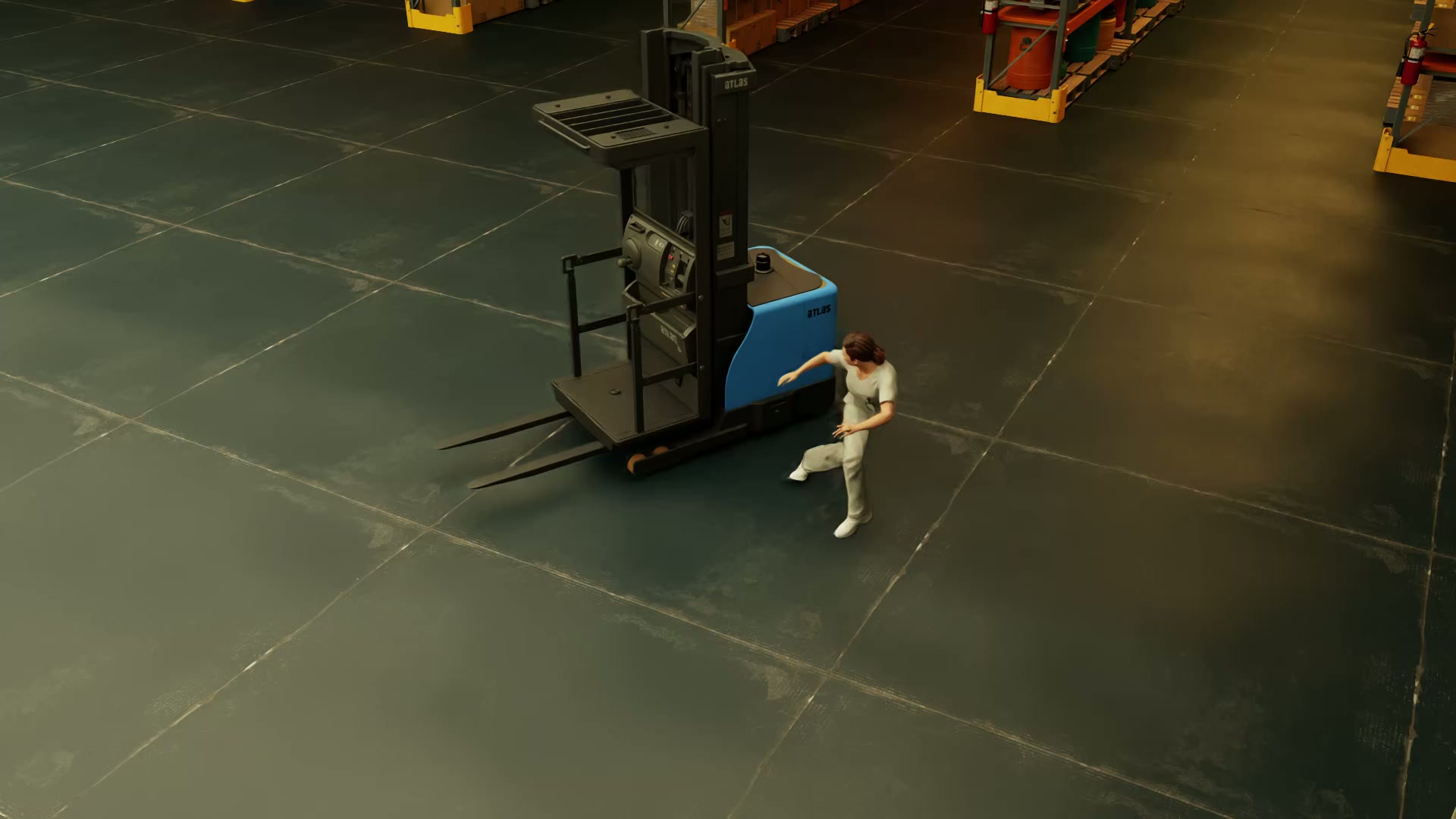}
            & \includegraphics[width=0.17\textwidth]{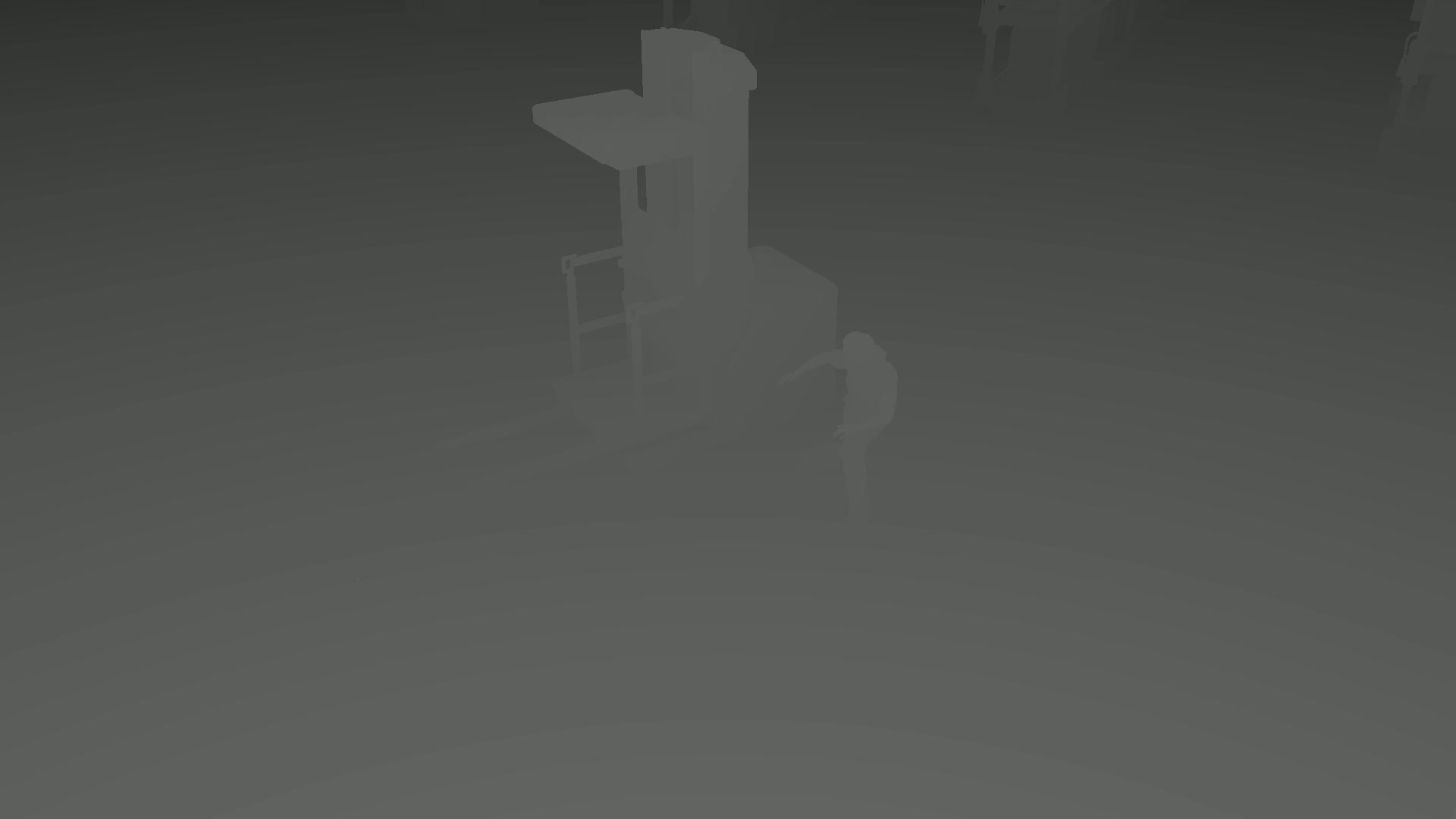}
            & \includegraphics[width=0.17\textwidth]{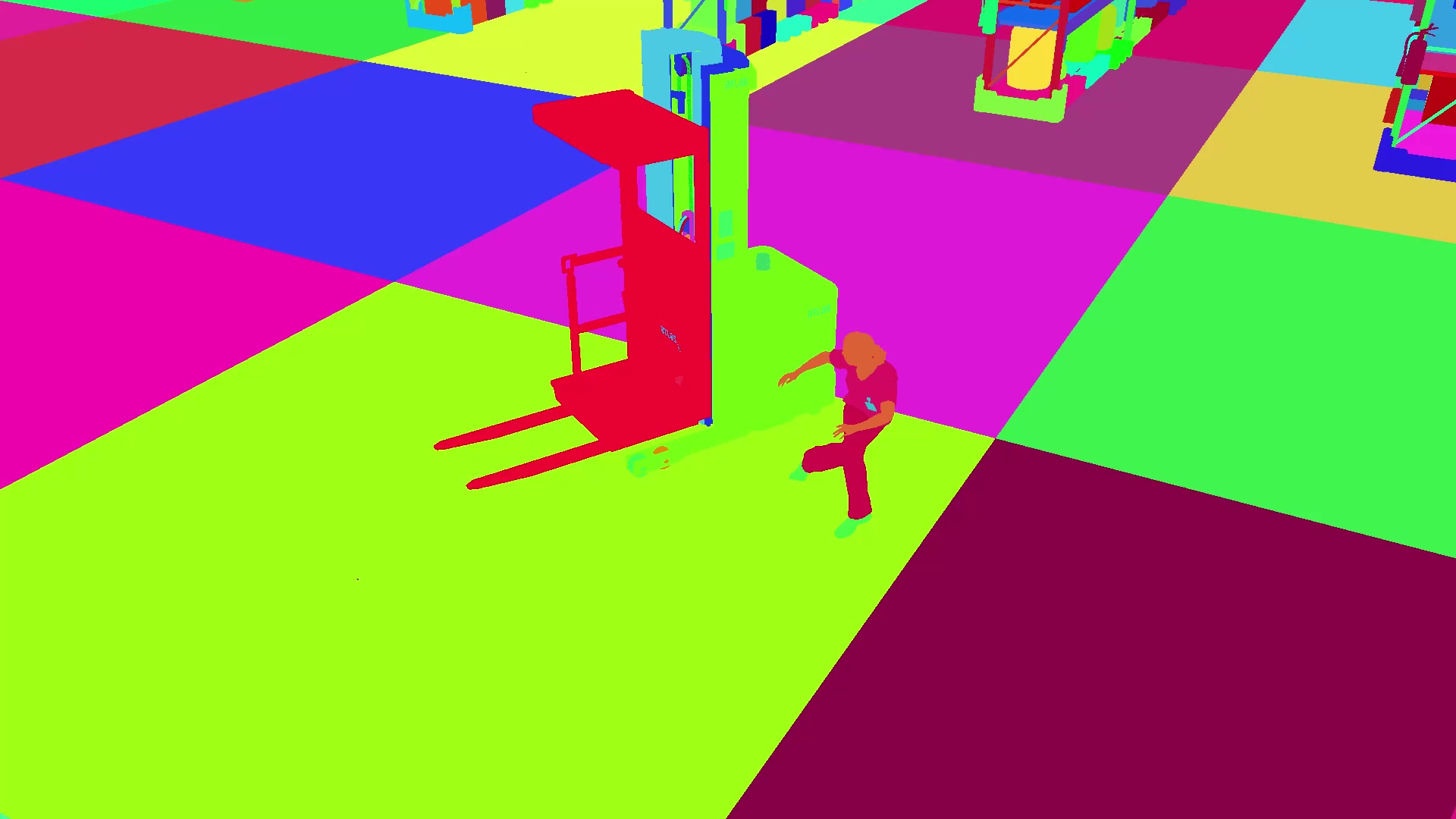}
            & \includegraphics[width=0.17\textwidth]{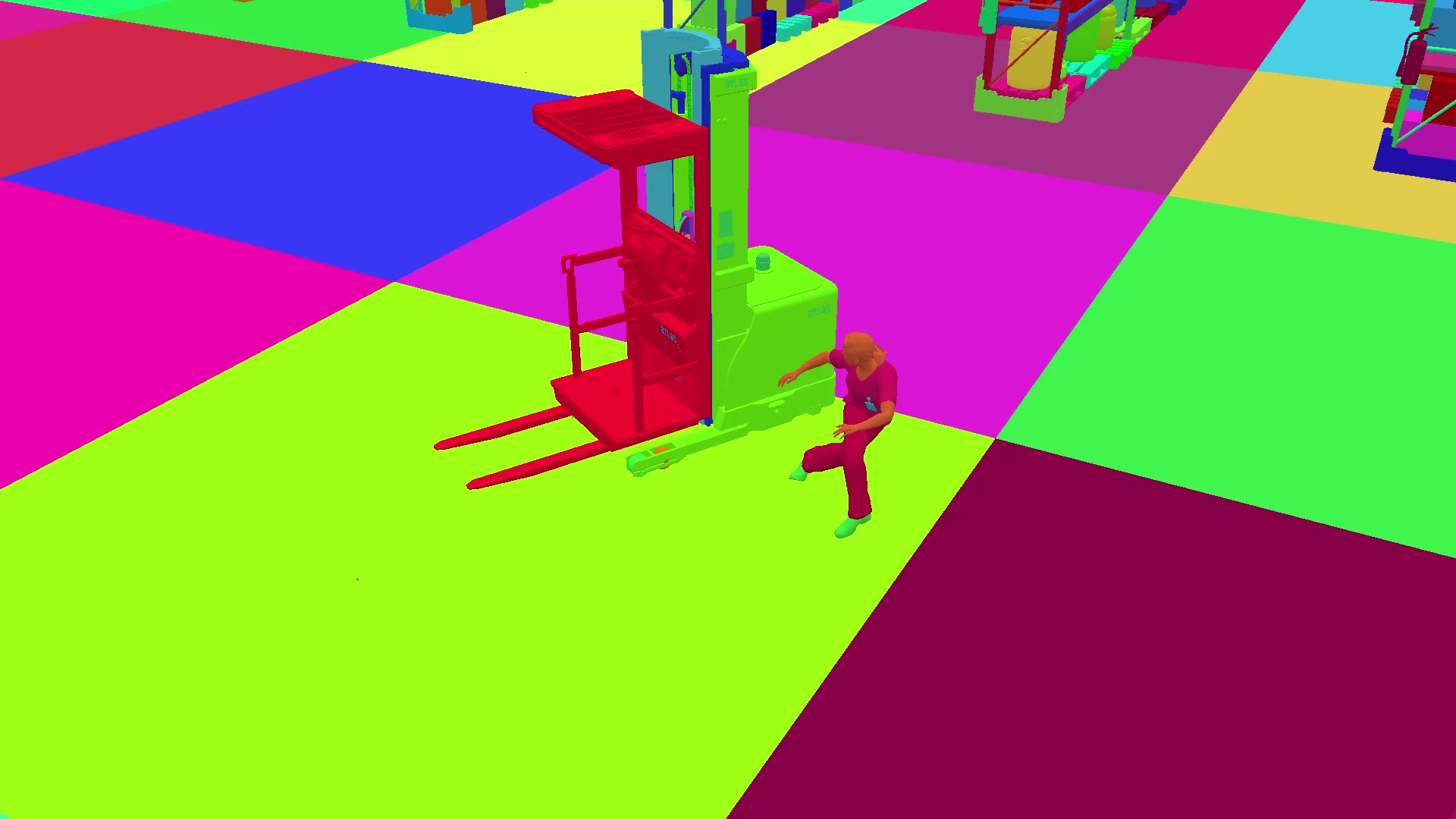}
            & \includegraphics[width=0.17\textwidth]{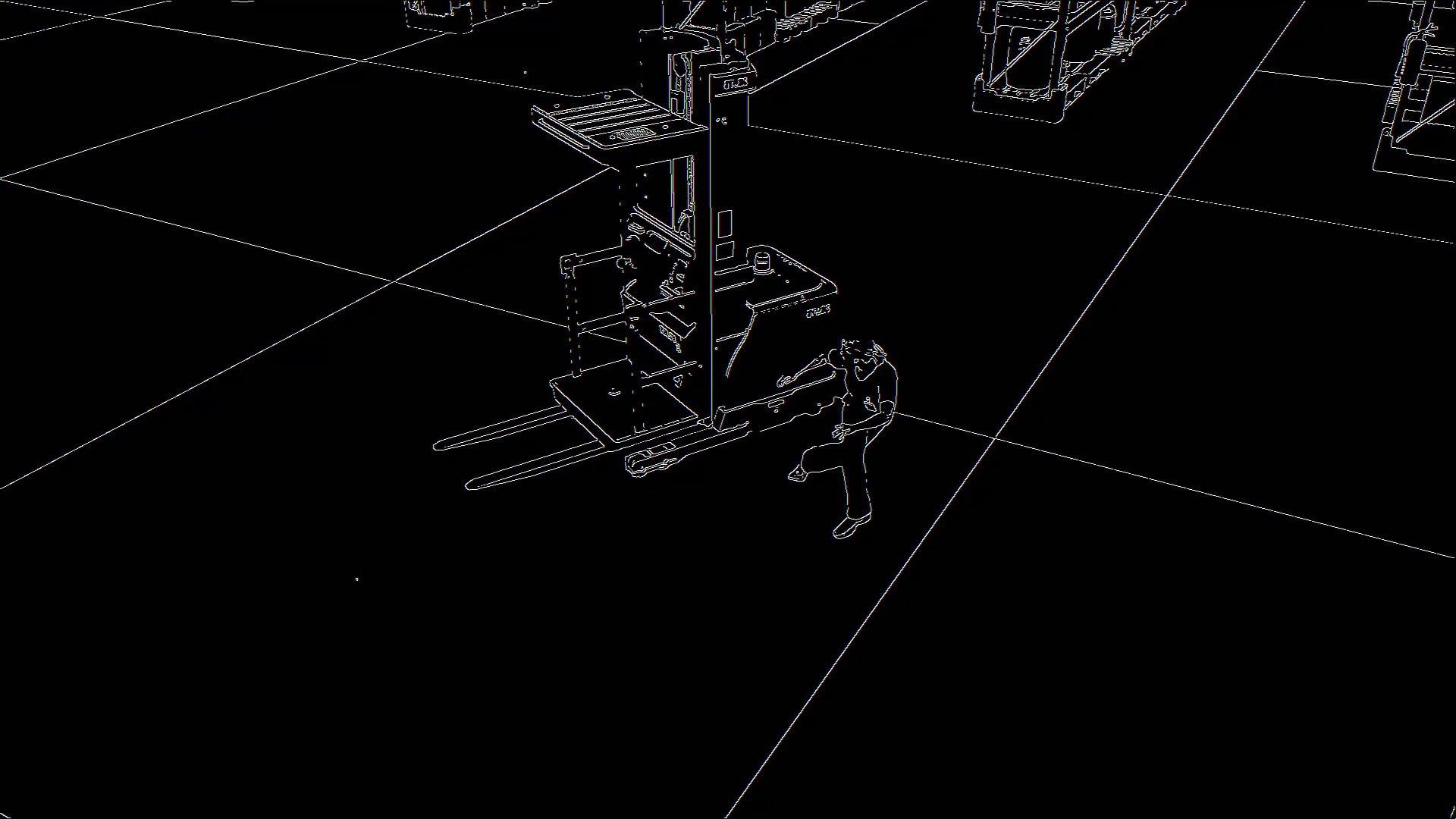} \\[3pt]
        \rotatebox{90}{Fire}
            & \includegraphics[width=0.17\textwidth]{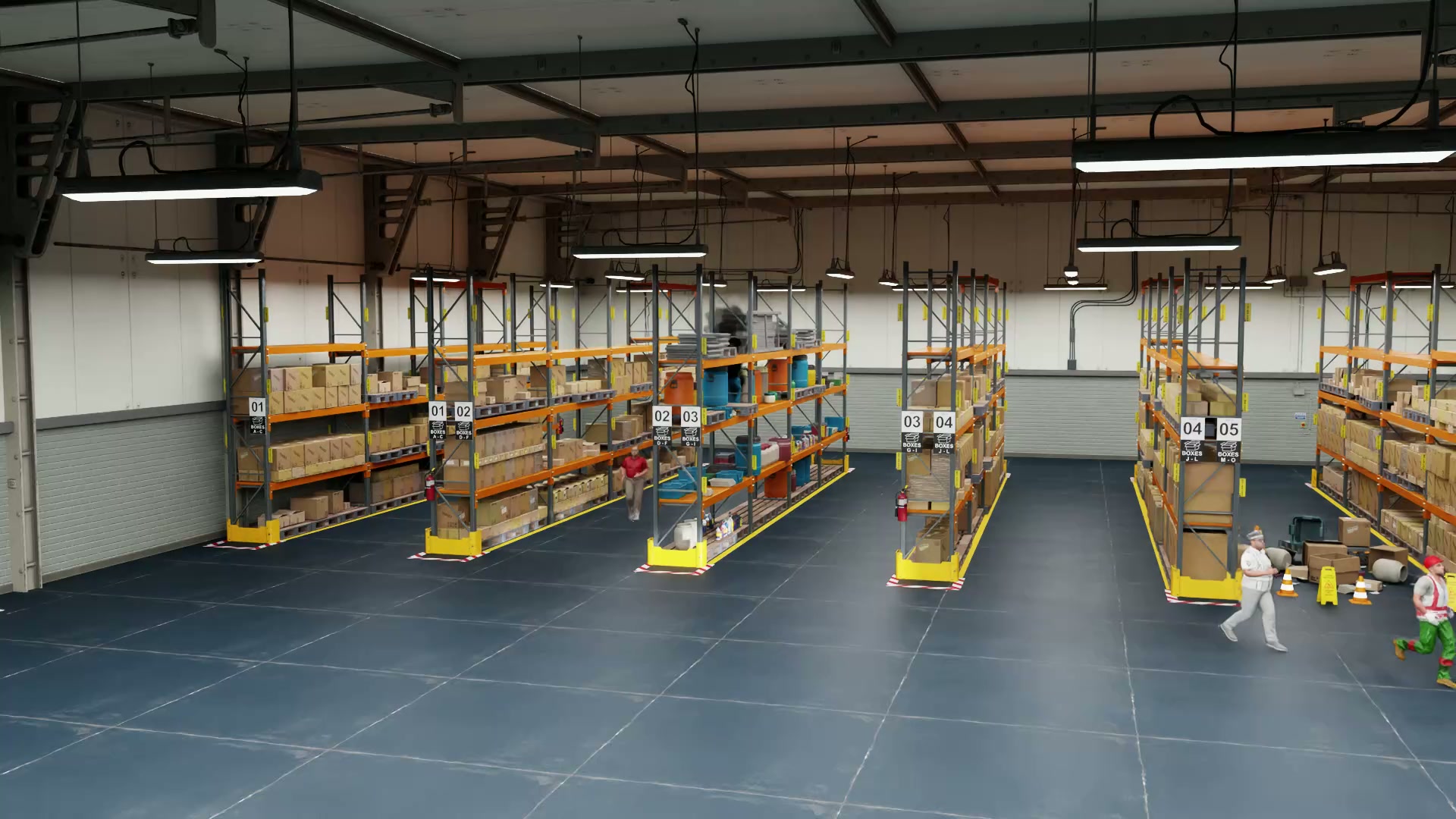}
            & \includegraphics[width=0.17\textwidth]{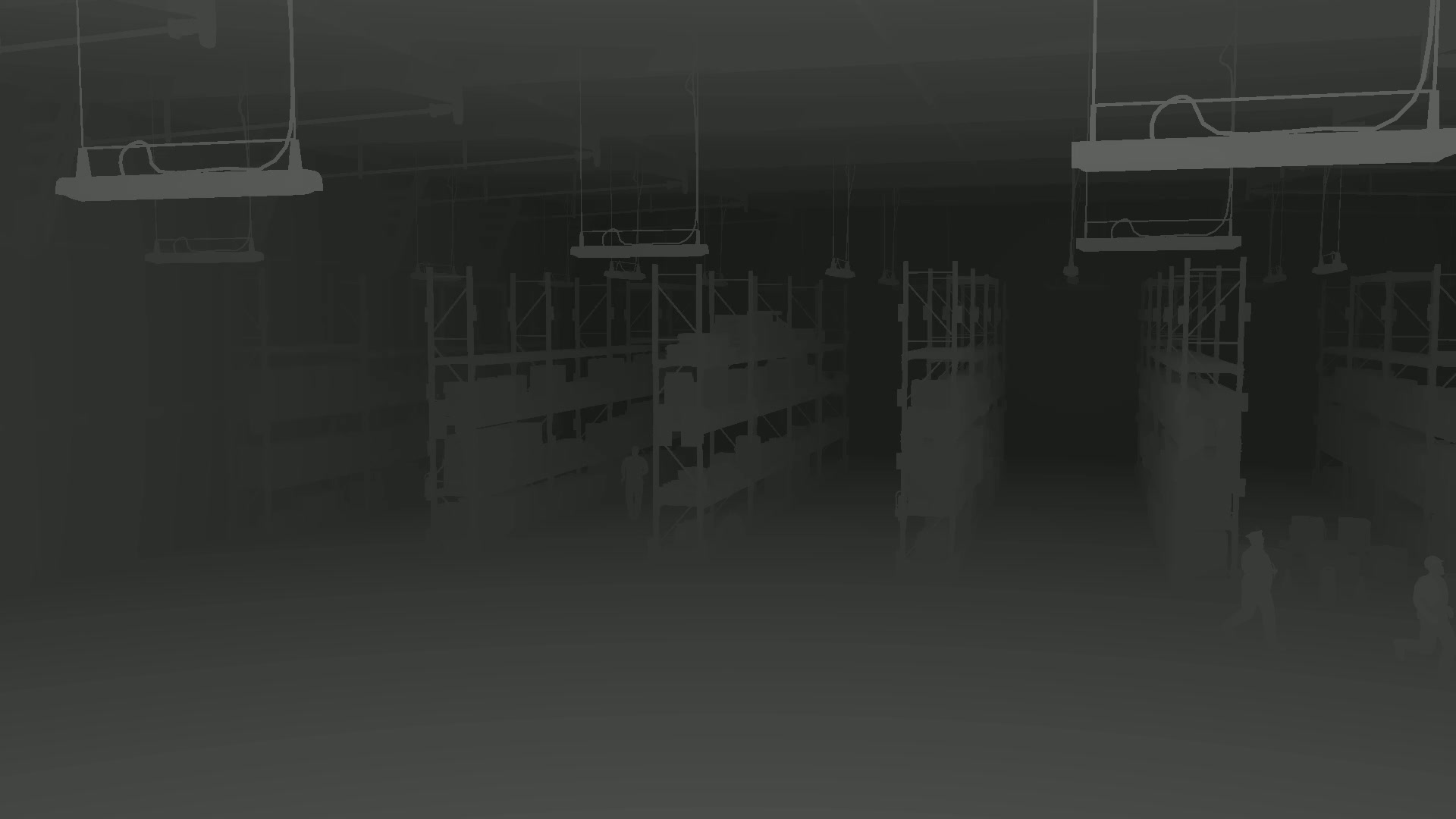}
            & \includegraphics[width=0.17\textwidth]{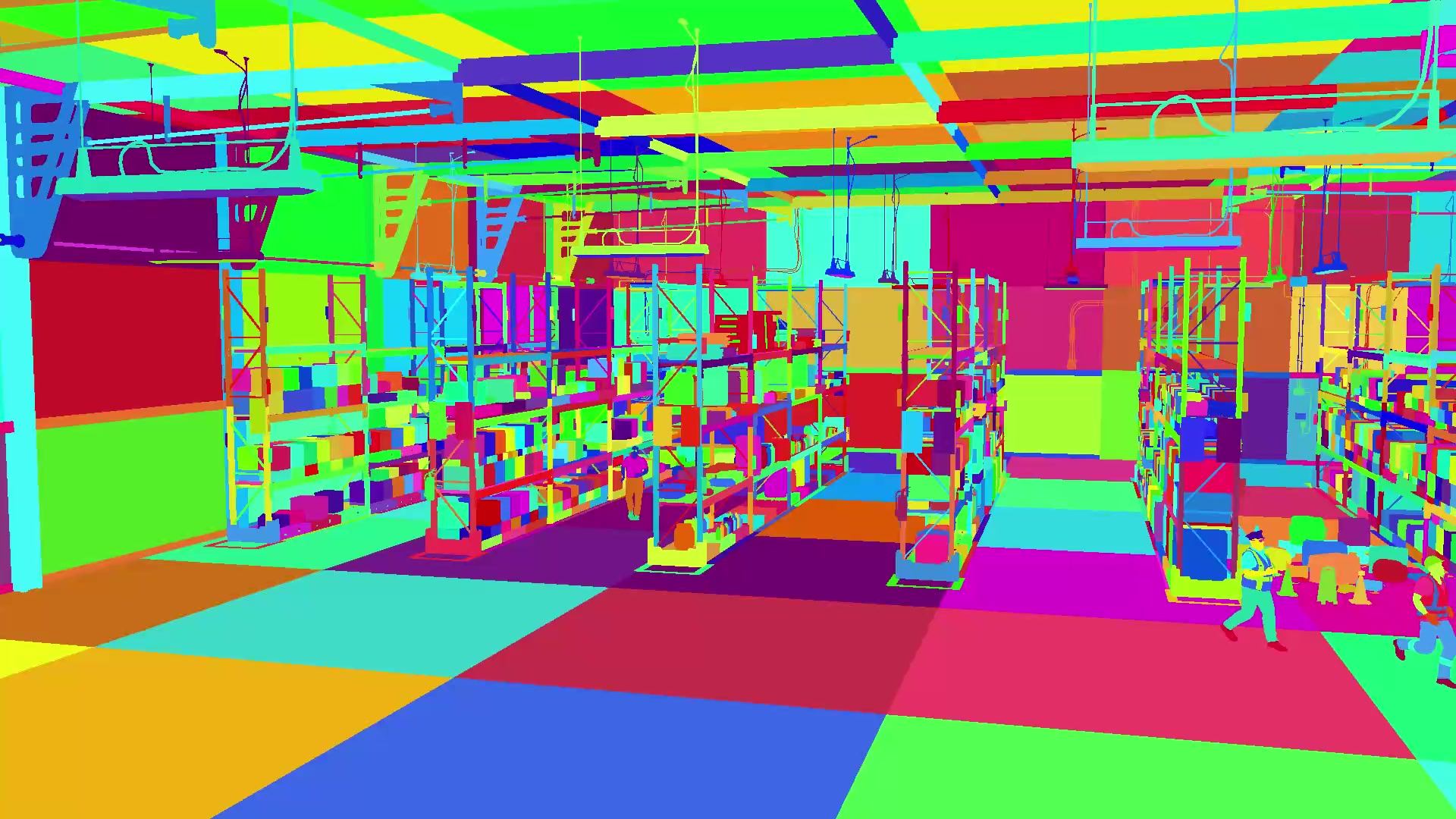}
            & \includegraphics[width=0.17\textwidth]{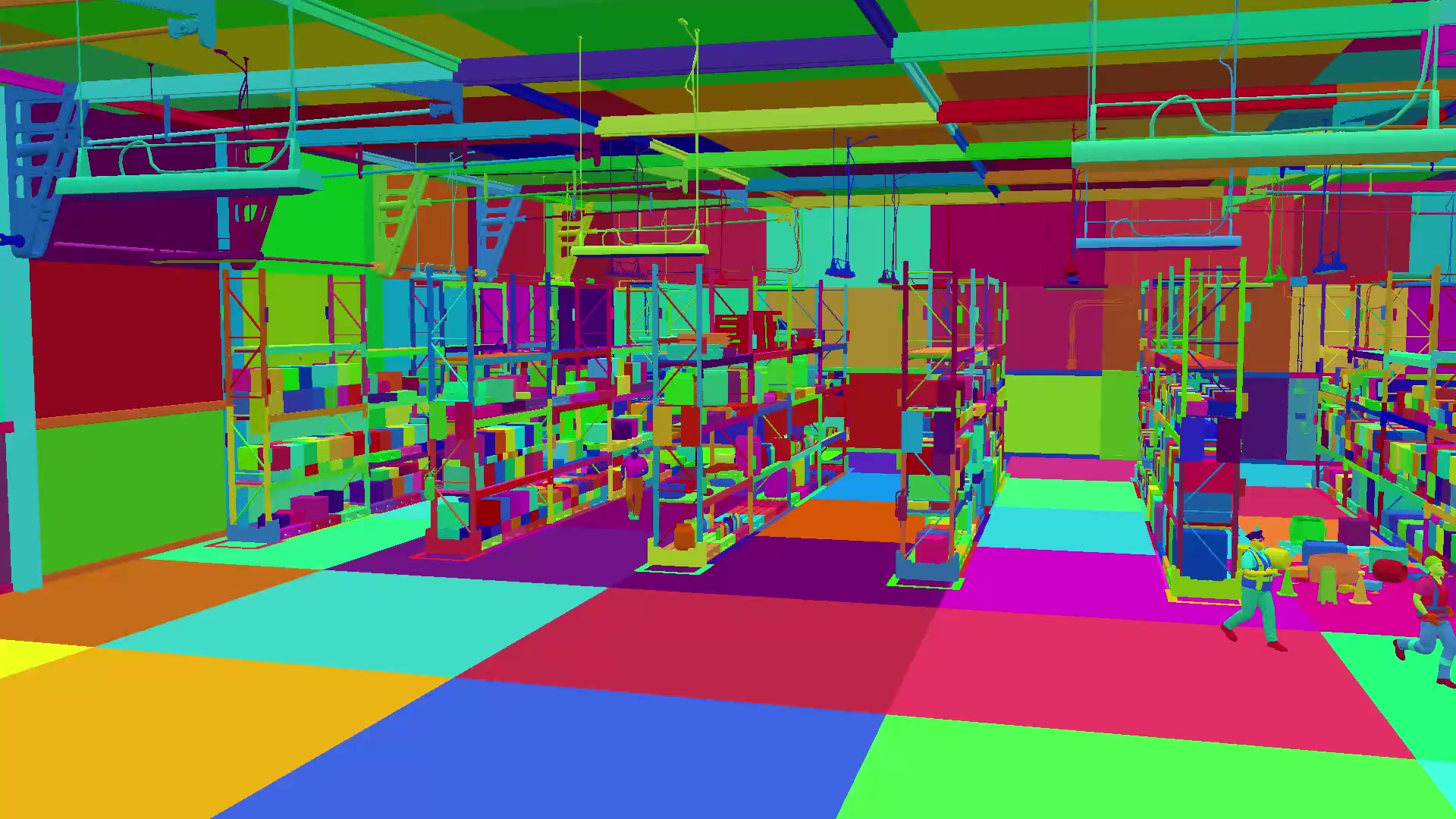}
            & \includegraphics[width=0.17\textwidth]{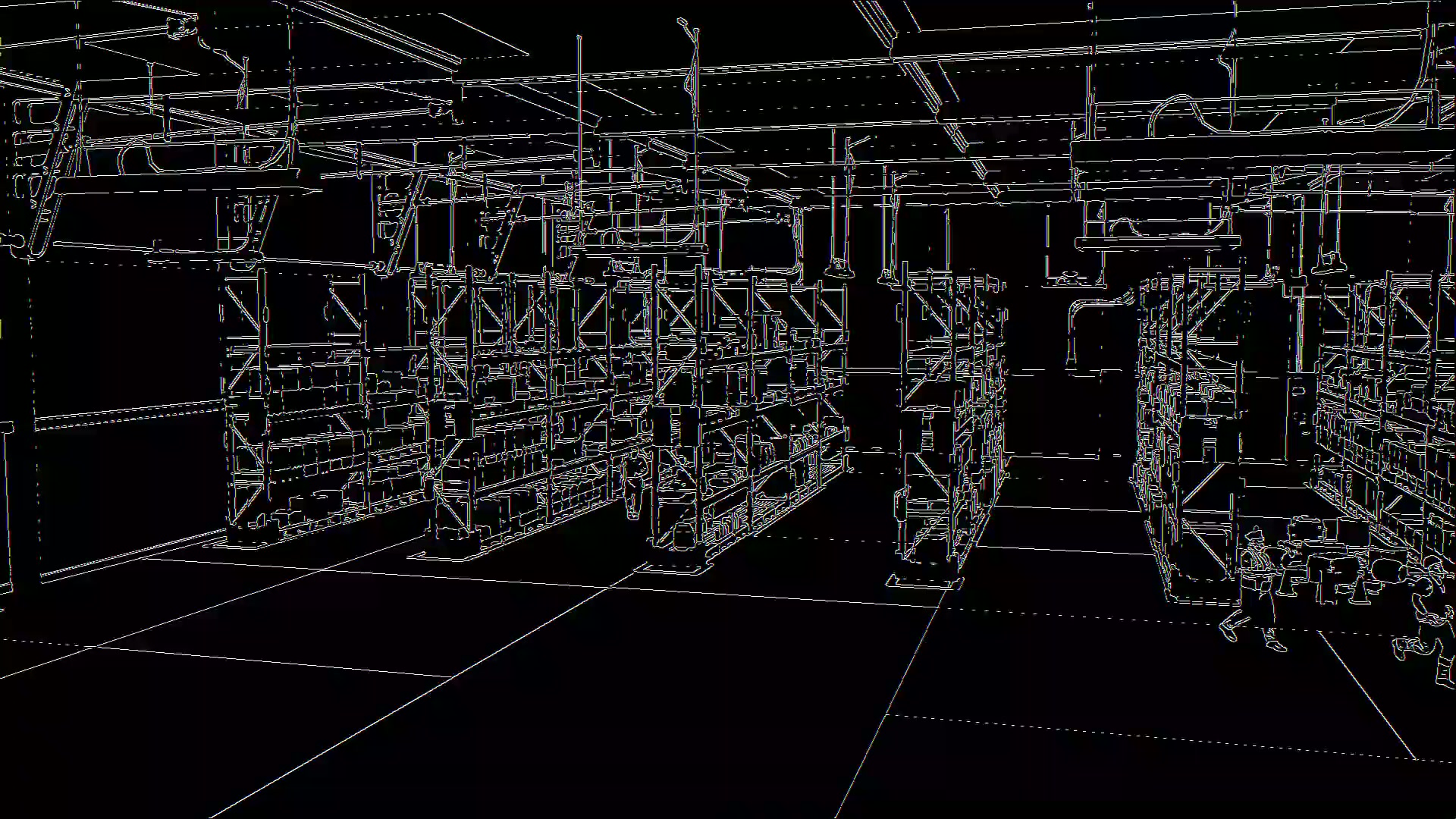} \\[3pt]
        \rotatebox{90}{Collision}
            & \includegraphics[width=0.17\textwidth]{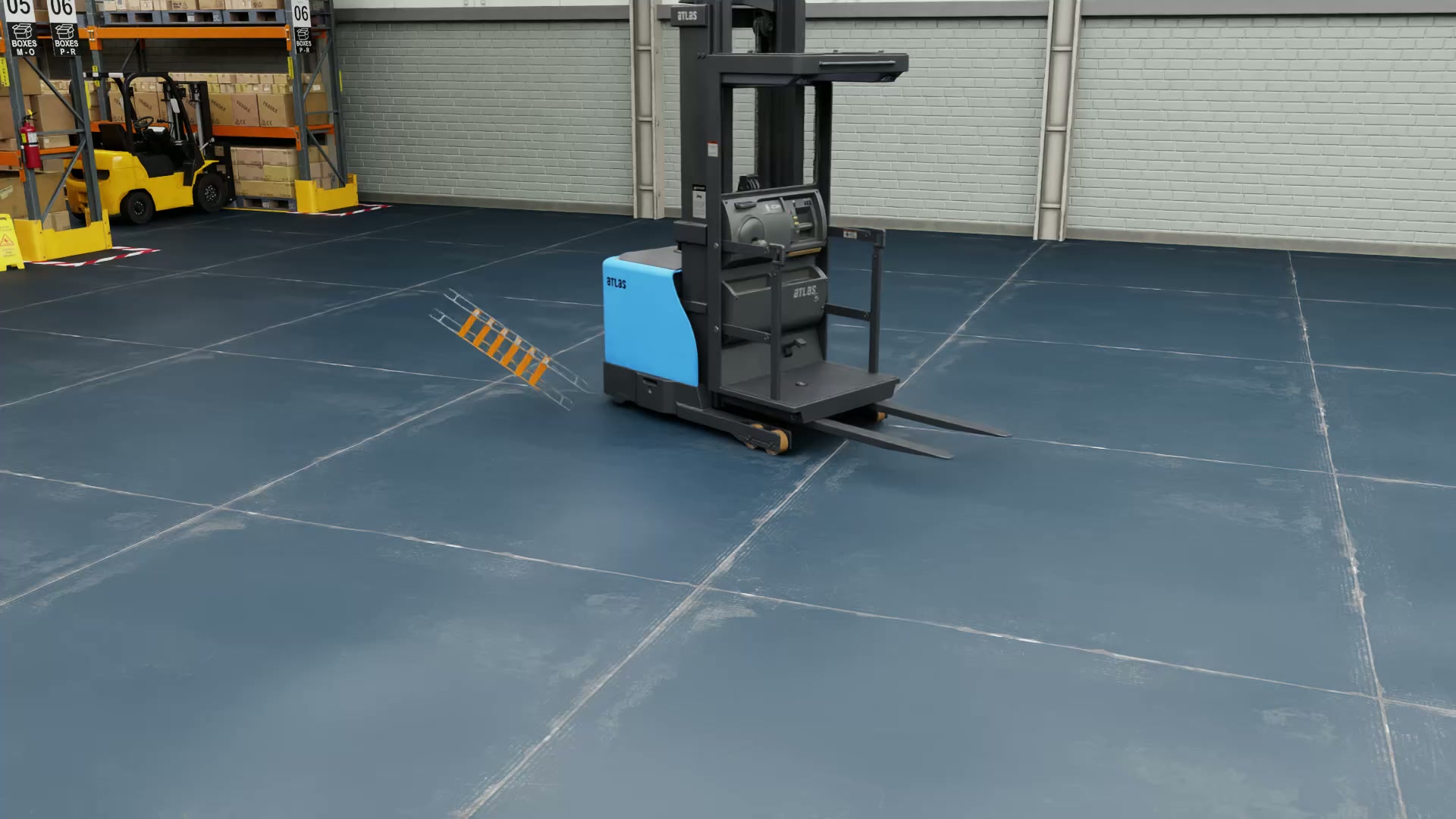}
            & \includegraphics[width=0.17\textwidth]{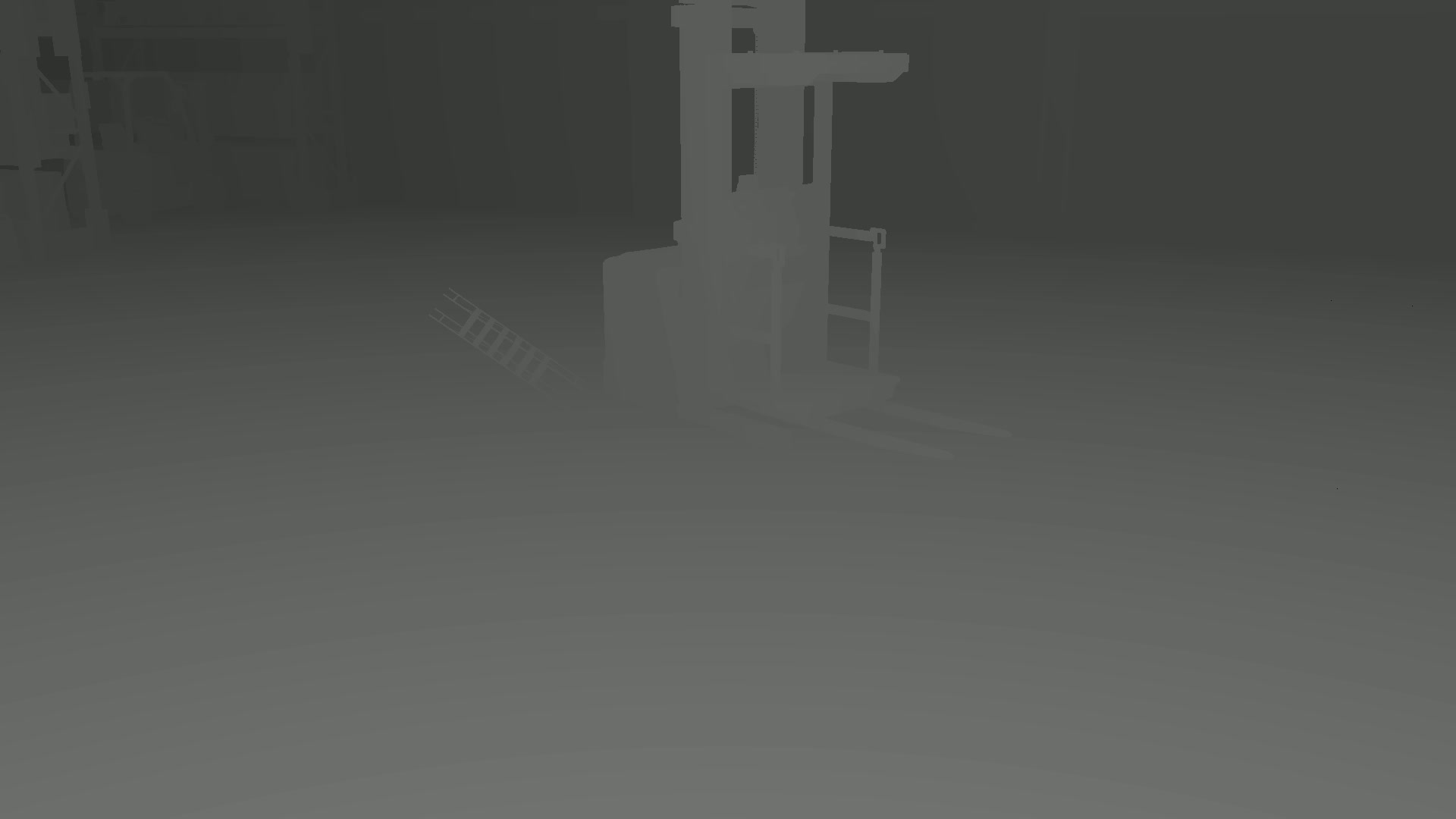}
            & \includegraphics[width=0.17\textwidth]{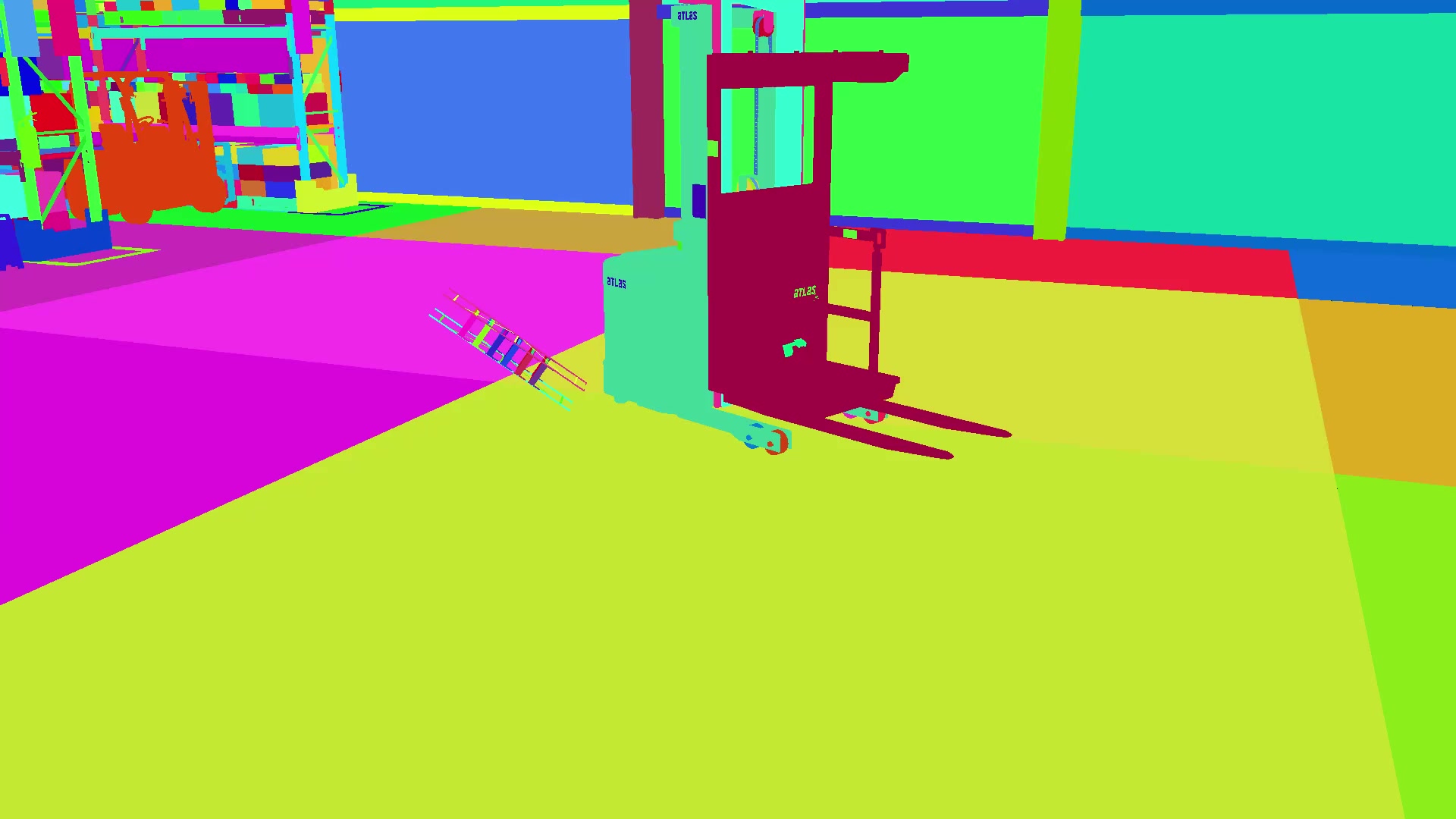}
            & \includegraphics[width=0.17\textwidth]{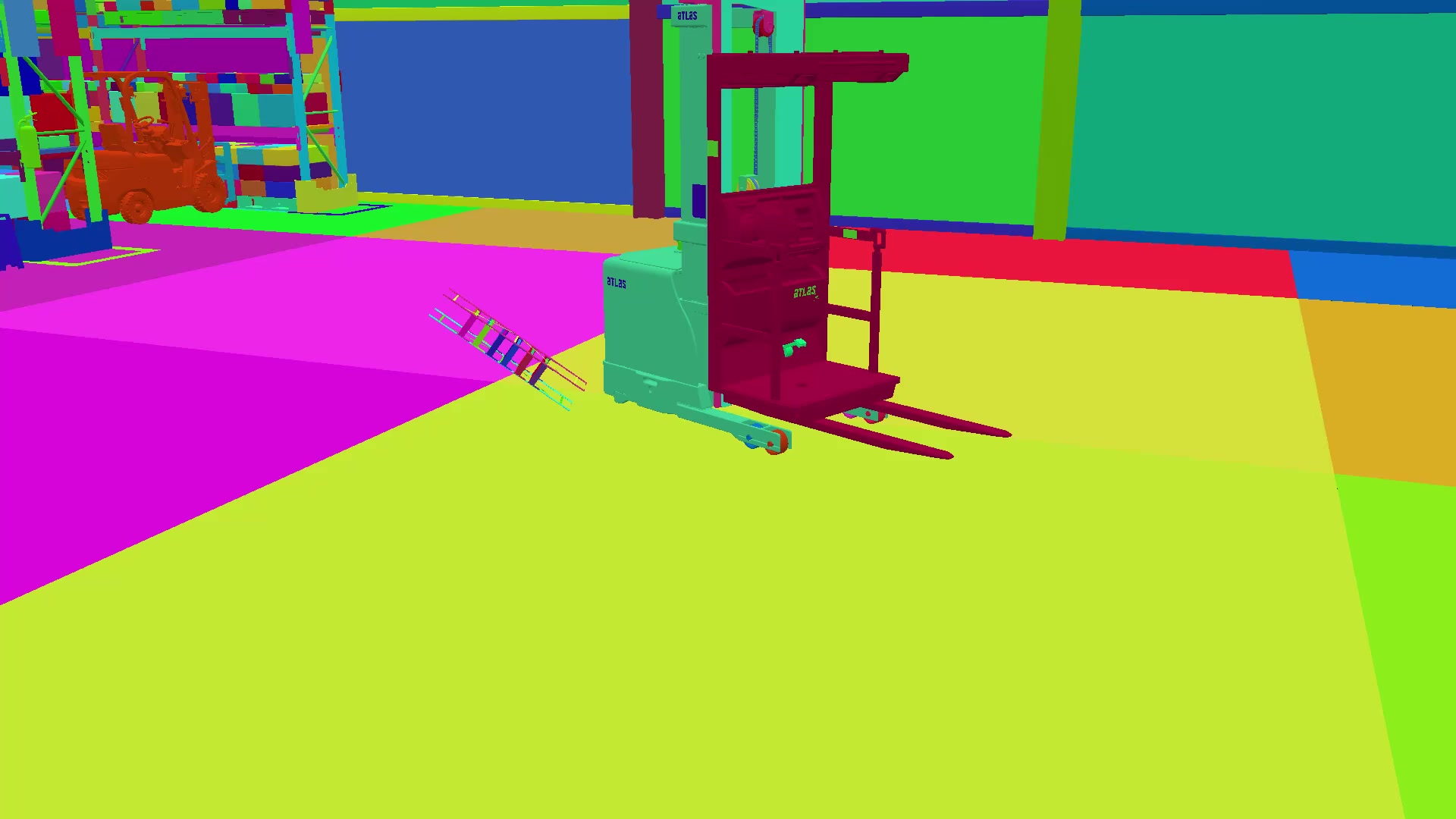}
            & \includegraphics[width=0.17\textwidth]{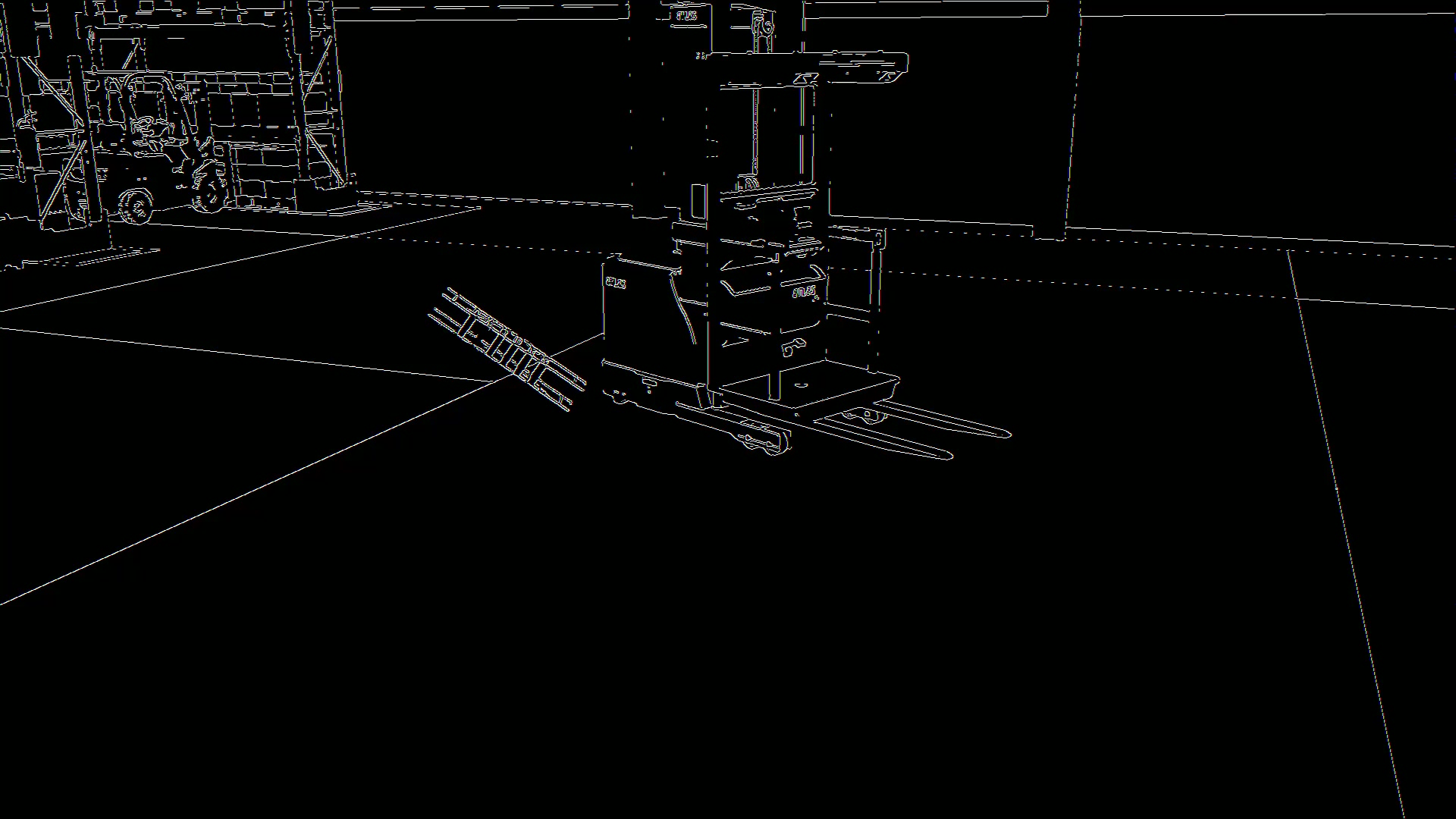} \\[3pt]
        \rotatebox{90}{Box pickup}
            & \includegraphics[width=0.17\textwidth]{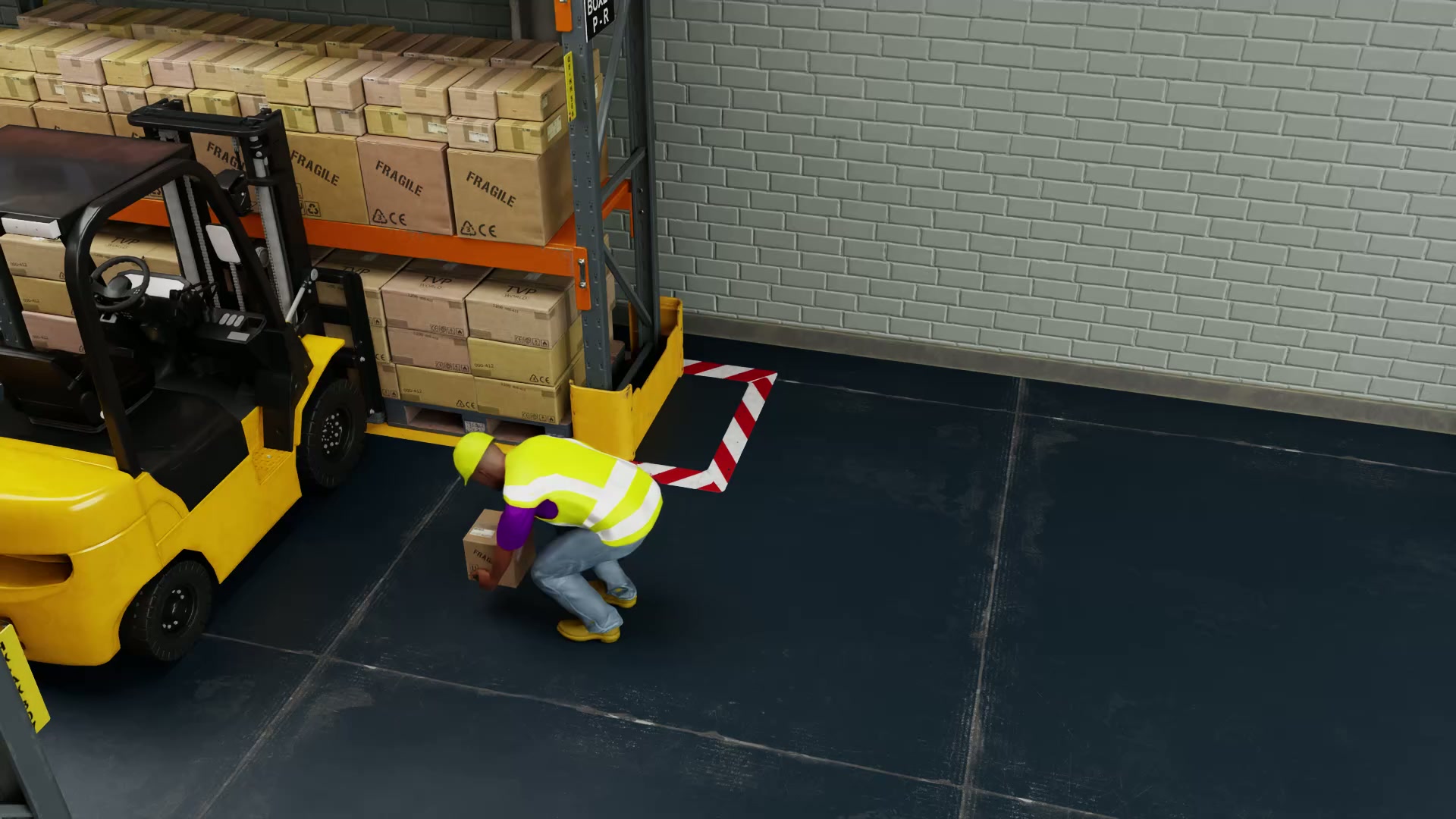}
            & \includegraphics[width=0.17\textwidth]{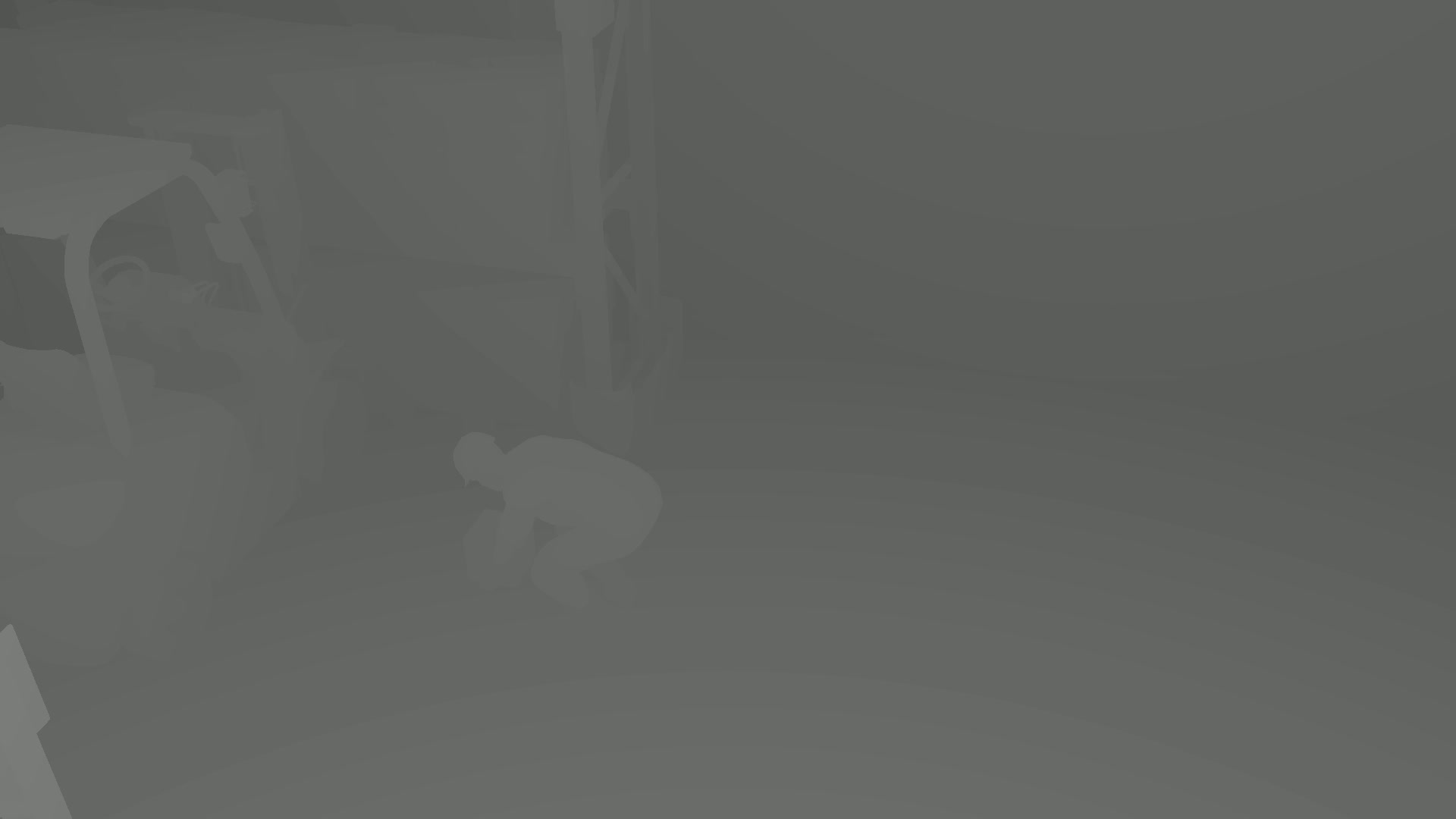}
            & \includegraphics[width=0.17\textwidth]{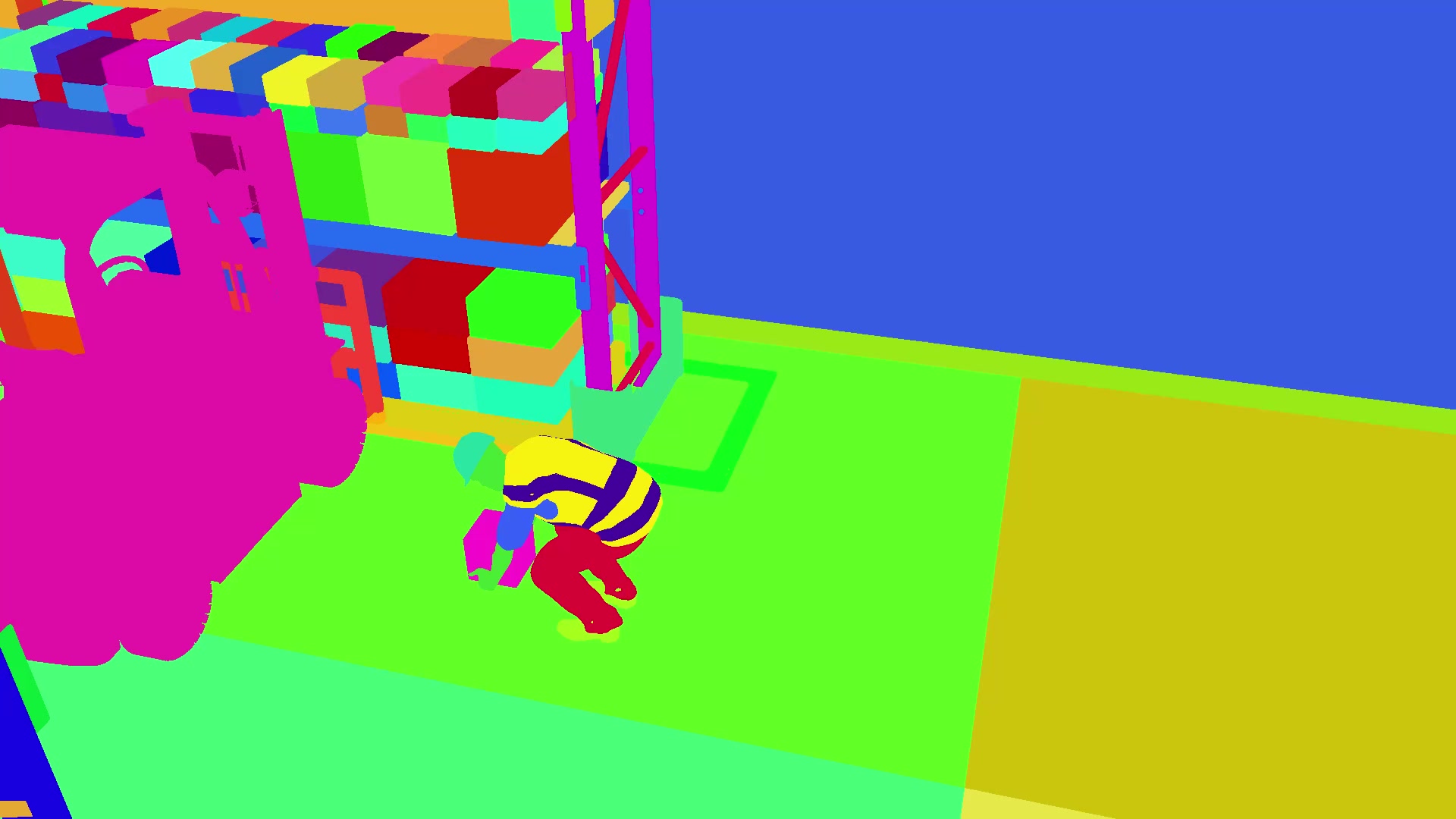}
            & \includegraphics[width=0.17\textwidth]{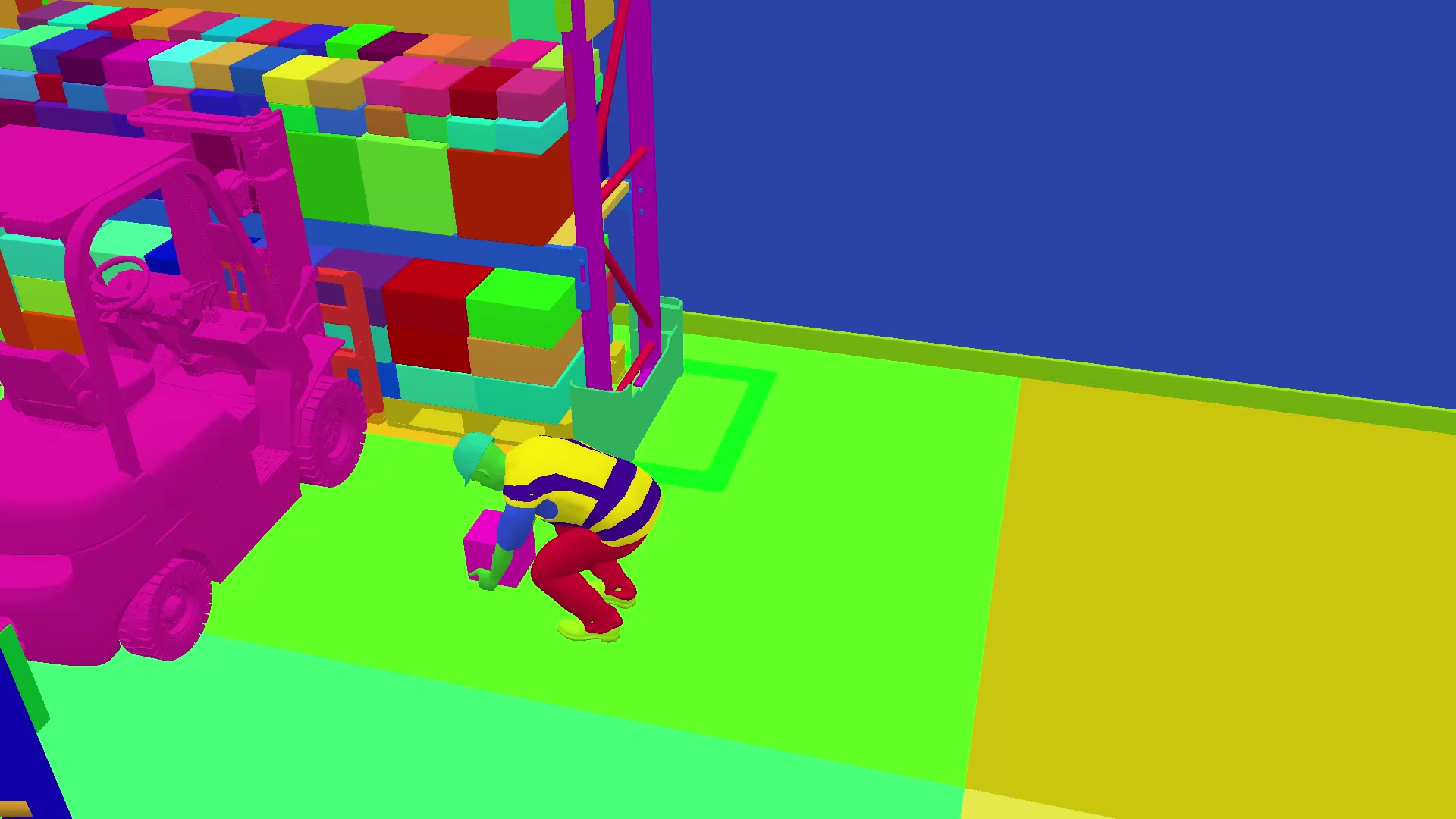}
            & \includegraphics[width=0.17\textwidth]{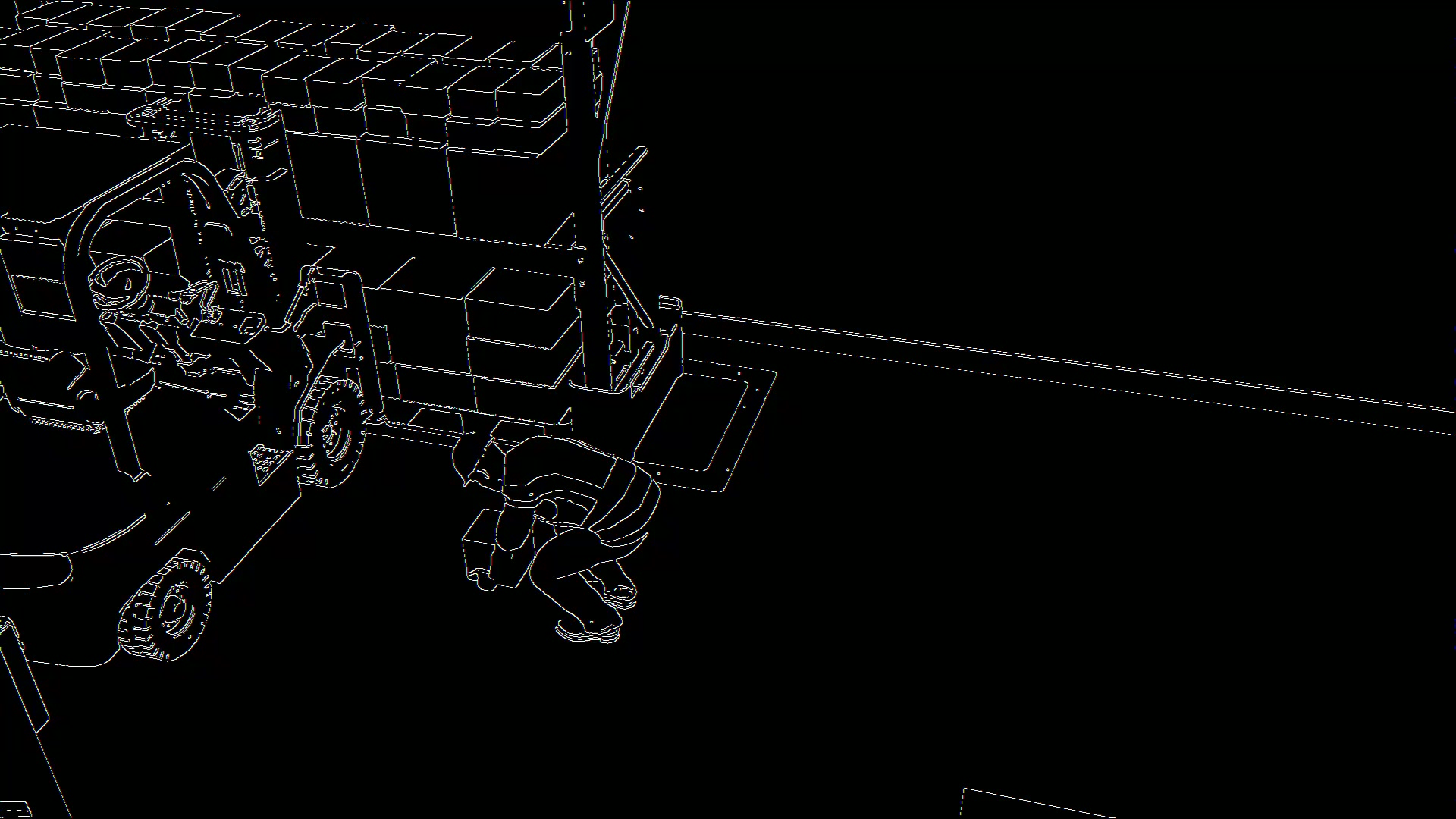} \\
    \end{tabular}
    \caption{\textbf{SDG-Warehouse dataset.} Sample view of the four scenarios and the annotations (RGB clips, metric depth, instance segmentation, shaded segmentation, and Canny edge).}
    \label{fig:sdg_warehouse_modalities}
\end{figure}

\paragraph{Annotation schema.}
Each camera viewpoint provides an H.264 RGB clip with synchronized per-frame annotations: metric depth (raw plus log-normalized colorized variant); instance segmentation (per-pixel IDs traceable to specific simulated objects, with a colorized variant); shaded segmentation (3D-aware rendering with normal-based shading); a Canny edge map computed on the shaded segmentation; 2D tight and loose axis-aligned bounding boxes plus 3D oriented bounding boxes for every tracked agent and prop; and per-frame camera intrinsics and extrinsics. Each annotation stream is also released as an encoded video alongside the RGB clip. Run-level structured metadata records scenario type, random seed, asset and agent inventory, event parameters (e.g., forklift dodge distance, fire ignition location, exit waypoints), and lighting randomization. \Cref{fig:sdg_warehouse_modalities} shows the modalities for one frame per scenario.

\subsection{Distribution of SDG Datasets}
\label{appendix:sdg_distribution}

\begin{figure}[t]
    \centering
    \begin{minipage}[t]{0.5\linewidth}
        \vspace{0pt}
        \centering
        \captionsetup{justification=raggedright, singlelinecheck=false}
        \includegraphics[width=\linewidth]{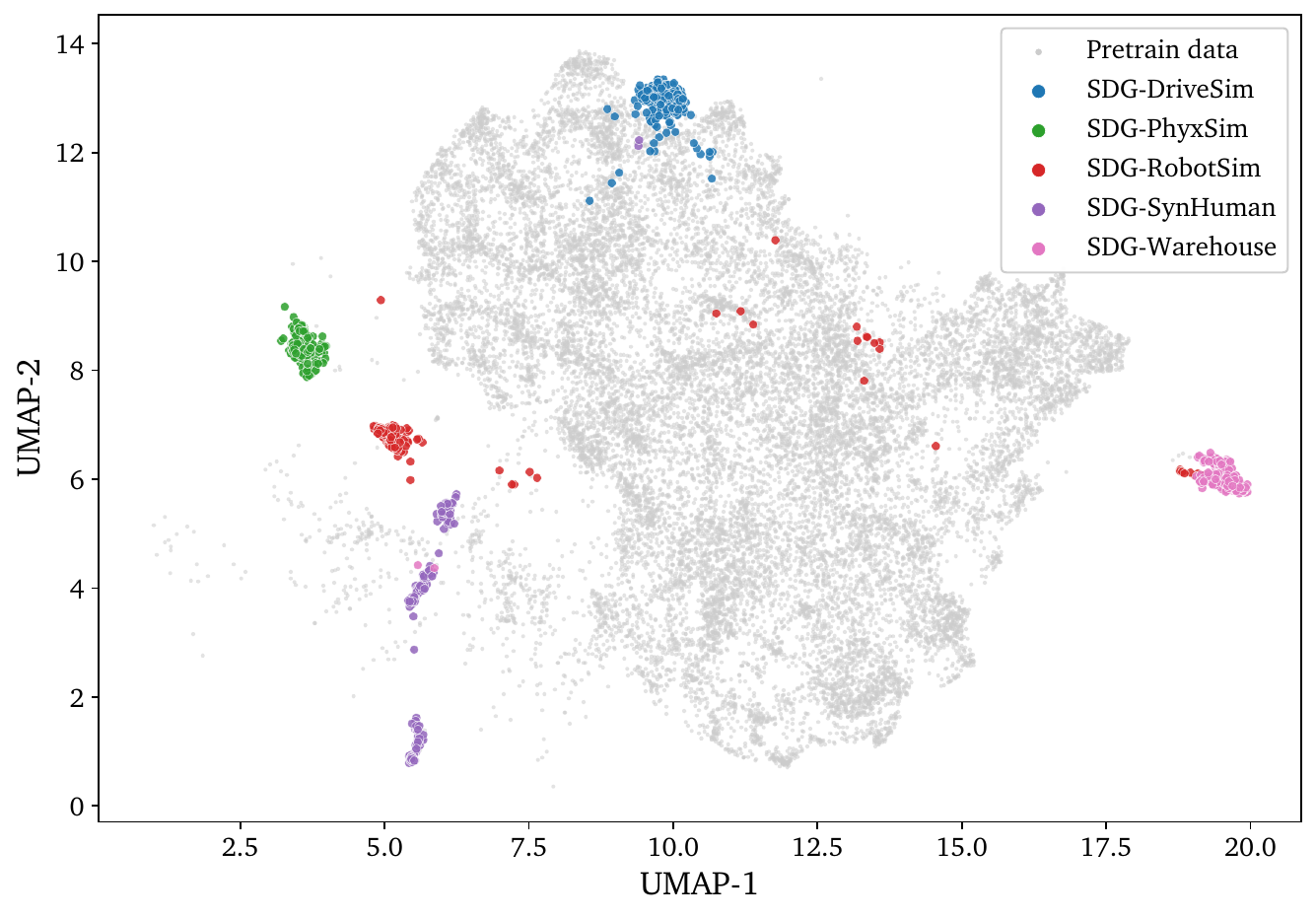}
        \captionof{figure}{\textbf{Joint embedding geometry of pre-train and SDG.}
        PCA$\rightarrow$UMAP projection of 20{,}000 pre-training cluster centroids
        (gray) and 200 randomly sampled clips from each SDG source. Each SDG source
        forms a distinct, tightly clustered region that overlaps only narrowly with
        the bulk pre-training distribution.}
        \label{fig:sdg_pretrain_umap}
    \end{minipage}
    \hfill
    \begin{minipage}[t]{0.43\linewidth}
        \vspace{0pt}
        \centering
        \captionsetup{justification=raggedright, singlelinecheck=false}
        \captionof{table}{\textbf{Distance from SDG to the pre-training manifold.}
        Computed on 100K random clips per source against the 20{,}000 centroids.
        \emph{CosSim} is the mean of $\max_j \cos(\mathbf{g}_i, \mathbf{c}_j)$.
        \emph{MMD$^2$} is an unbiased local Maximum Mean Discrepancy
        (RBF kernel, median-heuristic bandwidth). \emph{\#Local} is the number
        of distinct centroids the source's neighborhood spans. All SDG sources
        sit far from the pre-training distribution---a necessary complement rather
        than a redundant subset.}

        \vspace{1.2em}

        \label{tab:sdg_pretrain_dist}
        \footnotesize
        \setlength{\tabcolsep}{4pt}
        \renewcommand{\arraystretch}{1.15}
        \begin{tabular}{lccc}
            \toprule
            Source & CosSim $\uparrow$ & MMD$^2$ $\downarrow$ & \#Local \\
            \midrule
            \textit{Pre-train (ref.)} & \textbf{0.796} & \textbf{0.005} & 19{,}934 \\
            \midrule
            SDG-DriveSim   & 0.667 & 0.119 & 5{,}577 \\
            SDG-PhyxSim    & 0.589 & 0.221 & 6{,}644 \\
            SDG-RobotSim   & 0.627 & 0.162 & 9{,}760 \\
            SDG-SynHuman   & 0.650 & 0.191 & 4{,}975 \\
            SDG-Warehouse  & 0.712 & 0.361 & 1{,}920 \\
            \bottomrule
        \end{tabular}
    \end{minipage}
\end{figure}

To verify that synthetic content is genuinely complementary to the pre-training corpus, we analyze its position in the Cosmos-Embed1 video embedding space relative to clusters from the pre-training distribution. \cref{fig:sdg_pretrain_umap} visualizes the joint geometry of pre-training and SDG embeddings, while~\cref{tab:sdg_pretrain_dist} quantifies the distance from each SDG source to the pre-training manifold using an in-distribution self-reference as a baseline. Together, these results show that SDG occupies long-tail regions of the embedding space that are not sufficiently covered by web-scale pre-training alone.

\subsection{Ablation Study: Impact of SDG Datasets}
\label{appendix:sdg_ablation}

We study how different synthetic data generation (SDG) sources affect video generation quality and domain understanding by fine-tuning our pre-trained model (Cosmos3-Nano) on each source individually and jointly. We evaluate all variants using PAIBench-G T2V benchmark~\citep{zhou2025paibench}, which reports an overall score, a perceptual Quality score, and six domain-specific sub-scores: Common Sense, AV (autonomous vehicles), Robot,
Industry, Human, and Physics. Results are summarized in~\cref{tab:sdg_ablation}.

\newcommand{\dpos}[1]{\,{\scriptsize\textcolor{green!50!black}{$+$#1}}}
\newcommand{\dneg}[1]{\,{\scriptsize\textcolor{red!70!black}{$-$#1}}}
\newcommand{\dzer}{\,{\scriptsize\textcolor{gray}{$\pm$0.00}}}

\begin{table}[t]
\centering
\captionsetup{justification=raggedright, singlelinecheck=false}
\caption{\textbf{Ablation study on synthetic data generation (SDG) datasets, evaluated
on PAIBench-G T2V~\citep{zhou2025paibench}}. Each model is fine-tuned from the same pre-trained baseline
with data from one SDG source; SDG-All mixes all five sources. \textbf{Bold}
denotes improvement over the baseline; \underline{underline} marks the best
score in each column. Colored deltas show change relative to the baseline
(\textcolor{green!50!black}{green} = improvement,
\textcolor{red!70!black}{red} = degradation).}
\label{tab:sdg_ablation}
\resizebox{\textwidth}{!}{%
\begin{tabular}{lcccccccccc}
\toprule
\textbf{Model} &
\textbf{Overall} &
\textbf{Domain} &
\textbf{Quality} &
\textbf{Comm.\ Sense} &
\textbf{AV} &
\textbf{Robot} &
\textbf{Industry} &
\textbf{Human} &
\textbf{Physics} \\
\midrule
Baseline (pre-train)
  & 79.67\dzer & 86.87\dzer & 72.46\dzer
  & 91.89\dzer & 70.86\dzer & 87.37\dzer
  & 88.66\dzer & \underline{85.46}\dzer & 94.58\dzer \\
\midrule
+~SDG-DriveSim
  & \textbf{79.76}\dpos{0.09}
  & \textbf{86.97}\dpos{0.10}
  & \textbf{72.55}\dpos{0.09}
  & \textbf{92.41}\dpos{0.52}
  & 70.48\dneg{0.38}
  & \textbf{88.26}\dpos{0.89}
  & \textbf{88.92}\dpos{0.26}
  & 84.91\dneg{0.55}
  & 94.58\dzer \\
+~SDG-RobotSim
  & 79.66\dneg{0.01}
  & 86.60\dneg{0.27}
  & \underline{\textbf{72.72}}\dpos{0.26}
  & \textbf{92.22}\dpos{0.33}
  & 69.83\dneg{1.03}
  & \textbf{87.60}\dpos{0.23}
  & 87.44\dneg{1.22}
  & 84.99\dneg{0.47}
  & 94.50\dneg{0.08} \\
+~SDG-Warehouse
  & \textbf{79.74}\dpos{0.07}
  & \textbf{87.00}\dpos{0.13}
  & \textbf{72.48}\dpos{0.02}
  & \textbf{92.34}\dpos{0.45}
  & \textbf{71.15}\dpos{0.29}
  & \textbf{87.97}\dpos{0.60}
  & 88.63\dneg{0.03}
  & 85.01\dneg{0.45}
  & \underline{\textbf{94.79}}\dpos{0.21} \\
+~SDG-PhyxSim
  & 79.62\dneg{0.05}
  & 86.68\dneg{0.19}
  & \textbf{72.56}\dpos{0.10}
  & 91.57\dneg{0.32}
  & 69.44\dneg{1.42}
  & \textbf{88.17}\dpos{0.80}
  & \underline{\textbf{89.51}}\dpos{0.85}
  & 84.77\dneg{0.69}
  & \textbf{94.72}\dpos{0.14} \\
+~SDG-SynHuman
  & \underline{\textbf{79.79}}\dpos{0.12}
  & \underline{\textbf{87.16}}\dpos{0.29}
  & 72.41\dneg{0.05}
  & \underline{\textbf{92.55}}\dpos{0.66}
  & \underline{\textbf{71.33}}\dpos{0.47}
  & \underline{\textbf{88.60}}\dpos{1.23}
  & 88.51\dneg{0.15}
  & 85.08\dneg{0.38}
  & 94.56\dneg{0.02} \\
\midrule
+~SDG-All
  & \textbf{79.77}\dpos{0.10}
  & \textbf{86.97}\dpos{0.10}
  & \underline{\textbf{72.56}}\dpos{0.10}
  & \textbf{92.40}\dpos{0.51}
  & \textbf{71.19}\dpos{0.33}
  & \textbf{87.68}\dpos{0.31}
  & \textbf{88.79}\dpos{0.13}
  & 84.99\dneg{0.47}
  & \textbf{94.67}\dpos{0.09} \\
\bottomrule
\end{tabular}
}
\end{table}

\paragraph{Domain-specific improvements.}
A consistent pattern across all SDG variants is that each source lifts
\emph{different} domain-specific scores, reflecting the unique content
distribution of each simulator.
SDG-DriveSim yields the largest
gain in the Robot domain ($+0.89$) and a strong Common Sense improvement
($+0.52$), reflecting the rich structured dynamics of driving scenarios.
SDG-RobotSim improves perceptual Quality most among all sources ($+0.26$) and
moderately lifts the Robot score ($+0.23$). SDG-Warehouse provides broad
positive deltas across Overall, Domain, AV ($+0.29$), and Physics ($+0.21$),
while maintaining strong in Robot ($+0.60$). SDG-PhyxSim delivers the
largest single-domain gain: a $+0.85$ uplift in Industry and a $+0.80$ gain in
Robot, driven by its physics-grounded industrial and manipulation content.
SDG-SynHuman is the standout individual source, posting the best overall score
($79.79$) with the highest positive deltas in Domain ($+0.29$), Common Sense
($+0.66$), AV ($+0.47$), and Robot ($+1.23$). The breadth of human-centric synthetic content appears to
provide a strong general-purpose signal that transfers across multiple
evaluation domains.

\paragraph{Sim-to-real gap and domain trade-offs.}
The most consistent pattern of degradation across all SDG sources is the Human
domain score, which drops in every single variant without exception---ranging
from $-0.38$ (SDG-SynHuman) to $-0.69$ (SDG-PhyxSim). Notably, even
SDG-SynHuman, which is specifically built from synthetic human-centric scenes,
fails to recover this score. This suggests that the sim-to-real gap is
particularly pronounced for human-related visual content: current simulators do
not yet replicate the subtle appearance, motion, and behavioral nuances of real
humans with sufficient fidelity to benefit this evaluation domain. More broadly,
domain-specialized sources can also degrade orthogonal categories---SDG-RobotSim
hurts AV ($-1.03$) and Industry ($-1.22$), while SDG-PhyxSim's physics emphasis
comes with notable AV degradation ($-1.42$)---underscoring the need to mix
sources rather than rely on any single simulator.

\paragraph{Combined SDG-All achieves broad and balanced gains.}
Mixing all SDG sources (SDG-All) yields uniformly positive deltas across eight
of nine metrics, with only Human showing a residual dip ($-0.47$), consistent
with the sim-to-real gap observed above. It achieves the best Quality score
($72.56$) and consistent improvements across all other domain categories,
demonstrating that data diversity suppresses individual source biases. Based on
these results, our final model incorporates the SDG datasets \emph{together with real, high-quality videos} during a dedicated mid-training stage. This design allows the model to absorb the domain-specific physical
understanding encoded in synthetic data while retaining the perceptual fidelity
and visual realism it acquired from real footage in pre-training.

\section{Cosmos3-Edge LLM Model Training}
\label{app:training_edge}

Cosmos3-Edge uses a dense 2B backbone trained from scratch. Its training follows a two-stage curriculum: pre-training followed by supervised fine-tuning. The pre-training stage is further divided into base pre-training and long-context extension. We use BF16 precision throughout training; optimizer settings are given below.

\paragraph{Base pre-training.}
During base pre-training, we train the 2B Edge backbone from scratch on a total of 15T tokens from the Nemotron pre-training corpus, using a sequence length of 8,192 tokens. This stage consists of two sub-stages: general pre-training on a broad-coverage data mixture, followed by continued pre-training on a higher-quality mixture. The data mixture is hot-swapped during training: continued pre-training resumes from the general-pre-training checkpoint while preserving the optimizer state and learning-rate schedule, so only the data mixture changes. We use AdamW with peak learning rate $1.2\times 10^{-3}$, $(\beta_1,\beta_2)=(0.9,0.95)$, weight decay $0.1$, and gradient clipping at norm~1.0. We use a warmup-stable-decay (WSD) learning-rate schedule, aligning the data-mixture switch with the transition from the stable phase to the decay phase. The tokenizer is shared with the NVIDIA Nemotron-3 models \citep{nemotron3-family}.

\paragraph{Long-context extension.}
During the long-context extension phase, we extend the context window of the Cosmos3-Edge backbone to 128K tokens. Although the extended context window supports 128K-token training sequences, the primary goal of this stage is to improve robustness and quality at the deployed 32K-token sequence length. During the context extension phase, we train on 128K sequence length with an increased RoPE base of 1e8. We use a constant learning rate of $1.2\times 10^{-5}$ and the long-context phase is trained with 90B tokens. For the data blend in this phase, we downsample the pre-training blend to 80\% and add long document QA data as the remaining 20\% in the blend.

\paragraph{Supervised fine-tuning.}
We use the supervised fine-tuning (SFT) data from Nemotron-Cascade-2 \citep{yang2026nemotroncascade2}, which covers a broad set of domains including mathematics, coding, science, general chat, instruction following, tool use, and code-agent tasks. In total, the dataset contains approximately 26M SFT samples. We pack these examples into sequences of up to 128K tokens, yielding roughly 2.6M packed training samples. The model is trained in a single SFT stage with a global batch size of 32. We use the AdamW optimizer with a learning rate of $2\times10^{-5}$ and $(\beta_1,\beta_2)=(0.9,0.98)$. Empirically, model capability peaks after approximately 1.7 epochs, corresponding to 140K training steps.

We evaluate our SFT model on a set of text benchmarks spanning reasoning, science, instruction following, long context, and general capabilities: HMMT25 Feb \citep{hmmt2025feb}, GPQA \citep{rein2023gpqa}, MMLU-Pro \citep{wang2024mmlupro}, AA-LCR \citep{artificialanalysis2025lcr}, IFBench \citep{pyatkin2025generalizing}, and Scale AI Multi-Challenge \citep{sirdeshmukh2025multichallenge}. We compare against Qwen3.5-2B \citep{qwen3_5towards}, a strong baseline with the same model size.

As shown in ~\cref{tab:text_sft_results}, our text SFT model substantially outperforms Qwen3.5-2B on math reasoning and science benchmarks, such as HMMT25 Feb and GPQA. It yields comparable results on instruction-following and long-context evaluations, including IFBench and AA-LCR. However, it lags behind Qwen3.5-2B on general-domain benchmarks such as MMLU-Pro.

\begin{table}[h]
    \centering
    \caption{
    \textbf{Text benchmark results.} Comparing \textbf{Cosmos3-Edge} and Qwen3.5-2B across reasoning, science, instruction-following, and long-context evaluations. \textbf{Cosmos3-Edge} substantially improves mathematical and scientific reasoning capability on HMMT25 Feb and GPQA, while achieving comparable scores on IFBench and AA-LCR.
    }
    \label{tab:text_sft_results}

    \small
    \setlength{\tabcolsep}{8pt}
    \renewcommand{\arraystretch}{1.15}

    \begin{tabular}{lcccccc}
        \toprule
        \textbf{Model} &
        \textbf{\begin{tabular}[c]{@{}c@{}}HMMT25\\ Feb\end{tabular}} &
        \textbf{GPQA} &
        \textbf{\begin{tabular}[c]{@{}c@{}}MMLU\\ Pro\end{tabular}} &
        \textbf{AA-LCR} &
        \textbf{\begin{tabular}[c]{@{}c@{}}IFBench\\ (prompt)\end{tabular}} &
        \textbf{\begin{tabular}[c]{@{}c@{}}Scale AI\\ Multi-Challenge\end{tabular}} \\
        \midrule

        Qwen3.5-2B
        & 22.9
        & 51.6
        & \textbf{66.5}
        & \textbf{25.6}
        & 41.3
        & \textbf{33.7} \\

        \rowcolor{rowours}
        \textbf{Cosmos3-Edge}
        & \textbf{76.3}
        & \textbf{56.4}
        & 62.6
        & 22.8
        & \textbf{43.6}
        & 28.1 \\

        \bottomrule
    \end{tabular}
\end{table}

\section{Additional Ablation Study}
\label{appendix:training_ablations}

While the main experiments demonstrate the effectiveness of Cosmos 3 across a wide range of understanding and generation tasks, they do not fully isolate the contributions of individual design choices. To better understand the factors underlying the model's capabilities, we conduct a series of ablation studies examining key architectural, data, and training decisions. These studies investigate how the Reasoner and Generator interact within the Mixture-of-Transformers framework, the impact of multimodal training signals such as audio and action data, the effectiveness of temporal conditioning mechanisms, and the transferability of learned world and action representations across domains. Together, these analyses provide deeper insight into the design principles that enable Cosmos 3 to function as a unified omnimodal world model for Physical AI.s

\subsection{How the Reasoner Benefits the Generator}
In this study, we investigate how the Reasoner benefits the Generator model. We train two models using the Cosmos3-Nano architecture: one with Qwen3-VL-8B as the understanding tower, and the other with our Cosmos3-Nano Reasoner. In both cases, the Generator tower is trained from scratch. For this ablation, we restrict training to 256p and 480p resolutions with clip lengths of 0–200 frames, and set the sequence length to 25K. Following our large-scale run, we use joint image–video training. Both models are trained for 90K iterations on 256 GPUs. We report PAIBench scores for both below.

\newcommand{\roth}[1]{\rotatebox[origin=l]{90}{\small #1}}

\newcommand{\vraise}[1]{\raisebox{3.0em}{#1}}

\begin{table}[h]
\centering
\caption{\textbf{Understanding tower ablation.} We report domain and quality scores on PAIBench T2V and I2V for two variants that differ only in the pretrained model used to initialize the understanding tower while the Generator tower is trained from scratch: (1) Cosmos 3 Reasoner and (2) Qwen3-VL. Subject and background consistency are I2V-specific and not applicable to T2V. Initializing the understanding tower from the Cosmos3 Reasoner yields better domain scores on Physical AI domains than the Qwen3-VL variant.}
\setlength{\tabcolsep}{4pt}
\renewcommand{\arraystretch}{1.15}
\resizebox{\linewidth}{!}{%
\begin{tabular}{@{}ll | ccc | cccccc | cccccc | cc@{}}
\toprule
\multicolumn{2}{c}{} & \multicolumn{3}{c}{\textbf{Summary}} & \multicolumn{6}{c}{\textbf{Domain}} & \multicolumn{6}{c}{\textbf{Quality}} & \multicolumn{2}{c}{\textbf{I2V}} \\
\cmidrule(lr){3-5} \cmidrule(lr){6-11} \cmidrule(lr){12-17} \cmidrule(lr){18-19}
\textbf{Bench} & \textbf{Und. tower}
 & Overall & Domain & Quality
 & C.S. & AV & Rob. & Ind. & Hum. & Phy.
 & Subj. & Bg. & Motion & Aesth. & Imag. & Cons.
 & Subj. & Bg. \\
\midrule
\multirow{2}{*}{T2V}
& \cellcolor{rowours}Cosmos3 Reasoner  & \cellcolor{rowours}\textbf{74.3} & \cellcolor{rowours}\textbf{75.7} & \cellcolor{rowours}73.0          & \cellcolor{rowours}\textbf{81.2} & \cellcolor{rowours}\textbf{54.9} & \cellcolor{rowours}\textbf{71.3} & \cellcolor{rowours}\textbf{78.4} & \cellcolor{rowours}\textbf{76.2} & \cellcolor{rowours}\textbf{89.2} & \cellcolor{rowours}\textbf{96.0} & \cellcolor{rowours}\textbf{96.8} & \cellcolor{rowours}99.4          & \cellcolor{rowours}\textbf{55.2} & \cellcolor{rowours}71.4          & \cellcolor{rowours}\textbf{19.1} & \cellcolor{rowours}--            & \cellcolor{rowours}--            \\
& Qwen-3 VL & 73.3          & 73.7          & \textbf{73.0} & 80.5          & 52.6          & 66.5          & 77.4          & 74.0          & 88.7          & 95.8          & 96.6          & \textbf{99.4} & 55.0          & \textbf{72.2} & 19.0          & --            & --            \\
\midrule
\multirow{2}{*}{I2V}
& \cellcolor{rowours}Cosmos3 Reasoner  & \cellcolor{rowours}\textbf{79.4} & \cellcolor{rowours}\textbf{80.8} & \cellcolor{rowours}\textbf{78.1} & \cellcolor{rowours}89.0          & \cellcolor{rowours}59.4          & \cellcolor{rowours}\textbf{77.0} & \cellcolor{rowours}\textbf{84.8} & \cellcolor{rowours}\textbf{79.3} & \cellcolor{rowours}\textbf{91.6} & \cellcolor{rowours}\textbf{92.4} & \cellcolor{rowours}\textbf{94.7} & \cellcolor{rowours}99.4          & \cellcolor{rowours}\textbf{53.2} & \cellcolor{rowours}68.7          & \cellcolor{rowours}\textbf{20.2} & \cellcolor{rowours}\textbf{98.0} & \cellcolor{rowours}\textbf{98.0} \\
& Qwen-3 VL & 79.0          & 80.0          & 78.0          & \textbf{89.7} & \textbf{59.8} & 74.0          & 84.0          & 78.3          & 91.4          & 92.0          & 94.5          & \textbf{99.4} & 53.1          & \textbf{69.3} & 20.1          & 97.9          & 97.9          \\
\bottomrule
\end{tabular}%
}
\label{tab::ablation-understanding-tower}
\end{table}

As shown in Table~\ref{tab::ablation-understanding-tower}, replacing Qwen3-VL-8B with our Cosmos3-Nano Reasoner in the understanding tower yields consistent improvements in domain scores, particularly in Physical AI domains. On T2V, the Reasoner improves the overall Domain score from 73.7 to 75.7, with the largest gains concentrated in physically-grounded categories: Robot (+4.8, 66.5 → 71.3), Physics (+0.5, 88.7 → 89.2), AV (+2.3, 52.6 → 54.9), and Industry (+1.0, 77.4 → 78.4). A similar pattern holds on I2V, where the Domain score rises from 80.0 to 80.8, driven by gains in Common sense (+0.7), Industry (+0.8), Human (+1.0), and Physics (+0.2). These results suggest that the reasoner provides better embeddings for Physical AI domains for generator to learn. The quality scores for both models are comparable.

\subsection{Choice of FPS Control}

We study two complementary mechanisms for conditioning generation on a target frame rate: (i) MRoPE FPS modulation, which scales the temporal axis of the unified 3D MRoPE by the target FPS, and (ii) Text Control, which injects the target duration and FPS as natural-language text inside the structured JSON caption. For this ablation, we
restrict training to 256p and 480p resolutions with clip lengths of 0–200 frames, and set the sequence length
to 25K. Following our large-scale run, we use joint image–video training. We train 4 models for 130K iterations on 128 GPUs each, varying only whether each mechanism is active: Base (no control), Text Control (text only), MRoPE FPS Modulation (MRoPE only), and Text Control + MRoPE FPS Modulation (both). We curate an evaluation set with known source duration and FPS, $\sim$100 videos each of 10, 15, 24, 30 FPS ($\pm$2 FPS tolerance). We use the structured captions of these evaluation set videos as prompts and generate samples across 3 seeds per prompt at 480p (16:9) resolution. We produce a 5 s clip for each prompt.

We score each clip on Video Quality (VQ; DOVER perceptual quality score~\citep{dover}) and Dynamic Degree (DD; motion presence on a 0--1 scale~\citep{huang2023vbench}). From the three-seed DD scores per prompt ($p$), we compute a normalized motion control (MC) term
\begin{equation}
    \text{MC} = \mathbb{E}_{p}\!\left[\frac{\text{var}_p}{\text{var}_p + \text{mean}_p^2}\right],
\end{equation}
MC indicates the robustness of the control mechanism, \ie, the degree of variability per prompt for each control setting. We combine these into a Motion Fidelity score
\begin{equation}
    \text{MF} = (1 - |\text{DD} - \text{DD}_{\text{ref}}|)\,(1 - \text{MC}),
\end{equation}
where $\text{DD}_{\text{ref}}$ is the per-band mean DD over the real-video reference set. The final Composite Score is the product of Video Quality and Motion Fidelity, reflecting adherence to motion magnitude without impacting the perceptual quality of generated videos. This is computed as $\mathbb{E}_{\text{FPS-band}}\!\left[(\text{VQ}) \cdot \text{MF}\right]$, computed per FPS band and then averaged across bands.

\cref{tab:fps_control_ablation} reports averages across the four FPS bands. Both mechanisms improve over Base individually; MRoPE FPS modulation alone yields a larger gain than the Text Control ($+1.12$ vs.\ $+0.77$ in composite). Combining the two produces the best composite score ($+1.30$ over Base). VQ stays within a 0.2-point window across all four settings, indicating that the gains are concentrated in motion fidelity rather than video perceptual quality, \ie, the controls primarily improve temporal behavior. Based on these results, we adopt the Text Control + MRoPE FPS Modulation configuration for Cosmos 3 Generator.

\begin{table}[h]
\label{tab::fps_control}
    \centering
    \captionsetup{justification=raggedright, singlelinecheck=false}
    \caption{\textbf{FPS control ablation on Cosmos3-Nano.} Scores are averaged across four FPS bands (10, 15, 24, 30). VQ is the DOVER video quality score; MF is motion fidelity (0--1); Composite $= \mathbb{E}_{\text{band}}[(\text{VQ}) \cdot \text{MF}]$. \textbf{Bold} indicates best in the column. Text Control + MRoPE FPS Modulation is the most performant control setting that demonstrates adherence to reference FPS band motion while preserving perceptual quality.}
    \label{tab:fps_control_ablation}
    \setlength{\tabcolsep}{8pt}
    \renewcommand{\arraystretch}{1.25}
    \resizebox{\linewidth}{!}{
    \begin{tabular}{l l c c c}
        \toprule
        \textbf{Model} & \textbf{FPS Control Setting} & \textbf{Avg.\ VQ ($\uparrow$)} & \textbf{Avg.\ MF ($\uparrow$)} & \textbf{Avg.\ Composite ($\uparrow$)} \\
        \midrule
        Cosmos3-Nano & Base (No Control)  & 12.89 & 0.6626 & 8.51 \\
        Cosmos3-Nano & Text Control            & 12.99 & 0.7169 & 9.28 \\
        Cosmos3-Nano & MRoPE FPS Modulation & \textbf{13.03} & 0.7409 & 9.63 \\
        Cosmos3-Nano & \textbf{Text Control + MRoPE FPS Modulation} & 12.84 & \textbf{0.7649} & \textbf{9.81} \\
        \bottomrule
    \end{tabular}
    }
\end{table}

\subsection{Audio Data in Pre-Training}

We investigate the impact of including audio during continued pre-training on video metrics. Starting from the same pre-trained checkpoint, we train two Cosmos3-Nano variants on our pre-training dataset: one with video-only data and one with joint video-audio data. These ablations are run for 20k iterations on 128 GPUs. We restrict training to 256p and 480p resolutions.

\begin{table}[h]
\centering
\caption{\textbf{Effect of introducing audio data during pre-training.} We report PAIBench T2V and I2V scores after continued pre-training of two generator variants that differ only in whether audio is used during training. The video data is shared across both experiments.}
\setlength{\tabcolsep}{4pt}
\renewcommand{\arraystretch}{1.15}
\resizebox{\linewidth}{!}{%
\begin{tabular}{@{}ll | ccc | cccccc | cccccc | cc@{}}
\toprule
\multicolumn{2}{c}{} & \multicolumn{3}{c}{\textbf{Summary}} & \multicolumn{6}{c}{\textbf{Domain}} & \multicolumn{6}{c}{\textbf{Quality}} & \multicolumn{2}{c}{\textbf{I2V}} \\
\cmidrule(lr){3-5} \cmidrule(lr){6-11} \cmidrule(lr){12-17} \cmidrule(lr){18-19}
\textbf{Bench} & \textbf{Variant}
 & Overall & Domain & Quality
 & C.S. & AV & Rob. & Ind. & Hum. & Phy.
 & Subj. & Bg. & Motion & Aesth. & Imag. & Cons.
 & Subj. & Bg. \\
\midrule
\multirow{2}{*}{T2V} & Without Audio  & 78.6 & 83.8 & \textbf{73.4} & 90.7 & 64.7 & 81.9 & 84.4 & 82.7 & \textbf{94.8} & \textbf{95.9} & \textbf{97.1} & \textbf{99.5} & \textbf{57.3} & \textbf{70.6} & 19.8 & -- & -- \\
 & \cellcolor{rowours}With Audio  & \cellcolor{rowours}\textbf{79.1} & \cellcolor{rowours}\textbf{85.0} & \cellcolor{rowours}73.2 & \cellcolor{rowours}\textbf{91.5} & \cellcolor{rowours}\textbf{67.9} & \cellcolor{rowours}\textbf{83.8} & \cellcolor{rowours}\textbf{86.1} & \cellcolor{rowours}\textbf{83.7} & \cellcolor{rowours}93.8 & \cellcolor{rowours}95.5 & \cellcolor{rowours}96.9 & \cellcolor{rowours}\textbf{99.5} & \cellcolor{rowours}57.1 & \cellcolor{rowours}70.2 & \cellcolor{rowours}\textbf{19.9} & \cellcolor{rowours}-- & \cellcolor{rowours}-- \\
\midrule
\multirow{2}{*}{I2V} & Without Audio  & 81.7 & 85.1 & \textbf{78.4} & 93.2 & 67.5 & 82.0 & 86.0 & 83.2 & \textbf{95.1} & \textbf{93.3} & \textbf{95.1} & \textbf{99.5} & \textbf{55.0} & 67.6 & \textbf{20.4} & \textbf{98.4} & \textbf{98.3} \\
 & \cellcolor{rowours}With Audio  & \cellcolor{rowours}\textbf{82.2} & \cellcolor{rowours}\textbf{85.9} & \cellcolor{rowours}\textbf{78.4} & \cellcolor{rowours}\textbf{93.6} & \cellcolor{rowours}\textbf{68.6} & \cellcolor{rowours}\textbf{84.2} & \cellcolor{rowours}\textbf{86.5} & \cellcolor{rowours}\textbf{84.3} & \cellcolor{rowours}94.8 & \cellcolor{rowours}92.9 & \cellcolor{rowours}94.9 & \cellcolor{rowours}99.4 & \cellcolor{rowours}\textbf{55.0} & \cellcolor{rowours}\textbf{67.9} & \cellcolor{rowours}\textbf{20.4} & \cellcolor{rowours}98.2 & \cellcolor{rowours}98.2 \\
\bottomrule
\end{tabular}%
}
\label{tab::ablation-audio-pretraining}
\end{table}

The results in~\cref{tab::ablation-audio-pretraining} show that continued training without audio leads to a drop in both T2V and I2V scores, suggesting that joint video-audio pre-training does not degrade video generation quality and may provide a modest benefit even when evaluated purely on video-centric metrics.

\subsection{Synergy Between Action Modes}
\label{app:pusht_fd_id_policy_synergy}
As described in \cref{sec::generation_mode_def} and summarized in \cref{fig:action_modes}, Cosmos 3 supports three action generation modes: forward dynamics (FD), inverse dynamics (ID), and joint video-action prediction (policy).
We ablate whether a single joint action model can share useful structure across these modes.
We use the PushT dataset and Cosmos3-Edge for this experiment.
The model is trained for three single-mode action checkpoints for 2K steps each against one joint FD/ID/policy checkpoint trained for 6K steps, so that each mode is trained for the same amount of optimization steps.
We report the PSNR for FD and MSE for ID.
For policy mode, we report the policy coverage ratio of the T block and target region aggregated across 50 different initializations, with 10 rollouts from each starting point.
The results are summarized in \cref{tab:pusht_fd_id_policy_open_loop}.

\begin{table}[h]
\centering
\caption{\textbf{PushT action-mode synergy.} We compare single-mode FD, ID, and policy checkpoints trained for 2K steps each against one joint FD/ID/policy checkpoint trained for 6K steps. Lower is better for ID MSE; higher is better for FD PSNR and policy coverage. \textbf{Bold} indicates the better value in each column.}
\label{tab:pusht_fd_id_policy_open_loop}
\small
\setlength{\tabcolsep}{7pt}
\renewcommand{\arraystretch}{1.15}
\begin{tabular}{@{}lccc@{}}
\toprule
\textbf{Training setting} & \textbf{FD PSNR $\uparrow$} & \textbf{ID MSE $\downarrow$} & \textbf{Policy Coverage $\uparrow$} \\
\midrule
2K single-mode & \textbf{27.13} & $1.11 \times 10^{-3}$ & 74.1\% \\
6K joint FD/ID/policy & 26.22 & $\boldsymbol{3.09 \times 10^{-4}}$ & \textbf{77.3\%} \\
\bottomrule
\end{tabular}
\end{table}

The joint FD/ID/policy checkpoint improves the action-side metrics while preserving comparable forward-dynamics quality.
Compared with the single-mode checkpoints, ID MSE decreases from $1.11 \times 10^{-3}$ to $3.09 \times 10^{-4}$, a 72\% relative reduction, and policy coverage increases from 74.1\% to 77.3\%.
FD PSNR decreases from 27.13 to 26.22, indicating a modest tradeoff in reconstruction fidelity.
Overall, the joint checkpoint provides the best policy coverage and ID accuracy under the same per-mode optimization budget, suggesting that the action modes share useful structure even though FD quality benefits slightly from single-mode specialization.

\FloatBarrier

\subsection{Video-Action Consistency}
\label{app:robolab_video_action_consistency}
\begin{figure}[!htbp]
    \centering
    \includegraphics[width=\linewidth]{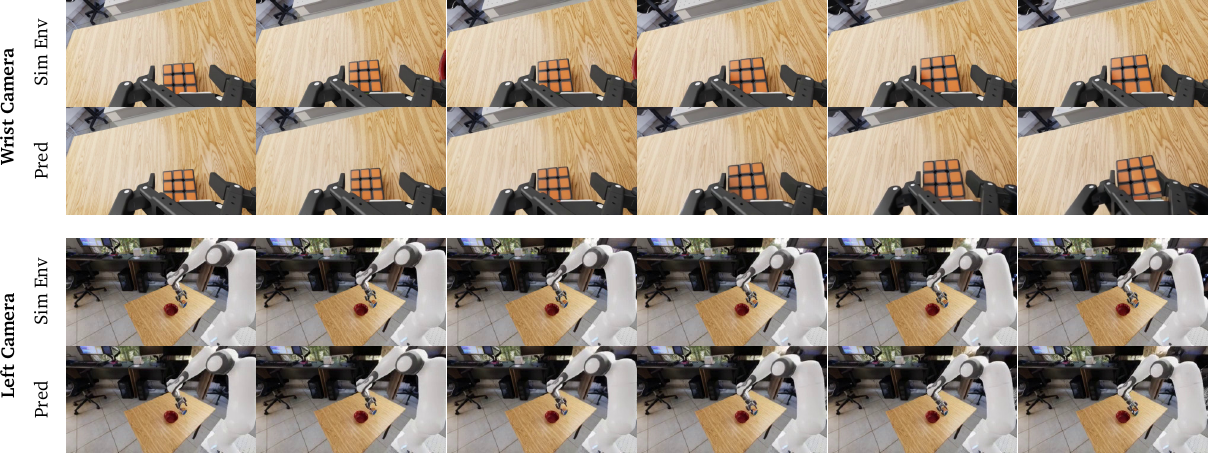}

    \caption{\textbf{Predicted video vs.\ simulator rollout on RoboLab.} For the wrist and left cameras, \textit{Sim Env} shows the video recorded by executing the predicted action chunk in the RoboLab simulator from the same initial state, while \textit{Pred} shows the video predicted by \textbf{Cosmos3-Nano-Policy-DROID} jointly with that action chunk. We observe that the predicted video closely matches the simulator rollout.}
    \label{fig:robolab_video_action_consistency_rubikscube}
    \label{fig:robolab_video_action_consistency}
\end{figure}

In addition to the closed-loop success rate reported in \cref{sec:robot_manip_policy}, we evaluate how well the video and action streams jointly predicted by Cosmos3-Nano-Policy-DROID remain aligned on RoboLab. 
Cosmos3-Nano-Policy-DROID is trained only on DROID, so RoboLab provides a held-out environment for testing this video-action consistency.
For each predicted action chunk, we execute the same chunk in the RoboLab simulator and compute PSNR between the model-predicted video and the resulting simulator rollout, then aggregate the scores across action chunks for the left and wrist camera views.
The left third-person view achieves 23.19\,dB, comparable to the PSNR levels achieved by our robotics forward-dynamics models on DROID (\cref{tab:action_posttraining_summary_wide}).
The wrist (eye-in-hand) view is more challenging because the camera moves with the end-effector and the model must infer newly revealed content; nevertheless, it reaches 17.33\,dB. 
These results indicate strong consistency between the predicted actions and predicted videos.
\Cref{fig:robolab_video_action_consistency} visualizes this consistency: the predicted frames closely match the corresponding frames from simulator rollouts initialized from the same state.

\FloatBarrier

\section{Cosmos-HumanEval Benchmark (Cosmos-HUE)}
\label{sec::cosmos_hue_bench}

\textbf{Cosmos-HumanEval} (Cosmos-HUE or HUE) is an evaluation benchmark for video generation introduced in the main results section. HUE provides a human-reference signal via atomic binary questions generated per video by a Vision-Language Model (VLM) pipeline and scored along the four HUE dimensions (Semantic Alignment, Physical Laws, Geometric Reasoning, Visual Integrity). This appendix complements that overview with the formal binary scoring scheme, the annotation protocol and reliability estimates, and the per-dimension and per-domain Text-to-Video (T2V) and Image-to-Video (I2V) leaderboards.

\paragraph{Binary response schema and scoring.} Each Layer~3 question elicits \textbf{Yes} (criterion met), \textbf{No} (clear violation), or \textbf{Unclear} (e.g., the region is
occluded or the motion is too fast to assess, or the video fails to depict the event the question presupposes, e.g., the prompt calls for a right turn but the lead vehicle never
turns, so the follow-up question about a smooth curved arc has no clear Yes or No). \emph{Unclear is treated as No}: a model is not rewarded for failing to produce the prompted action,
and a non-confident annotator is conservatively counted against the model. Concretely, Unclear does not contribute to the Yes numerator but remains in the denominator, so a video that
is mostly Unclear cannot score above one with the same number of explicit Yes answers. Let $\mathcal{Q}_v$ denote the set of questions issued for video $v$; the per-video
HUE score is
\begin{equation}
  \mathrm{HUE}(v) \;=\; \frac{\sum_{q \in \mathcal{Q}_v} \mathbf{1}\!\left[\mathrm{ans}(v,q) = \textsc{Yes}\right]}{\lvert \mathcal{Q}_v \rvert} \times 100\%,
  \label{eq:hue_video}
\end{equation}
with $\mathrm{HUE}(v) \in [0, 100]\%$ since ``Yes'' is always the desirable outcome. Model-level scores aggregate over $V$ test videos:
\begin{equation}
  \mathrm{HUE}(m) \;=\; \frac{\sum_{v=1}^{V} \sum_{q \in \mathcal{Q}_v} \mathbf{1}\!\left[\mathrm{ans}(v,q) = \textsc{Yes}\right]}{\sum_{v=1}^{V} \lvert \mathcal{Q}_v \rvert} \times
100\%.
  \label{eq:hue_model}
\end{equation}
The grand mean weights every answered (video, question) observation equally, avoiding inflation from videos that happen to receive fewer applicable questions. Dimension-level scores
are computed identically by restricting the sums to questions belonging to each dimension.

\paragraph{Annotation protocol.} Each video receives up to 16 atomic binary questions from the three-layer pipeline above, and each (video, question) pair is independently rated by \emph{two} human annotators. If the two annotators agree, the consensus answer is recorded as the canonical response; if they disagree, the question is escalated to a third \emph{quality-control (QC) reviewer}, whose answer is final. This double-rating + QC-tiebreaker workflow yields a single canonical answer per (video, question) pair, bounds the contribution of any single annotator, and produces the auditable response stream that the scoring formulas in Equations~\ref{eq:hue_video}--\ref{eq:hue_model} operate on.

\paragraph{Question design and iteration.} Authoring and refining atomic questions targets two goals: \textbf{low variance} (re-running the same model on the same prompts produces a similar score), and \textbf{GT saturation} (real videos paired with the prompts score at or near $100\%$). We use these criteria to evaluate candidate questions and to iterate the question bank, revising or dropping items that fail any criterion. The process is ongoing: the GT score remains below $100\%$ (\cref{tab:hue_t2v_leaderboard,tab:hue_i2v_leaderboard}), so question-bank refinement continues.

\paragraph{Reliability and confidence intervals.} The current T2V evaluation pool samples 100 prompts from PAIBench-G, with 5 random-seed generations per prompt and up to 20 questions
per video, yielding up to 10{,}000 binary observations per model checkpoint. Treating each as an independent Bernoulli trial with underlying Yes-rate $p = \mathrm{HUE}(m)/100$, the
95\% confidence interval on the model-level HUE score (in percentage points) is $\mathrm{HUE}(m) \pm 100 \cdot 1.96\sqrt{p(1-p)/N}$ with $N$ the observation count; in practice this
gives CI$_{95}$ widths around $\pm 0.6$ points for top-tier scores on the scale of \cref{tab:hue_t2v_leaderboard}. Atomic binary judgments reduce within-annotator variance compared to
Likert-scale grading, where the annotator must integrate multiple observations, weigh them, and map onto an arbitrary scale. Each of these three operations introduces noise. The
\emph{Real video GT} row scores 93.6 on T2V prompts and 94.4 on I2V prompts, reflecting (i) minor residual prompt--video mismatches in the sampled PAIBench-G pairs despite manual
review, and (ii) imperfections in the auto-generated question set, where overly strict, ambiguous, or off-target questions can yield No or Unclear on a real video. We treat the gap
from GT to $100\%$ as a north-star for ongoing question-bank refinement.

\paragraph{T2V leaderboard.} \cref{tab:hue_t2v_leaderboard} reports current T2V results across our model panel and the strongest external T2V baselines. Veo-3.1 leads overall (91.3) and Seedance-1.5-Pro is second (90.0); Cosmos3-Super is the best open-source generator at 89.3, ahead of Wan2.2-A14B (88.2) and Cosmos3-Nano (87.6). The per-dimension picture is more favorable to Cosmos3-Super: it is the best open-source model on 9 of 12 axes, and beats every generator, including closed-source, on AV (87.7) and Physics (91.5). The gap from the strongest generator to real video (\emph{Real video GT}, 93.6) is $\sim$2.3 points overall, leaving meaningful headroom for the next round.

\paragraph{I2V leaderboard.} \cref{tab:hue_i2v_leaderboard} reports the parallel I2V evaluation. Veo-3.1 leads overall (89.7), with Cosmos3-Super trailing by only $0.1$ points (89.6);
Cosmos3-Nano (88.6) edges Wan2.2-A14B (88.4) and Seedance-1.5-Pro (87.6). Cosmos3-Super beats every generator outright on Visual Integrity (94.2), Robotics (91.1), and Miscellaneous
(94.8), and ties Veo-3.1 for the lead on Semantic Alignment (both 90.3); Cosmos3-Nano wins AV (87.6) outright; Wan2.2-A14B wins Physics (91.9). The full I2V slate spans $\sim$9 points
across generators.

\definecolor{rowours}{HTML}{E8F4E0}      
\definecolor{rowourslight}{HTML}{F2F8EC} 
\begin{table}[H]
  \centering
  \caption{\textbf{Cosmos HUE T2V leaderboard.} Per-dimension and per-domain breakdowns (\%; higher is better). Left block: Overall HUE plus the four dimensions (Semantic
Alignment, Physical Laws, Geometric Reasoning, Visual Integrity). Right block (after the rule): Overall HUE restricted to each PAIBench-G prompt domain (\textbf{C.S.}, \textbf{AV},
\textbf{Rob.}, \textbf{Ind.}, \textbf{Hum.}, \textbf{Phy.}, \textbf{Misc.}\ abbreviate Common Sense, Autonomous Vehicle, Robotics, Industry, Human, Physics, Miscellaneous).
\textcolor{nvidiagreen!60!black}{\textbf{Cosmos 3 generators (ours)}} are shaded green; \emph{Real video GT} is the upper reference. \textbf{Bold} marks the best in the column,
\underline{underline} the second best. \textbf{Cosmos3-Super} leads open-source on 9 of 12 axes and beats every generator, including closed-source, on AV (87.7) and Physics (91.5).}
  \label{tab:hue_t2v_leaderboard}
  \footnotesize
  \setlength{\tabcolsep}{3pt}
  \renewcommand{\arraystretch}{1.15}
  \begin{tabular}{l l c cccc | ccccccc}
      \toprule
      \textbf{Model} & \textbf{Type} & \textbf{Overall}
      & \makecell{\textbf{Sem.}\\\textbf{Align.}} & \makecell{\textbf{Phys.}\\\textbf{Laws}} & \makecell{\textbf{Geo.}\\\textbf{Reas.}} & \makecell{\textbf{Vis.}\\\textbf{Integ.}}
      & \textbf{C.S.} & \textbf{AV} & \textbf{Rob.} & \textbf{Ind.} & \textbf{Hum.} & \textbf{Phy.} & \textbf{Misc.} \\
      \midrule
      \rowcolor{black!8}
      \emph{Real video GT (PAI-Bench)} & --- & 93.6 & 95.0 & 90.5 & 94.1 & 94.8 & 92.0 & 93.7 & 94.9 & 94.4 & 93.2 & 93.1 & 93.6 \\
      \midrule
      \rowcolor{rowours}
      \textbf{Cosmos3-Super} & Open-sourced & 89.3 & \underline{92.4} & 85.4 & 86.6 & \underline{93.7} & 89.5 & \textbf{87.7} & 88.4 & 89.4 & 88.6 & \textbf{91.5} & 93.0 \\
      \rowcolor{rowours}
      \textbf{Cosmos3-Nano} & Open-sourced & 87.6 & 91.5 & 83.6 & 84.1 & 92.5 & 89.5 & \underline{87.0} & 86.6 & 85.8 & 86.2 & 87.9 & \underline{94.0} \\
      \midrule
      Wan2.2-A14B & Open-sourced & 88.2 & 92.1 & 83.4 & 85.6 & 92.6 & 87.8 & 84.9 & 83.7 & 91.1 & 89.9 & 87.5 & 93.4 \\
      HunyuanVideo-1.5 & Open-sourced & 86.5 & 88.4 & 81.8 & 83.6 & 93.6 & 88.3 & 81.0 & 81.3 & 88.0 & 87.9 & 87.4 & 93.3 \\
      Wan2.1-14B & Open-sourced & 84.0 & 87.7 & 77.1 & 78.5 & 92.7 & 85.3 & 79.1 & 78.2 & 87.0 & 85.1 & 85.8 & 90.7 \\
      Wan2.2-5B & Open-sourced & 80.8 & 86.9 & 73.8 & 73.6 & 89.9 & 83.4 & 73.5 & 76.0 & 84.4 & 81.3 & 82.8 & 87.9 \\
      Cosmos-Predict2.5-14B & Open-sourced & 82.1 & 88.1 & 74.6 & 75.8 & 90.7 & 81.3 & 84.4 & 76.8 & 82.4 & 81.3 & 84.5 & 90.2 \\
      Cosmos-Predict2.5-2B & Open-sourced & 81.8 & 88.1 & 74.0 & 75.1 & 90.6 & 81.5 & 81.8 & 79.5 & 82.7 & 80.3 & 82.7 & 89.8 \\
      \midrule
      Veo-3.1 & Closed-sourced & \textbf{91.3} & \textbf{94.3} & \underline{87.7} & \textbf{91.0} & \textbf{93.9} & \textbf{92.7} & 85.6 & \textbf{91.5} & \textbf{94.7} &
\underline{90.3} & 90.4 & \textbf{95.8} \\
      Seedance-1.5-Pro & Closed-sourced & \underline{90.0} & 91.2 & \textbf{88.4} & \underline{89.8} & 92.8 & \underline{90.5} & 83.6 & \underline{89.0} & \underline{92.9} &
\textbf{90.7} & \underline{91.4} & 91.7 \\
      \bottomrule
  \end{tabular}
\end{table}

\begin{table}[H]
  \centering
  \caption{\textbf{Cosmos HUE I2V leaderboard.} Per-dimension and per-domain breakdowns (\%; same protocol as \cref{tab:hue_t2v_leaderboard}, with image conditioning). Layout,
domain abbreviations, color shading, and \textbf{bold}/\underline{underline} conventions match \cref{tab:hue_t2v_leaderboard}.}
  \label{tab:hue_i2v_leaderboard}
  \footnotesize
  \setlength{\tabcolsep}{3pt}
  \renewcommand{\arraystretch}{1.15}
  \begin{tabular}{l l c cccc | ccccccc}
      \toprule
      \textbf{Model} & \textbf{Type} & \textbf{Overall}
      & \makecell{\textbf{Sem.}\\\textbf{Align.}} & \makecell{\textbf{Phys.}\\\textbf{Laws}} & \makecell{\textbf{Geo.}\\\textbf{Reas.}} & \makecell{\textbf{Vis.}\\\textbf{Integ.}}
      & \textbf{C.S.} & \textbf{AV} & \textbf{Rob.} & \textbf{Ind.} & \textbf{Hum.} & \textbf{Phy.} & \textbf{Misc.} \\
      \midrule
      \rowcolor{black!8}
      \emph{Real video GT (PAI-Bench)} & --- & 94.4 & 94.8 & 93.2 & 95.1 & 95.4 & 94.2 & 96.0 & 94.9 & 93.4 & 92.1 & 96.3 & 98.1 \\
      \midrule
      \rowcolor{rowours}
      \textbf{Cosmos3-Super} & Open-sourced & \underline{89.6} & \textbf{90.3} & \underline{87.5} & \underline{87.0} & \textbf{94.2} & \underline{90.7} & \underline{86.2} &
\textbf{91.1} & 89.6 & \underline{87.1} & \underline{91.5} & \textbf{94.8} \\
      \rowcolor{rowours}
      \textbf{Cosmos3-Nano} & Open-sourced & 88.6 & \underline{89.2} & 86.4 & 86.3 & \underline{93.5} & 90.6 & \textbf{87.6} & \underline{90.6} & 88.0 & 84.7 & 91.0 & 93.9 \\
      \midrule
      Wan2.2-A14B & Open-sourced & 88.4 & 88.6 & 86.0 & 85.7 & \underline{93.5} & 90.1 & 84.7 & 84.8 & \underline{92.0} & 86.5 & \textbf{91.9} & 92.5 \\
      HunyuanVideo-1.5 & Open-sourced & 85.6 & 87.1 & 80.8 & 83.0 & 92.5 & 90.0 & 81.7 & 77.2 & 89.6 & 85.2 & 89.5 & 91.8 \\
      Wan2.1-14B & Open-sourced & 83.9 & 85.0 & 80.2 & 80.6 & 91.2 & 88.1 & 76.8 & 74.8 & 89.9 & 84.3 & 85.8 & 93.2 \\
      Wan2.2-5B & Open-sourced & 80.4 & 83.4 & 74.6 & 75.7 & 89.5 & 86.1 & 74.3 & 68.6 & 88.4 & 79.6 & 84.8 & 90.5 \\
      Cosmos-Predict2.5-14B & Open-sourced & 83.0 & 85.0 & 77.9 & 77.5 & 92.4 & 88.1 & 85.2 & 78.4 & 83.0 & 78.5 & 84.8 & 93.2 \\
      Cosmos-Predict2.5-2B & Open-sourced & 82.6 & 86.1 & 77.3 & 78.2 & 90.2 & 85.6 & 86.0 & 79.0 & 85.6 & 77.1 & 85.1 & 92.4 \\
      \midrule
      Veo-3.1 & Closed-sourced & \textbf{89.7} & \textbf{90.3} & \textbf{87.7} & \textbf{89.2} & 93.2 & \textbf{92.3} & 86.0 & 88.6 & \textbf{93.4} & \textbf{87.2} & 91.2 &
\underline{94.3} \\
      Seedance-1.5-Pro & Closed-sourced & 87.6 & 87.7 & 85.4 & 86.5 & 91.8 & 89.5 & 82.9 & 85.7 & 90.6 & 85.7 & 90.6 & 93.0 \\
      \bottomrule
  \end{tabular}
\end{table}

\newpage
\section{Contributors and Acknowledgments}
\label{sec::contributors}

Contributors in each group are listed alphabetically by last name.

\subsection{Contributors}

\paragraph{Supervision} \mbox{}
\vspace{-0.32\baselineskip}
\begingroup
\footnotesize
\begin{tasks}[label=\relax, label-width=0pt, item-indent=4pt](5)
\task Ming-Yu Liu
\end{tasks}
\endgroup

\paragraph{Model Architecture} \mbox{}
\vspace{-0.32\baselineskip}
\begingroup
\footnotesize
\begin{tasks}[label=\relax, label-width=0pt, item-indent=4pt](5)
\task Yogesh Balaji
\task Yu-Wei Chao 
\task Prithvijit Chattopadhyay
\task Siddharth Gururani
\task Suneel Indupuru
\task George Kurian
\task Zhaoshuo Li 
\task Tsung-Yi Lin
\task Ming-Yu Liu
\task Qianli Ma
\task Kaichun Mo 
\task Min Shi
\task Jiaxiang Tang 
\task Wei-Cheng Tseng 
\end{tasks}
\endgroup

\paragraph{Reasoner Pre-Training Data} \mbox{}
\vspace{-0.32\baselineskip}
\begingroup
\footnotesize
\begin{tasks}[label=\relax, label-width=0pt, item-indent=4pt](5)
\task Yin Cui 
\task Sameer Dharur 
\task Yifan Ding 
\task Yufan Huang 
\task Kuno Kim 
\task Chia-Wen Kuo 
\task Xuan Li 
\task Tsung-Yi Lin 
\task Ruipu Luo 
\task Yatian Pang 
\task Hao Yuan 
\task Xiaohui Zeng 
\task Haotian Zhang 
\task Jing Zhang 
\end{tasks}
\endgroup

\paragraph{Reasoner Post-Training Data} \mbox{}
\vspace{-0.32\baselineskip}\paragraph{\footnotesize Spatial Understanding} \mbox{}
\vspace{-0.32\baselineskip}
\begingroup
\footnotesize
\begin{tasks}[label=\relax, label-width=0pt, item-indent=4pt](5)
\task Eric Cameracci
\task Yan Chang
\task Xiaotong Chen
\task An-Chieh Cheng
\task Aleksandr Efitorov
\task Ryan Ji
\task Jingyi Jin
\task Sifei Liu
\task Hesam Rabeti
\task Marilyn Reeb
\task Yichu Yang
\task Shun Zhang
\end{tasks}
\endgroup

\vspace{-0.32\baselineskip}\paragraph{\footnotesize Temporal Understanding} \mbox{}
\vspace{-0.32\baselineskip}
\begingroup
\footnotesize
\begin{tasks}[label=\relax, label-width=0pt, item-indent=4pt](5)
\task Zaid Pervaiz Bhat
\task Yin Cui
\task Zekun Hao 
\task Arihant Jain
\task Boyi Li 
\task Xuan Li
\task Marco Pavone 
\task Varun Praveen
\task Shitao Tang 
\end{tasks}
\endgroup

\vspace{-0.32\baselineskip}\paragraph{\footnotesize 2D Grounding} \mbox{}
\vspace{-0.32\baselineskip}
\begingroup
\footnotesize
\begin{tasks}[label=\relax, label-width=0pt, item-indent=4pt](5)
\task Prithvijit Chattopadhyay 
\task An-Chieh Cheng
\task Yin Cui 
\task Siddharth Gururani 
\task Jaehun Jung 
\task Zhiqi Li 
\task Sifei Liu
\task Varun Praveen 
\task Shihao Wang 
\task Yu Wang 
\task Zhiding Yu 
\end{tasks}
\endgroup

\vspace{-0.32\baselineskip}\paragraph{\footnotesize Robotics} \mbox{}
\vspace{-0.32\baselineskip}
\begingroup
\footnotesize
\begin{tasks}[label=\relax, label-width=0pt, item-indent=4pt](5)
\task Yan Chang 
\task Prithvijit Chattopadhyay 
\task Aigul Dzhumamuratova 
\task Aleksandr Efitorov 
\task Ryan Ji 
\task Jingyi Jin 
\task Jaehun Jung 
\task Zhaoshuo Li 
\task Zhiqi Li 
\task Kaichun Mo 
\task Soha Pouya 
\task Hesam Rabeti 
\task Haoxiang Wang 
\task Shihao Wang 
\task Yichu Yang 
\task Zhiding Yu 
\task Shun Zhang 
\end{tasks}
\endgroup

\vspace{-0.32\baselineskip}\paragraph{\footnotesize Driving} \mbox{}
\vspace{-0.32\baselineskip}
\begingroup
\footnotesize
\begin{tasks}[label=\relax, label-width=0pt, item-indent=4pt](5)
\task Niket Agarwal 
\task Mohammad Qazim Bhat 
\task Yulong Cao 
\task Ke Chen 
\task Wenhao Ding 
\task Yifan Ding 
\task Amol Fasale 
\task Yufan Huang 
\task Boris Ivanovic 
\task Jingyi Jin 
\task Marco Pavone 
\task Yan Wang 
\task Xinshuo Weng 
\task Tianjun Xiao 
\task Jiashu Xu 
\task Xiaodong Yang 
\end{tasks}
\endgroup

\vspace{-0.32\baselineskip}\paragraph{\footnotesize Smart Infrastructure} \mbox{}
\vspace{-0.32\baselineskip}
\begingroup
\footnotesize
\begin{tasks}[label=\relax, label-width=0pt, item-indent=4pt](5)
\task Zaid Pervaiz Bhat 
\task Yifan Ding 
\task Vikram Fugro 
\task Prashant Gaikwad 
\task Tomasz Kornuta 
\task Xiaolong Li 
\task Piyush Shekdar 
\task Vignesh Srinivasakumar 
\task Paris Zhang 
\task Yilin Zhao 
\end{tasks}
\endgroup

\vspace{-0.32\baselineskip}\paragraph{\footnotesize Healthcare} \mbox{}
\vspace{-0.32\baselineskip}
\begingroup
\footnotesize
\begin{tasks}[label=\relax, label-width=0pt, item-indent=4pt](5)
\task Yufan He
\task Nic Ma
\task Daguang Xu
\task Dong Yang
\end{tasks}
\endgroup

\vspace{-0.32\baselineskip}\paragraph{\footnotesize Visual Critics} \mbox{}
\vspace{-0.32\baselineskip}
\begingroup
\footnotesize
\begin{tasks}[label=\relax, label-width=0pt, item-indent=4pt](5)
\task Zekun Hao 
\task Haotian Zhang 
\end{tasks}
\endgroup

\vspace{-0.32\baselineskip}\paragraph{\footnotesize Prompt Upsampling} \mbox{}
\vspace{-0.32\baselineskip}
\begingroup
\footnotesize
\begin{tasks}[label=\relax, label-width=0pt, item-indent=4pt](5)
\task Yifan Ding
\task Hamid Eghbalzadeh
\task Francesco Ferroni
\task Siddharth Gururani
\task Xuan Li
\task Seungjun Nah
\task Andrew Z. Wang
\task Boxiang Wang
\task Jiashu Xu
\task Mengyao Xu
\task Xingqian Xu
\end{tasks}
\endgroup

\vspace{-0.32\baselineskip}\paragraph{\footnotesize Filtering and Cleanup} \mbox{}
\vspace{-0.32\baselineskip}
\begingroup
\footnotesize
\begin{tasks}[label=\relax, label-width=0pt, item-indent=4pt](5)
\task Sameer Dharur
\task Yifan Ding
\task Kuno Kim
\task Xuan Li
\end{tasks}
\endgroup

\paragraph{Generator Data -- Image} \mbox{}
\paragraph{\footnotesize Deduplication and Filtering} \mbox{}
\vspace{-0.32\baselineskip}
\begingroup
\footnotesize
\begin{tasks}[label=\relax, label-width=0pt, item-indent=4pt](5)
\task Sameer Dharur
\task Jiaojiao Fan
\task Francesco Ferroni
\task Jinfeng Li
\task Stella Shi
\task Mengyao Xu
\task Haotian Zhang
\task Fengzhe Zhou
\end{tasks}
\endgroup

\vspace{-0.32\baselineskip}\paragraph{\footnotesize Captioning} \mbox{}
\vspace{-0.32\baselineskip}
\begingroup
\footnotesize
\begin{tasks}[label=\relax, label-width=0pt, item-indent=4pt](5)
\task Seungjun Nah
\task Shitao Tang
\task Andrew Z. Wang
\task Boxiang Wang
\task Jiashu Xu
\task Mengyao Xu
\task Xingqian Xu
\end{tasks}
\endgroup

\vspace{-0.32\baselineskip}\paragraph{\footnotesize Synthetic Data} \mbox{}
\vspace{-0.32\baselineskip}
\begingroup
\footnotesize
\begin{tasks}[label=\relax, label-width=0pt, item-indent=4pt](5)
\task Francesco Ferroni
\task Seungjun Nah
\task Stella Shi
\task Xingqian Xu
\end{tasks}
\endgroup

\vspace{-0.32\baselineskip}\paragraph{\footnotesize SFT Data} \mbox{}
\vspace{-0.32\baselineskip}
\begingroup
\footnotesize
\begin{tasks}[label=\relax, label-width=0pt, item-indent=4pt](5)
\task Vanni Brighella 
\task Jiaxin Cao 
\task Chieh-Yun Chen 
\task Magdalena Dadela 
\task Marco Di Lucca 
\task Jiaojiao Fan 
\task Francesco Ferroni 
\task Xiao Fu 
\task Jinwei Gu 
\task Miguel Guerrero 
\task Zekun Hao 
\task Cyrus Hogg 
\task Scott Kassekert 
\task Freya Li 
\task Ling Li 
\task Ming-Yu Liu 
\task Hyejin Moon 
\task Seungjun Nah 
\task Sehwi Park 
\task Morteza Ramezanali 
\task Shitao Tang 
\task Andrew Z. Wang 
\task Ting-Chun Wang 
\task Jiashu Xu 
\task Mengyao Xu 
\task Xingqian Xu 
\task Xiaodong Yang 
\task Jenny Zhang 
\end{tasks}
\endgroup

\paragraph{Generator Data -- Video} \mbox{}
\vspace{-0.32\baselineskip}\paragraph{\footnotesize Deduplication and Filtering} \mbox{}
\vspace{-0.32\baselineskip}
\begingroup
\footnotesize
\begin{tasks}[label=\relax, label-width=0pt, item-indent=4pt](5)
\task Yogesh Balaji
\task Prithvijit Chattopadhyay
\task Yin Cui
\task Sameer Dharur
\task Imad El Hanafi
\task Jiaojiao Fan
\task Jinwei Gu
\task Zekun Hao
\task Jacob Huffman
\task Jinfeng Li
\task Chen-Hsuan Lin
\task Alice Luo
\task Yatian Pang
\task Stella Shi
\task Shitao Tang
\task Andrew Z. Wang
\task Ting-Chun Wang
\task Mengyao Xu
\task Haotian Zhang
\task Jing Zhang
\task Fengzhe Zhou
\end{tasks}
\endgroup

\vspace{-0.32\baselineskip}\paragraph{\footnotesize Captioning} \mbox{}
\vspace{-0.32\baselineskip}
\begingroup
\footnotesize
\begin{tasks}[label=\relax, label-width=0pt, item-indent=4pt](5)
\task Xuan Li
\task Seungjun Nah
\task Shitao Tang
\task Andrew Z. Wang
\task Jiashu Xu
\task Mengyao Xu
\task Xingqian Xu
\end{tasks}
\endgroup

\vspace{-0.32\baselineskip}\paragraph{\footnotesize Synthetic Data} \mbox{}
\vspace{-0.32\baselineskip}
\begingroup
\footnotesize
\begin{tasks}[label=\relax, label-width=0pt, item-indent=4pt](5)
\task Martin Antolini 
\task Adeline Aubame 
\task Eric Cameracci 
\task Mark Carlson 
\task Carlos Casanova 
\task Yan Chang 
\task Xiaotong Chen 
\task Xiu Chia 
\task Nalin Dadhich 
\task Rodrigo Vieira Del Monte 
\task Robert Denomme 
\task Aleksandr Efitorov 
\task Imad El Hanafi
\task Jinwei Gu 
\task Ankur Handa 
\task Chris Helvig 
\task Ryan Ji 
\task Jingyi Jin 
\task Sunny Kim 
\task JF Lafleche 
\task Jayjun Lee 
\task Jiajun Li 
\task Shangru Li 
\task Hai Loc Lu 
\task Xiangyu Lu 
\task Alice Luo
\task Louis Marcoux 
\task Miguel Martin 
\task Durra Mohsin 
\task Kirill Motkov 
\task Thabang Ngazimbi 
\task Julian Ouyang 
\task Sehwi Park 
\task Soha Pouya 
\task Hesam Rabeti 
\task Morteza Ramezanali 
\task Marilyn Reeb 
\task Stella Shi
\task Kayley Ting 
\task Qiao Wang 
\task Mengyao Xu
\task Hans Yang 
\task Shun Zhang 
\task Charles Zhou 
\end{tasks}
\endgroup

\vspace{-0.32\baselineskip}\paragraph{\footnotesize SFT Data} \mbox{}
\vspace{-0.32\baselineskip}
\begingroup
\footnotesize
\begin{tasks}[label=\relax, label-width=0pt, item-indent=4pt](5)
\task Aditi 
\task Arslan Ali 
\task Vanni Brighella 
\task Tiffany Cai 
\task Ting-Yun Chang 
\task Prithvijit Chattopadhyay 
\task Chieh-Yun Chen 
\task Xiaotong Chen 
\task Jeana Choi 
\task Magdalena Dadela 
\task Marco Di Lucca 
\task Naomi Eigbe 
\task Jiaojiao Fan 
\task Amol Fasale 
\task Francesco Ferroni 
\task Xiao Fu 
\task Katelyn Gao 
\task Yihuai Gao 
\task Akash Gokul 
\task Jinwei Gu 
\task Miguel Guerrero 
\task Zekun Hao 
\task Nathan Hayes-Roth 
\task Cyrus Hogg 
\task Yufan Huang 
\task DeLesley Hutchins 
\task Ryan Ji 
\task Yanan Jian 
\task Jingyi Jin 
\task Scott Kassekert 
\task Ashna Khetan 
\task Gwanghyun Kim 
\task Xin Kong 
\task Omar Laymoun 
\task Gabriele Leone 
\task Freya Li 
\task Ling Li 
\task Zhaoshuo Li 
\task Chen-Hsuan Lin 
\task Ming-Yu Liu 
\task Alice Luo 
\task Qianli Ma 
\task Ashkan Mirzaei 
\task Hyejin Moon 
\task Seungjun Nah 
\task Sehwi Park 
\task Morteza Ramezanali 
\task Min Shi 
\task Stella Shi 
\task Amir Sotoodeh 
\task Shitao Tang 
\task Tolou Tavakkoli 
\task Wei-Cheng Tseng 
\task Andrew Z. Wang 
\task Boxiang Wang 
\task Shijie Wang 
\task Ting-Chun Wang 
\task David Wehr 
\task Fangyin Wei 
\task Jiashu Xu 
\task Mengyao Xu 
\task Xingqian Xu 
\task Yao Xu 
\task Hans Yang 
\task Xiaodong Yang 
\task Jenny Zhang 
\task Jing Zhang 
\task Liangkai Zhang 
\task Xuanmeng Zhang 
\task Fengzhe Zhou 
\end{tasks}
\endgroup

\paragraph{Generator Data -- Audio} \mbox{}
\begingroup
\footnotesize
\begin{tasks}[label=\relax, label-width=0pt, item-indent=4pt](5)
\task Sreyan Ghosh
\task Siddharth Gururani
\task Tingle Li
\end{tasks}
\endgroup

\paragraph{Generator Data -- Action} \mbox{}
\vspace{-0.32\baselineskip}\paragraph{\footnotesize Robotics} \mbox{}
\vspace{-0.32\baselineskip}
\begingroup
\footnotesize
\begin{tasks}[label=\relax, label-width=0pt, item-indent=4pt](5)
\task Yu-Wei Chao
\task Qizhi Chen
\task Yihuai Gao
\task Joel Jang
\task Gwanghyun Kim
\task Xin Kong
\task Zhaoshuo Li
\task Kaichun Mo
\task Delin Qu
\task Wei-Cheng Tseng
\task Yichu Yang
\end{tasks}
\endgroup

\vspace{-0.32\baselineskip}\paragraph{\footnotesize Egocentric} \mbox{}
\vspace{-0.32\baselineskip}
\begingroup
\footnotesize
\begin{tasks}[label=\relax, label-width=0pt, item-indent=4pt](5)
\task Ling Li
\task Zhaoshuo Li
\task Qianli Ma
\end{tasks}
\endgroup

\vspace{-0.32\baselineskip}\paragraph{\footnotesize Autonomous Vehicle} \mbox{}
\vspace{-0.32\baselineskip}
\begingroup
\footnotesize
\begin{tasks}[label=\relax, label-width=0pt, item-indent=4pt](5)
\task Wenjie Luo
\task Jiaxiang Tang
\task Yan Wang
\task Xiaodong Yang
\task Yurong You
\end{tasks}
\endgroup

\vspace{-0.32\baselineskip}\paragraph{\footnotesize Camera Motion} \mbox{}
\vspace{-0.32\baselineskip}
\begingroup
\footnotesize
\begin{tasks}[label=\relax, label-width=0pt, item-indent=4pt](5)
\task Chen-Hsuan Lin
\task Jiaxiang Tang
\task Shitao Tang
\end{tasks}
\endgroup

\paragraph{Generator Data -- Transfer} \mbox{}
\vspace{-0.32\baselineskip}
\begingroup
\footnotesize
\begin{tasks}[label=\relax, label-width=0pt, item-indent=4pt](5)
\task Francesco Ferroni
\task Qianli Ma
\task Min Shi
\task Ting-Chun Wang
\task Fangyin Wei
\end{tasks}
\endgroup

\paragraph{Training Recipe} \mbox{}
\vspace{-0.32\baselineskip}\paragraph{\footnotesize Edge LLM Training} \mbox{}
\vspace{-0.32\baselineskip}
\begingroup
\footnotesize
\begin{tasks}[label=\relax, label-width=0pt, item-indent=4pt](5)
\task Wenliang Dai
\task Dongfu Jiang
\task Kezhi Kong
\task Zihan Liu
\task Deepak Narayanan
\task Mostofa Patwary
\task Wei Ping
\task Shrimai Prabhumoye
\task Mohammad Shoeybi
\task Roger Waleffe
\task Boxiang Wang
\end{tasks}
\endgroup

\vspace{-0.32\baselineskip}\paragraph{\footnotesize Reasoner Pre-Training} \mbox{}
\vspace{-0.32\baselineskip}
\begingroup
\footnotesize
\begin{tasks}[label=\relax, label-width=0pt, item-indent=4pt](5)
\task Jiaxin Cao
\task Liang Feng
\task Kuno Kim
\task Tsung-Yi Lin
\task Ruipu Luo
\end{tasks}
\endgroup

\vspace{-0.32\baselineskip}\paragraph{\footnotesize Reasoner SFT} \mbox{}
\vspace{-0.32\baselineskip}
\begingroup
\footnotesize
\begin{tasks}[label=\relax, label-width=0pt, item-indent=4pt](5)
\task Xuan Li
\task Zhaoshuo Li
\task Tsung-Yi Lin
\task Xiaohui Zeng
\end{tasks}
\endgroup

\vspace{-0.32\baselineskip}\paragraph{\footnotesize Generator Pre-training} \mbox{}
\vspace{-0.32\baselineskip}
\begingroup
\footnotesize
\begin{tasks}[label=\relax, label-width=0pt, item-indent=4pt](5)
\task Yogesh Balaji
\task Prithvijit Chattopadhyay
\task Siddharth Gururani
\task Suneel Indupuru
\task George Kurian
\task Ting-Chun Wang
\task Jiashu Xu
\task Jing Zhang
\end{tasks}
\endgroup

\vspace{-0.32\baselineskip}\paragraph{\footnotesize Generator Mid-training} \mbox{}
\vspace{-0.32\baselineskip}
\begingroup
\footnotesize
\begin{tasks}[label=\relax, label-width=0pt, item-indent=4pt](5)
\task Yogesh Balaji 
\task Prithvijit Chattopadhyay 
\task Francesco Ferroni
\task Siddharth Gururani
\task Suneel Indupuru
\task George Kurian
\task Zhaoshuo Li
\task Qianli Ma
\task Kaichun Mo
\task Trung Pham
\task Min Shi
\task Ting-Chun Wang
\task Fangyin Wei
\task Jing Zhang
\end{tasks}
\endgroup

\vspace{-0.32\baselineskip}\paragraph{\footnotesize Generator Post-training} \mbox{}
\vspace{-0.32\baselineskip}
\begingroup
\footnotesize
\begin{tasks}[label=\relax, label-width=0pt, item-indent=4pt](5)
\task Yu-Wei Chao
\task Francesco Ferroni
\task Zekun Hao
\task Hao Liang
\task Kaichun Mo
\task Seungjun Nah
\task Xingqian Xu
\task Yichu Yang
\end{tasks}
\endgroup

\paragraph{Infrastructure} \mbox{}
\vspace{-0.32\baselineskip}\paragraph{\footnotesize Data} \mbox{}
\vspace{-0.32\baselineskip}
\begingroup
\footnotesize
\begin{tasks}[label=\relax, label-width=0pt, item-indent=4pt](5)
\task Imad El Hanafi
\task Francesco Ferroni
\task Siddharth Gururani 
\task Jacob Huffman
\task Alice Luo
\task David Page
\task Stella Shi
\task Sergei Vasilev
\task Pengcuo Zeren
\task Jing Zhang
\end{tasks}
\endgroup

\vspace{-0.32\baselineskip}\paragraph{\footnotesize Training} \mbox{}
\vspace{-0.32\baselineskip}
\begingroup
\footnotesize
\begin{tasks}[label=\relax, label-width=0pt, item-indent=4pt](5)
\task Alisson Azzolini
\task Junjie Bai
\task Maciej Bala
\task Jiaxin Cao
\task Hassan Eslami
\task Liang Feng
\task Vivek Goel
\task Rama Govindaraju
\task Elfie Guo
\task Ali Hassani
\task Yufan Huang
\task Suneel Indupuru
\task Kuno Kim
\task Hui Kuang
\task George Kurian
\task Himangshu Lahkar
\task Pengcheng Li
\task Hao Liang
\task Maosheng Liao
\task Xiangyu Lu
\task Ruipu Luo
\task Martin Ding Ma
\task Sunil Srinivasa
\task Bartosz Stefaniak
\task Shangkun Sun
\task Yangyang Tang
\task Yan Wang
\task Dinghao Yang
\task Hao Yuan
\task Pengcuo Zeren
\task Jing Zhang
\task Xuanmeng Zhang
\task Yuliya Zhautouskaya
\task Dima Zhylko
\end{tasks}
\endgroup

\vspace{-0.32\baselineskip}\paragraph{\footnotesize Serving} \mbox{}
\vspace{-0.32\baselineskip}
\begingroup
\footnotesize
\begin{tasks}[label=\relax, label-width=0pt, item-indent=4pt](5)
\task Jon Allen
\task Maciej Bala
\task Yogesh Balaji
\task Aarti Basant
\task Yu-Wei Chao
\task Chieh-Yun Chen
\task Jeana Choi
\task Joyjit Daw
\task Sameer Dharur
\task Yuzhu Dong
\task Hamid Eghbalzadeh
\task Benedikt Falk
\task Sergiy Fefilatyev
\task Liang Feng
\task Francesco Ferroni
\task Rama Govindaraju
\task Zekun Hao
\task Suneel Indupuru
\task Atharva Joshi
\task Julia Kiczka
\task Slawek Kierat
\task Xin Kong
\task Egor Krivov
\task Chia-Wen Kuo
\task George Kurian
\task Wojciech Kutak
\task Zhaoshuo Li
\task Maosheng Liao
\task Tsung-Yi Lin
\task Dawid Majchrowski
\task Qing Miao
\task Shreyas Misra
\task Saeid Motiian
\task Seungjun Nah
\task Yatian Pang
\task Wojciech Rymer
\task Mateusz Sieniawski
\task Rahul Heinrich Steiger
\task Jiaxiang Tang
\task Yangyang Tang
\task Krzysztof Tomala
\task Andrew Z. Wang
\task Kedi Wu
\task Hongchi Xia
\task Ruqing Xu
\task Cindy Zha
\task Haotian Zhang
\task Jing Zhang
\task Liangkai Zhang
\task Xuanmeng Zhang
\task Yuliya Zhautouskaya
\task Shilin Zhu
\task Artur Zolkowski
\end{tasks}
\endgroup

\vspace{-0.32\baselineskip}\paragraph{\footnotesize Benchmark} \mbox{}
\vspace{-0.32\baselineskip}
\begingroup
\footnotesize
\begin{tasks}[label=\relax, label-width=0pt, item-indent=4pt](5)
\task Mukesh Beladiya
\task Dan Blick
\task Tiffany Cai
\task Roshan Chaudhari
\task Ke Ding
\task Yifan Ding
\task Yuzhu Dong
\task Hamid Eghbalzadeh
\task Naomi Eigbe
\task Francesco Ferroni
\task Aryaman Gupta
\task Nathan Hayes-Roth
\task Yanan Jian
\task Nikhilesh Joshi
\task Ashna Khetan
\task Saurav Kumar
\task Hao Liang
\task Martin Ding Ma
\task Qianli Ma
\task Miguel Martin
\task Ashkan Mirzaei
\task Trung Pham
\task Tolou Tavakkoli
\task Jibin Varghese
\task Thomas Volk
\task Ting-Chun Wang
\end{tasks}
\endgroup

\paragraph{Results and Benchmarks} \mbox{}
\vspace{-0.32\baselineskip}\paragraph{\footnotesize Reasoning} \mbox{}
\vspace{-0.32\baselineskip}
\begingroup
\footnotesize
\begin{tasks}[label=\relax, label-width=0pt, item-indent=4pt](5)
\task Zaid Pervaiz Bhat
\task Dan Blick
\task Wenyan Cong
\task Yin Cui
\task Ke Ding
\task Yifan Ding
\task Naomi Eigbe
\task Zekun Hao
\task Ryan Ji
\task Weiwei Kang
\task Kuno Kim
\task Tomasz Kornuta
\task Boyi Li
\task Xuan Li
\task Tsung-Yi Lin
\task Xiangyu Lu
\task Miguel Martin
\task Varun Praveen
\task Shitao Tang
\task Andrew Z. Wang
\task Shihao Wang
\task Yan Wang
\task Yu Wang
\task Mengyao Xu
\task Xiaodong Yang
\task Zhiding Yu
\task Paris Zhang
\task Shun Zhang
\task Yilin Zhao
\end{tasks}
\endgroup

\vspace{-0.32\baselineskip}\paragraph{\footnotesize Visual Generation} \mbox{}
\vspace{-0.32\baselineskip}
\begingroup
\footnotesize
\begin{tasks}[label=\relax, label-width=0pt, item-indent=4pt](5)
\task Arslan Ali
\task Yogesh Balaji 
\task Mukesh Beladiya
\task Tiffany Cai
\task Prithvijit Chattopadhyay 
\task Yifan Ding
\task Yilun Du 
\task Hamid Eghbalzadeh 
\task Jiaojiao Fan
\task Francesco Ferroni
\task Akash Gokul
\task Jinwei Gu
\task Aryaman Gupta
\task Siddharth Gururani 
\task Zekun Hao
\task Yufan Huang
\task Joel Jang 
\task Yanan Jian
\task Gwanghyun Kim 
\task Saurav Kumar
\task Qianli Ma
\task Ashkan Mirzaei 
\task Seungjun Nah
\task Mahesh Patekar
\task Amir Sotoodeh
\task Tolou Tavakkoli
\task Jibin Varghese
\task Raju Wagwani
\task Andrew Z. Wang
\task Ting-Chun Wang 
\task Mengyao Xu
\task Xingqian Xu
\task Haotian Zhang
\end{tasks}
\endgroup

\vspace{-0.32\baselineskip}\paragraph{\footnotesize Action Generation} \mbox{}
\vspace{-0.32\baselineskip}
\begingroup
\footnotesize
\begin{tasks}[label=\relax, label-width=0pt, item-indent=4pt](5)
\task Yu-Wei Chao 
\task Qizhi Chen 
\task Alperen Degirmenci 
\task Yuzhu Dong 
\task Aigul Dzhumamuratova 
\task Yihuai Gao 
\task Hugo Hadfield 
\task Suneel Indupuru 
\task Ryan Ji 
\task Jingyi Jin 
\task Gwanghyun Kim 
\task Xin Kong 
\task George Kurian 
\task Zhaoshuo Li 
\task Hao Liang 
\task Jiangran Lyu 
\task Qianli Ma 
\task Kaichun Mo 
\task Sehwi Park 
\task Tianwei Shen 
\task Sunil Srinivasa 
\task Jiaxiang Tang 
\task Wei-Cheng Tseng 
\task David Wehr 
\task Xiaodong Yang 
\task Xuning Yang 
\task Yichu Yang 
\task Liangkai Zhang 
\task Shun Zhang 
\task Zhizheng Zhang 
\end{tasks}
\endgroup

\vspace{-0.32\baselineskip}\paragraph{\footnotesize Audio Generation} \mbox{}
\vspace{-0.32\baselineskip}
\begingroup
\footnotesize
\begin{tasks}[label=\relax, label-width=0pt, item-indent=4pt](5)
\task Mukesh Beladiya
\task Vanni Brighella 
\task Magdalena Dadela 
\task Sreyan Ghosh
\task Miguel Guerrero 
\task Siddharth Gururani
\task Cyrus Hogg 
\task Tingle Li
\task Morteza Ramezanali 
\end{tasks}
\endgroup

\vspace{-0.32\baselineskip}\paragraph{\footnotesize Transfer Generation} \mbox{}
\vspace{-0.32\baselineskip}
\begingroup
\footnotesize
\begin{tasks}[label=\relax, label-width=0pt, item-indent=4pt](5)
\task Francesco Ferroni
\task Qianli Ma
\task Trung Pham
\task Min Shi
\task Ting-Chun Wang
\end{tasks}
\endgroup

\paragraph{Other Contributions} \mbox{}
\vspace{-0.32\baselineskip}
\begingroup
\footnotesize
\begin{tasks}[label=\relax, label-width=0pt, item-indent=4pt](5)
\task Josh Bapst 
\task Han Cai 
\task Junyu Chen 
\task Wenkai Chen
\task Yu Chen
\task Click Cheng 
\task Chaeyeon Chung 
\task Nicole Drumheller 
\task Jim Fan 
\task Sanja Fidler 
\task TJ Galda 
\task Wenhang Ge 
\task Arushi Goel 
\task Song Han 
\task Mohammad Harrim 
\task Madison Huang 
\task Michael Huang 
\task Sophia Huang 
\task Pranjali Joshi 
\task Andy Ju 
\task Jan Kautz 
\task Zhifeng Kong 
\task Sanggil Lee 
\task Pawel Morkisz 
\task Yashraj Narang 
\task Shubham Pachori 
\task Xuanchi Ren 
\task Kristen Rumley 
\task Jun Saito 
\task Yeongho Seol 
\task John Shao
\task Humphrey Shi 
\task Kevin Shih 
\task Shuran Song 
\task Alexander Sotelo 
\task Yue Tang 
\task Rohit Watve 
\task Jay Zhangjie Wu 
\task Summer Xiao
\task Kevin Xie 
\task Simon Yuen 
\task Ann Zhao
\task Yuke Zhu 
\end{tasks}
\endgroup

\subsection{Acknowledgments}
\begingroup
\footnotesize
\begin{tasks}[label=\relax, label-width=0pt, item-indent=4pt](5)
\task Pranav Atreya
\task Vince Auletta
\task Kameron Balutch
\task Stan Birchfield
\task Drew Bishop
\task Valts Blukis
\task Jinju Chu
\task Demi Cruz
\task Jenna Diamond
\task John Dickinson
\task Santanu Dutta
\task Henry Estela
\task Mohamed Fawzy
\task Gail Frederick
\task Chad Geisler
\task Jeffrey Glick
\task Nikhil Gupta
\task Christine Harvey
\task Greg Heinrich
\task Emily Hill
\task Chris Holguin
\task Mike Houston
\task Spencer Huang
\task Rizwan Khan
\task Jim King
\task Elena Lantz
\task Ben Levin
\task Jordan Mackenzie
\task Amanda Moran
\task Ankit Patel
\task Greg Pauloski
\task Lindsey Pavao
\task Erin Potter
\task Ryan Punamiya 
\task Karan Sapra 
\task Jane Polak Scowcroft
\task Stas Sergienko
\task Brandon Soubasis
\task Kevin Su
\task Sai Swarup
\task Allyson Tabas
\task Andrew Tao 
\task Andy Tran
\task Jonathan Tremblay
\task Derek Tzeng
\task Isabelle Udell
\task Gandhi Vaithilingam
\task Jie Wang
\task Jing Wang
\task Qi Wang
\task Bob Wise
\task Danfei Xu 
\task Jay Yang
\task Yuting Yang
\task Aileen Zaman
\end{tasks}
\endgroup

\clearpage
\begingroup
\raggedright
\sloppy
\makeatletter
\renewcommand{\bibsection}{%
  \par\noindent{\headingfont\refname}\par\vspace{5pt}%
}
\setlength{\bibhang}{0pt}
\renewcommand\@bibsetup[1]{%
  \setlength{\leftmargin}{0pt}%
  \setlength{\itemindent}{0pt}%
  \setlength{\labelwidth}{0pt}%
  \setlength{\labelsep}{0pt}%
  \setlength{\listparindent}{0pt}%
  \setlength{\itemsep}{\bibsep}%
  \setlength{\parsep}{\z@}%
}
\makeatother
\setlength{\emergencystretch}{3em}
\Urlmuskip=0mu plus 1mu\relax
\bibliographystyle{plainnat}
\bibliography{main}
\endgroup

\end{document}